%% file: main.tex
\documentclass{article}
\pdfoutput=1

\usepackage[dvipsnames]{xcolor}
\usepackage[utf8]{inputenc} 
\usepackage[T1]{fontenc}    
\usepackage{hyperref}       
\usepackage{url}            
\usepackage{booktabs}       
\usepackage{amsfonts}       
\usepackage{nicefrac}       
\usepackage{microtype}      
\usepackage{wrapfig}
\usepackage{subcaption}
\usepackage{graphicx}
\usepackage{multirow}

\usepackage[inline]{enumitem}
\usepackage{bm}
\usepackage{amsfonts,amssymb,amsmath}
\usepackage[capitalise]{cleveref}
\newtheorem{theorem}{Proposition}

\DeclareMathAlphabet{\pazocal}{OMS}{zplm}{m}{n}

\usepackage[accepted]{icml2021}

\input{commands.tex}

\icmltitlerunning{On Disentangled Representations Learned from Correlated Data}

\begin{document}

\twocolumn[
\icmltitle{On Disentangled Representations Learned from Correlated Data}
\icmlsetsymbol{equal}{*}

\begin{icmlauthorlist}
\icmlauthor{Frederik Tr\"auble}{mpi}
\icmlauthor{Elliot Creager}{to}
\icmlauthor{Niki Kilbertus}{hai}
\icmlauthor{Francesco Locatello}{amzn}
\icmlauthor{Andrea Dittadi}{dtu}
\icmlauthor{Anirudh Goyal}{mila}
\icmlauthor{Bernhard Sch\"olkopf}{mpi}
\icmlauthor{Stefan Bauer}{mpi,cifar}
\end{icmlauthorlist}

\icmlaffiliation{to}{University of Toronto and Vector Institute}
\icmlaffiliation{mpi}{Max Planck Institute for Intelligent Systems, Tübingen, Germany}
\icmlaffiliation{hai}{Helmholtz AI, Munich}
\icmlaffiliation{dtu}{Technical University of Denmark}
\icmlaffiliation{mila}{Mila and Université de Montréal}
\icmlaffiliation{amzn}{Amazon (work partly done when FL was at ETH Zurich and MPI-IS)}
\icmlaffiliation{cifar}{CIFAR Azrieli Global Scholar}

\icmlcorrespondingauthor{Frederik Träuble}{frederik.traeuble@tuebingen.mpg.de}

\icmlkeywords{Machine Learning, ICML, representation learning, disentanglement, correlations}

\vskip 0.3in
]

\printAffiliationsAndNotice{}

\begin{abstract}
\looseness=-1
The focus of disentanglement approaches has been on identifying independent factors of variation in data. However, the causal variables underlying real-world observations are often not statistically independent.
In this work, we bridge the gap to real-world scenarios by analyzing the behavior of the most prominent disentanglement approaches on correlated data in a large-scale empirical study (including 4260 models). We show and quantify that systematically induced correlations in the dataset are being learned and reflected in the latent representations, which has implications for downstream applications of disentanglement such as fairness. We also demonstrate how to resolve these latent correlations, either using weak supervision during training  or by post-hoc correcting a pre-trained model with a small number of labels.

\end{abstract}

\section{Introduction}
\label{sec:intro}
Disentangled representations promise generalization to unseen scenarios \citep{higgins2017darla,locatello2020weakly}, increased interpretability \citep{adel2018discovering, higgins2018scan}, domain generalization \citep{miladinovic2019disentangled} or better sim2real transfer \citep{dittadi2021transfer} as well as a connection to robustness and causal inference \citep{suter2018interventional}.  

\begin{figure}
\begin{center}
\vspace{1cm}
\includegraphics[width=0.75\columnwidth]{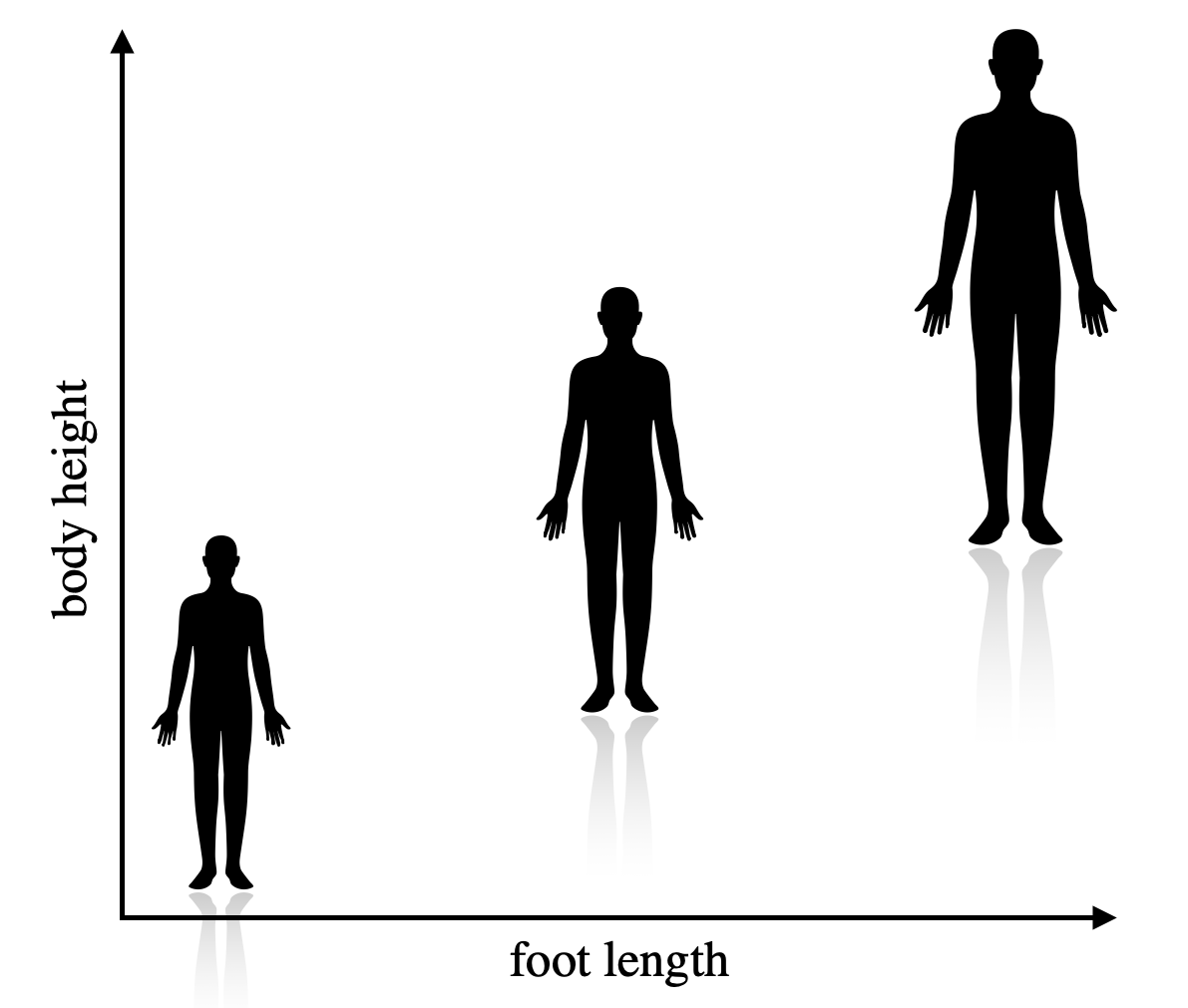}
\end{center}
\caption{While in principle we would like to consider foot length and body height to be independent features of our model, they exhibit strong positive correlation in observed data.}
\label{fig:causal-structure}
\end{figure}

While the advantages of disentangled representations have been well established, they generally assume the existence of natural factors that vary independently within the given dataset, which is rarely the case in real-world settings. As an example, consider a dataset containing images of various persons (see \cref{fig:causal-structure}). Higher-level factors of this representation, such as foot length and body height are in fact found to be statistically correlated \citep{agnihotri2007estimation, grivas2008correlation}. Humans can still conceive varying both factors independently across their entire range. However, only given a correlated dataset, learning systems may be tempted to encode both factors simultaneously in a single \emph{size} factor. It is thus argued that what we actually want to infer are independent (causal) mechanisms \citep{goyal2019recurrent,leeb2020structured, scholkopf2021toward}. 

A complex generative model can be thought of as the composition of independent mechanisms, which generate high-dimensional observations (such as images or videos). In the causality community, this is often considered a prerequisite to achieve representations that are robust to interventions upon variables determined by such models \citep{PetJanSch17}. The notion of \emph{disentangled} representation is one particular instantiation of this idea \citep{bengio2013representation}. The goal of disentanglement learning is to find a representation of the data which captures all the ground-truth factors of variation (FoV) independently. 

Despite the recent growth of the field, state-of-the-art disentanglement learners have mostly been trained and evaluated on datasets in which all FoV indeed vary independently from each other. Their performance remains unknown for more realistic settings where FoV are correlated during training. Given the potential societal impact in the medical domain \citep{chartsias2018factorised} or fair decision making \citep{locatello2019fairness,madras2018learning, creager2019flexibly}, the evaluation of the usefulness of disentangled representations trained on correlated data is of high importance.

The idealized settings employed for validating disentanglement learning so far would typically build on a dataset that has as many persons with small feet and small body height as small feet and large body height. Realistic unlabeled datasets are rather unlikely to be of this type.
It is thus an open question to what degree existing inductive biases from the encoder/decoder architecture, but most importantly these dataset biases, affect the learned representation.
In this work, we introduce dataset correlations in a controlled manner to understand in detail to what degree state-of-the-art approaches can cope with such correlations.

To this end, we report a large-scale empirical study to systematically assess the effect of induced correlations between pairs of factors of variation in training data on the learned representations.
To provide a qualitative and quantitative evaluation, we investigate multiple datasets with access to ground-truth labels.
Moreover, we study the generalization abilities of the representations learned on correlated data as well as their performance in particular for the downstream task of fair decision making.\looseness=-1
    
\xhdr{Contributions.} We summarize our contributions:
\begin{itemize}[leftmargin=+.2in,topsep=0pt]
 \setlength\itemsep{0.01em}
\item We present the first large-scale empirical study (4260 models)\footnote{Each model was trained for 300,000 iterations on Tesla V100 GPUs. Reproducing these experiments requires approximately 0.79 GPU years.} that examines how modern disentanglement learners perform when ground truth factors of the observational data are \emph{correlated}.
\item We find that, in this case, factorization-based inductive biases are insufficient to learn disentangled representations.
Existing methods fail to disentangle correlated factors of variation as the latent space dimensions become statistically entangled, which has implications for downstream applications of disentanglement such as fairness.
We corroborate this theoretically by showing that, in the correlated setting, all generative models that match the marginal likelihood of the ground truth model are not disentangled.
\item We propose a simple post-hoc procedure that can align entangled latents with only very few labels and investigate the usefulness of weakly-supervised approaches to resolve latent entanglement already at train time when only correlated data is available.
\end{itemize}

\section{Background}
\label{sec:methods}

Current state-of-the-art disentanglement approaches use the framework of variational autoencoders (VAEs) \citep{kingma2013auto,rezende2014stochastic}.
The (high-dimensional) observations $\mathbf{x}$ are modeled as being generated from latent features $\mathbf{z}$ according to the probabilistic model $p_\theta(\mathbf{x}|\mathbf{z})p(\mathbf{z})$.
Typically, the prior $p(\zb)$ is fixed, while the generative model $p_\theta(\mathbf{x}|\mathbf{z})$ and the approximate posterior $q_\phi(\mathbf{z}|\mathbf{x})$ are parameterized by neural networks with parameters $\theta$ and $\phi$ respectively. These are optimized by maximizing the variational lower bound (ELBO) of the log likelihood $\log p(\mathbf{x}^{(1)},\dots,\mathbf{x}^{(N)})$ of data $D=\{ \mathbf{x}^{(i)} \}_{i=1}^N $:
\begin{align}
\label{eq:elbo_standard}
\begin{split}
\mathcal{L}_{ELBO} = \sum_{i=1}^N \Big(
  & \mathbb{E}_{q_\phi(\mathbf{z}|\mathbf{x}^{(i)})} [ \log p_\theta (\mathbf{x}^{(i)} | \mathbf{z}) ] \\
  & - D_{KL}\bigl( q_\phi(\mathbf{z}|\mathbf{x}^{(i)}) \| p (\mathbf{z}) \bigr) \Big)
\end{split}
\end{align}
This objective does not enforce any structure on the latent space, except for similarity (in KL divergence) to the prior $p (\mathbf{z})$ (typically chosen as an isotropic Gaussian).
Thus, no specific structure and semantic meaning of latent representations is encouraged by this objective.
Consequently, various works propose new evaluation metrics to quantify different notions of disentanglement of the learned representations, as well as new disentanglement learning methods, such as $\beta$-VAE, AnnealedVAE, FactorVAE, $\beta$-TCVAE, and DIP-VAE, that incorporate suitable structure-imposing regularizers \citep{higgins2016beta, burgess2018understanding, kim2018disentangling, chen2018isolating,  kumar2017variational, eastwood2018framework, mathieu2019disentangling}. 

Since unsupervised disentanglement by optimizing the marginal likelihood in a generative model is impossible \citep[Theorem 1]{locatello2018challenging}, inductive biases like grouping information \citep{bouchacourt2017multi} or access to labels \citep{locatello2019disentangling} is required. To address this theoretical limitation, methods have been proposed that only require weak label information \citep{locatello2020weakly,shu2019weakly}. Changes in natural environments, which typically  correspond to changes of only a few underlying FoV, can provide a weak supervision signal for representation learning algorithms \citep{goyal2019recurrent,foldiak1991learning, schmidt2007learning, bengio2019meta,ke2019learning, klindt2020towards}. 
In the absence of correlations it has been shown on common disentanglement datasets that this weak supervision facilitates learning disentangled representations \citep{locatello2020weakly, shu2019weakly}.

Most popular datasets in the disentanglement literature exhibit perfect independence in their FoV such as \textit{dSprites}~\citep{higgins2016beta}, \textit{Cars3D}~\citep{reed2015deep}, \textit{SmallNORB}~\citep{lecun2004learning}, \textit{Shapes3D}~\citep{kim2018disentangling} or \textit{MPI3D} variants \citep{gondal2019transfer}. 
At some level this is sensible as it reflects the underlying assumption in the inductive bias being studied.
However, this assumption is unlikely to hold in practice as is also shown by \citet{li2019learning}.

\section{The Problem with Correlated Data}
Most work on learning disentangled representations assumes that there is an underlying set of independent ground truth variables that govern the generative process of observable data.
These methods are hence predominantly trained and evaluated on datasets that obey independence in the true factors of variation by design, which we then consider to be the correct factorization. Formally, disentanglement methods typically assume $\mathbf{x} \sim \int_{\cb}p^{*}(\mathbf{x}|\mathbf{c})p^{*}(\mathbf{c}) d\cb$ where the prior over ground truth factors $\mathbf{c}$ factorizes as
\begin{equation}
    p^{*}(c_1,c_2,\ldots , c_n) = \prod_{i=1}^{n} p^*(c_i).
\end{equation}
Note that this is the dataset-generating prior and distinct from the latent prior $p(\mathbf{z})$ of the model.

In the real world, however, we generally expect correlations in the collected datasets, i.e., the joint distribution of the ground truth factors $\{c_i\}_{i=1}^{n}$ does not factorize:
\begin{equation}
    p^*(c_1,c_2,\ldots , c_n) \neq \prod_{i=1}^{n} p^*(c_i).
\end{equation}
In this case, we speak of dependence between the random variables, also commonly referred to as correlation. 
Even though there might be correlation between multiple variables simultaneously, for a principled systematic analysis we will mostly consider pairwise correlations.
Correlation between two variables can stem from various sources, e.g., from a direct causal relationship (one variable affects the other), or from unobserved circumstances (confounders) affecting both. Real-world datasets display many of these ``spurious'' and often a priori unknown correlations \citep{geirhos2020shortcut, li2019learning}.
In the introductory example, various (potentially hidden) confounders such as age or sex may indeed affect both foot length and body height. Even though we expect such a dataset to exhibit strong correlations in these factors, we would like to model them independently of each other. 

To establish some intuition for why such correlations pose problems for generative models attempting to learn a disentangled representation,
assume two FoV $c_{1}$ and $c_{2}$ that are correlated in the dataset. In a perfectly disentangled representation, a single latent dimension $z_{c1}$ would model factor $c_{1}$ and $z_{c2}$ would model factor $c_{2}$. However, because our model has an independent latent prior, the resulting generative model would generate all combinations of the two correlated factors, i.e., put probability mass also outside of the training distribution. Hence, perfect disentanglement would lead to a sub-optimal log likelihood.
We can show this more formally as follows (see \cref{app:proof} for a proof).
\begin{theorem}\label{thm}
Consider the latent variable model
$p_\theta(\xb) = \int_{\zb} p_\theta(\xb \given \zb) p(\zb) d\zb$
and let the true data distribution be
$p^*(\xb) = \int_{\cb} p^*(\xb \given \cb) p^*(\cb) d\cb$
where $\zb \in \Z$ and $\cb \in \C$.
Assume that the fixed prior $p(\zb) = \prod_i p(z_i)$ factorizes, the true FoV prior $p^*(\cb) \neq \prod_i p^*(c_i)$ does not, and that there exists a smooth bijection $f^*: \C \rightarrow \Z$ with smooth inverse (i.e., a diffeomorphism) that transforms $p^*(\cb)$ into $p(\zb)$. 
Then, the likelihood $p_\theta(\xb)$ can be equal to the optimal likelihood $p^*(\xb)$ if and only if the representations $\zb$ are entangled w.r.t. the true factors of variation $\cb$.
\end{theorem}
Note that we define a representation to be \emph{disentangled} if the Jacobian of $f^*$ is diagonal up to a permutation, in line with \citet{locatello2018challenging} \& \citet{locatello2020weakly}.
Under the assumptions of this proposition, we established that a generative model that matches the true marginal likelihood $p^*(\xb)$ cannot be disentangled.
This suggests that, if some FoV are correlated in the dataset and the prior $p(\zb)$ factorizes, methods that optimize a lower bound to the log likelihood might have a bias \textit{against} disentanglement.\footnote{While the impossibility result of \citet{locatello2018challenging} states that there may be many generative models achieving the optimal likelihood for a ground-truth model with independent FoVs, Proposition 1 states that if the ground-truth prior is correlated, a disentangled representation will never achieve this optimal likelihood and therefore entangled representations are preferred.}
The primary goal of this work is thus to empirically assess to what extent the additional inductive biases of state-of-the-art disentanglement methods still suffice to learn disentangled representations when the training data exhibits correlations.

\xhdr{Why not encode train-time correlations?}
The prior discussion raises the question why we would like to learn representations that do not encode data correlations.
Why not encode foot size and body height in a single latent dimension when we are only shown data in which they are strongly correlated?
Maximizing the log likelihood, which amounts to matching the data distribution at train time, would embrace such correlations, and it would later allow us to sample novel examples from this distribution.
We now provide three major reasons why this may not always be the goal in real-world applications.

First, if we want to build models that generalize well across multiple tasks and dataset distributions the standard approach is myopic. A key promise of disentanglement is to learn meaningful and compact representations, in which relevant semantic aspects of the data are structurally disentangled and can be analyzed or controlled independently from each other during inference or generation. From this perspective, resolving instead of exploiting correlations in the train data is desirable at test time, where such representations would remain robust and meaningful under distribution shifts.
Returning to our running example, assume training data is constrained to a fixed sex and age bracket, both being confounders for foot length and body height \citep{grivas2008correlation}. If we could disentangle these factors despite the correlation, the model would arguably be more robust to distribution shifts, for example when testing on a different sex or different ages.

Second, we would like to be able to sample out-of-distribution (OOD) examples or intervene on individual factors independently of the others. Should a good model not be able to meaningfully reason about and imagine someone's foot length changing independently of their body height?

Lastly, disentangled representation are relevant in fairness settings, where representations are used in downstream tasks for consequential decisions that ought to be fair with respect to sensitive (protected) variables \citep{locatello2019fairness,creager2019flexibly}. Undesirable data correlations with such protected variables are a major problem in fairness applications. Therefore, it is crucial to evaluate the validity of factorization-based inductive biases to learn disentangled representations from correlated observational data.

\xhdr{How can we resolve these latent correlations?}
To resolve potential latent correlations, we will investigate two approaches. First, we explore the scenario in which few labels are given after training with the specific purpose of resolving these correlations. For this, we propose a simple approach in Section~\ref{sec:fast-adaptation}, which we find to be effective. Second, we employ the performance of the weakly-supervised approach of \citet{locatello2020weakly}. They showed (on uncorrelated data) that access to pairs of observations which display differences in a known number of FoV (without knowing which ones specifically) suffices to learn disentangled representations.
These additional weak assumptions render the generative model identifiable in contrast to unsupervised disentanglement and may arguably be indeed available in certain practical settings, e.g., in temporally close frames from a video of a moving robot arm where some factors remain unchanged.\footnote{
On the other hand, in applications with fairness concerns it may be impossible to intervene on FoV representing sensitive attributes of individuals (sex, race, etc.); we refer to \cite{Kilbertusetal17,madras2019fairness} for a more complete discussion.}
We test the Ada-GVAE algorithm from \citet{locatello2020weakly}, which requires a pair of observations that differ in an unknown number of factors.
To impose this structure in the latent space, the latent factors that are shared by the two observations are estimated from the $k$ largest coordinate-wise KL divergences $D_{KL}( q_\phi(z_i|\mathbf{x}^{(1)}) \| q_\phi(z_i|\mathbf{x}^{(2)}))$. They then maximize the following modified $\beta$-VAE \citep{higgins2016beta} objective for the pair of observations
\begin{align}
  \sum_{i\in\{1,2\}}  &\mathbb{E}_{(\mathbf{x}^{(1)}, \mathbf{x}^{(2)})} \Big[ 
  \mathbb{E}_{\tilde q_\phi^{(i)}(\hat\rvz | \mathbf{x}^{(1)},\mathbf{x}^{(2)})} \log(p_\theta(\mathbf{x}^{(i)}|\hat\rvz)) \nonumber \\
    &\quad - \beta D_{KL} \left(\tilde q_\phi^{(i)}(\hat\rvz | \mathbf{x}^{(1)}, \mathbf{x}^{(2)})\|p(\hat\rvz)\right)
    \Big],\label{eq:mod-elbo}
\end{align}
with $\tilde q_\phi^{(i)}(\hat z_j | \mathbf{x}^{(1)},\mathbf{x}^{(2)}) = q_\phi(\hat z_j | \mathbf{x}^{(i)})$ for the latent dimensions $z_j$ that are inferred to be changing and $\tilde q_\phi^{(i)}(\hat z_j | \mathbf{x}^{(1)}, \mathbf{x}^{(2)}) = a(q_\phi(\hat z_j | \mathbf{x}^{(1)}), q_\phi(\hat z_j | \mathbf{x}^{(2)}))$ for those inferred to be shared. The averaging function $a$ forces the approximate posterior of the shared latent variable to be the same in the two observations. 
\section{Unsupervised Disentanglement under Correlated Factors}
\label{sec:section4}
In this section, we present the key findings from our empirical study of unsupervised disentanglement learning approaches on a particular variant of correlated datasets. We start by outlining the experimental design of our studies in \cref{sec:experiment_design}. Based on this, in \cref{sec:unsupervised_result} we present a latent space analysis of unsupervised disentanglement learners and find that factorization-based inductive biases are insufficient to learn disentangled representations from correlated observational data. Although these methods fail to disentangle correlated factors, in \cref{sec:generalization} we qualitatively show that they still generalize to FoV combinations that are out-of-distribution with respect to the training data.

\begin{figure*}
\centering
\minipage{0.9\textwidth}
\minipage{0.64\textwidth}
  \includegraphics[width=\textwidth]{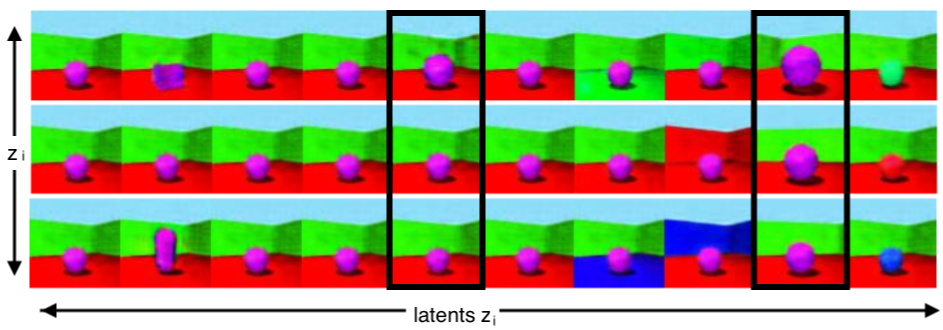}
\endminipage\hfill
\minipage{0.3\textwidth}
\includegraphics[width=\textwidth]{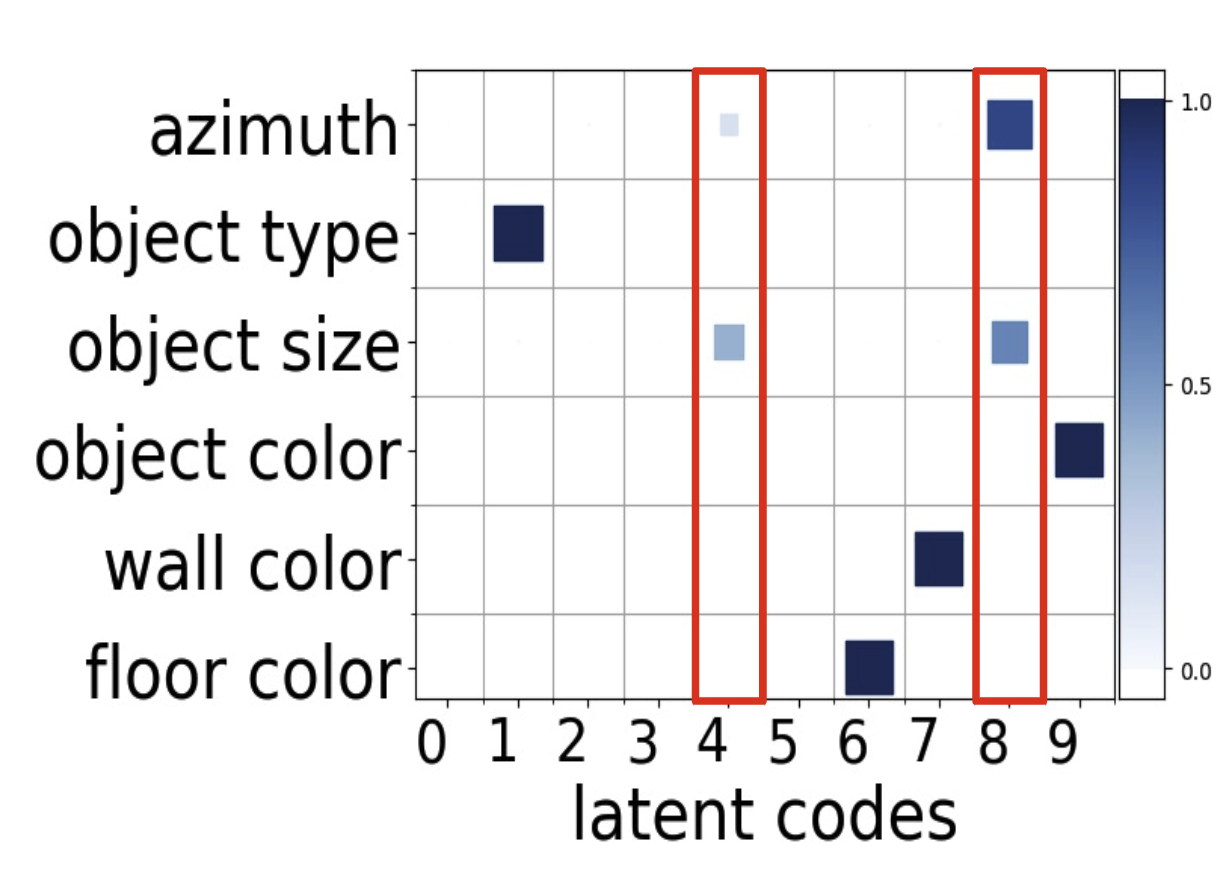}
\endminipage
\endminipage
\caption{\textbf{Left:} Latent traversals for the model with best DCI score (commonly used metric for measuring disentanglement \citep{eastwood2018framework}) among all 180 models trained on strongly correlated ($\sigma = 0.2$) Shapes3D data (A).
The traversals in latent dimensions 5 and 9 (highlighted in black) encode a mixture of azimuth and object size, reflecting the main correlation line of the joint distribution and a smaller, locally orthogonal one.
\textbf{Right:} A heat map of the GBT feature importance matrix of this model indicates an entanglement of azimuth and object size encoded into both latent codes.}
\label{fig:representation_structure_under_correlation_traversals}
\end{figure*}

\subsection{Experimental Design}
\label{sec:experiment_design}
\begin{table}
    \centering
    \resizebox{\columnwidth}{!}{
    \begin{tabular}{c|l|c|c}
        \toprule
        corr. dataset & data source &  1st correlated FoV &  2nd correlated FoV\\
        \midrule
        A & Shapes3D & object size & azimuth\\
        B & dSprites & orientation & x-position\\
        C & MPI3D &  1st DoF & 2nd DoF\\
        D & Shapes3D & object color & object size\\
        E & Shapes3D & object color & azimuth\\
        \bottomrule
    \end{tabular}}
    \caption{Dataset variants with introduced correlations under investigation within this empirical study.}
    \label{tab:dataset_variants}
\end{table}
To begin to systematically understand the unknown behavior of SOTA disentanglement approaches on correlated data, we focus on linear correlations with Gaussian noise between pairs of FoV, denoted by $c_1$ and $c_2$.\footnote{We emphasize that this seemingly narrow class of correlations already captures most relevant effects of more general correlations between FoV at train time on the learned representation. This is not to say that one can draw general conclusions about highly nonlinear correlated settings. However, rigorously understanding the linear case in a wide range of controlled settings already represents a considerable set of experiments and is a vital step in bridging the gap between highly idealized settings and real-world applications.} For our experiments we introduce five correlated dataset variants comprising dSprites, Shapes3D and MPI3D with correlations between single pairs of FoV. \Cref{tab:dataset_variants} shows the correlated FoV in each dataset variant and the names we refer to them by. 

We then parameterize correlations by sampling the training dataset from the joint distribution 
\begin{equation*}
  p(c_{1},c_{2}) \propto \exp\biggl(-\frac{(c_1 - \alpha c_2)^2}{2\sigma^2}\biggr)
\end{equation*}
with $c_1 \in \{0, \dots, c_1^{\max}\}$ and $c_2 \in \{0, \dots, c_2^{\max}\}$ and $\alpha = c_{2}^{\max}  / c_{1}^{\max}$. The correlation strength can be tuned by $\sigma$, for which we choose 0.2, 0.4, 0.7 in normalized units with respect to the range of values in $\{c_1,c_2\}$. Lower $\sigma$ indicates stronger correlation. See 
\cref{fig:generalization_reconstruction} for an example of $p(c_{1},c_{2})$ for correlating azimuth and object size in Shapes3D.
Additionally, we study the uncorrelated limit ($\sigma=\infty$), which amounts to the case typically considered in the literature. All remaining factors are sampled uniformly at random. 

We train the same six VAE methods as \citet{locatello2018challenging}, i.e., $\beta$-VAE, FactorVAE, AnnealedVAE, DIP-VAE-I, DIP-VAE-II and $\beta$-TC-VAE, each with 6 hyperparameter settings and 5 random seeds. These 180 models are trained on datasets A, B, and C with $\sigma \in \{0.2, 0.4, 0.7, \infty\}$ and datasets D and E with $\sigma \in \{0.2, 0.4\}$, totaling 2880 models.
\cref{sec:implementation_details} describes additional implementation details.\footnote{Code for reproducing experiments is available under \url{https://github.com/ftraeuble/disentanglement_lib}}

\subsection{Unsupervised Methods for Correlated Data}
\label{sec:unsupervised_result}
Here we assess the applicability of SOTA unsupervised learning approaches to correlated data. Are the factorization-based inductive biases introduced by these methods enough to disentangle correlated FoV although it is sub-optimal from a standard VAE perspective?

\xhdr{Latent structure and pairwise entanglement.}
We start by visually inspecting latent traversals of some trained models on Shapes3D (A). 
For strong correlations ($\sigma = 0.2$ and $\sigma = 0.4$), we typically observe trained models with two latent codes encoding the two correlated variables simultaneously.
In these cases, one of the latent codes corresponds to data along the major axis of the correlation line whereas the other latent code dimension manifests in an orthogonal change of the two variables along the minor axis.
Perhaps unsurprisingly, a full traversal of the code corresponding to the minor axis often seems to cover only observations within the variance of the correlation line, i.e., in the training distribution.
\cref{fig:representation_structure_under_correlation_traversals} (left) shows this effect for the latent space of a model trained on Shapes3D (A) with strongest correlation ($\sigma = 0.2$). 
To quantify this observation, we analyze the importance of individual latent codes in predicting the value of a given ground truth FoV.
An importance weight for each pair of \{FoV, latent dimension\} is computed by training a gradient boosting tree (GBT) classifier to predict the ground truth labels from the latents (10,000 examples).
In the right panel of \cref{fig:representation_structure_under_correlation_traversals}, we show these importance weights for the model used to generate traversals in the left panel.
The corresponding evaluation for a model trained on the same dataset with a much weaker correlation of $\sigma = 0.7$ does not reveal this feature visually (see \cref{fig:representation_structure_under_correlation_traversals_appendix} in \cref{sec:unsupervised_result_appendix}
).

\begin{table}
\minipage{\columnwidth}
\resizebox{\textwidth}{!}{
  \begin{tabular}{l|lcccc}
\toprule
Corr. strength &  & $\sigma=0.2$    & $\sigma=0.4$    & $\sigma=0.7$    & $\sigma=\infty$ (uc)   \\
 \midrule
\multirow{2}{*}{Shapes3D (A)} & \textcolor{BrickRed}{object size - azimuth} & \textcolor{BrickRed}{0.38}   & \textcolor{BrickRed}{0.26}    & \textcolor{BrickRed}{0.13}   & \textcolor{blue}{0.08}           \\
 & median uncorrelated pairs                     & 0.09         & 0.09         & 0.09         & 0.08 \\
\midrule
\multirow{2}{*}{dSprites (B)} & \textcolor{BrickRed}{orientation - position x} & \textcolor{BrickRed}{0.17}   & \textcolor{BrickRed}{0.16}   & \textcolor{BrickRed}{0.14}   & \textcolor{blue}{0.11}           \\
 & median uncorrelated pairs                        & 0.13         & 0.13         & 0.13         & 0.13                 \\
 \midrule
\multirow{2}{*}{MPI3D (C)} & \textcolor{BrickRed}{First DOF - Second DOF} & \textcolor{BrickRed}{0.2}    & \textcolor{BrickRed}{0.19}   & \textcolor{BrickRed}{0.17}   & \textcolor{blue}{0.16}           \\
& median uncorrelated pairs                      & 0.16         & 0.16         & 0.15         & 0.15 \\
 \bottomrule
\end{tabular}}
\endminipage
\caption{The means of the pairwise entanglement scores for the correlated pair (red) and the median of the uncorrelated pairs. Stronger correlation leads to statistically more entangled latents compared to the baseline score without correlation (blue), thus uncovering still existent correlations in the latent representation.}
\label{fig:pairwise_scores_thresholds_and_unfairness}
\end{table}

To support this claim empirically across the full study with all datasets, we calculate a pairwise entanglement score that allows us to measure how difficult it is to separate two factors of variation from their latent codes. This computation involves grouping FoV into pairs based on an ordering of their pairwise mutual information or GBT feature importance between latents and FoV; we refer to \cref{sec:implementation_details} for a detailed description of this procedure.
\Cref{fig:pairwise_scores_thresholds_and_unfairness} shows that across all datasets the pair of correlated FoV has a substantially higher score than the median of all other pairs, indicating that they are harder to disentangle.
This score decreases with weaker correlation, i.e., the pair becomes easier to disentangle for $\sigma \geq 0.7$.
We conclude that models still disentangle weakly correlated factors, but inductive biases become insufficient to avoid latent entanglements for stronger correlations.
In \cref{sec:unsupervised_result_appendix} we further consolidate our claims by showing that the correlated pair is more entangled in the latent representation across all unsupervised experiments and datasets. We also further corroborate our finding that standard disentanglement metrics when evaluated on all FoV are insufficient to reveal this entanglement.

\begin{figure}
\begin{center}
  \includegraphics[width=0.32\columnwidth]{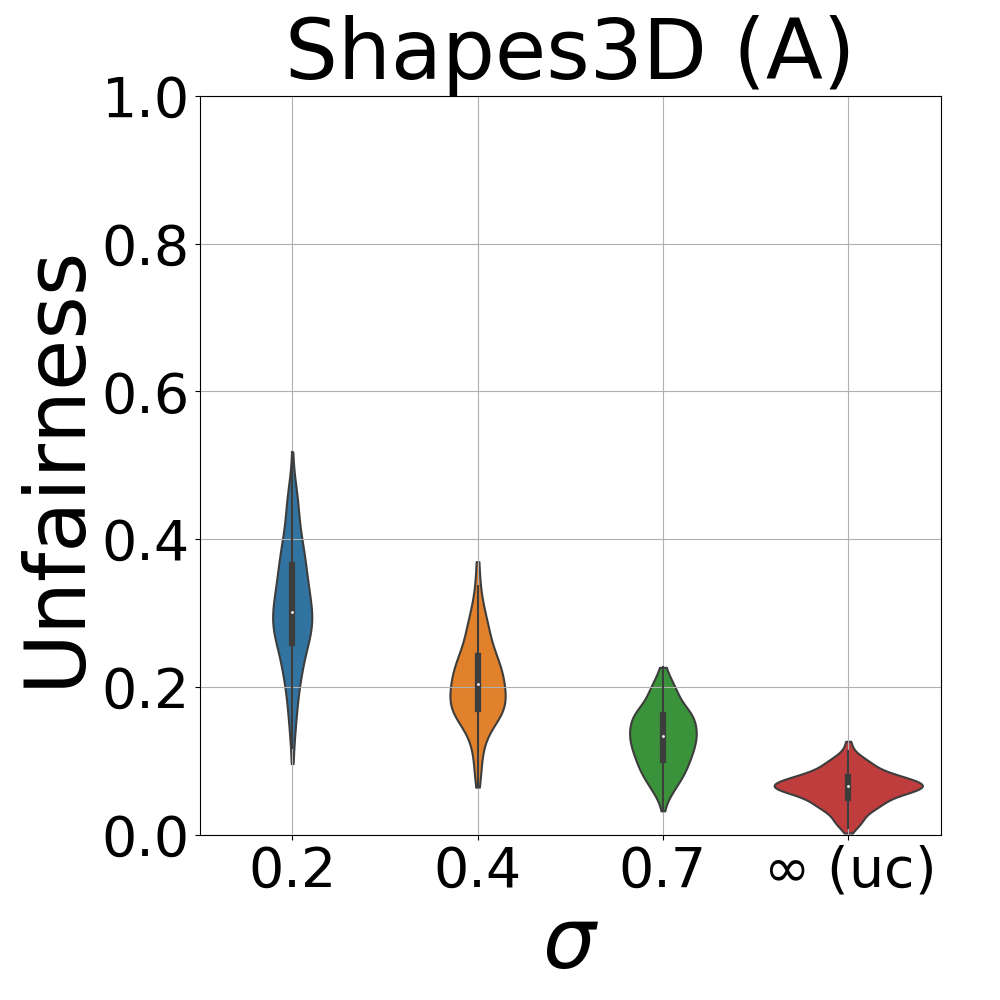}
  \includegraphics[width=0.33\columnwidth]{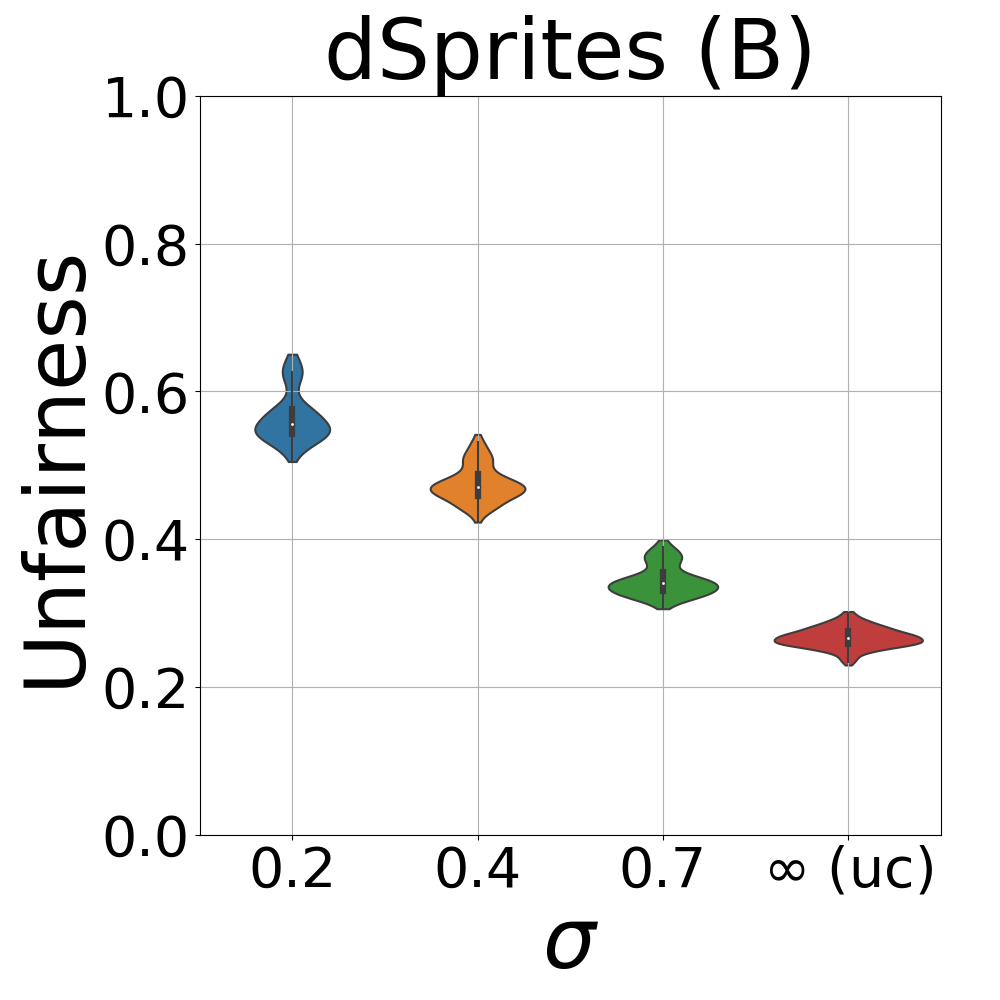}
  \includegraphics[width=0.33\columnwidth]{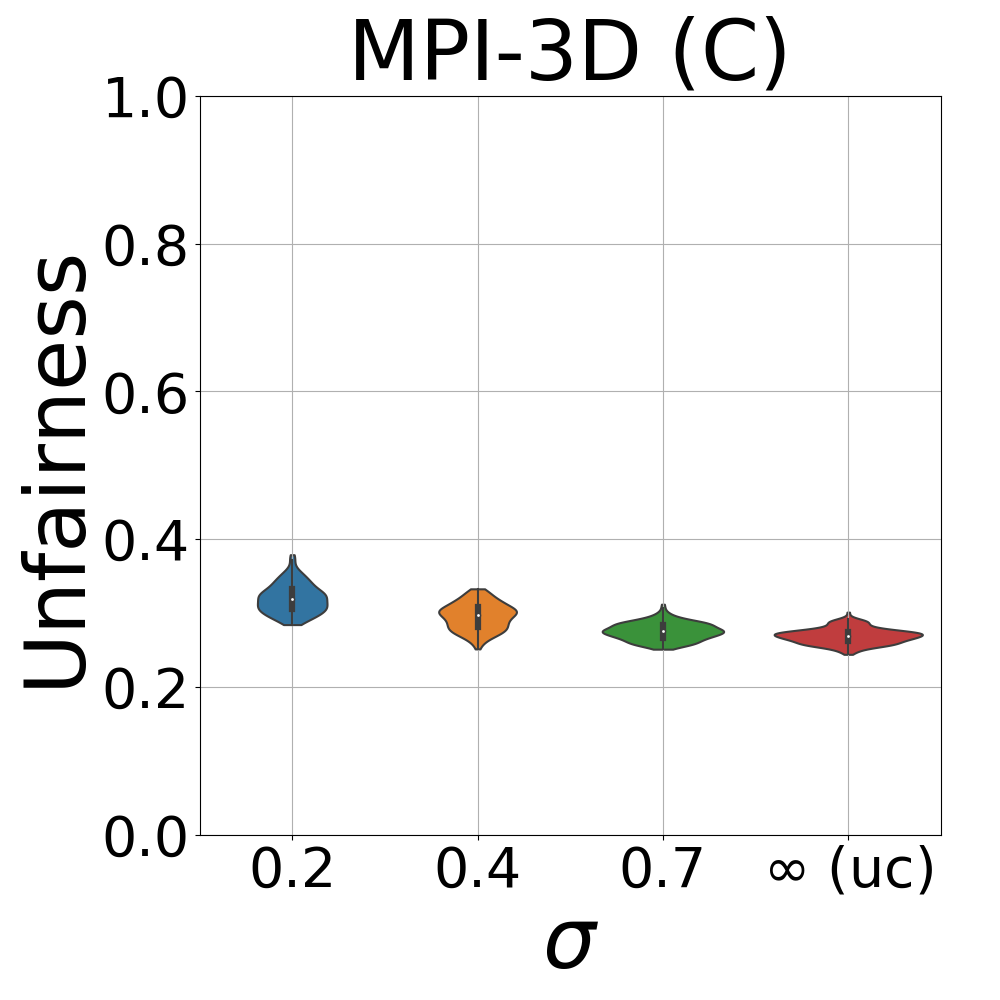}
\end{center}
\caption{Stronger correlations in the train data lead to substantially elevated unfairness scores for the correlated pair of factors.}
\label{fig:unfairness}
\end{figure}

\xhdr{Does this matter for downstream tasks?} Correlations between variables are of crucial importance in fairness applications, motivating an additional investigation on ramifications of these entangled latent spaces. In this setting we are interested in the unfairness of predicting one of the two correlated variables when the other represents a protected or sensitive attribute.
In the following, we use a variant of demographic parity \citep{dwork2012fairness} that computes pairwise mutual information between latents and FoV \citep{locatello2019fairness}.
In \cref{fig:unfairness} we evaluate this score when correlations are present across all unsupervised experiments and datasets. 
Unfairness tracks correlation strength in this scenario. 
\citet{locatello2019fairness} show that representations learned without supervision may exhibit unfairness even without correlations. Our results suggest that the problem is substantially aggravated in the presence of correlations between a sensitive attribute and another ground truth FoV.

\xhdr{Summary.} Factorization-based inductive biases are insufficient to learn disentangled representations from correlated observational data in the unsupervised case. We observed persisting pairwise entanglement in the latent space, both visually and quantitatively using appropriate metrics, which might be particularly problematic for fairness applications.

\subsection{Generalization Under Changing Correlations}
\label{sec:generalization}

\begin{figure}
\minipage{\columnwidth}
\centering
\includegraphics[width=0.95\textwidth]{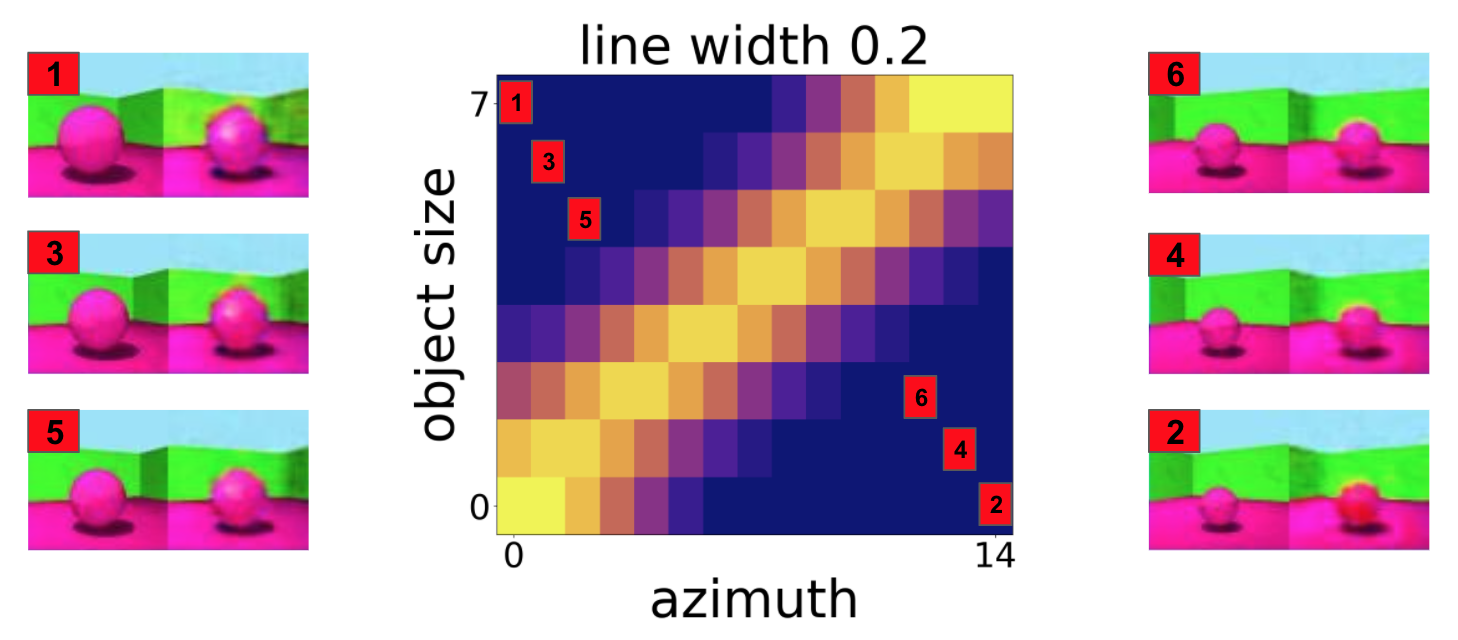}
\endminipage
\vskip 0.1in
\minipage{0.49\columnwidth}
  \includegraphics[width=\columnwidth]{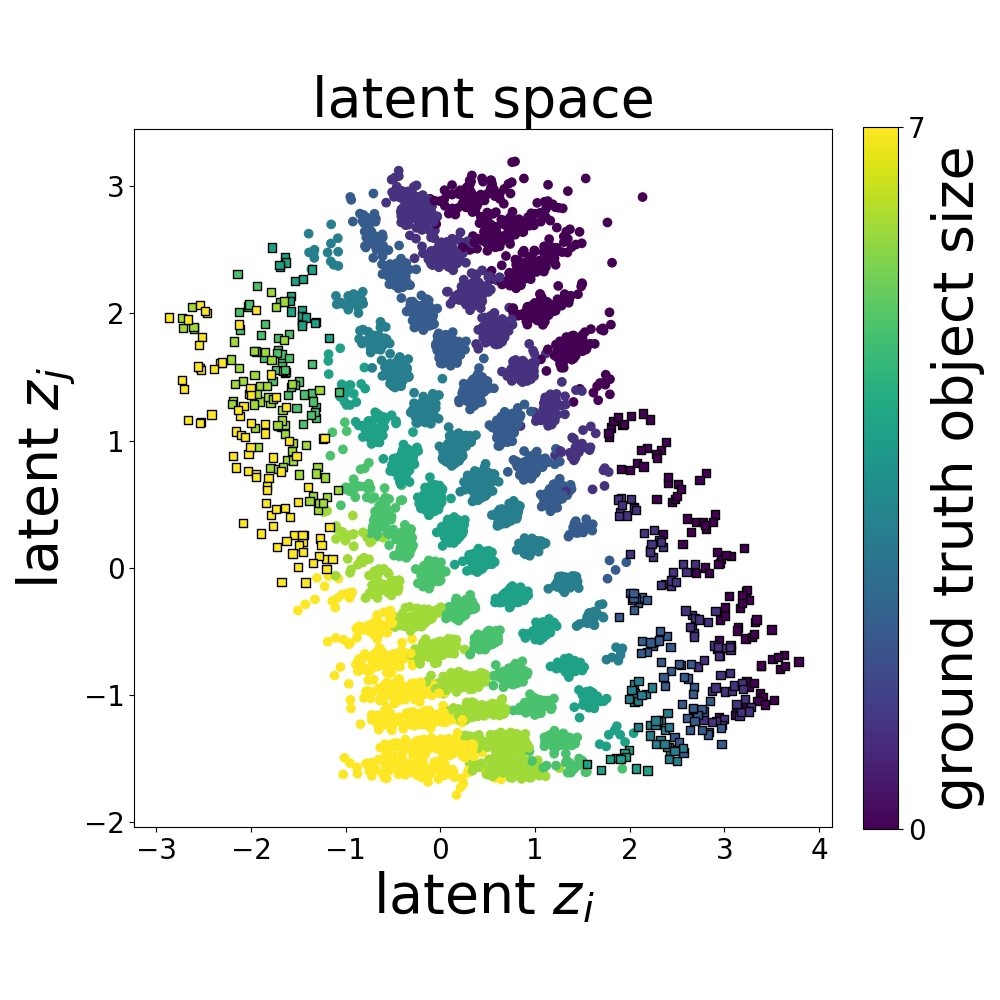}
\endminipage
\minipage{0.49\columnwidth}
  \includegraphics[width=\columnwidth]{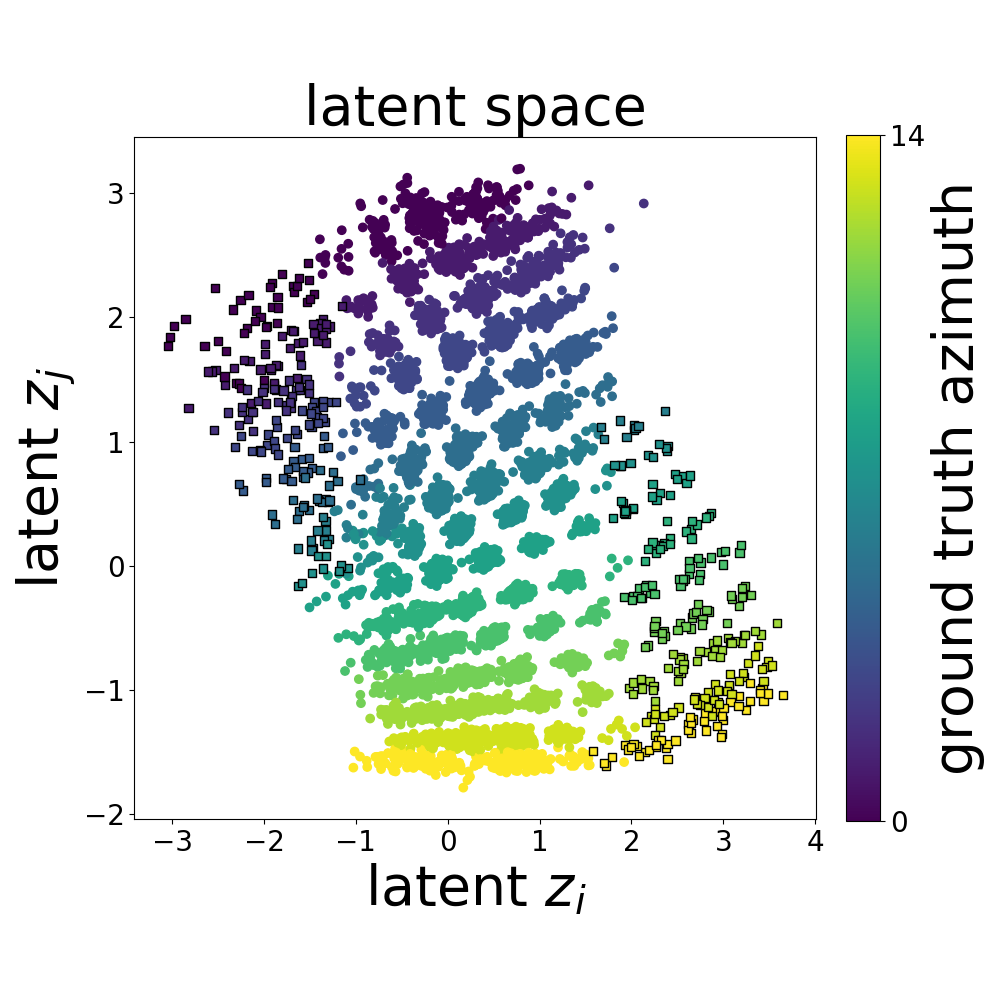}
\endminipage
\caption{Generalization to out-of-distribution (OOD) test data. \textbf{Top:} Reconstructions of observations the model has never seen during training. \textbf{Bottom:} Latent space distribution of the two entangled dimensions. 
Circles without edges represents latent encodings of (correlated) training examples. Circles with edges are OOD examples that break the correlation pattern.}
\label{fig:generalization_reconstruction}
\end{figure}

We now aim to understand the robustness and generalization capabilities of these models on data that is sampled not iid from the correlated train distribution but instead out-of-distribution (OOD) from a low-probability region given the correlation structure, e.g., large feet but short body height.
In our Shapes3D experiments, this amounts to large object size and small azimuth which has \textit{zero} probability under the correlated train distribution.
We focus on the model from \cref{fig:representation_structure_under_correlation_traversals}, which has disentangled all non-correlated factors well, allowing us to only focus on the two latent dimensions encoding the entangled variables.

First, we analyze examples with object size and azimuth values that have numerically zero probability under the correlated data distribution as shown as numbers 1 to 6 in \cref{fig:generalization_reconstruction} (top panel) in the FoV space.
The model can reconstruct these examples despite never having seen such configurations during training.
This suggests that OOD examples are meaningfully encoded into the existing structure of the latent space \emph{and} that the decoder is equally capable of generating observations from such unseen representations. Note that in contrast to \citet{zhao2018bias} who test this combinatorial generalization in the interpolation regime from uncorrelated, unbiased, but sparsely populated train data, to the best of our knowledge this extrapolation OOD generalization has not been studied on strongly correlated data so far.
To test this hypothesis further, we analyzed latent traversals originating from these OOD points 
and observe that changes in the remaining factors reliably yield the expected reconstructions.
Traversals with respect to the two entangled latent codes continue to encode object size and azimuth.

To fully understand these models' generalization properties we visualize the latent space encoding of the training distribution projected onto the two identified entangled dimensions (i.e., marginalizing over all others) in \cref{fig:generalization_reconstruction} (bottom panel).
We depict the ground truth value of each correlated variable via a color gradient.
The two sets of depicted points are (1) latent codes for samples from the correlated training data (no edges) and (2) latent codes for samples with (object size, azimuth) configurations that have zero probability under the correlated training distribution (edges).
As expected, contours of equal color (ground truth) are not aligned with the latent axes, indicating that the two latent dimensions encode both FoV at the same time.
The generalization capabilities of this model away from the training data at least for the decoder can be described as follows: If we imagine the contour lines of equal color in both plots for the training data, i.e., only circles without edges, and linearly extend those lines, they form a grid in the latent space beyond where the training data lies.
Latent encodings of unseen combinations of ground truth FoV happen to meaningfully extend and complete this grid structure in the latent space.
Note that in this example, there is a natural ordering on both correlated FoV (numerical values for azimuth and size).
Even though for categorical factors such as object color in datasets D and E we cannot expect the ad-hoc ordering of colors to be preserved in the latent space, the latent encodings still extrapolate meaningfully for examples with unseen combinations of ground truth FoV.
In \cref{sec:generalization_properties_appendix}, we show some of these characteristic latent space visualizations with similar extrapolation and generalization capabilities.

\xhdr{Summary.} We conclude from these results that even though the models fail to disentangle the correlated FoV, they are still incorporating enough structure in the latent space to generalize towards unseen FoV configurations that are OOD w.r.t.\ the correlated train distribution.
\section{Recovering Factorized Latent Codes}
\label{sec:weakly_supervised_study_results}
We now investigate the usefulness of semi- and weakly-supervised approaches to resolve latent entanglement.

\xhdr{Post-hoc alignment correction with few labels.}\label{sec:fast-adaptation}
When a limited number of FoV labels $\{ \mathbf{c}_i\}_{i=1}^M$ can be accessed, a reasonable option for resolving entangled dimensions of the latent code is by \emph{fast adaptation}.
To identify the two entangled dimensions $\mathbf{z}_{fa}:=(z_{i} , z_{j})$ we look at the maximum feature importance for a given FoV from a GBT trained using these labels only.
We then train a \emph{substitution function} $f_{\theta}: \mathbb{R}^2 \to \{0,\ldots,c_1^{\max}\} \times \{0,\ldots,c_2^{\max}\}$ via supervised learning to infer the two ground truth FoV $c_1, c_2$ from the entangled latent codes $z_i, z_j$, $f_{\theta}(\mathbf{z}_{fa}) = (c_1, c_2)$.
We then use this prediction to replace these two dimension of the latent codes.
Crucially, both steps of this procedure rely on the same $M$ FoV labels, of which we assume only very few being available in practice.

\cref{fig:pairwise_disentanglement_linear_regression_spotlight} shows the pairwise entanglement score of the correlated FoV under this fast adaptation with a linear regression as the substitution function, which succeeds with as few as 100 labels (we only need the correlated factor labels), corresponding to less than 0.02\% of all ground truth labels in Shapes3D.
However, fast adaptation with linear regression substitution fails in some settings: when no two latent dimensions encode the applied correlation isolated from the other latent codes, or when the correlated variables do not have a unique natural ordering (e.g., categorical variables).
Accordingly, a nonlinear substitution function such as an MLP can further reduce pairwise entanglement in these cases (see additional results in \cref{subsec:fast_adaptation_appendix}).

\begin{table}
\begin{center}
\resizebox{1.0\columnwidth}{!}{
\begin{tabular}{l|lcccc}
\toprule
 \# Labels &                                  &    0 &   100 &   1000 &   10000 \\
\midrule
\multirow{2}{*}{Shapes3D (A) $\sigma = 0.2$}&\textcolor{BrickRed}{object size - azimuth} & \textcolor{BrickRed}{0.38} &  \textcolor{blue}{0.17} &   \textcolor{blue}{0.15} &    \textcolor{blue}{0.15} \\
    & median uncorrelated pairs              & 0.09 &  0.08 &   0.07 &    0.07 \\

\midrule
\multirow{2}{*}{Shapes3D (A) $\sigma = 0.4$} & \textcolor{BrickRed}{object size - azimuth} & \textcolor{BrickRed}{0.26} &  \textcolor{blue}{0.1}  &   \textcolor{blue}{0.1}  &    \textcolor{blue}{0.1}  \\
 & median uncorrelated pairs              & 0.09 &  0.08 &   0.08 &    0.08 \\
\bottomrule
\end{tabular}}
\caption{\textbf{Fast adaption with few labels:} Pairwise entanglement scores for correlated FoV pair in Shapes3D (A). The correlated pair is highlighted (red). Zero labels reflects the unsupervised baseline without any fast adaptation. Growing number of labels show that fast adaption using linear regression reduces these correlations with as little as 100 labels (blue). Reported pairwise scores are averaged over 180 models per correlation strength.}
\label{fig:pairwise_disentanglement_linear_regression_spotlight}
\end{center}
\end{table}

\begin{figure}
\minipage{0.31\columnwidth}
  \includegraphics[width=\columnwidth]{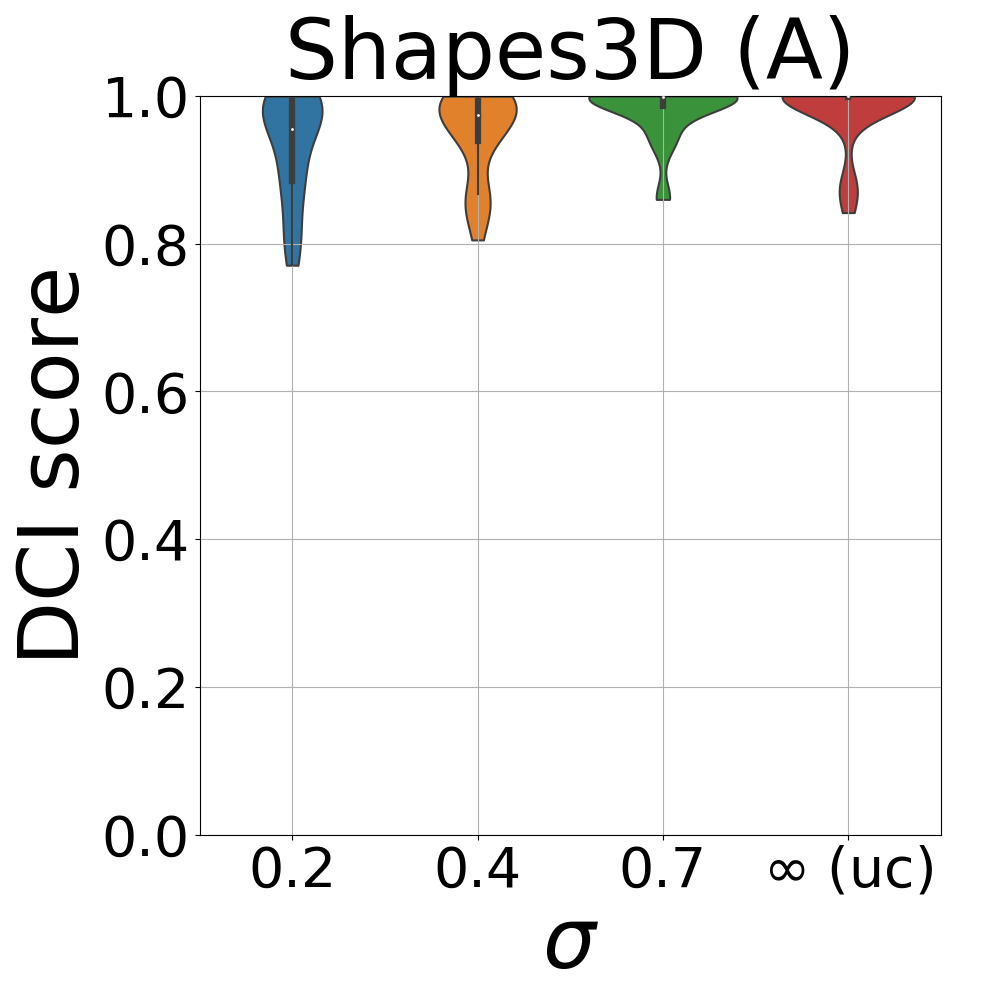}
\endminipage\hfill
\minipage{0.32\columnwidth}
  \includegraphics[width=\columnwidth]{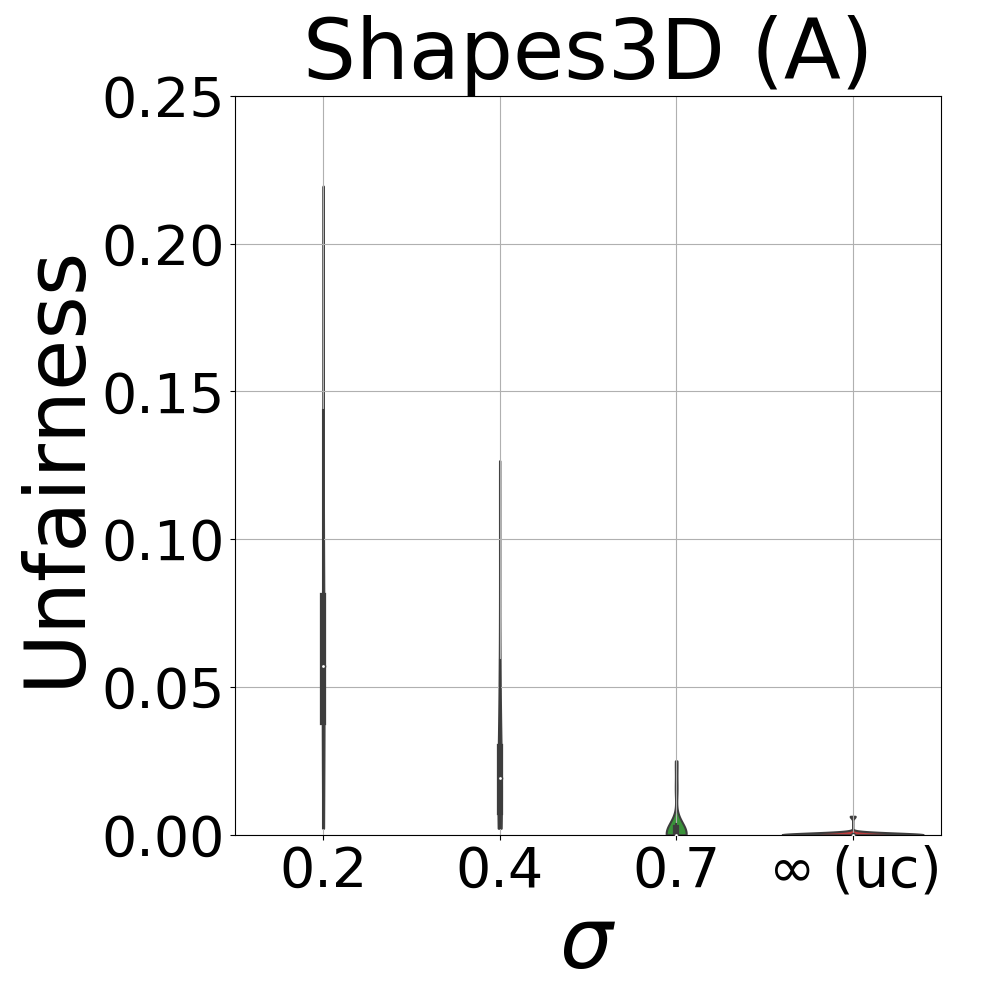}
\endminipage\hfill
\minipage{0.36\columnwidth}
  \includegraphics[width=\columnwidth]{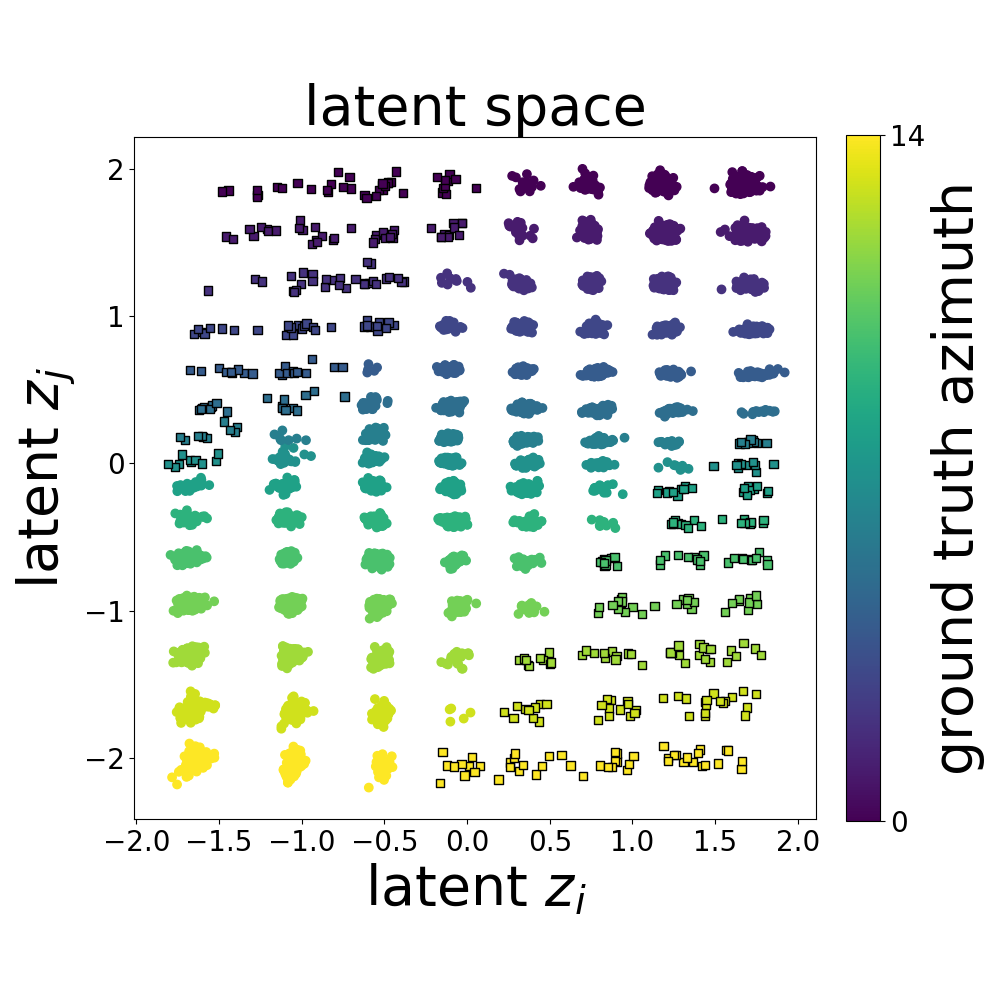}
\endminipage
\vskip 0.1in
\centering
\minipage{\columnwidth}
 \includegraphics[width=\columnwidth]{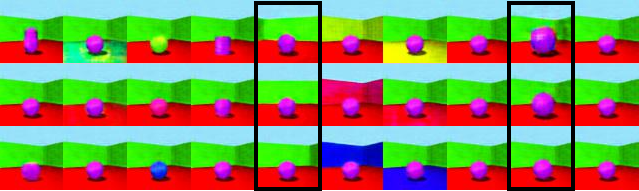}
\endminipage
\vskip 0.15in
\caption{
\textbf{Top left:} We show the disentanglement (DCI score) for models trained on Shapes3D \emph{with weak supervision} when object size and azimuth are correlated with different strength. Weak supervision helps to recover improved, often perfect, disentanglement. \textbf{Top middle:} Unfairness scores between correlated FoV are drastically reduced when using weak supervision (see scale) across all correlation strengths. \textbf{Top right:} Latent dimensions of the best DCI model with strong correlation ($\sigma=0.2$) when using weak supervision. Representations are axis-aligned with respect to both of the correlated ground truth factors. \textbf{Lower:} Latent traversals of this model trained on strong correlation ($\sigma=0.2$).
}
\label{fig:weakly_supervised_study_dci_unfairness_latent_35}
\end{figure}

\xhdr{Weak supervision mitigates learning latent entanglement.}
We now return to the weakly-supervised method from \cref{sec:methods} and evaluate its applicability when training data is correlated.
Specifically, we study the variant where the difference in the observation pair is present in one random ground truth factor. Each time we construct such a pair, all underlying ground truth factors share the same values except for one factor. Whenever this happens to be one of the correlated factors, the values of this factor within each pair are drawn from the marginal probability distribution conditioned on the other correlated factor. In these cases the difference in this factor is typically very small and depends on the correlation strength. Note that this procedure assures that constructed pairs are consistent with the observational data such that the correlation is never broken.
In this part of our study, we consider datasets A, D, and E, and train $\beta$-VAEs with the same 6 hyperparameters and 5 random seeds as in the unsupervised study, yielding 360 models. 
\cref{fig:weakly_supervised_study_dci_unfairness_latent_35} summarizes the weak supervision results on Shapes3D (A) when imposing correlations in object size and azimuth. We consistently observe much better disentangled models, often achieving perfect DCI score irrespective of correlations in the dataset. The latent spaces tend to strongly align their coordinates with the ground truth label axis. Finally, weak-supervision reduces unfairness relative to the unsupervised baseline from \cref{fig:unfairness}, and occasionally even achieves zero unfairness score. Additional results on the other datasets can be found in \cref{subsec:appendix_weak_supervision} including two additional training regimes in which the weak-supervision pairs are sampled from an interventional (but still correlated) distribution, representing an additional 1020 models. We consistently observe the same strong trends regarding disentangled correlations  in all of the above studies using weak supervision.

These results suggest that weak supervision can provide a strong inductive bias capable of finding the right factorization and resolving spurious correlations for datasets of unknown degree of correlation. As a prominent example, this is an issue in the fairness context where real-world datasets often display unknown discriminatory correlations.

\xhdr{Summary.} When ground truth information about the correlated FoV is available, our fast adaptation method can resolve latent correlations using only few labels of correlated FoV. When no labels are available, recently proposed approaches in weakly supervised disentanglement learning applied on correlated data offer an alternative to overcoming pairwise correlations at train time. We have thus shown how to leverage even weak supervision signals to learn disentangled representations from correlated data.

\section{Other Related Work}
\looseness=-1 \xhdr{ICA and Disentanglement.} Ideas related to disentangling the factors of variation date back to the non-linear ICA literature \citep{comon1994independent, jutten2003advances, hyvarinen1999nonlinear, hyvarinen2018nonlinear, hyvarinen2016unsupervised}. Recent work combines non-linear ICA with disentanglement \citep{khemakhem2020variational, sorrenson2020disentanglement}. 

\xhdr{Entangled and rotated latent spaces.} \citet{zhou2020evaluating} measure disentanglement based on topological similarity of latent submanifolds, arguing that this may help uncover latent entanglements. While having orthogonal goals to ours, it complements our empirical findings regarding the problematic structure of entangled latent spaces.
\citet{rolinek2018variational} contributed a related theoretical result showing that VAEs promote latent spaces pursued by (locally) orthogonal PCA embeddings due to the role of the diagonal covariance of the latent prior. They confirm this experimentally using the uncorrelated dSprites dataset. \citet{stuhmer2019isavae} use a family of rotationally asymmetric distributions as the latent prior, which can help learning disentangled subspaces. In contrast to the modeling perspective, we studied the effect of dependencies from the data perspective with strong correlations in the data generating prior $p^*(\mathbf{c})$. 

\xhdr{Studies on correlated data.} \looseness=-1The literature so far is missing a systematic large-scale empirical study how popular inductive biases such as factorized priors behave when actually learning from correlated datasets, although several smaller experiments along these lines can be acknowledged.
\citet{chen2018isolating} studied correlated 3DFaces \citep{paysan20093d} by fixing all except three factors in which the authors conclude that the $\beta$-TC-VAE regularizer can help to disentangle imposed correlations based on the MIG metric. However, the latent structure was not studied in detail; our findings suggest that global disentanglement metrics are insufficient to fully diagnose models learned from correlated data. 
Based on the observation that the assumption of independent factors is unlikely to hold in practice and certain factors might be correlated in data \citet{li2019learning} propose methods based on a pairwise independence assumption instead. \citet{brekelmans2019exact} show that Echo noise results in superior disentanglement compared to standard $\beta$-VAE in a small experiment on a downsampled dSprites variant.
\citet{creager2019flexibly} based some of the evaluations of a proposed new autoencoder architecture in the fairness context on a biased dSprites variant and \citet{yang2020causalvae} study a linear SCM in a VAE architecture on datasets with dependent variables.
However, their studies focused on representation learners that require strong supervision via FoV labels at train time.

\section{Conclusion}
\label{sec:conclusion}
We have taken first steps to understand the gap between idealized datasets and realistic applications of modern disentanglement learners by presenting the first large-scale empirical study examining how such models cope with correlated observational data. We find that existing methods fail to learn disentangled representations for strongly correlated factors of variation.
We discuss and quantify practical implications for downstream tasks like fair decision making. 
Despite these shortcomings, the learned latent space structure of some models naturally accommodates unseen examples via meaningful extrapolation, leading to out-of-distribution generalization capabilities. 
Based on these findings, we ultimately demonstrate how to correct for latent entanglement via fast adaptation and other weakly supervised training methods.
Future work is needed to address open question surrounding whether these results extend to high resolution inputs \citep{miladinovic2021spatial} and their impact on downstream tasks in the real world \citep{ahmed2020causalworld, dittadi2021representation}. 
Finally, our findings draw additional attention towards how inductive biases and weak supervision can be combined for successful disentanglement and under what circumstances this leads to strong out-of-distribution generalization.

\section*{Acknowledgments}
Frederik Träuble would like to thank Felix Leeb, Georgios Arvanitidis, Dominik Zietlow and Julius von Kügelgen for fruitful discussions. This work was supported by the German Federal Ministry of Education and Research (BMBF) through the Tübingen AI Center (FKZ: 01IS18039B) and the Deutsche Forschungsgemeinschaft (DFG, German Research Foundation) under Germany’s Excellence Strategy – EXC number 2064/1 – Project number 390727645. The authors thank the International Max Planck Research School for Intelligent Systems (IMPRS-IS) for supporting Frederik Träuble and CIFAR.

\bibliography{main}
\bibliographystyle{icml2021}
\clearpage
\appendix
\icmltitlerunning{On Disentangled Representations Learned from Correlated Data - Supplemental Material}
\input{appendix.tex}

\end{document}

%% file: commands.tex
\newcommand{\xhdr}[1]{\textbf{#1}\;}

\DeclareMathOperator{\E}{\mathbb{E}}

\newcommand{\given}{\,|\,}

\newcommand{\xb}{\mathbf{x}}
\newcommand{\zb}{\mathbf{z}}
\newcommand{\cb}{\mathbf{c}}

\newcommand{\Z}{\mathcal{Z}}
\newcommand{\C}{\mathcal{C}}
\newcommand{\HH}{\mathcal{H}}
\newcommand{\kl}{D_\mathrm{KL}}

\def\rvz{{\mathbf{z}}}

%% file: appendix.tex
\section{Proof of Proposition \ref{thm}}\label{app:proof}

Let $f^*: \C \rightarrow \Z$ be a diffeomorphism that transforms $p^*(\cb)$ (the true prior distribution of the factors of variation) into the fixed model prior $p(\zb)$:
\begin{equation}
    p^{*(\cb)}(\cb) = p^{(\zb)}( f^*(\cb) ) \left| \det \nabla_{\cb} f^*(\cb) \right|\ .
\end{equation}
Superscripts are included here to clarify which random variable a distribution is defined over, but will be often omitted.
The existence of $f^*$ implies that the true generative model $p^*(\xb \given \zb)$ can be expressed as a composition of a deterministic transformation, $(f^*)^{-1}$, and a stochastic one, $p^*(\xb \given \cb)$.

Similarly, let $f_{\theta}: \C \rightarrow \Z$ be a diffeomorphism, parameterized by $\theta$, that defines a distribution $p_{\theta}(\cb)$ (in general different from $p^*(\cb)$) with support $\C$:
\begin{equation}
    p_{\theta}^{(\cb)}(\cb) = p^{(\zb)}( f_{\theta}(\cb) ) \left| \det \nabla_{\cb} f_{\theta}(\cb) \right|\ .
\end{equation}
We will assume that the learned generative model $p_{\theta}(\xb \given \zb)$ can be expressed as a composition of the learned deterministic transformation $f_{\theta}^{-1}$ and the true $p^*(\xb \given \cb)$:
\begin{align}
    p_{\theta}(\xb) &= \int_{\zb} p_{\theta}(\xb \given \zb) p(\zb) d\zb \nonumber\\
    &= \int_{\cb} p_{\theta}(\xb \given f_{\theta}(\cb)) p_{\theta}(\cb) d\cb \nonumber\\
    &= \int_{\cb} p^*(\xb \given \cb) p_{\theta}(\cb) d\cb
\end{align}
where $\cb = f_{\theta}^{-1}(\zb)$. Note that the distribution $p_{\theta}(\cb)$ is implicitly learned by learning $f_{\theta}$, and it represents the learned prior distribution over the true factors of variation. 

The expected log likelihood we wish to maximize is
\begin{equation*}
    \E_{p^*(\xb)} [\log p_{\theta}(\xb)] = -\HH(p^*(\xb)) - \kl(p^*(\xb) \| p_{\theta}(\xb))
\end{equation*}
where the differential entropy $\HH(p^*(\xb))$ is constant with respect to the model parameters, and can therefore be ignored. The KL term can be rewritten as
\begin{align}
    \kl(& p^*(\xb) \| p_{\theta}(\xb)) 
    = \int_{\xb} p^*(\xb) \log \frac{p^*(\xb)}{p_{\theta}(\xb)} d\xb \nonumber\\
    &= \int_{\xb} \int_{\cb} p^*(\xb \given \cb) p^*(\cb) \log \frac{p^*(\cb) \frac{p^*(\xb \given \cb)}{p^*(\cb \given \xb)}}{p_{\theta}(\cb) \frac{p^*(\xb \given \cb)}{p^*(\cb \given \xb)}} d\cb\ d\xb \nonumber\\
    &= \int_{\xb} \int_{\cb} p^*(\xb \given \cb) p^*(\cb) \log \frac{p^*(\cb)}{p_{\theta}(\cb)} d\cb\ d\xb \nonumber\\
    &= \int_{\cb}  p^*(\cb) \log \frac{p^*(\cb)}{p_{\theta}(\cb)} \int_{\xb}  p^*(\xb \given \cb) d\xb\ d\cb \nonumber\\
    &= \kl(p^*(\cb) \| p_{\theta}(\cb)) \ .
\end{align}
Note that, since the KL divergence is always non-negative, the maximum likelihood corresponds to $\kl(p^*(\xb) \| p_{\theta}(\xb)) = 0$.

Let a matrix be $\sigma$-diagonal if there exists a permutation $\sigma$ that makes it diagonal.
Since by assumption $p^*(\cb)$ does not factorize while $p(\zb)$ does, it follows that $\nabla_{\cb} f^*(\cb)$ (the Jacobian of $f^*$) is not $\sigma$-diagonal.\footnote{
If $p(\zb)$ factorizes and $\nabla_{\cb} f^*(\cb)$ is $\sigma$-diagonal, then $p^*(\cb)$ also factorizes. Thus, since by assumption $p^*(\cb)$ does \textit{not} factorize, either $\nabla_{\cb} f^*(\cb)$ is not $\sigma$-diagonal or $p(\zb)$ does not factorize. Because the latter is false by assumption, it must be that $\nabla_{\cb} f^*(\cb)$ (the Jacobian of $f^*$) is \textit{not} $\sigma$-diagonal.}
However, if the representations $\zb$ are disentangled w.r.t. the true factors $\cb$ then the Jacobian of $f_{\theta}$ is $\sigma$-diagonal.
Thus, $f_{\theta}(\cb)$ cannot be equal to $f^*(\cb)$ almost everywhere. This in turn means that $\kl(p^*(\cb) \| p_{\theta}(\cb)) > 0$, hence $\E_{p^*(\xb)} [\log p_{\theta}(\xb)] < \E_{p^*(\xb)} [\log p^*(\xb)]$. 
This proves that if the generative model is disentangled w.r.t. the true factors then its expected likelihood is less than the optimal likelihood.

On the other hand, in the general case in which the representations are not necessarily disentangled, we can choose ${\theta}$ such that $f_{\theta}(\cb) = f^*(\cb)$ almost everywhere, which implies that $\kl(p^*(\cb) \| p_{\theta}(\cb))=0$. Thus, there exists an entangled model that has optimal likelihood.

We have proved that (i) if the generative model is constrained to be disentangled then the optimal likelihood cannot be achieved, and (ii) if it is \textit{not} constrained to be disentangled then the optimal likelihood \textit{can} be achieved. Equivalently, the optimal likelihood can be attained if and only if the generative model is entangled w.r.t. the true generative factors.

\section{Implementation details}
\label{sec:implementation_details}

\paragraph{Unsupervised Disentanglement methods.} For the sake of comparison, the considered disentanglement methods in this work cover the full collection of state-of-the-art approaches in \texttt{disentanglement\textunderscore lib} from \citet{locatello2018challenging} based on representations learned by VAEs.
The set contains six different methods that enforce disentanglement of the representation by equipping the loss with different regularizers that aim at enforcing the special structure of the posterior aggregate encoder distribution.
A detailed description of the regularizer forms used in this work, specifically $\beta$-VAE \citep{higgins2016beta}, FactorVAE \citep{kim2018disentangling}, AnnealedVAE \citep{burgess2018understanding}, DIP-VAE-I, DIP-VAE-II \citep{kumar2017variational} and $\beta$-TC-VAE \citep{chen2018isolating} is provided in \citet{locatello2018challenging}.
We use the same encoder and decoder architecture with 10 latent dimensions for every model.

\begin{figure}[t]
\minipage{0.245\columnwidth}
  \includegraphics[width=\columnwidth]{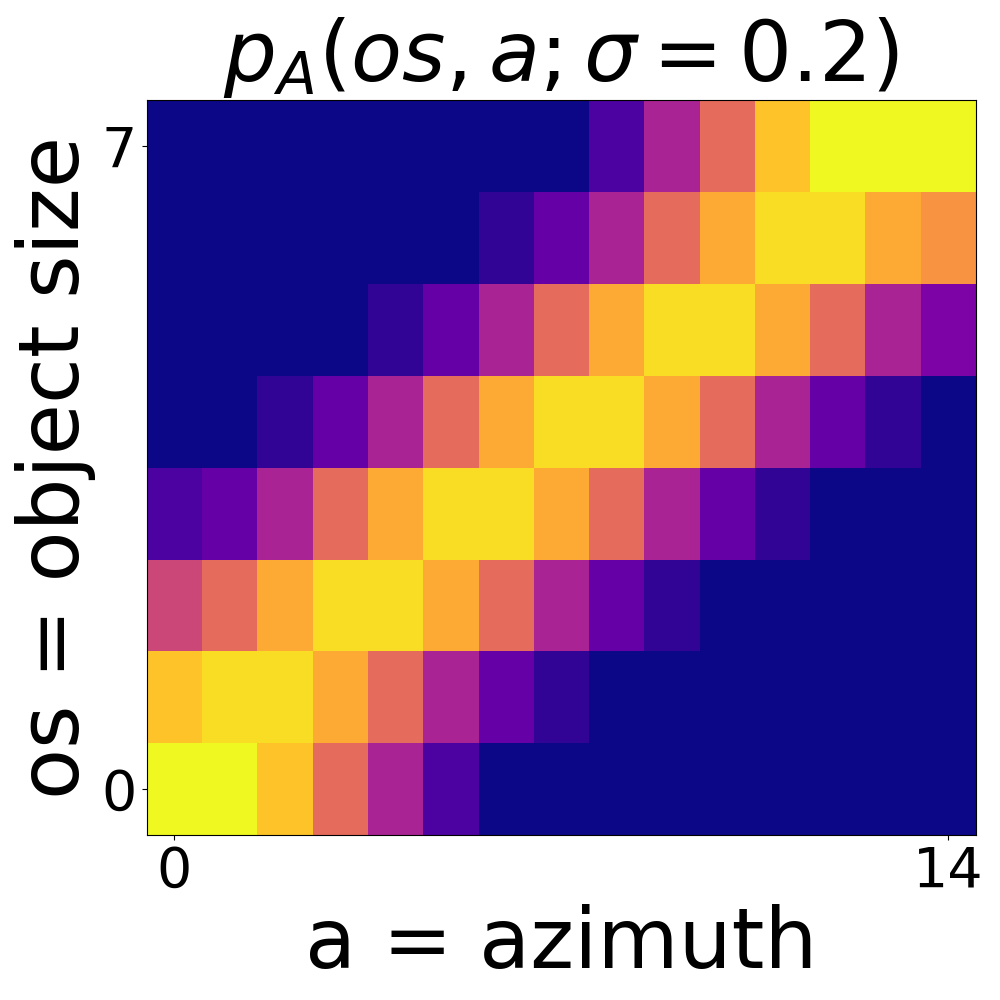}
\endminipage\hfill
\minipage{0.245\columnwidth}
  \includegraphics[width=\columnwidth]{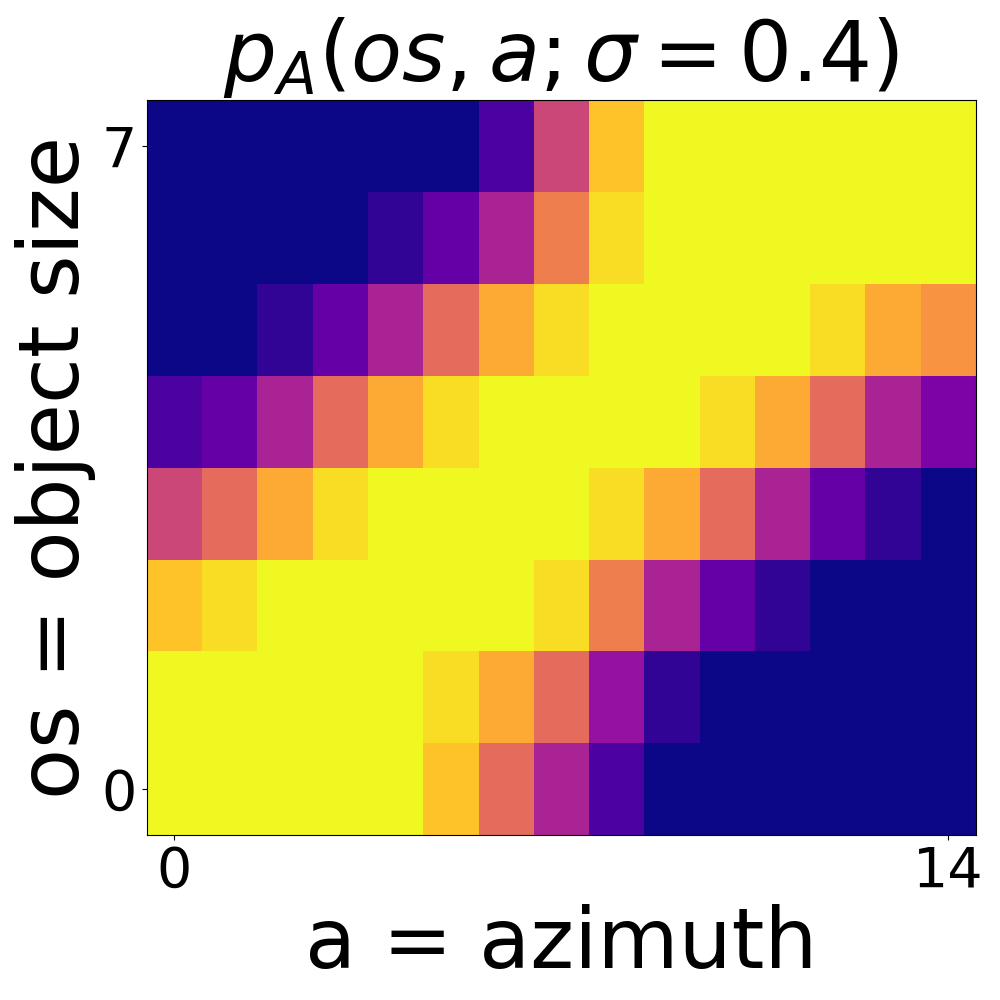}
\endminipage\hfill
\minipage{0.245\columnwidth}
  \includegraphics[width=\columnwidth]{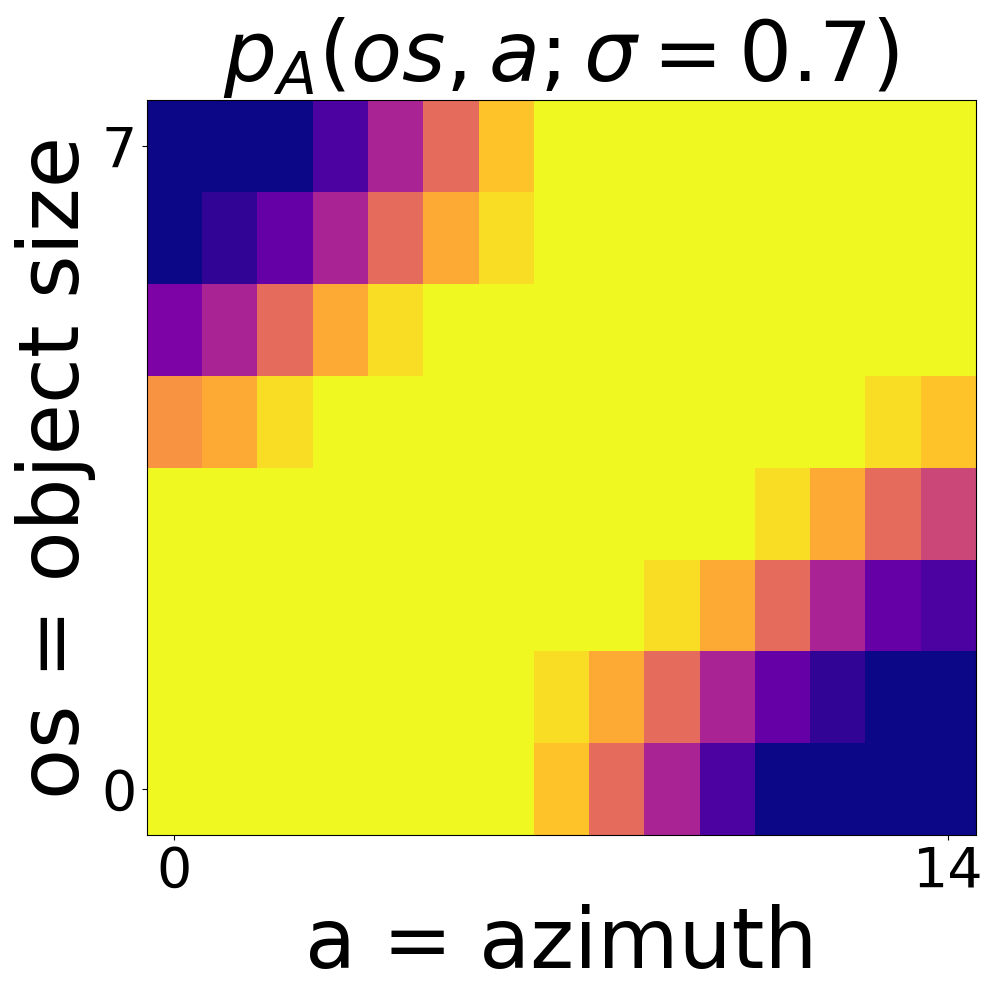}
\endminipage\hfill
\minipage{0.245\columnwidth}
  \includegraphics[width=\columnwidth]{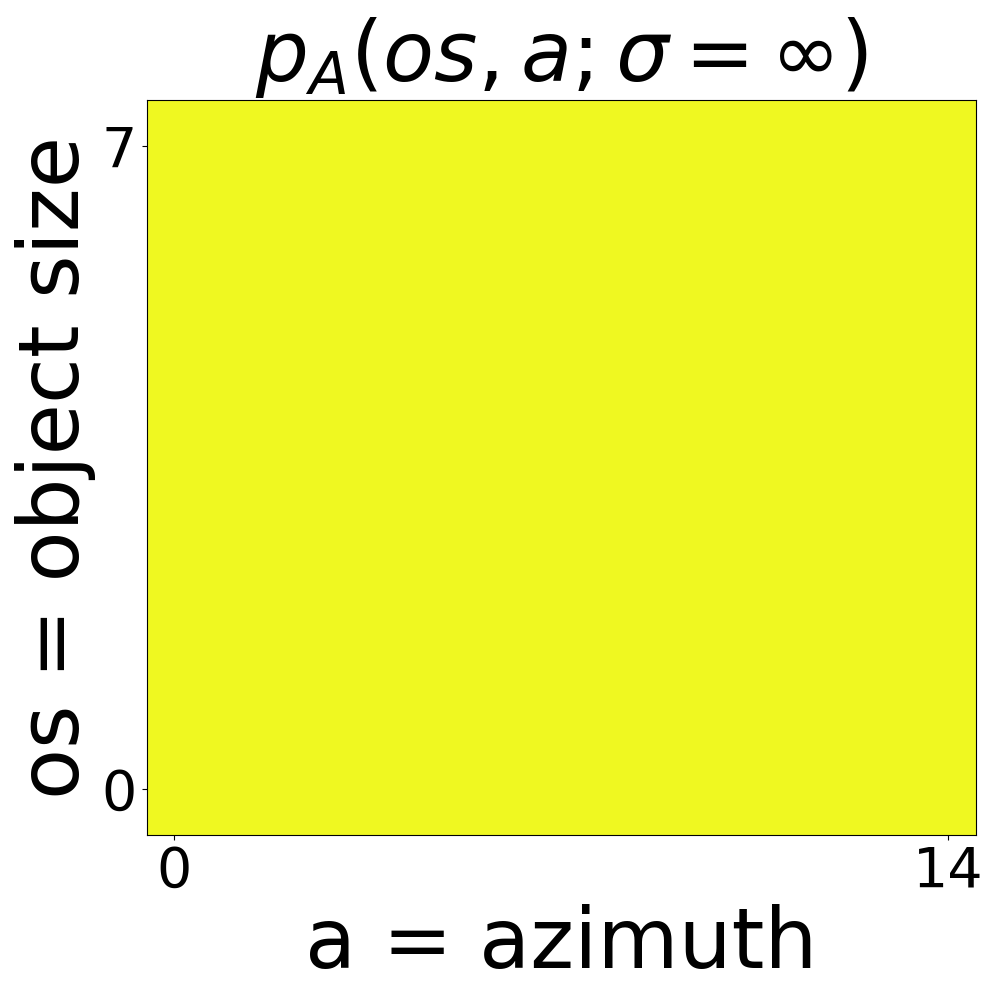}
\endminipage
\vskip 0.0in
\minipage{0.245\columnwidth}
  \includegraphics[width=\columnwidth]{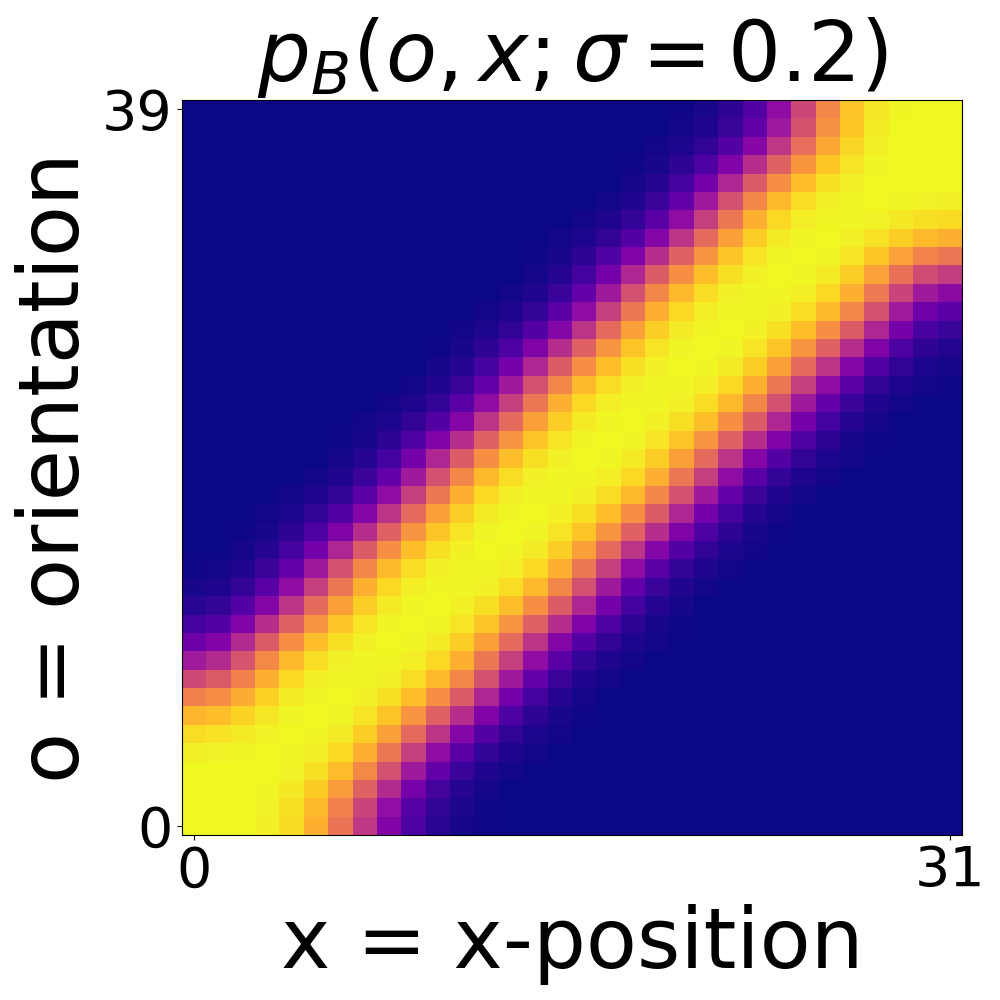}
\endminipage\hfill
\minipage{0.245\columnwidth}
  \includegraphics[width=\columnwidth]{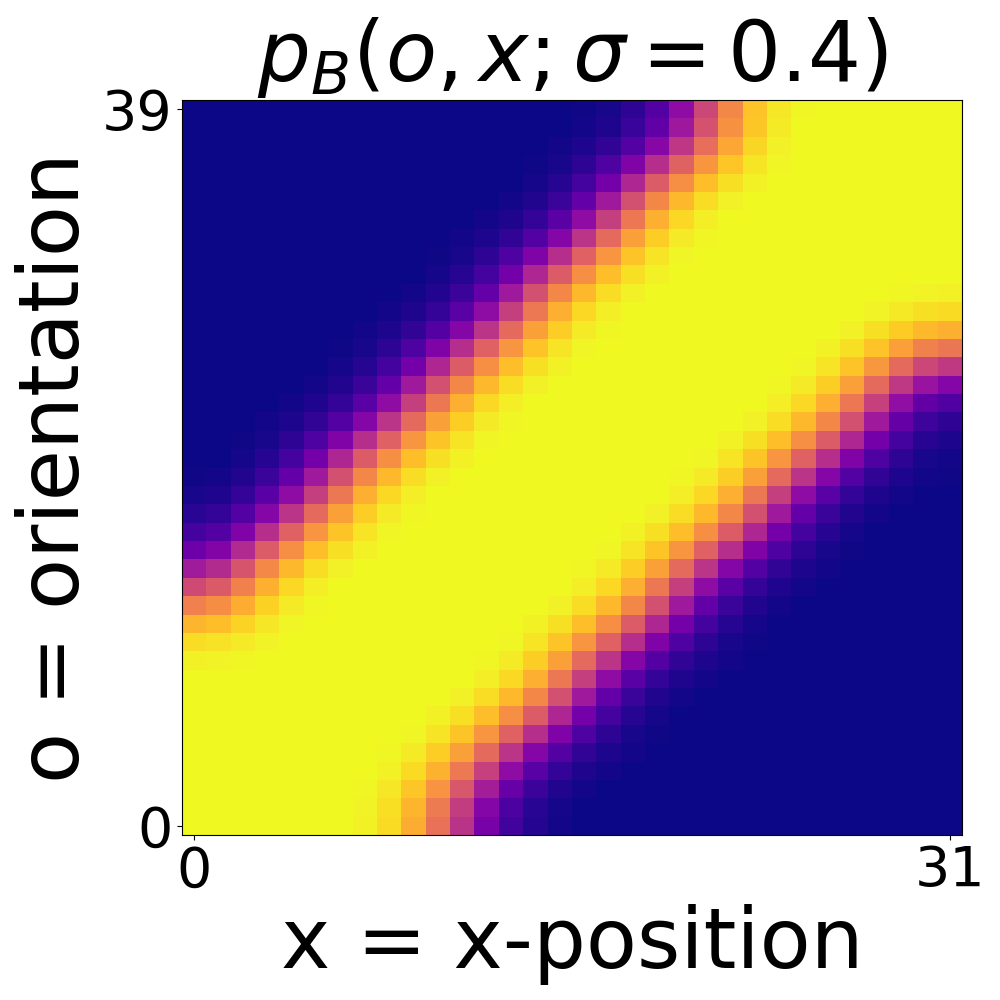}
\endminipage\hfill
\minipage{0.245\columnwidth}
  \includegraphics[width=\columnwidth]{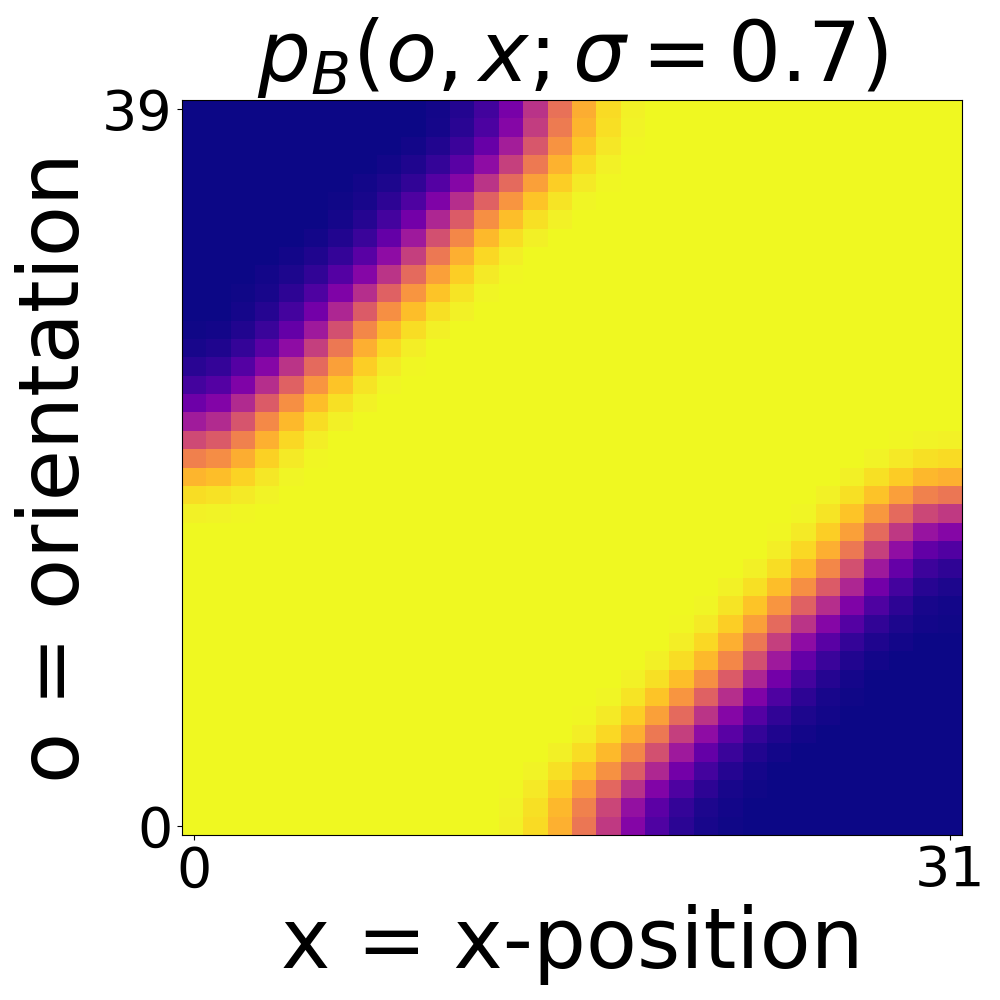}
\endminipage\hfill
\minipage{0.245\columnwidth}
  \includegraphics[width=\columnwidth]{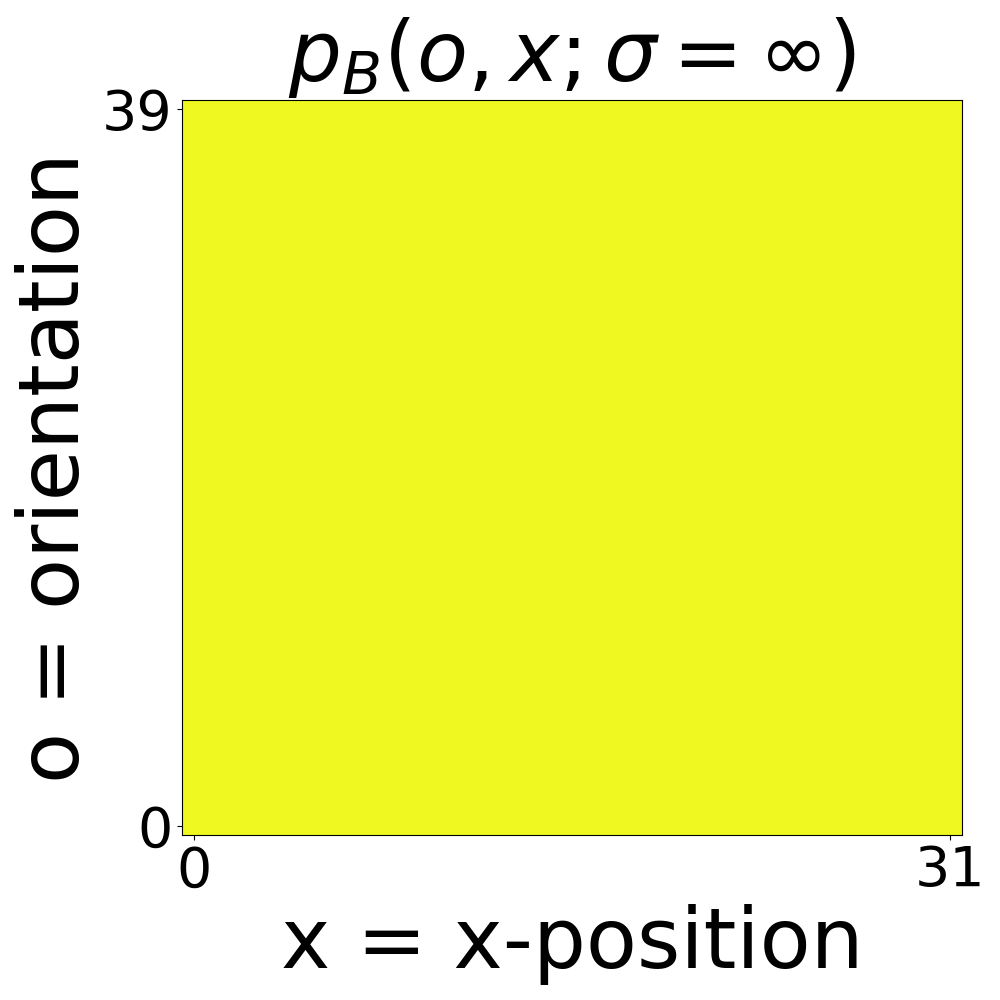}
\endminipage
\vskip 0.0in
\minipage{0.245\columnwidth}
  \includegraphics[width=\columnwidth]{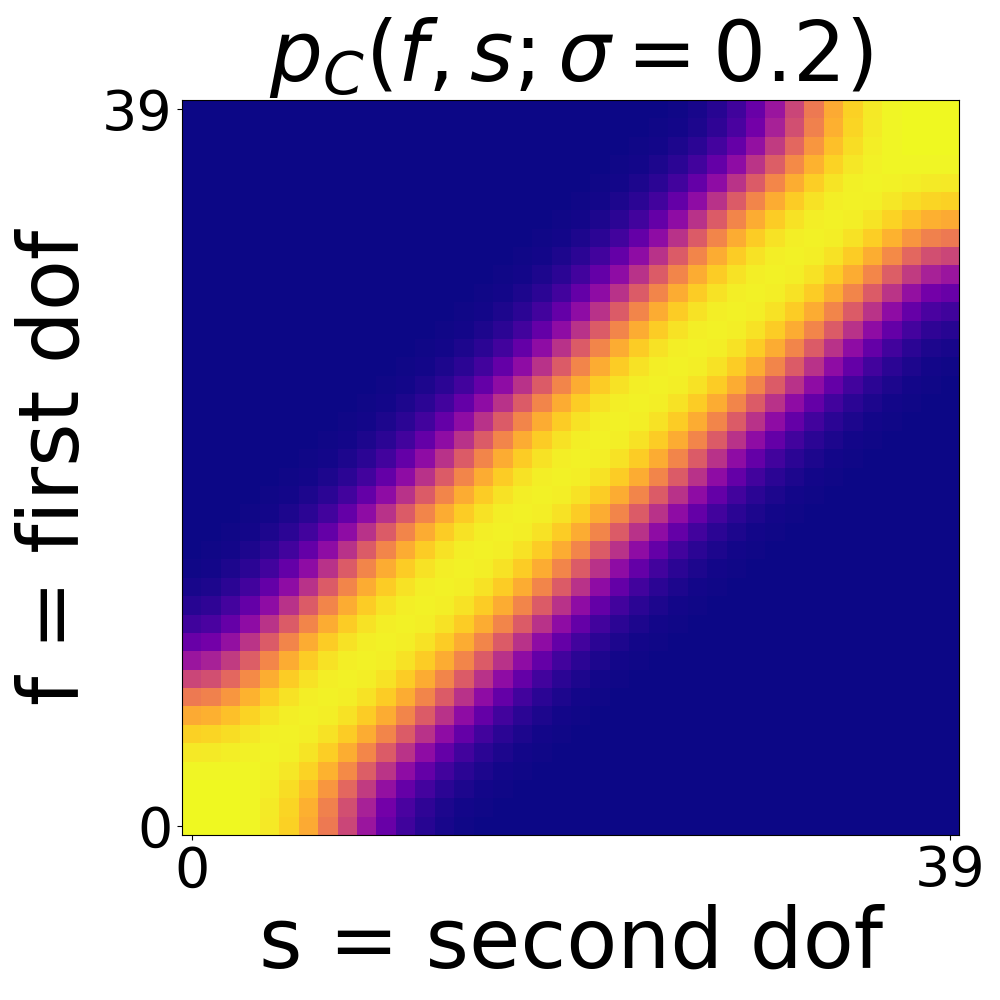}
\endminipage\hfill
\minipage{0.245\columnwidth}
  \includegraphics[width=\columnwidth]{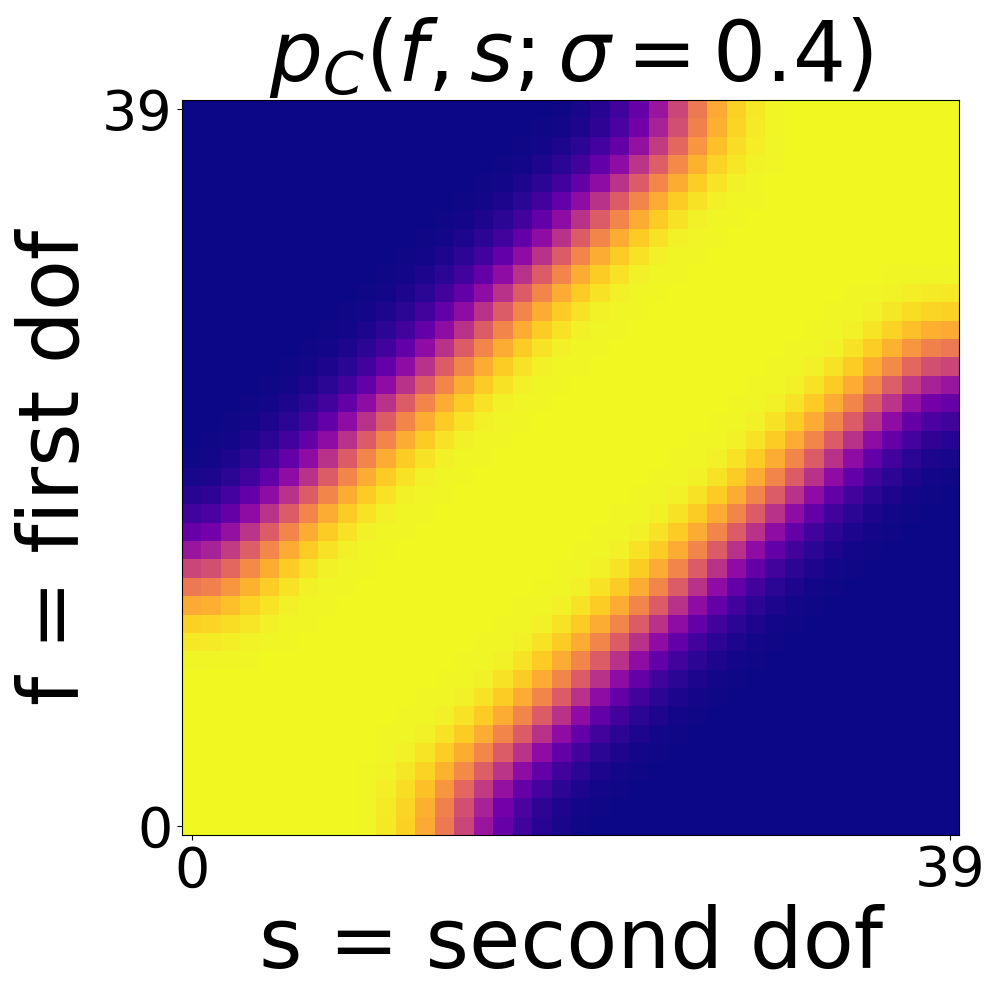}
\endminipage\hfill
\minipage{0.245\columnwidth}
  \includegraphics[width=\columnwidth]{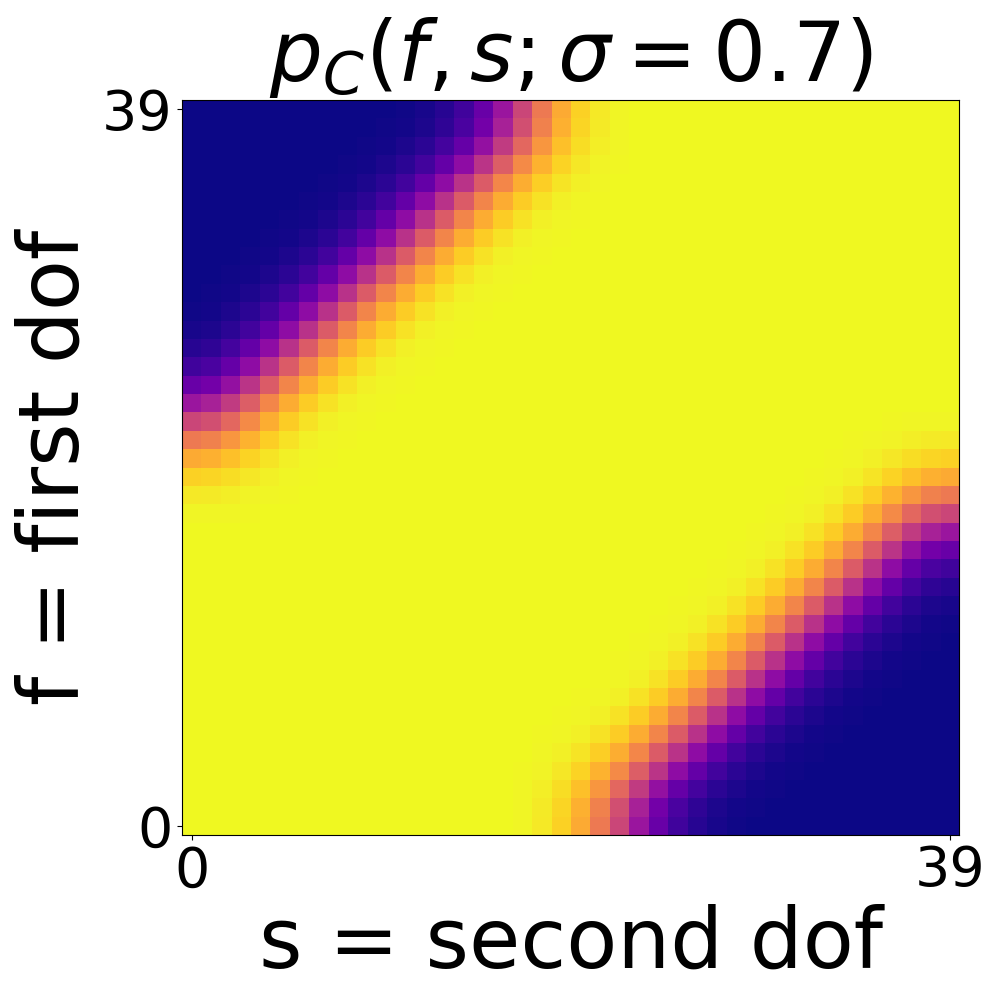}
\endminipage\hfill
\minipage{0.245\columnwidth}
  \includegraphics[width=\columnwidth]{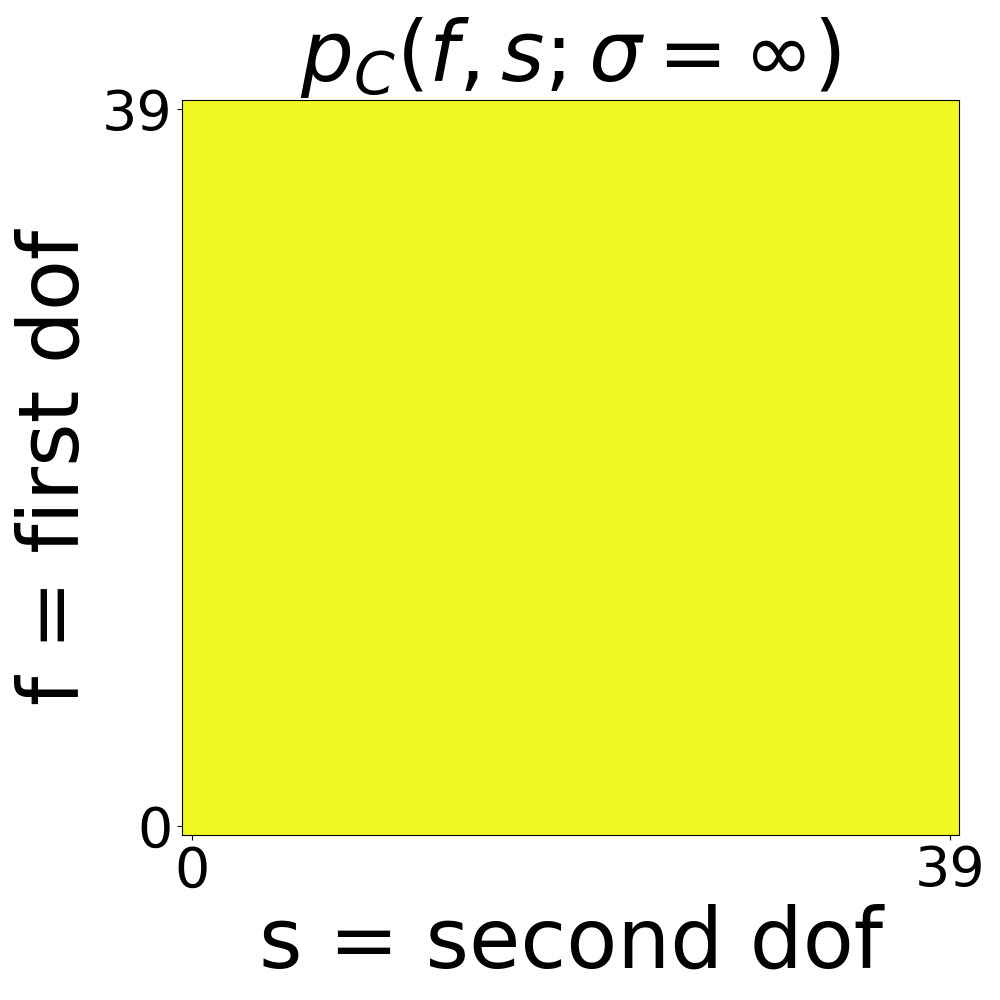}
\endminipage
\vskip 0.0in
\minipage{0.245\columnwidth}
  \includegraphics[width=\columnwidth]{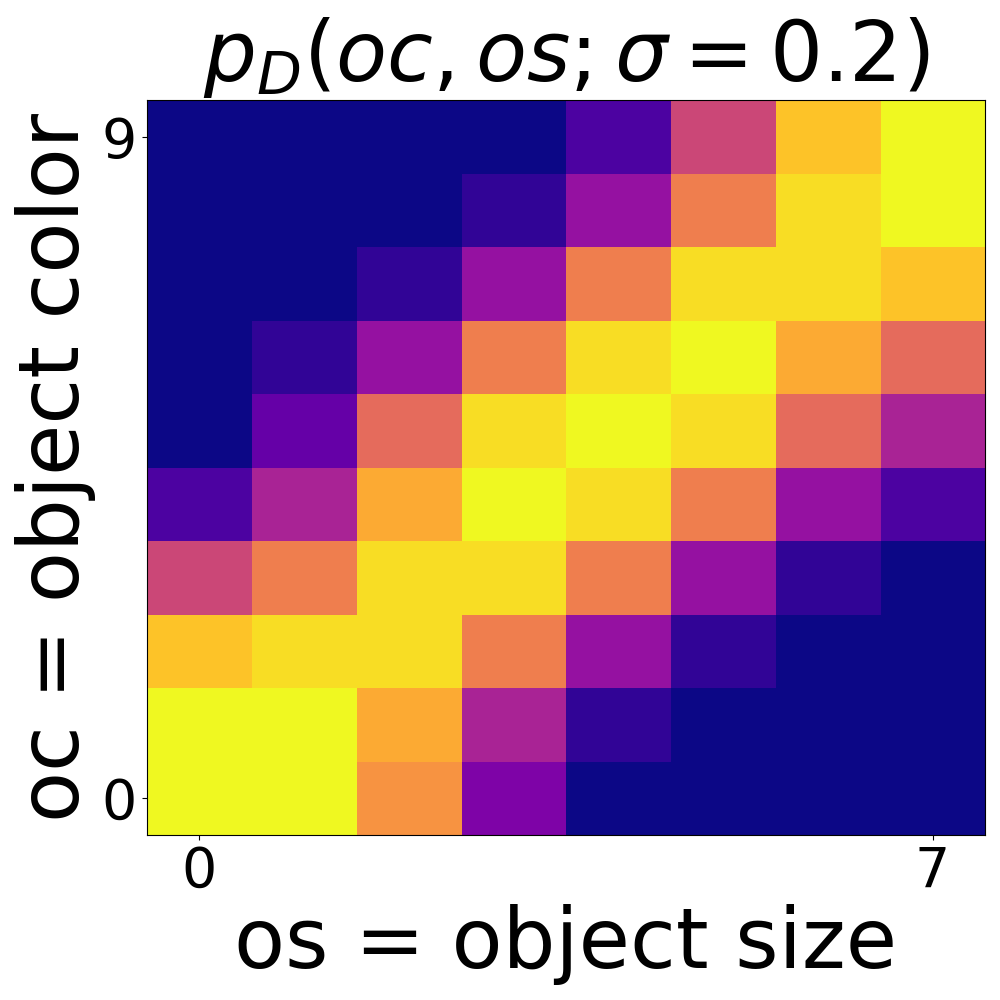}
\endminipage\hfill
\minipage{0.245\columnwidth}
  \includegraphics[width=\columnwidth]{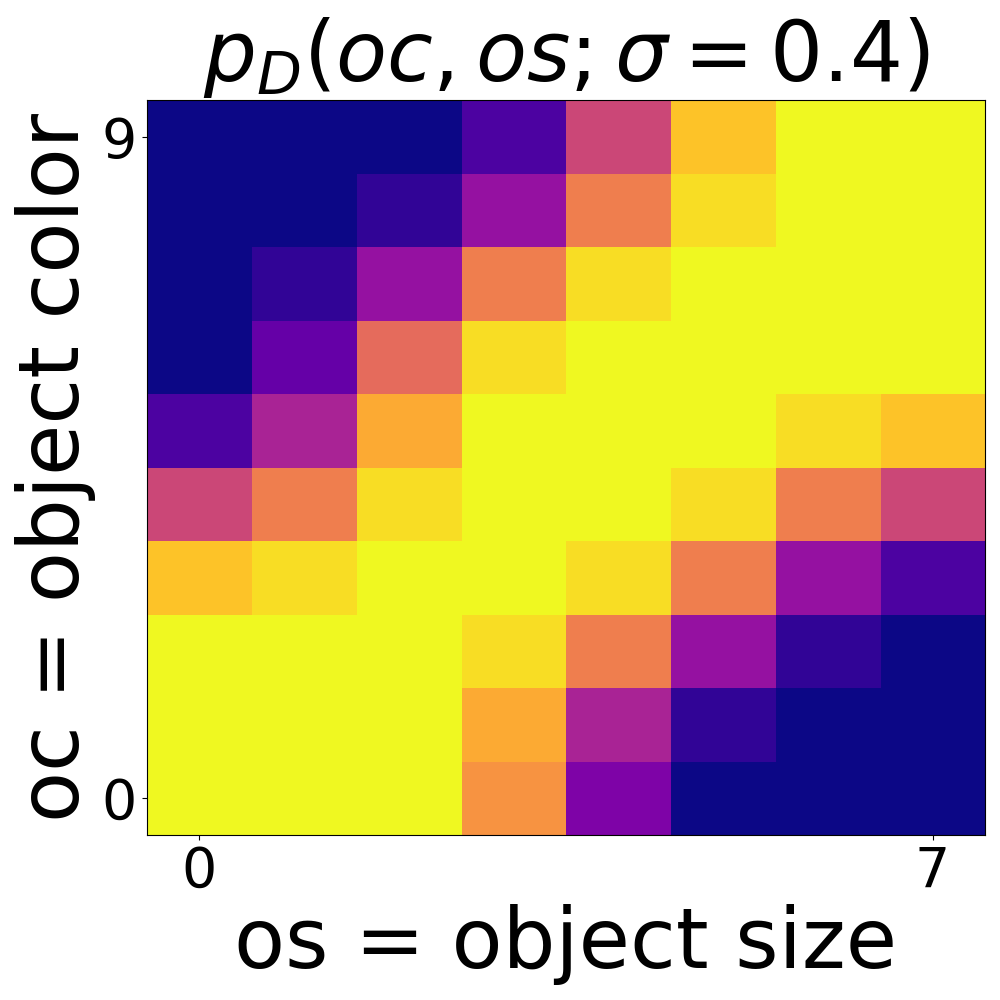}
\endminipage\hfill
\minipage{0.245\columnwidth}
  \includegraphics[width=\columnwidth]{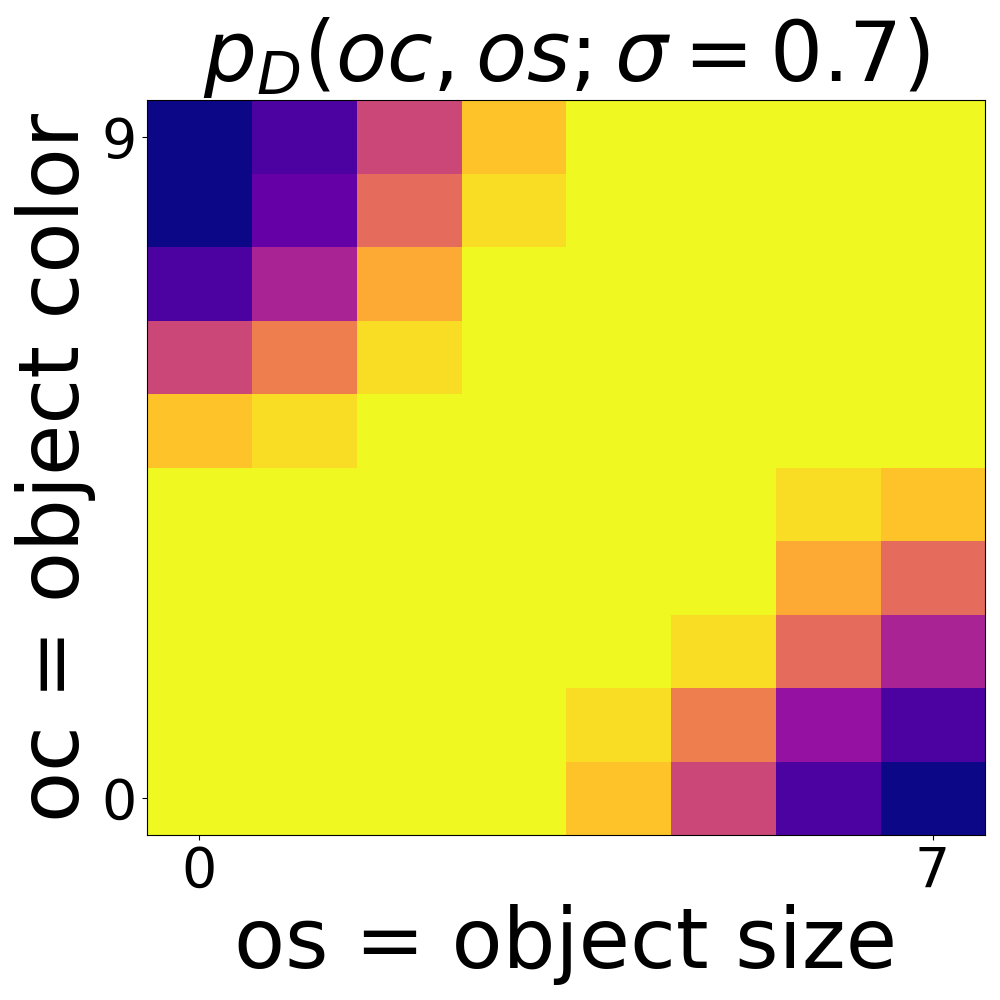}
\endminipage\hfill
\minipage{0.245\columnwidth}
  \includegraphics[width=\columnwidth]{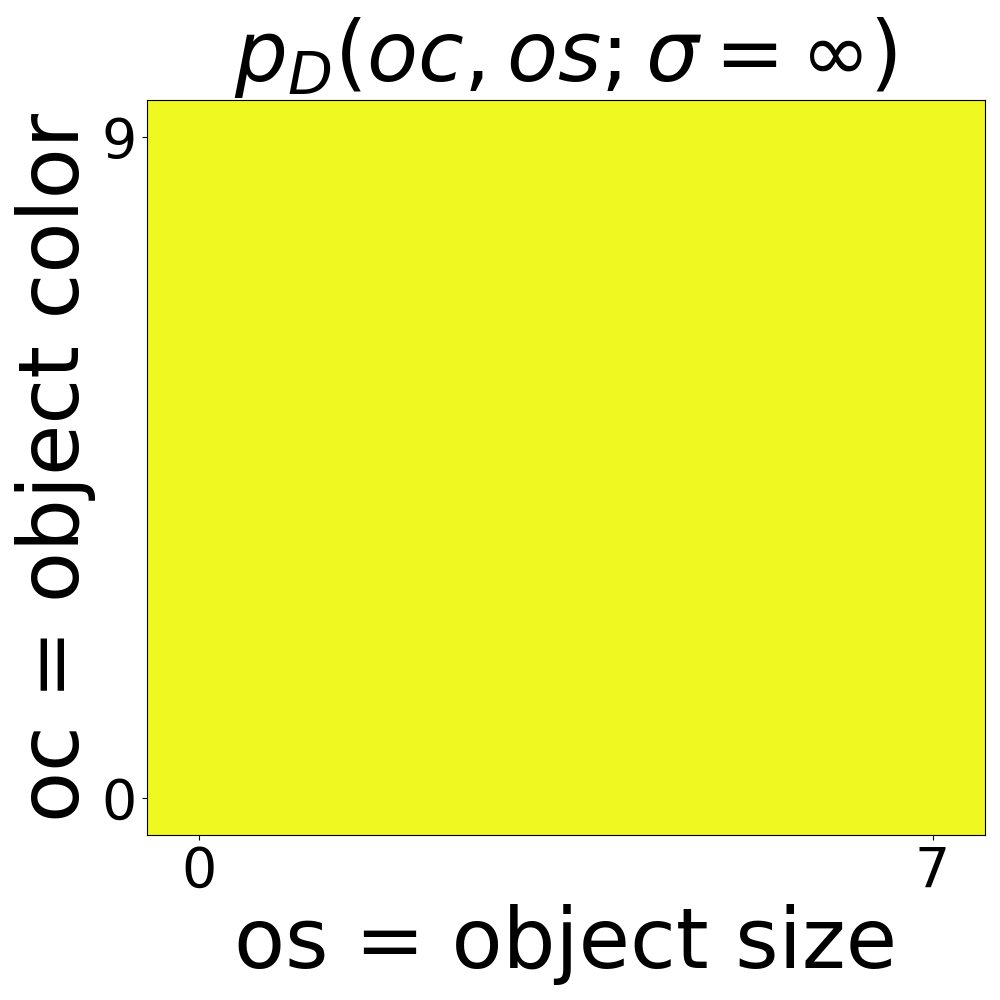}
\endminipage
\vskip 0.0in
\minipage{0.245\columnwidth}
  \includegraphics[width=\columnwidth]{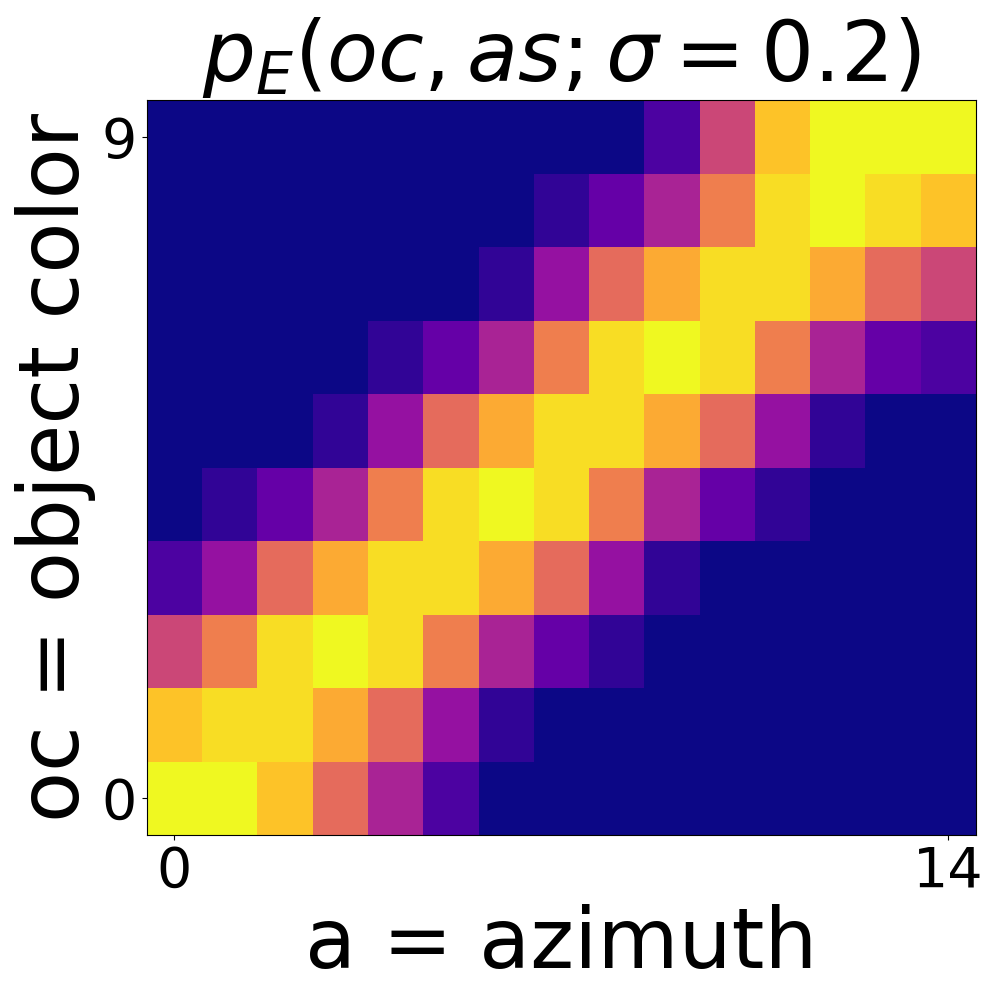}
\endminipage\hfill
\minipage{0.245\columnwidth}
  \includegraphics[width=\columnwidth]{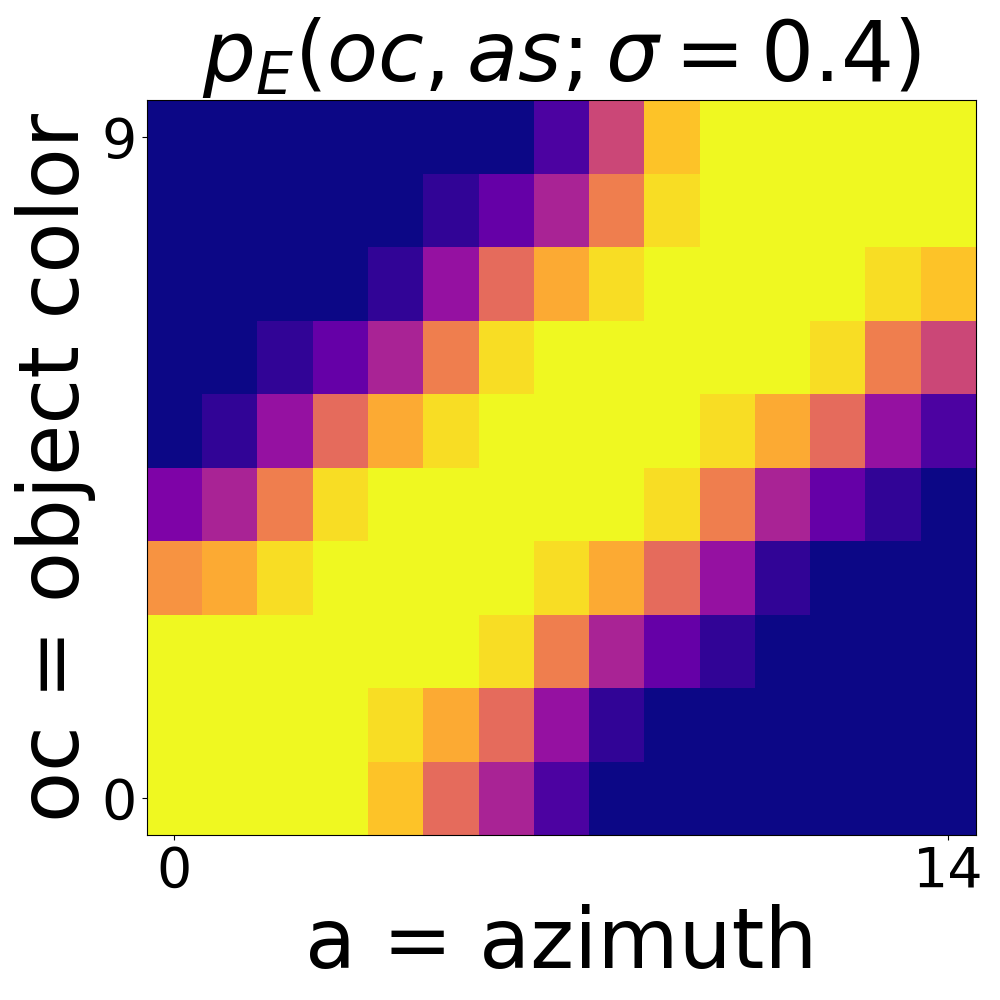}
\endminipage\hfill
\minipage{0.245\columnwidth}
  \includegraphics[width=\columnwidth]{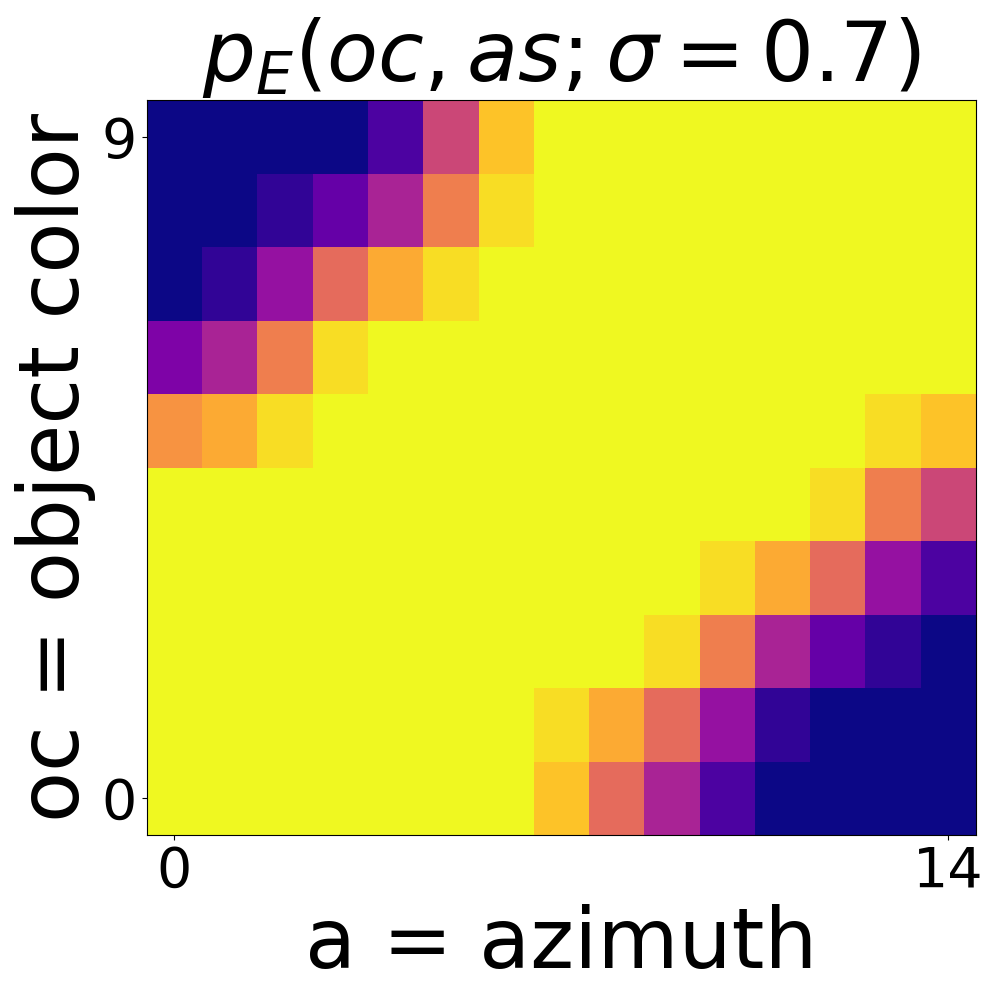}
\endminipage\hfill
\minipage{0.245\columnwidth}
  \includegraphics[width=\columnwidth]{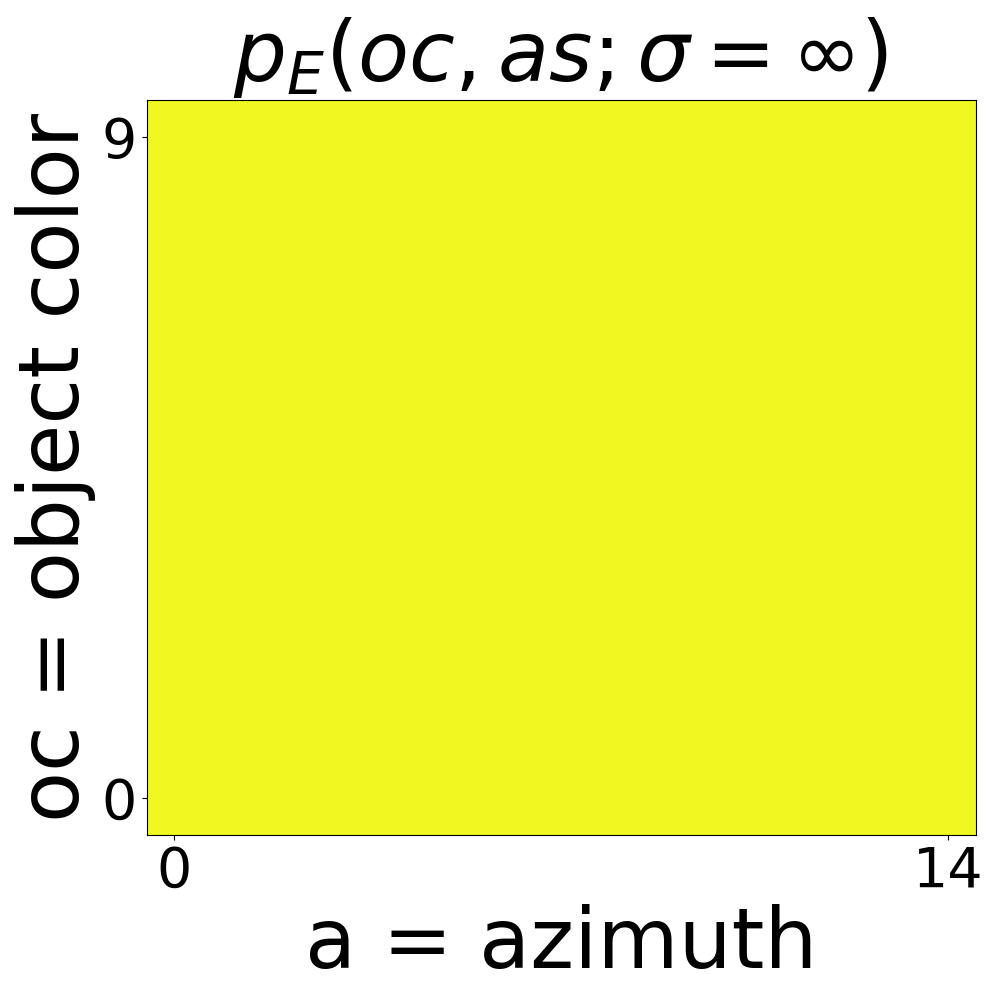}
\endminipage\hfill
\caption{Probability distributions for sampling training data in the correlated pair of FoVs in the respective datasets (A, B, C, D, E) considering correlation strengths of $\sigma = 0.2$, $\sigma = 0.4$, $\sigma = 0.7$ and $\sigma = \infty$, the uncorrelated limit (from left to right).}
\label{fig:prob_distribution_shapes3d_35}
\end{figure}

\paragraph{Joint distributions of correlated factors in datasets.} In \cref{fig:prob_distribution_shapes3d_35} we show the joint probability distributions of the correlated pair of FoV for all datasets and correlation strengths considered in this study. Dataset A, B and C were designed with correlated factors of variation that are ordinal for a natural visual interpretation of the traversals. In contrast, datasets D and E contain a correlated factor of variation that has no such natural ordering.

\paragraph{Pairwise entanglement metric.} 
We base the computation of the pairwise entanglement metric on a procedure developed in \citet{locatello2020sober}: Starting from the GBT feature importance matrix between encoded FoV and latent codes, a bipartite graph between latent codes and FoVs is constructed such that the edge weights are the corresponding matrix elements. After deleting all edges that have weight smaller than a given threshold, one counts the number of disconnected sub graphs having more than one vertex and how many FoV are still connected with at least one latent. When repeating this computation for different thresholds, one can track at which particular threshold a given pair of FoV is merged with each other in this bipartite graph, resulting in the metric we are reporting. A pair of FoV that are being merged at a higher threshold are statistically more related to each other via shared latent code dimensions. This computation can be not only based on the GBT feature importance matrix but likewise on weight matrices inferred from the mutual information.

\paragraph{Unfairness between a pair of FoV.} The scores reported are based on a notion of demographic parity for predicting a target variable $y$ given a protected and sensitive variable $s$. Both $y$ and $s$ can be associated with a factor of variation here. Rather than using the global total variation average as defined in \citet{locatello2019fairness}, we report the individual demographic parities for the correlated factors specifically.

\paragraph{Disentanglement metrics.} To measure disentanglement of a learned representation, various metrics have been proposed, each requiring access to the ground truth labels.
The BetaVAE score is based upon the prediction of a fixed factor from the disentangled representation using a linear classifier \citep{higgins2016beta}.
The FactorVAE score is intended to correct for some failures of the former by utilizing majority vote classifiers based on a normalized variance of each latent dimension \citep{kim2018disentangling}.
The SAP score represents the mean distance between the classification errors of the two latent dimensions that are most predictable \citep{kumar2017variational}.
For the MIG score, one computes the mutual information between the latent representation and the ground truth factors and calculates the final score using a normalized gap between the two highest MI entries for each factor.
Finally, a disentanglement score proposed by \citet{eastwood2018framework}, often referred to as DCI score, is calculated from a dimension-wise entropy reflecting the usefulness of the dimension to predict a single factor of variation.

\begin{table}
\minipage{1.0\columnwidth}
\resizebox{\textwidth}{!}{
  \begin{tabular}{llcccc}
\toprule
 corr. strength & & $\sigma=0.2$  & $\sigma=0.4$ & $\sigma=0.7$ & $\sigma=\infty$ (uc)   \\
 \midrule
 Shapes3D (A) & \textcolor{BrickRed}{object size - azimuth} & \textcolor{BrickRed}{0.38} (\textcolor{BrickRed}{0.28})   & \textcolor{BrickRed}{0.26} (\textcolor{BrickRed}{0.25})    & \textcolor{BrickRed}{0.13} (\textcolor{BrickRed}{0.2})  & \textcolor{blue}{0.08} (\textcolor{blue}{0.17})          \\
 & median other pairs                     & 0.09 (0.2)         & 0.09 (0.2)        & 0.09 (0.19)        & 0.08 (0.18) \\
\midrule
 dSprites (B) & \textcolor{BrickRed}{orientation - position x} &\textcolor{BrickRed}{0.17} (\textcolor{BrickRed}{0.34})   & \textcolor{BrickRed}{0.16} (\textcolor{BrickRed}{0.31})   & \textcolor{BrickRed}{0.14} (\textcolor{BrickRed}{0.24})  & \textcolor{blue}{0.11} (\textcolor{blue}{0.14})           \\
 & median other pairs                        & 0.13 (0.16)         & 0.13 (0.18)        & 0.13 (0.19)        & 0.13 (0.15)                \\
 \midrule
 MPI3D (C) & \textcolor{BrickRed}{First DOF - Second DOF} &\textcolor{BrickRed}{0.2} (\textcolor{BrickRed}{0.54})   & \textcolor{BrickRed}{0.19} (\textcolor{BrickRed}{0.52})  & \textcolor{BrickRed}{0.17} (\textcolor{BrickRed}{0.5})  & \textcolor{blue}{0.16} (\textcolor{blue}{0.49})          \\
 & median other pairs                      & 0.16 (0.25)        & 0.16 (0.25)        & 0.15 (0.26)       & 0.15 (0.25) \\
\midrule
 Shapes3D (D) & \textcolor{BrickRed}{object color - object size} &  \textcolor{BrickRed}{0.29} (\textcolor{BrickRed}{0.38}) &  \textcolor{BrickRed}{0.28} (\textcolor{BrickRed}{0.31}) & - & -  \\
 & median uncorrelated pairs                   &  0.07 (0.11) &  0.07 (0.11) & - & - \\
\midrule
 Shapes3D (E) & \textcolor{BrickRed}{object color - azimuth} &  \textcolor{BrickRed}{0.25} (\textcolor{BrickRed}{0.43}) &  \textcolor{BrickRed}{0.23} (\textcolor{BrickRed}{0.3}) & - & -  \\
 & median uncorrelated pairs               &  0.1 (0.15)  &  0.09 (0.15) & - & -  \\
\bottomrule
\end{tabular}}
\endminipage
\caption{Pairwise entanglement scores help to uncover still existent correlations in the latent representation.
Mean of the pairwise entanglement scores for the correlated pair (red) and the median of the uncorrelated pairs. We see that stronger correlation leads to statistically more entanglement latents across all datasets studied compared to their baseline pairwise entanglement where the data exhibits no correlations (blue).
Each score is the mean across 180 models for each dataset and correlation strength.
Scores are based on GBT feature importance; scores in brackets are based on the Mutual Information.}
\label{fig:pairwise_scores_thresholds_extended}
\end{table}

\section{Additional Results \cref{sec:section4}}\label{sec:appendix_additional_results}

\subsection{\cref{sec:unsupervised_result}}
\label{sec:unsupervised_result_appendix}

\begin{figure*}
\begin{center}
\hfill
\minipage{0.55\textwidth}
  \includegraphics[width=\columnwidth]{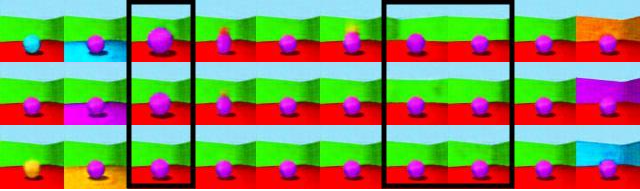}
\endminipage
\hfill
\minipage{0.3\textwidth}
\includegraphics[width=\columnwidth]{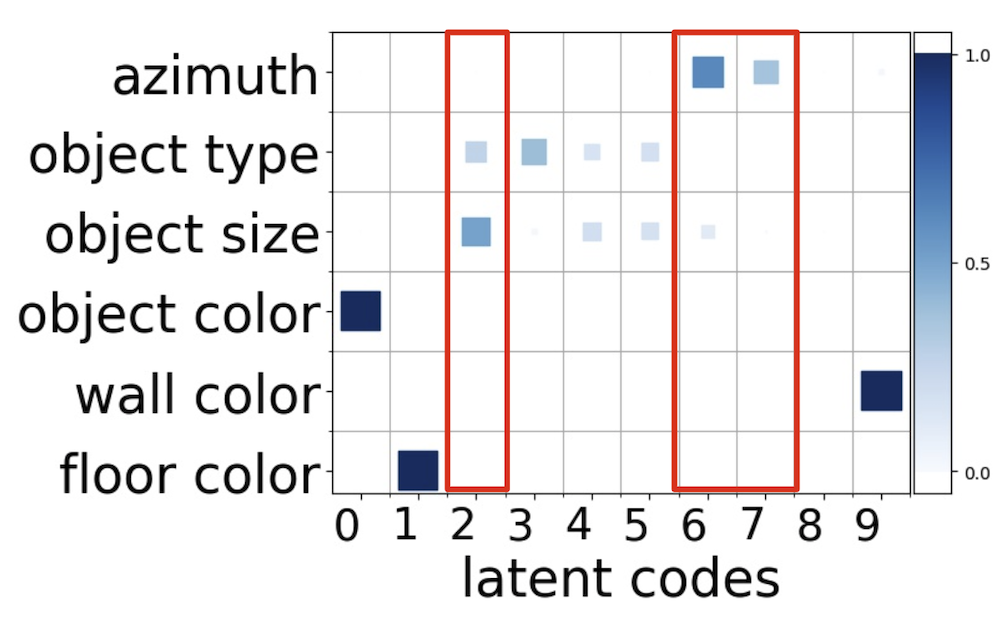}
\endminipage
\hfill
\end{center}
\caption{
We show latent traversals (left) of the best DCI score model among all 180 trained models with weak correlation ($\sigma = 0.7$) in object size and azimuth.
The traversals in latent codes 2, 6, and 7 (highlighted in black), suggest that these dimensions encode no mixture of azimuth and object size compared to the models with stronger correlation. This is supported by the GBT feature importance matrix of this model (right).} 
\label{fig:representation_structure_under_correlation_traversals_appendix}
\end{figure*}

\begin{table*}
\tiny
\begin{subtable}{.5\textwidth}
\caption{$\beta-$VAE}
  \begin{tabular}{llcccc}
\toprule
 corr. strength & & 0.2  & 0.4 & 0.7 & $\infty$ (uc)   \\
 \midrule
 Shapes3D (A) & \textcolor{BrickRed}{object size - azimuth} &  0.38 &  0.25 &  0.14 &            0.08 \\
 & median uncorrelated pairs              &  0.08 &  0.09 &  0.09 &            0.07 \\
\midrule
 dSprites (B) & \textcolor{BrickRed}{orientation - position x} & 0.18 &  0.16 &  0.14 &            0.12 \\
 & median uncorrelated pairs                 &  0.14 &  0.14 &  0.14 &            0.12 \\
 \midrule
 MPI3D (C) & \textcolor{BrickRed}{First DOF - Second DOF} & 0.22 &  0.19 &  0.18 &            0.16 \\
 & median uncorrelated pairs               &  0.16 &  0.15 &  0.15 &            0.14 \\
\midrule
 Shapes3D (D) & \textcolor{BrickRed}{object color - object size} &  0.28 &  0.3  & - & - \\
 & median uncorrelated pairs                   &  0.08 &  0.07 & - & - \\
\midrule
 Shapes3D (E) & \textcolor{BrickRed}{object color - azimuth} &  0.24 &  0.25 & - & - \\
 & median uncorrelated pairs               &  0.11 &  0.09 & - & - \\
\bottomrule
\end{tabular}
\end{subtable}
\hfill
\tiny
\begin{subtable}{.5\textwidth}
\caption{Factor-VAE}
  \begin{tabular}{llcccc}
\toprule
 corr. strength & & 0.2  & 0.4 & 0.7 & $\infty$ (uc)   \\
 \midrule
 Shapes3D (A) & \textcolor{BrickRed}{object size - azimuth} &  0.48 &  0.3  &  0.1  &            0.03 \\
 & median uncorrelated pairs              &  0.07 &  0.06 &  0.07 &            0.04 \\
\midrule
 dSprites (B) & \textcolor{BrickRed}{orientation - position x} & 0.23 &  0.2  &  0.16 &            0.12 \\
 & median uncorrelated pairs                 &  0.14 &  0.15 &  0.14 &            0.13 \\
 \midrule
 MPI3D (C) & \textcolor{BrickRed}{First DOF - Second DOF} &  0.23 &  0.22 &  0.19 &            0.18 \\
 & median uncorrelated pairs               &  0.15 &  0.15 &  0.14 &            0.15 \\
\midrule
 Shapes3D (D) & \textcolor{BrickRed}{object color - object size} & 0.36 &  0.33 & - & - \\
 & median uncorrelated pairs                   &  0.02 &  0.03 & - & - \\
\midrule
 Shapes3D (E) & \textcolor{BrickRed}{object color - azimuth} &  0.3 &  0.28 & - & - \\
 & median uncorrelated pairs               &   0.1 &  0.08 & - & -\\
\bottomrule
\end{tabular}
\end{subtable}
\vskip 0.15in
\tiny
\begin{subtable}{.5\textwidth}
\caption{Annealed-VAE}
  \begin{tabular}{llcccc}
\toprule
 corr. strength & & 0.2  & 0.4 & 0.7 & $\infty$ (uc)   \\
 \midrule
 Shapes3D (A) & \textcolor{BrickRed}{object size - azimuth} &  0.32 &  0.25 &  0.13 &            0.11 \\
 &median uncorrelated pairs              &  0.09 &  0.09 &  0.11 &            0.1  \\
\midrule
 dSprites (B) & \textcolor{BrickRed}{orientation - position x} & 0.17 &  0.16 &  0.14 &            0.1  \\
 &median uncorrelated pairs                 &  0.14 &  0.15 &  0.14 &            0.15 \\
 \midrule
 MPI3D (C) & \textcolor{BrickRed}{First DOF - Second DOF} &  0.17 &  0.17 &  0.15 &            0.15 \\
 &median uncorrelated pairs               &  0.15 &  0.15 &  0.15 &            0.14 \\
\midrule
 Shapes3D (D) & \textcolor{BrickRed}{object color - object size} & 0.33 &  0.28 & - & -  \\
 & median uncorrelated pairs                   &  0.07 &  0.08 & - & -  \\
\midrule
 Shapes3D (E) & \textcolor{BrickRed}{object color - azimuth} &  0.25 &  0.19 & - & -  \\
 & median uncorrelated pairs               &  0.1  &  0.1 & - & -   \\
\bottomrule
\end{tabular}
\end{subtable}
\hfill
\begin{subtable}{.5\textwidth}
\caption{$\beta$-TC-VAE}
  \begin{tabular}{llcccc}
\toprule
 corr. strength & & 0.2  & 0.4 & 0.7 & $\infty$ (uc)   \\
 \midrule
 Shapes3D (A) & \textcolor{BrickRed}{object size - azimuth} &  0.41 &  0.26 &  0.09 &            0.05 \\
 &median uncorrelated pairs              &  0.07 &  0.09 &  0.06 &            0.05 \\
\midrule
 dSprites (B) & \textcolor{BrickRed}{orientation - position x} & 0.18 &  0.15 &  0.13 &            0.11 \\
 &median uncorrelated pairs                 &  0.14 &  0.14 &  0.13 &            0.12 \\
 \midrule
 MPI3D (C) & \textcolor{BrickRed}{First DOF - Second DOF} &  0.24 &  0.22 &  0.19 &            0.17 \\
 &median uncorrelated pairs               &  0.18 &  0.17 &  0.15 &            0.15 \\
\midrule
 Shapes3D (D) & \textcolor{BrickRed}{object color - object size} & 0.3  &  0.29 & - & -  \\
 &median uncorrelated pairs                   &  0.05 &  0.06 & - &  \\
\midrule
 Shapes3D (E) & \textcolor{BrickRed}{object color - azimuth} &  0.23 &  0.23 & - & -  \\
 &median uncorrelated pairs               &  0.09 &  0.07 & - &  \\
\bottomrule
\end{tabular}
\end{subtable}
\vskip 0.15in
\begin{subtable}{.5\textwidth}
\caption{Dip-VAE-I}
  \begin{tabular}{llcccc}
\toprule
 corr. strength & & 0.2  & 0.4 & 0.7 & $\infty$ (uc)  \\
 \midrule
 Shapes3D (A) & \textcolor{BrickRed}{object size - azimuth} &  0.38 &  0.24 &  0.14 &            0.07 \\
 &median uncorrelated pairs              &  0.1  &  0.1  &  0.11 &            0.09 \\
\midrule
 dSprites (B) & \textcolor{BrickRed}{orientation - position x} & 0.13 &  0.13 &  0.12 &            0.11 \\
 &median uncorrelated pairs                 &  0.11 &  0.11 &  0.11 &            0.11 \\
 \midrule
 MPI3D (C) & \textcolor{BrickRed}{First DOF - Second DOF} &  0.16 &  0.15 &  0.14 &            0.14 \\
 &median uncorrelated pairs               &  0.13 &  0.13 &  0.13 &            0.13 \\
\midrule
 Shapes3D (D) & \textcolor{BrickRed}{object color - object size} & 0.27 &  0.25  & - & -\\
 &median uncorrelated pairs                   &  0.06 &  0.06  & - & - \\
\midrule
 Shapes3D (E) & \textcolor{BrickRed}{object color - azimuth} &  0.22 &  0.22  & - & -\\
 &median uncorrelated pairs               &  0.1  &  0.1  & - & -  \\
\bottomrule
\end{tabular}
\end{subtable}
\hfill
\begin{subtable}{.5\textwidth}
\caption{Dip-VAE-II}
  \begin{tabular}{llcccc}
\toprule
 corr. strength & & 0.2  & 0.4 & 0.7 & $\infty$ (uc)  \\
 \midrule
 Shapes3D (A) & \textcolor{BrickRed}{object size - azimuth} &  0.28 &  0.23 &  0.19 &            0.14 \\
 &median uncorrelated pairs              &  0.12 &  0.11 &  0.11 &            0.11 \\
\midrule
 dSprites (B) & \textcolor{BrickRed}{orientation - position x} & 0.14 &  0.14 &  0.13 &            0.12 \\
 &median uncorrelated pairs                 &  0.14 &  0.14 &  0.13 &            0.13 \\
 \midrule
 MPI3D (C) & \textcolor{BrickRed}{First DOF - Second DOF} &  0.22 &  0.21 &  0.18 &            0.17 \\
 &median uncorrelated pairs               &  0.15 &  0.14 &  0.15 &            0.15 \\
\midrule
 Shapes3D (D) & \textcolor{BrickRed}{object color - object size} & 0.23 &  0.2 & - & -  \\
 &median uncorrelated pairs                   &  0.13 &  0.12   & - & -  \\
\midrule
 Shapes3D (E) & \textcolor{BrickRed}{object color - azimuth} &  0.23 &  0.19   & - & -  \\
 &median uncorrelated pairs               &  0.12 &  0.13   & - & -  \\
\bottomrule
\end{tabular}
\end{subtable}

\caption{Pairwise entanglement scores from \cref{fig:pairwise_scores_thresholds_extended} separated along each disentanglement regularizer. Mean of the pairwise entanglement scores for the correlated pair (red) and the median of the uncorrelated pairs. We see that stronger correlation leads to statistically more entanglement latents across all datasets and regularizers studied compared to their baseline pairwise entanglement where the data exhibits no correlations.
Each pairwise score is the mean across 30 models for each dataset and correlation strength.
Scores are based on GBT feature importance.}
\label{fig:pairwise_scores_thresholds_extended_per_regularizer}
\end{table*}

\paragraph{Latent structure and pairwise entanglement.} 
Our hypothesis that the latent representations are less correlated if the correlation strength is weaker is shown for a model on Shapes3D (A) with weak correlation in \cref{fig:representation_structure_under_correlation_traversals_appendix}. Here the latent traversals do not mirror the major and minor axis of the correlated joint distribution.

\begin{figure}
\minipage{0.3\columnwidth}
  \includegraphics[width=\columnwidth]{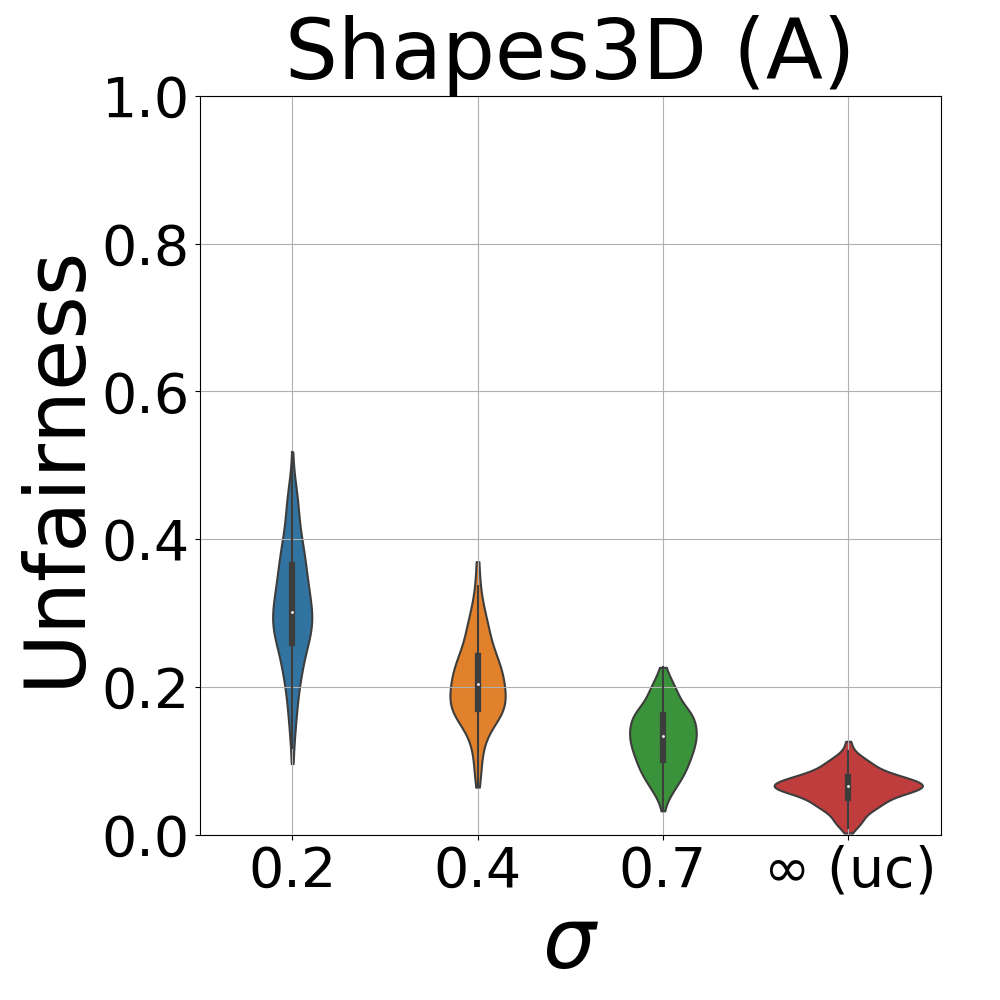}
\endminipage\hfill
\minipage{0.3\columnwidth}
  \includegraphics[width=\columnwidth]{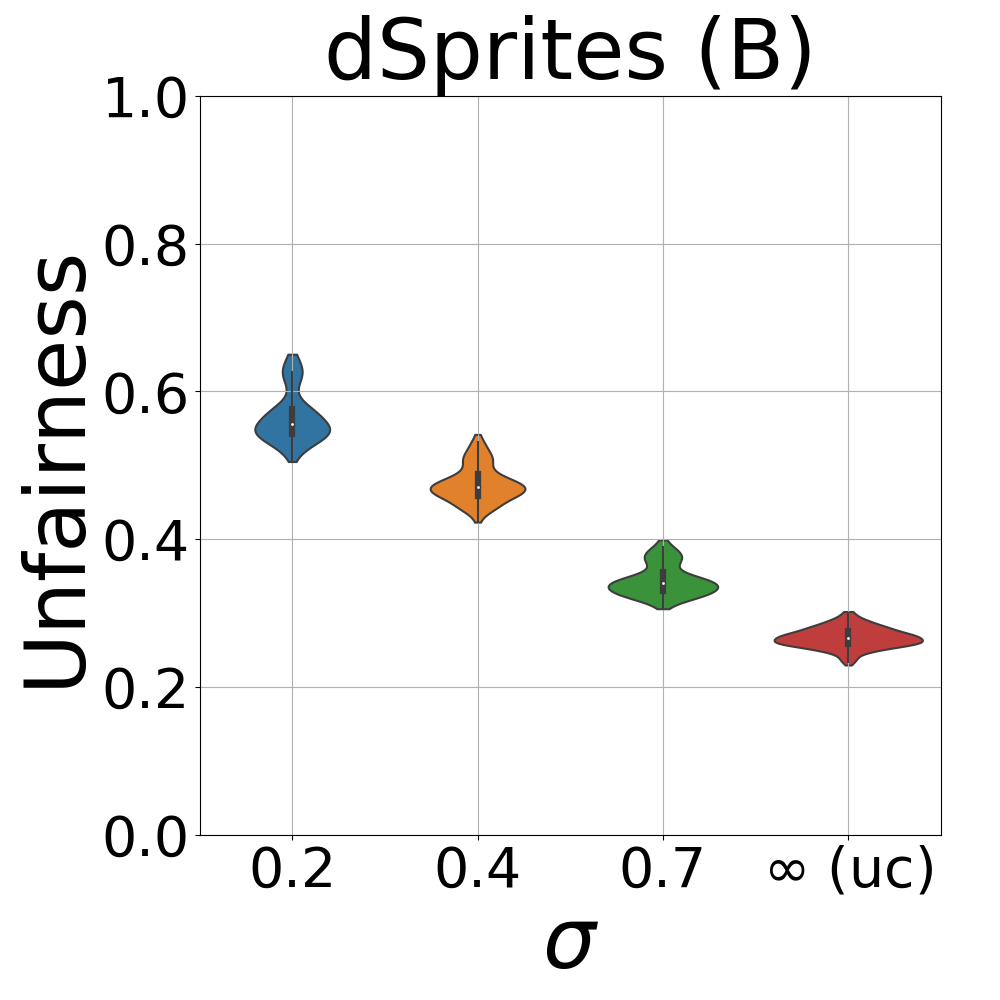}
\endminipage\hfill
\minipage{0.3\columnwidth}
  \includegraphics[width=\columnwidth]{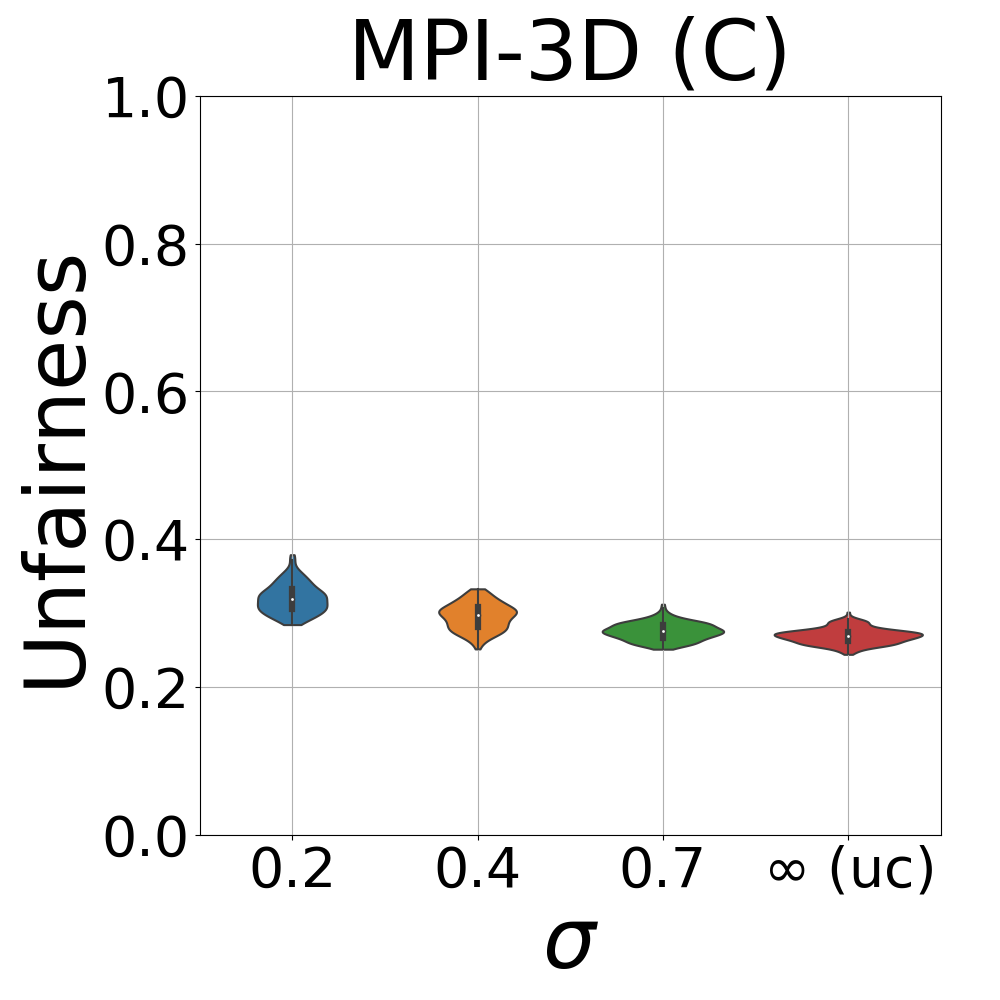}
\endminipage
\caption{Disentangled representations trained on correlated data are anti-correlated with higher fairness properties. The plots show the mean unfairness scores between the correlated factors with decreasing correlation strength for Shapes3D (A), dSprites (B) and MPI3D-real (C).}
\label{fig:unfairness_scores_wo_fa}
\end{figure}

To make our conclusion more sound we perform an empirical analysis of the pairwise entanglement metrics for the correlated pair vs. the median of all other pairs across the entire unsupervised study on all datasets and models trained. \cref{fig:pairwise_scores_thresholds_extended} shows the results of this analysis, aggregated across all disentanglement models. To avoid missing any particular disentanglement regularizer that might disentangle the correlated pair but is hidden among the combined aggregation, we also separately report the thresholds for each of the six disentanglement models in \cref{fig:pairwise_scores_thresholds_extended_per_regularizer}. We can clearly see that the correlated pair has a much higher entanglement than the rest of the pairs in the trained models across the full board, thus confirming our conclusion that inductive bias of current SOTA unsupervised disentanglement learners is insufficient.
Another pairwise metric that tracks the correlation strength in our scenario is the unfairness score between the correlated pair of factors that is being shown for datasets A, B and C in \cref{fig:unfairness_scores_wo_fa}.

\begin{figure}
\minipage{0.19\columnwidth}
  \includegraphics[width=\columnwidth]{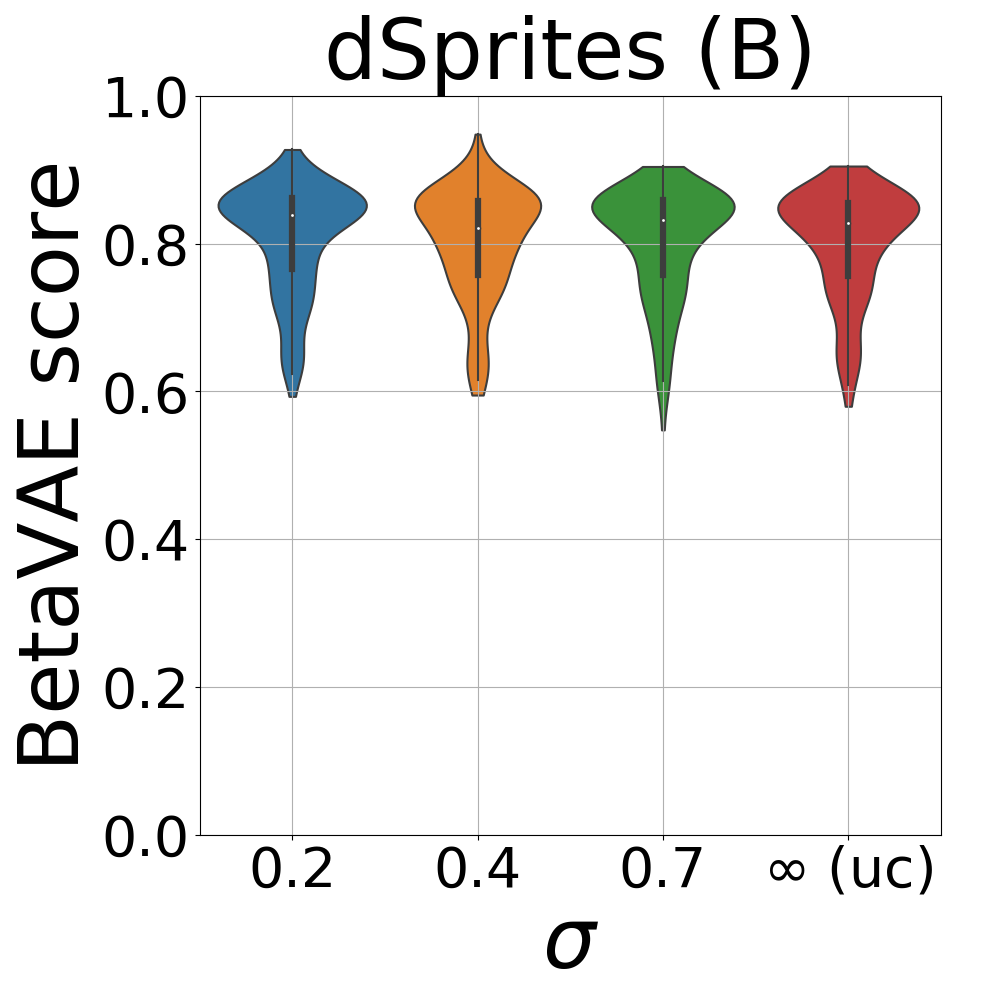}
\endminipage\hfill
\minipage{0.19\columnwidth}
  \includegraphics[width=\columnwidth]{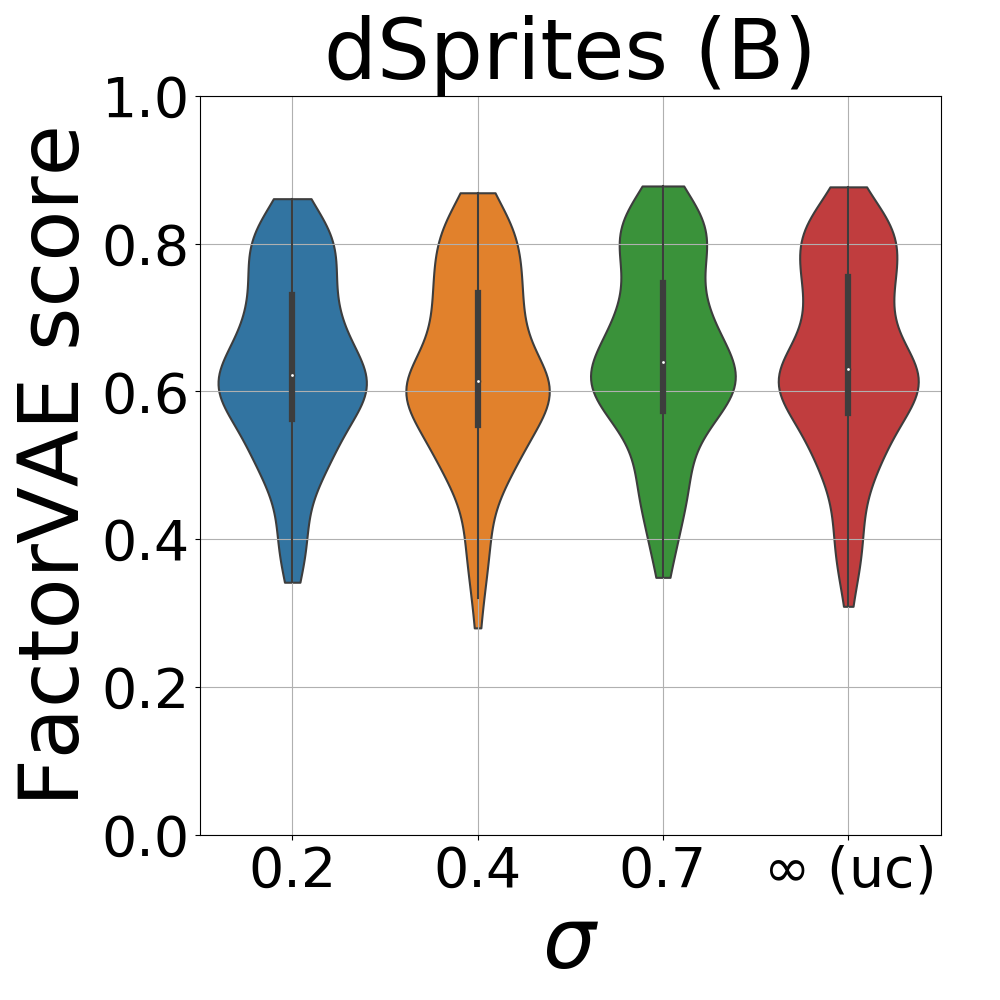}
\endminipage\hfill
\minipage{0.19\columnwidth}
  \includegraphics[width=\columnwidth]{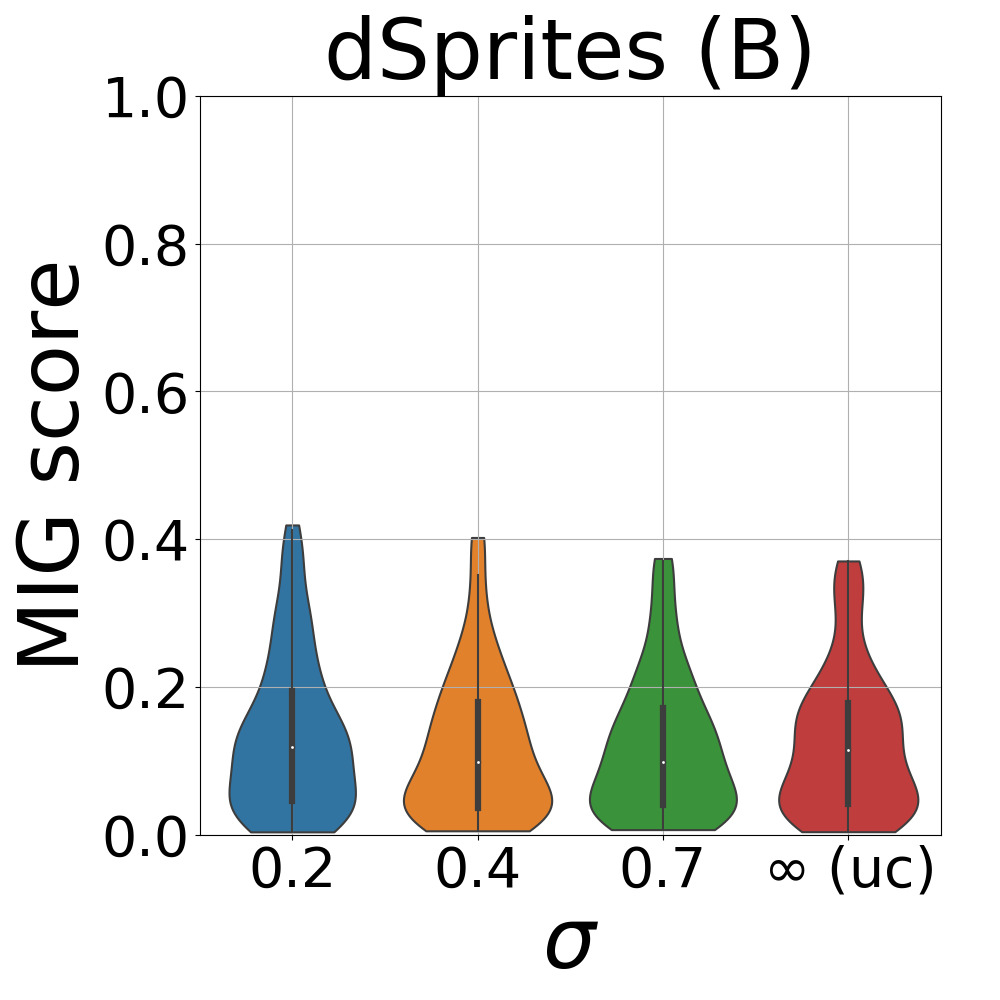}
\endminipage\hfill
\minipage{0.19\columnwidth}
  \includegraphics[width=\columnwidth]{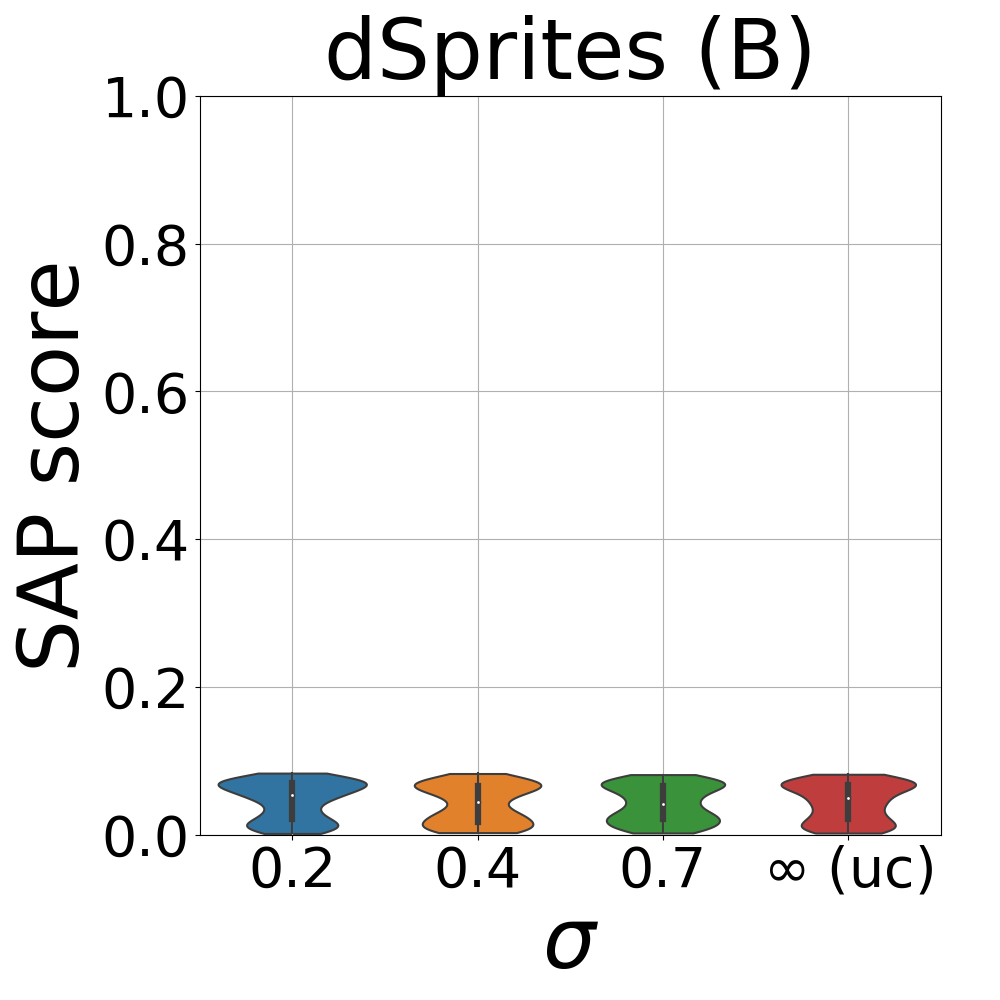}
\endminipage\hfill
\minipage{0.19\columnwidth}
  \includegraphics[width=\columnwidth]{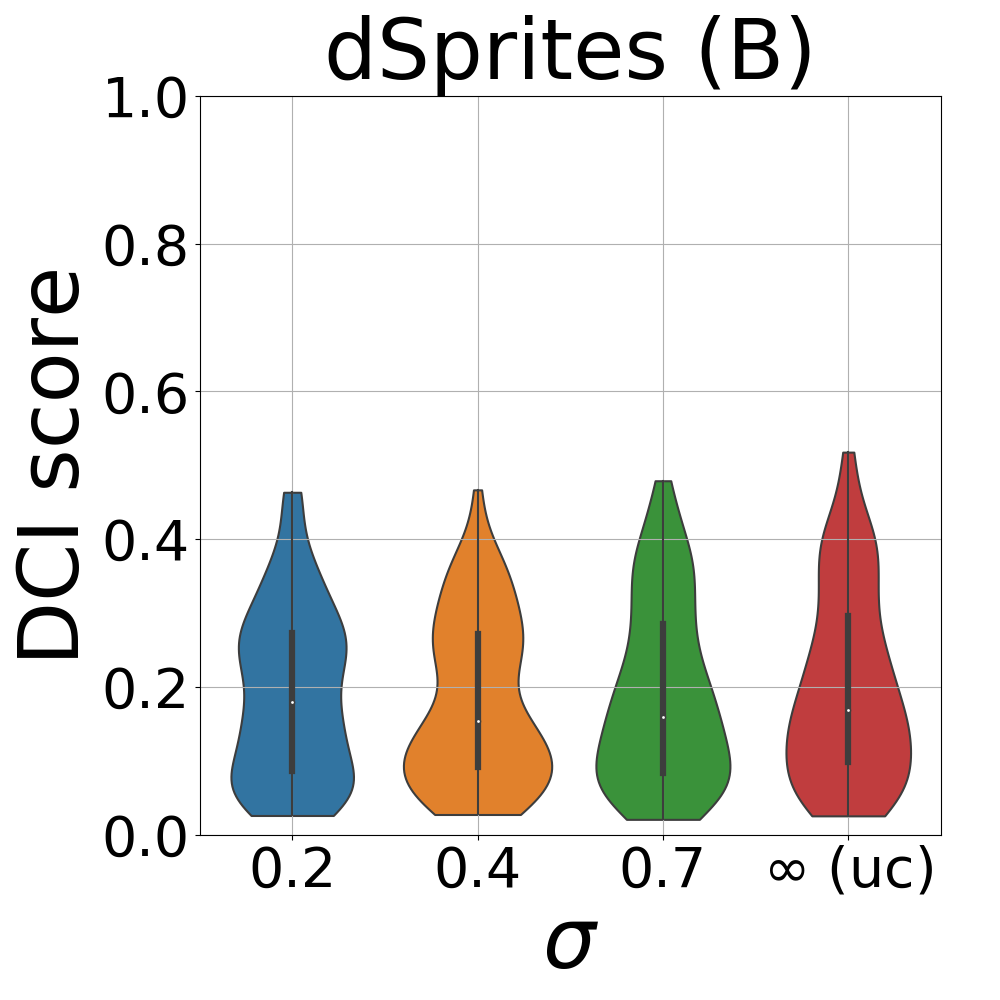}
\endminipage
\vskip 0.1in
\minipage{0.19\columnwidth}
  \includegraphics[width=\columnwidth]{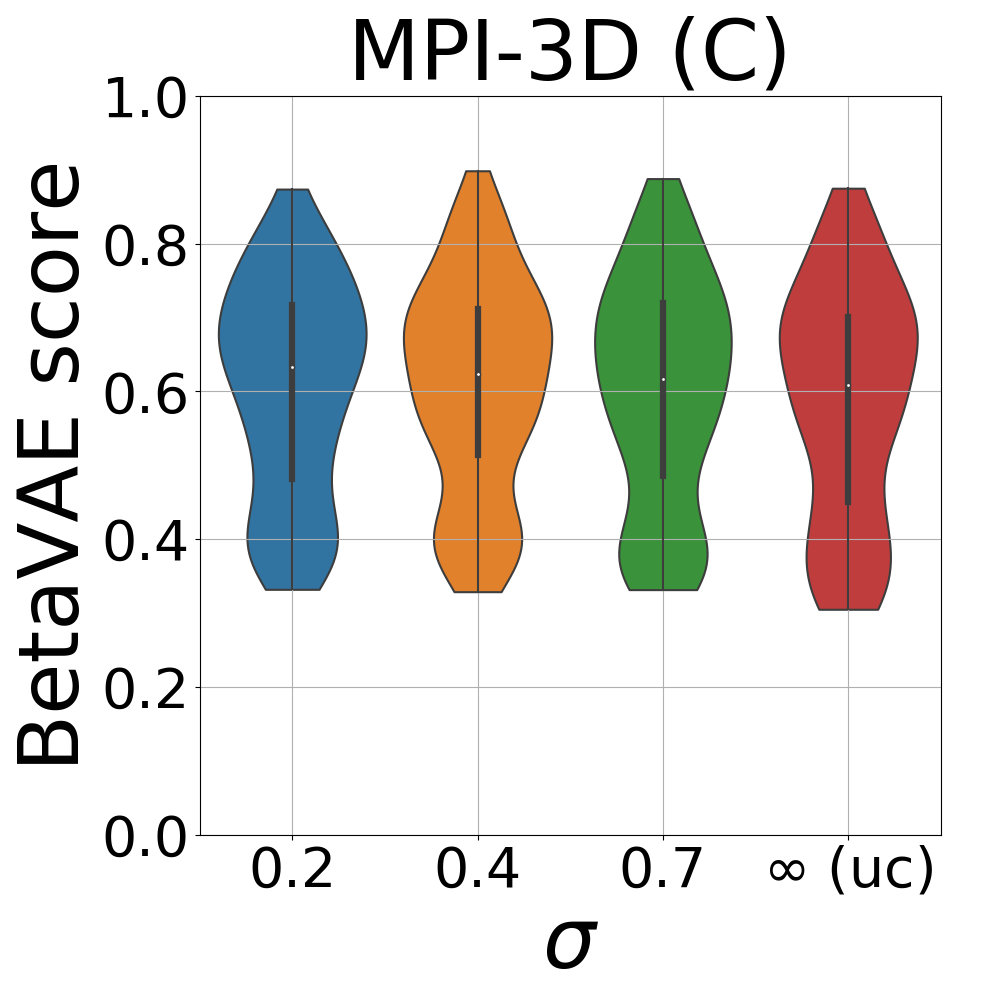}
\endminipage\hfill
\minipage{0.19\columnwidth}
  \includegraphics[width=\columnwidth]{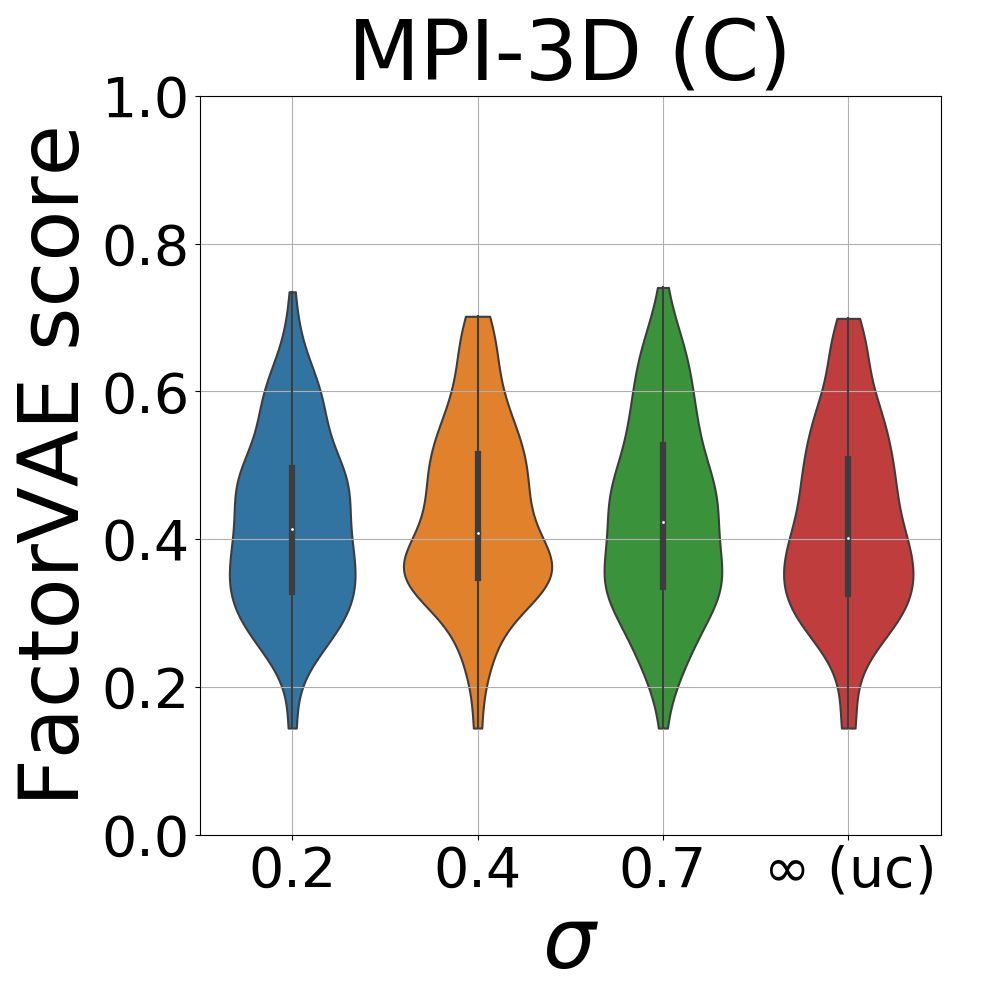}
\endminipage\hfill
\minipage{0.19\columnwidth}
  \includegraphics[width=\columnwidth]{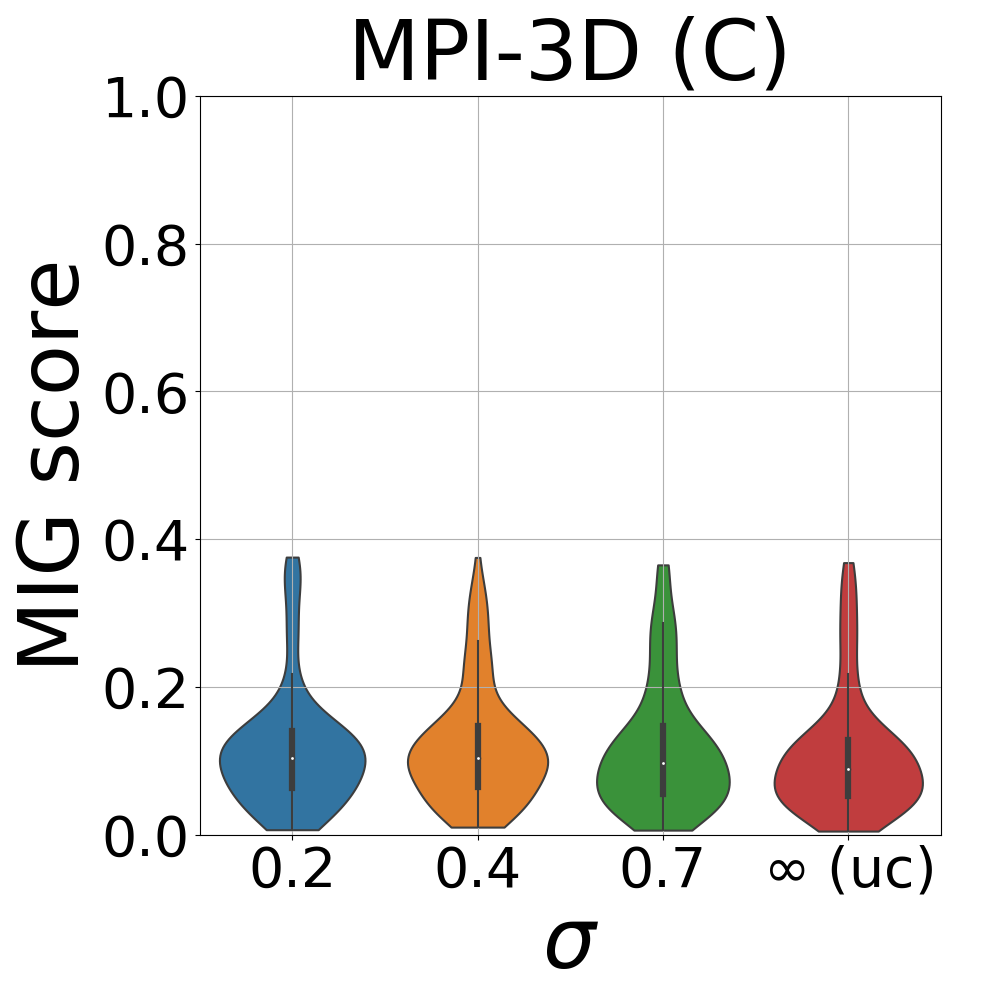}
\endminipage\hfill
\minipage{0.19\columnwidth}
  \includegraphics[width=\columnwidth]{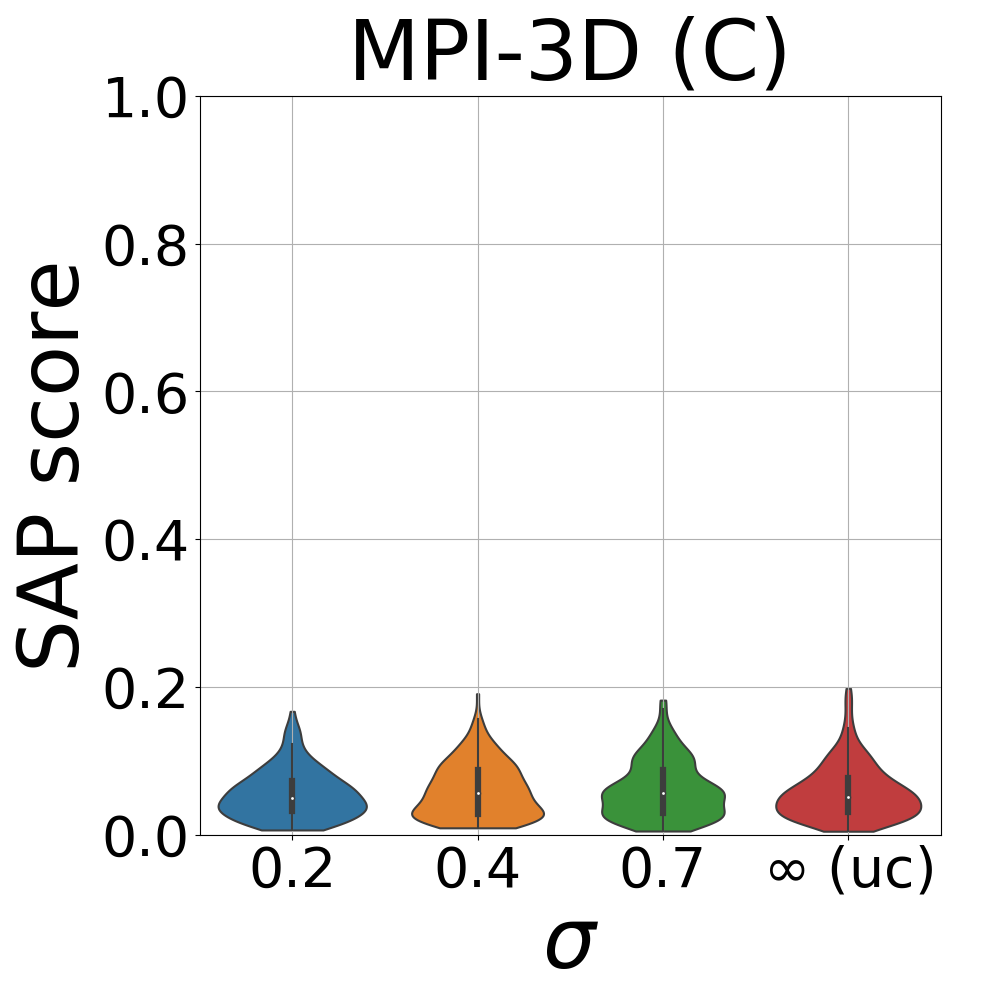}
\endminipage\hfill
\minipage{0.19\columnwidth}
  \includegraphics[width=\columnwidth]{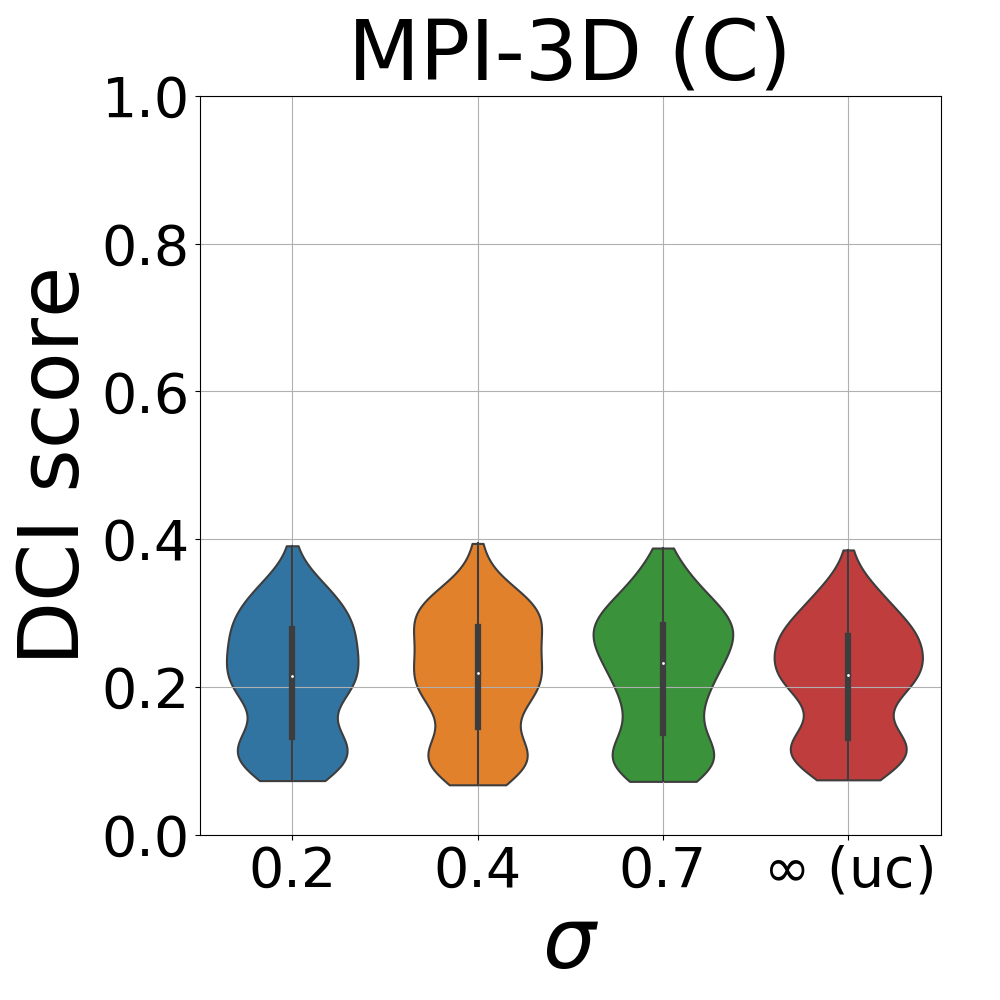}
\endminipage
\vskip 0.1in
\minipage{0.19\columnwidth}
  \includegraphics[width=\columnwidth]{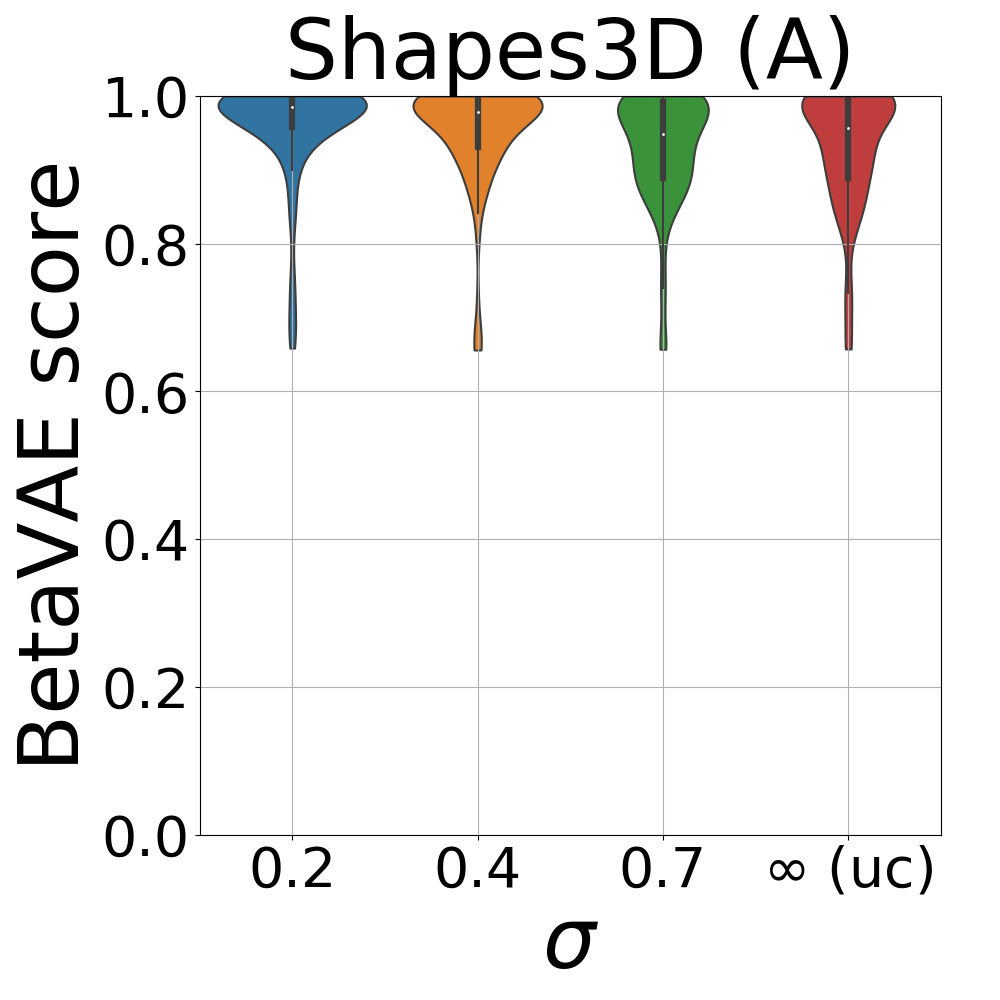}
\endminipage\hfill
\minipage{0.19\columnwidth}
  \includegraphics[width=\columnwidth]{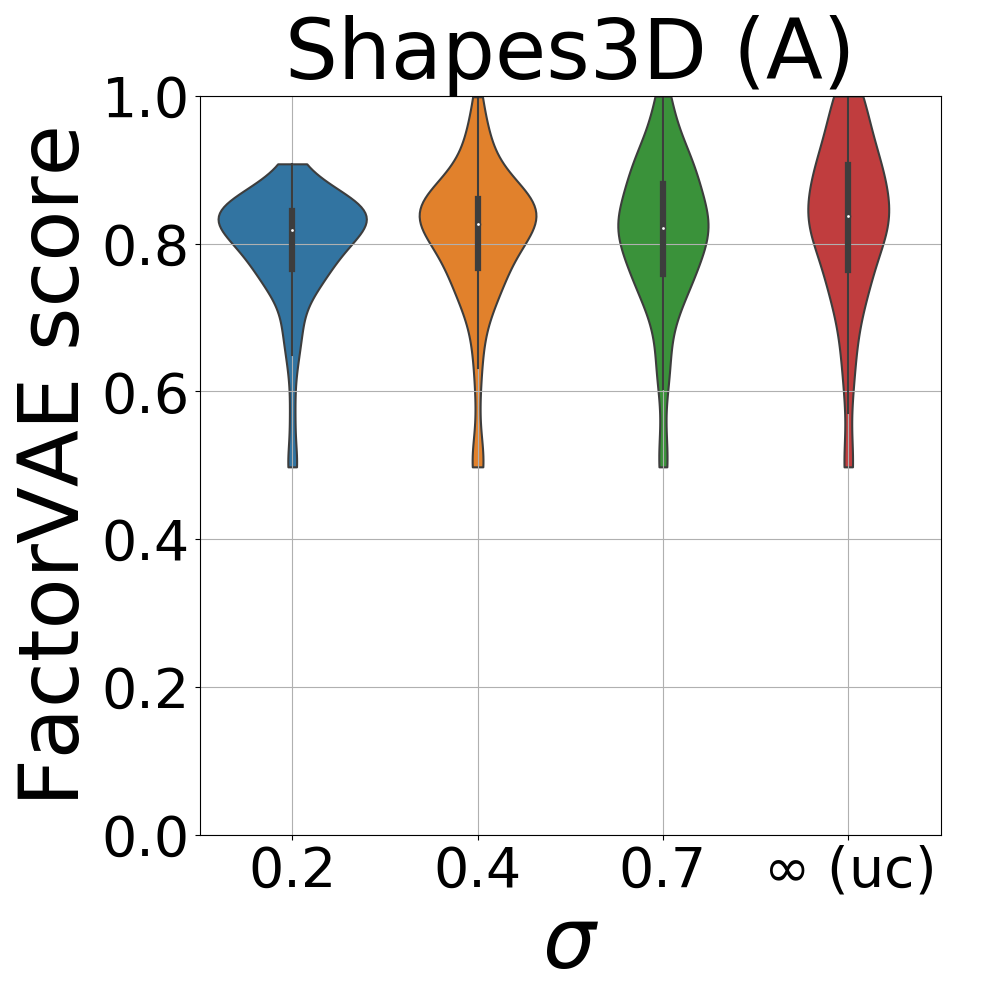}
\endminipage\hfill
\minipage{0.19\columnwidth}
  \includegraphics[width=\columnwidth]{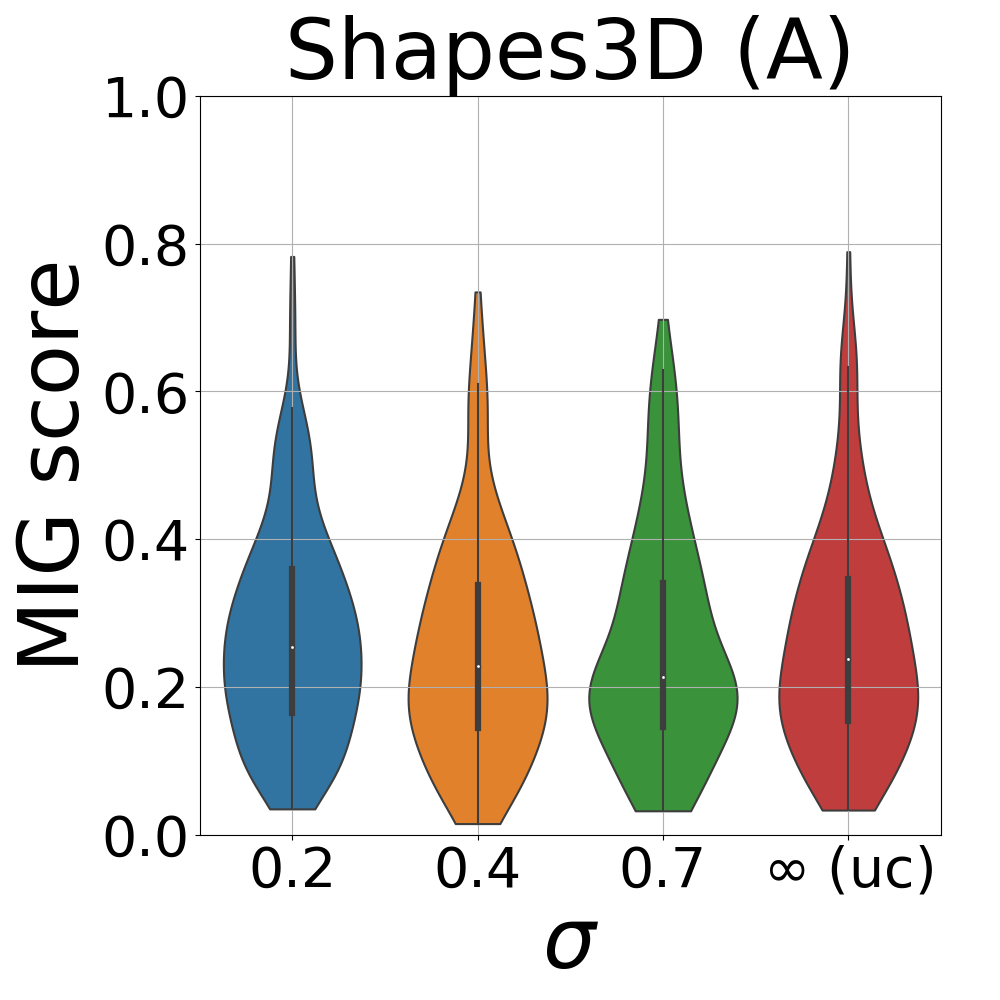}
\endminipage\hfill
\minipage{0.19\columnwidth}
  \includegraphics[width=\columnwidth]{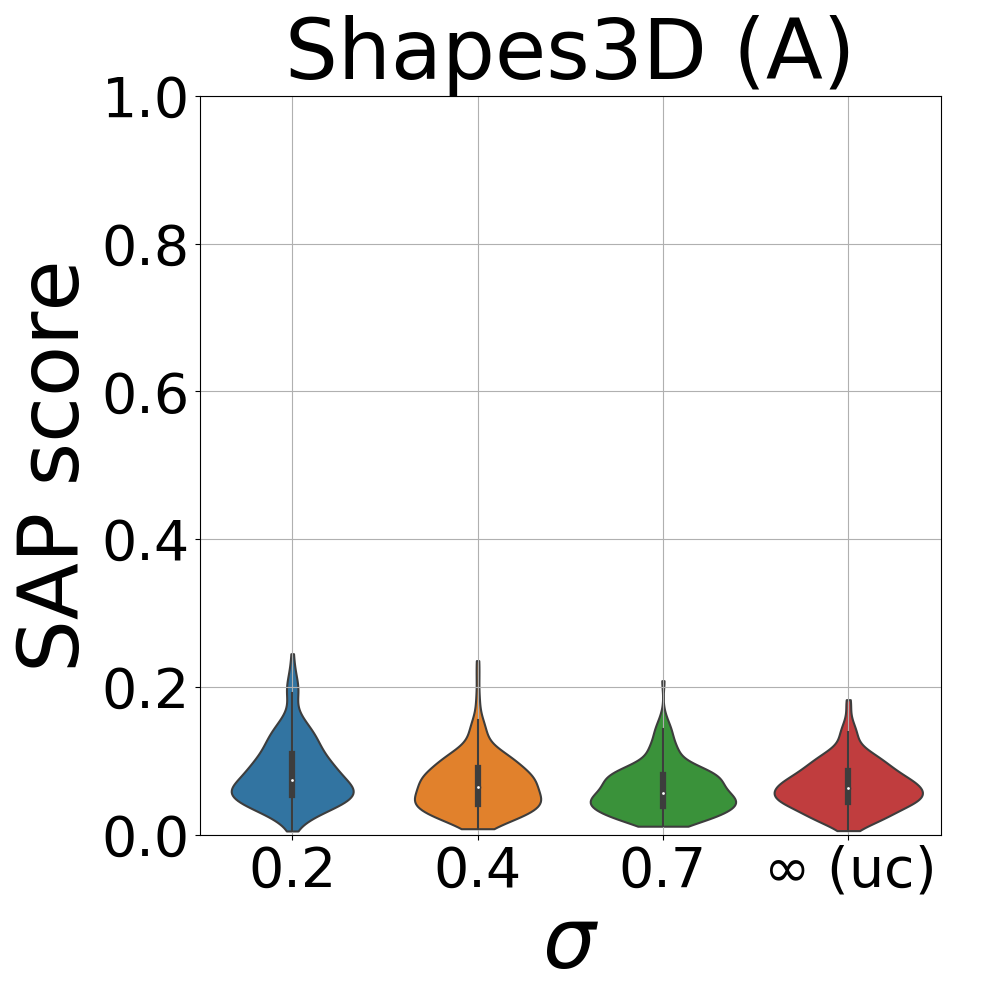}
\endminipage\hfill
\minipage{0.19\columnwidth}
  \includegraphics[width=\columnwidth]{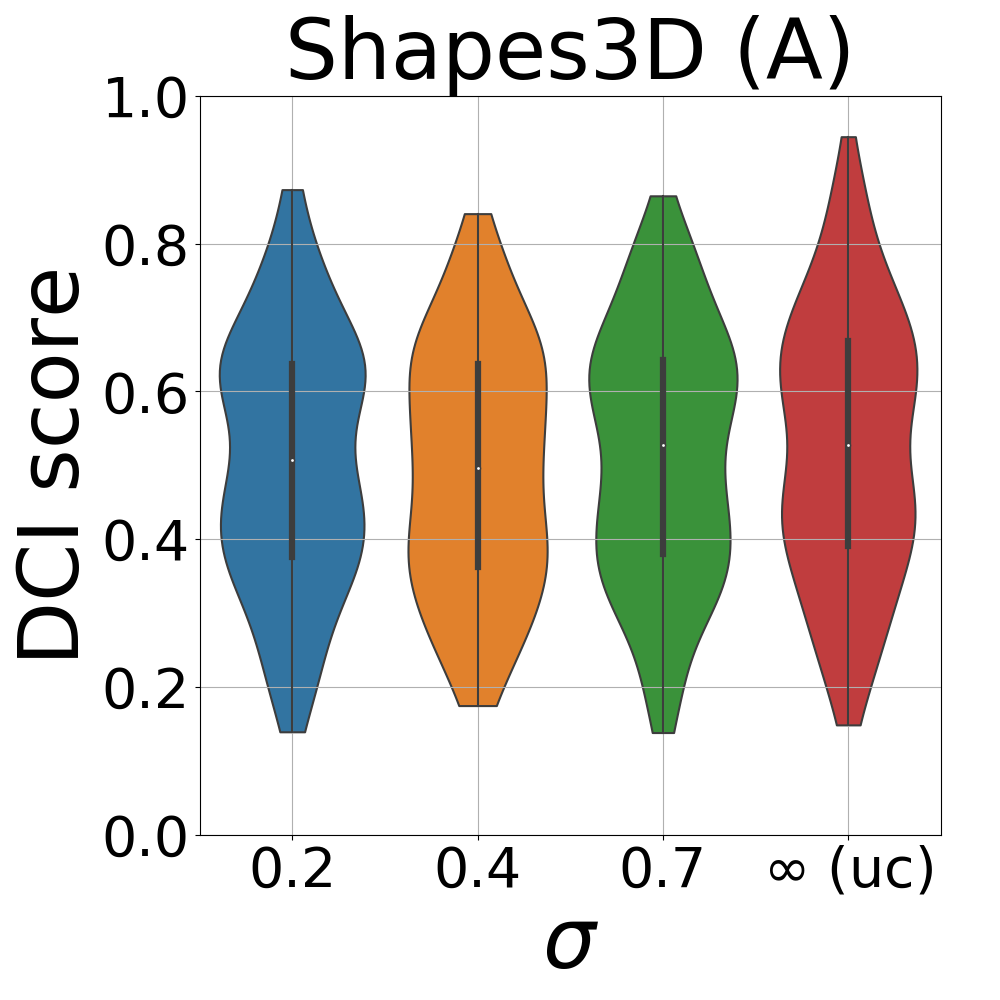}
\endminipage
\caption{Standard global disentanglement metrics evaluated on the correlated training set showing no clear trend for different correlation strengths.}
\label{fig:disentanglement_metrics_unsupervised_appendix}
\end{figure}

\begin{figure}
\minipage{0.19\columnwidth}
  \includegraphics[width=\columnwidth]{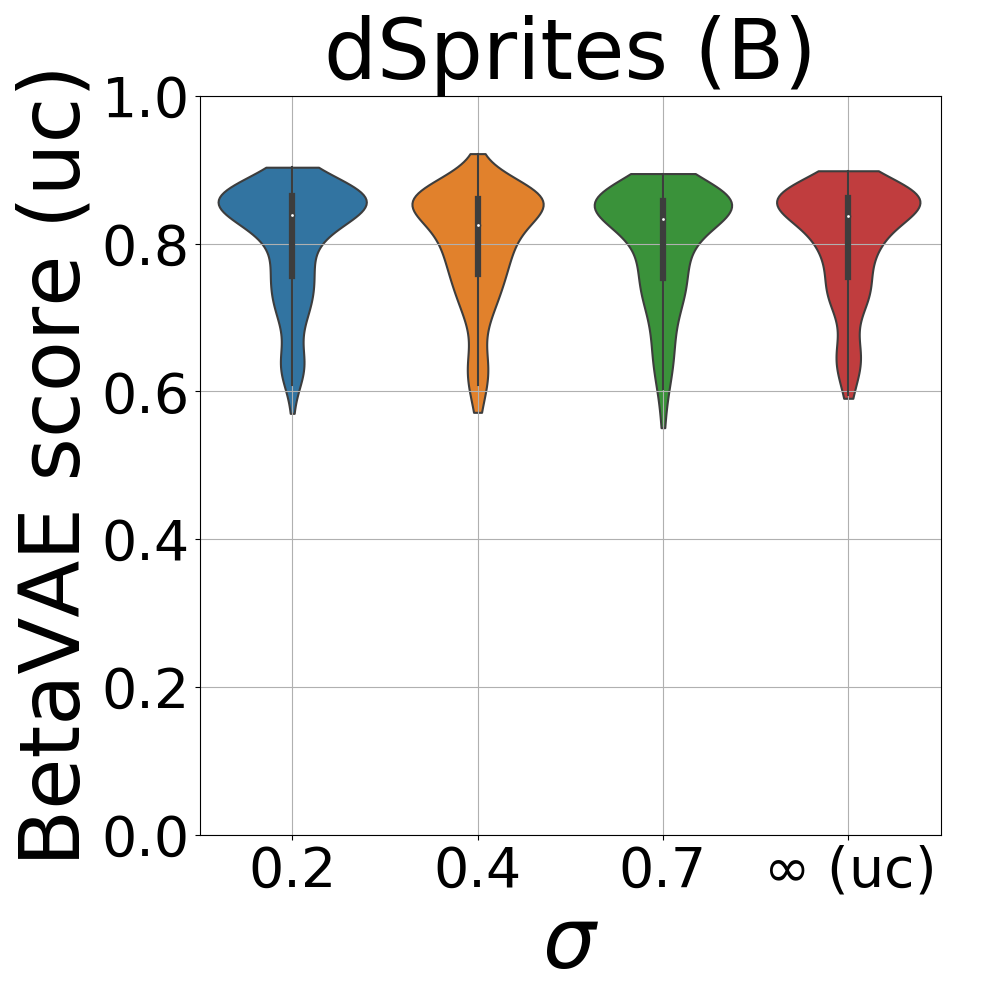}
\endminipage\hfill
\minipage{0.19\columnwidth}
  \includegraphics[width=\columnwidth]{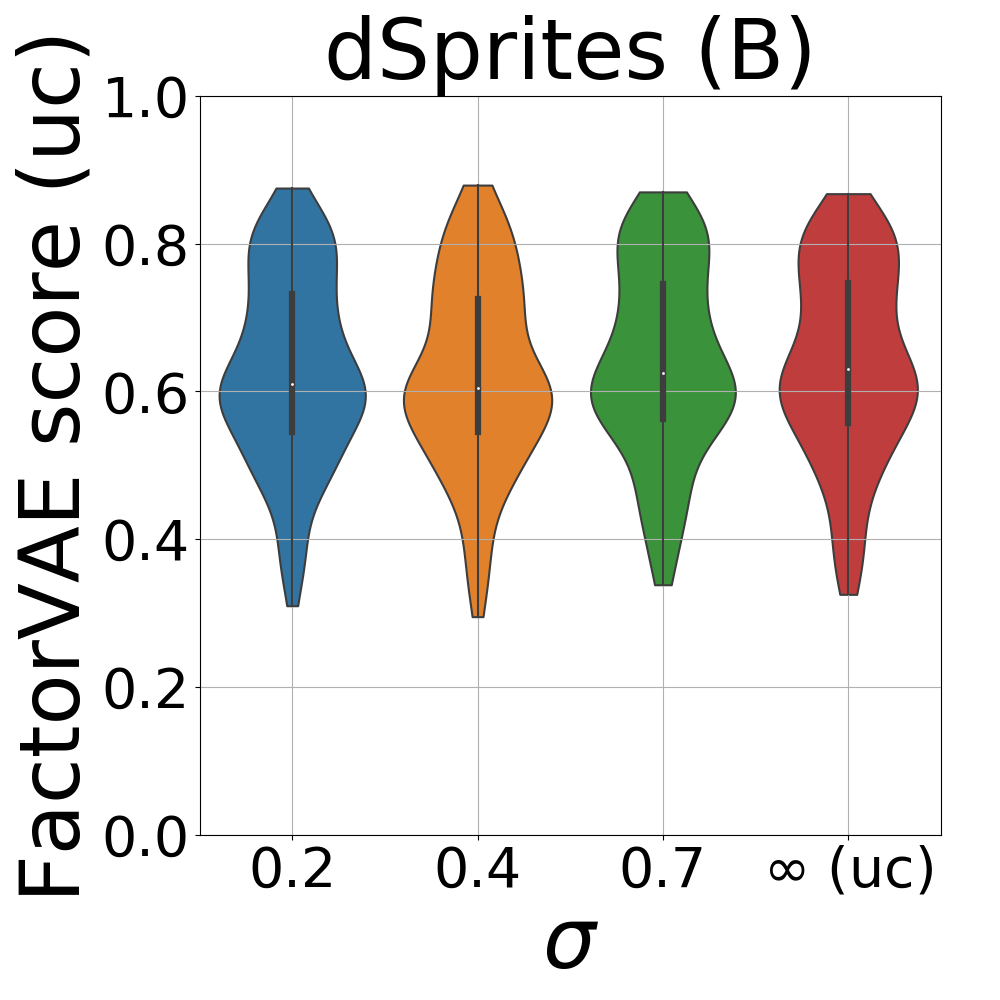}
\endminipage\hfill
\minipage{0.19\columnwidth}
  \includegraphics[width=\columnwidth]{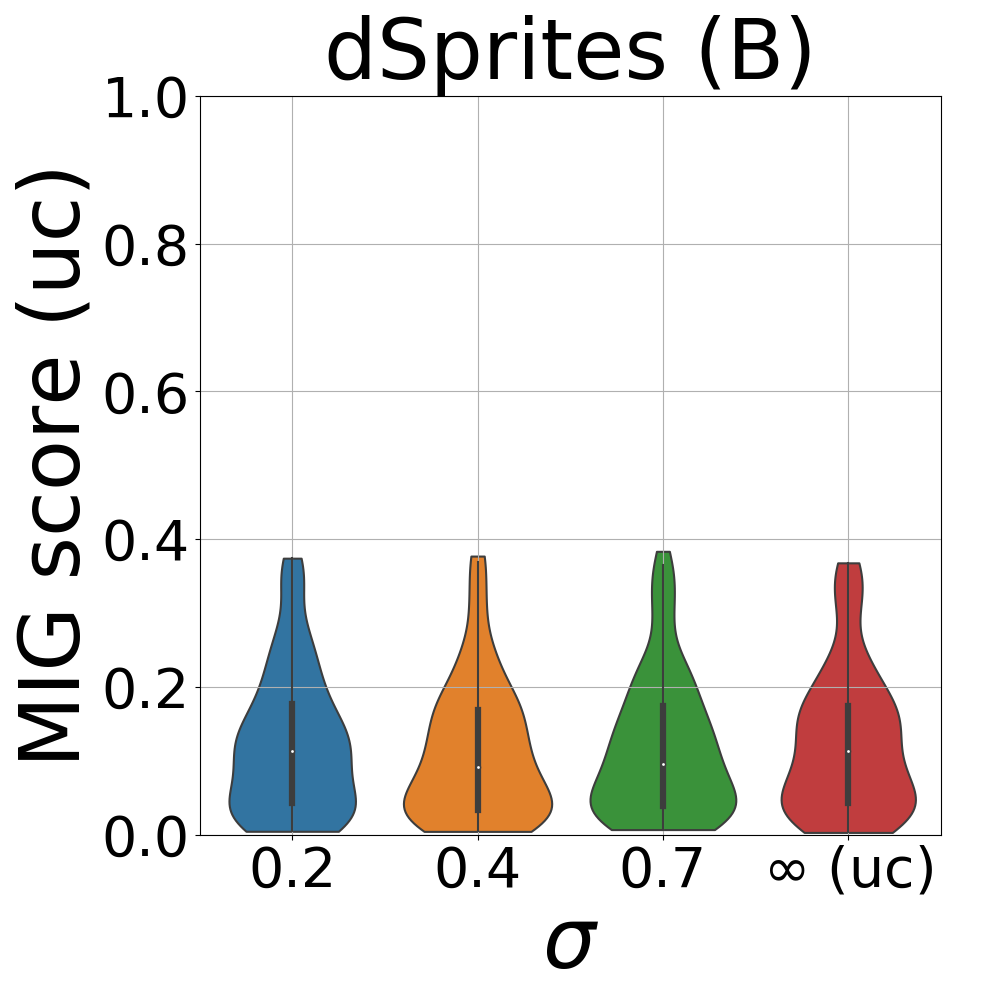}
\endminipage\hfill
\minipage{0.19\columnwidth}
  \includegraphics[width=\columnwidth]{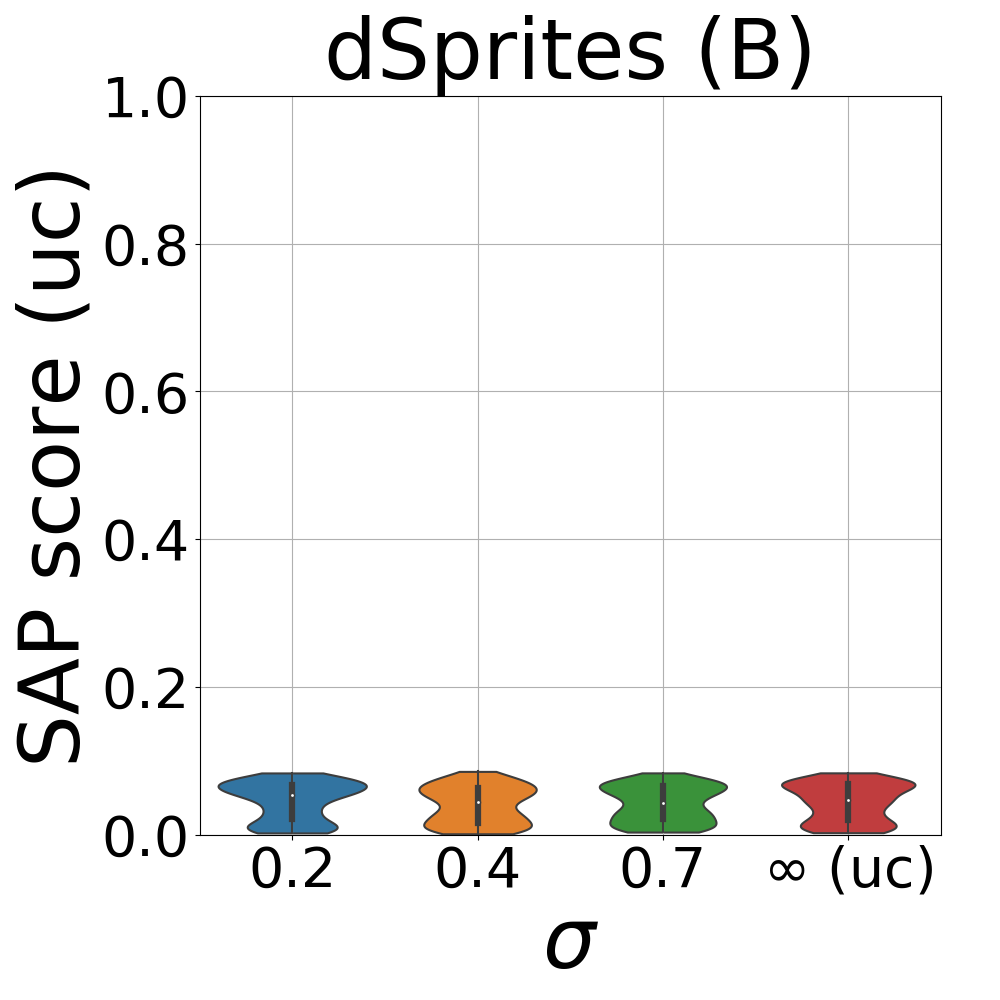}
\endminipage\hfill
\minipage{0.19\columnwidth}
  \includegraphics[width=\columnwidth]{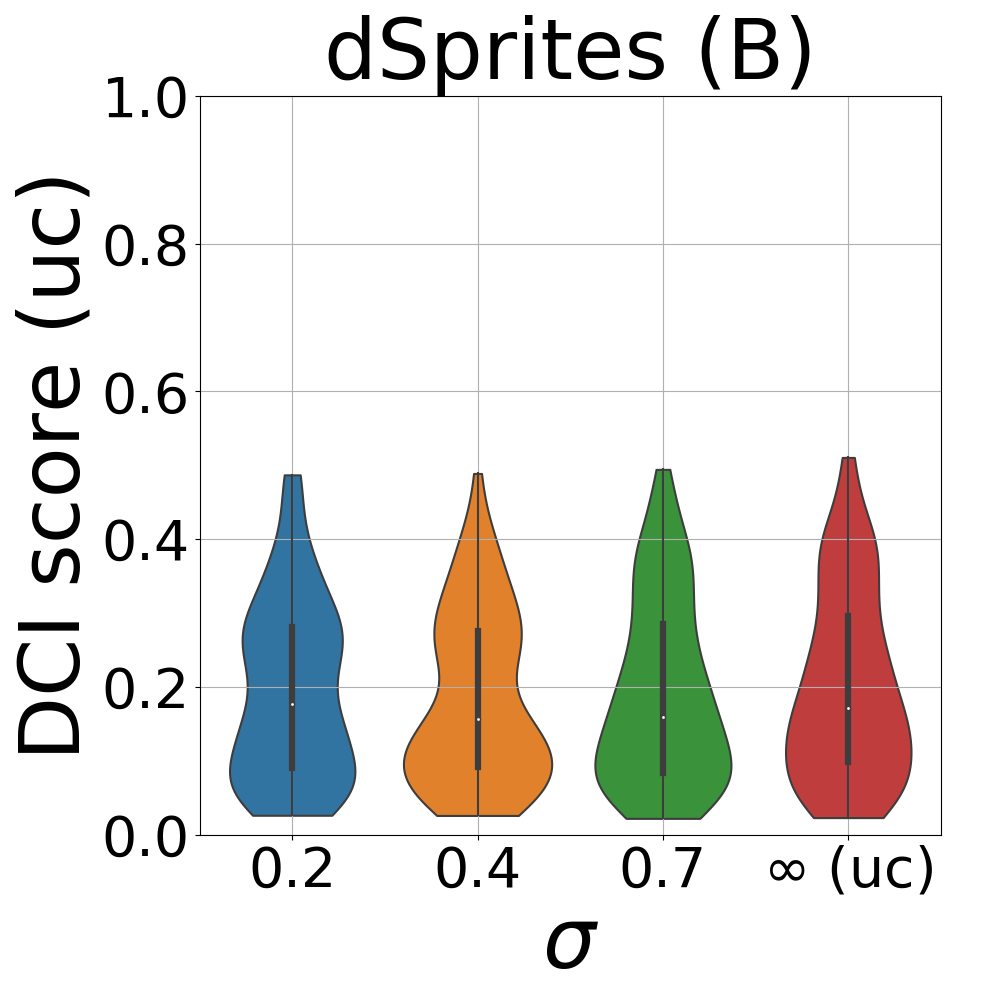}
\endminipage
\vskip 0.1in

\minipage{0.19\columnwidth}
  \includegraphics[width=\columnwidth]{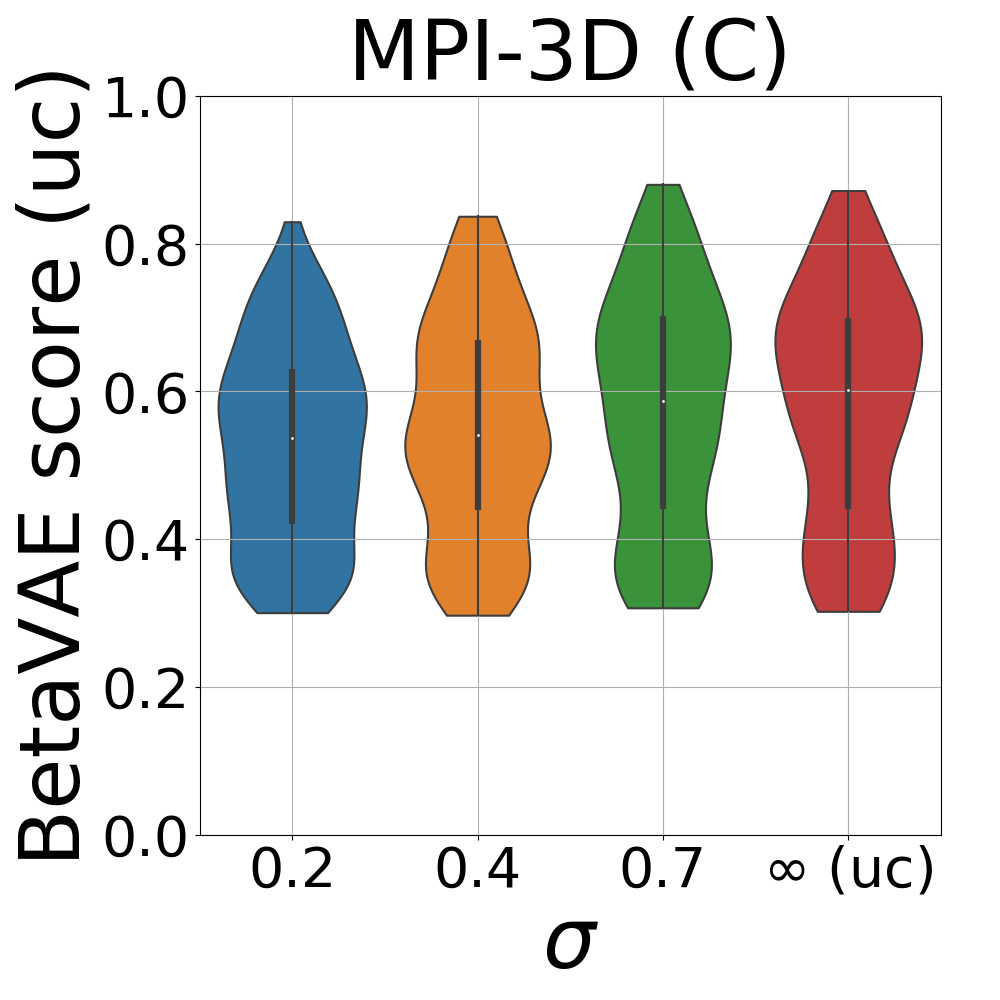}
\endminipage\hfill
\minipage{0.19\columnwidth}
  \includegraphics[width=\columnwidth]{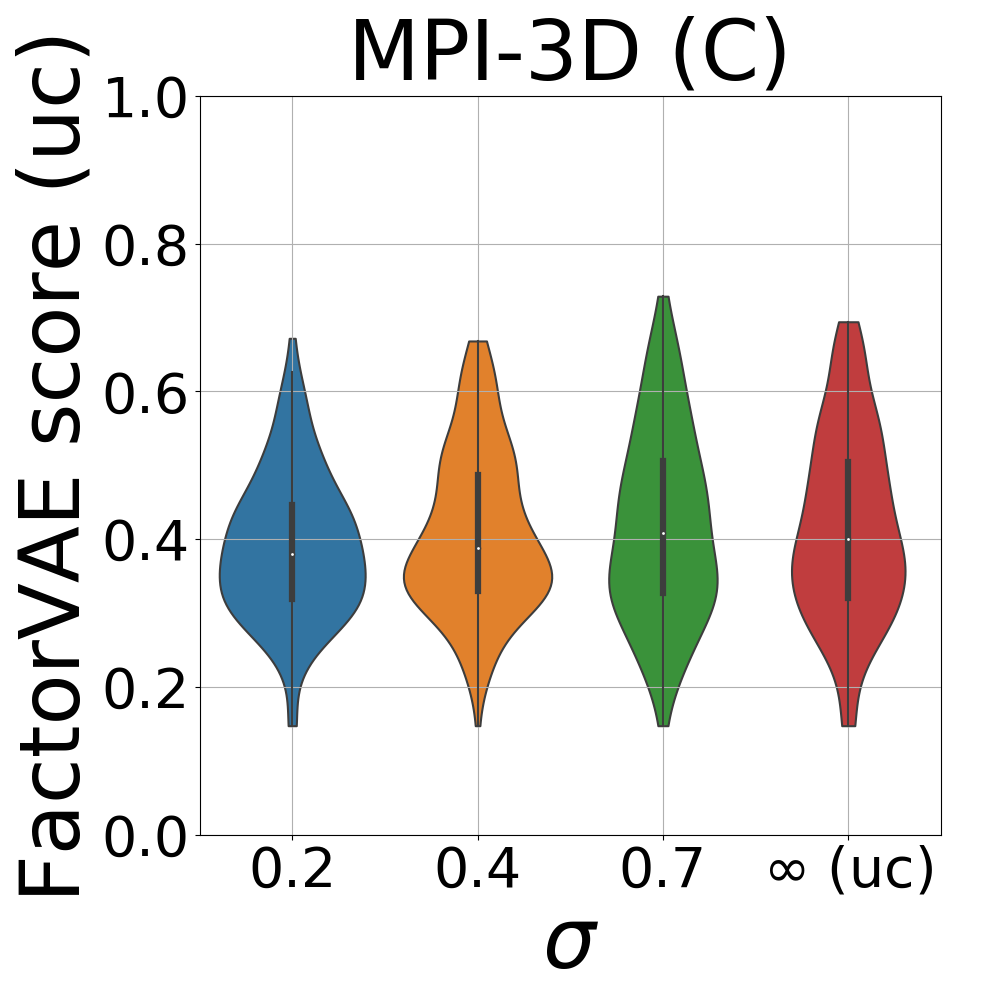}
\endminipage\hfill
\minipage{0.19\columnwidth}
  \includegraphics[width=\columnwidth]{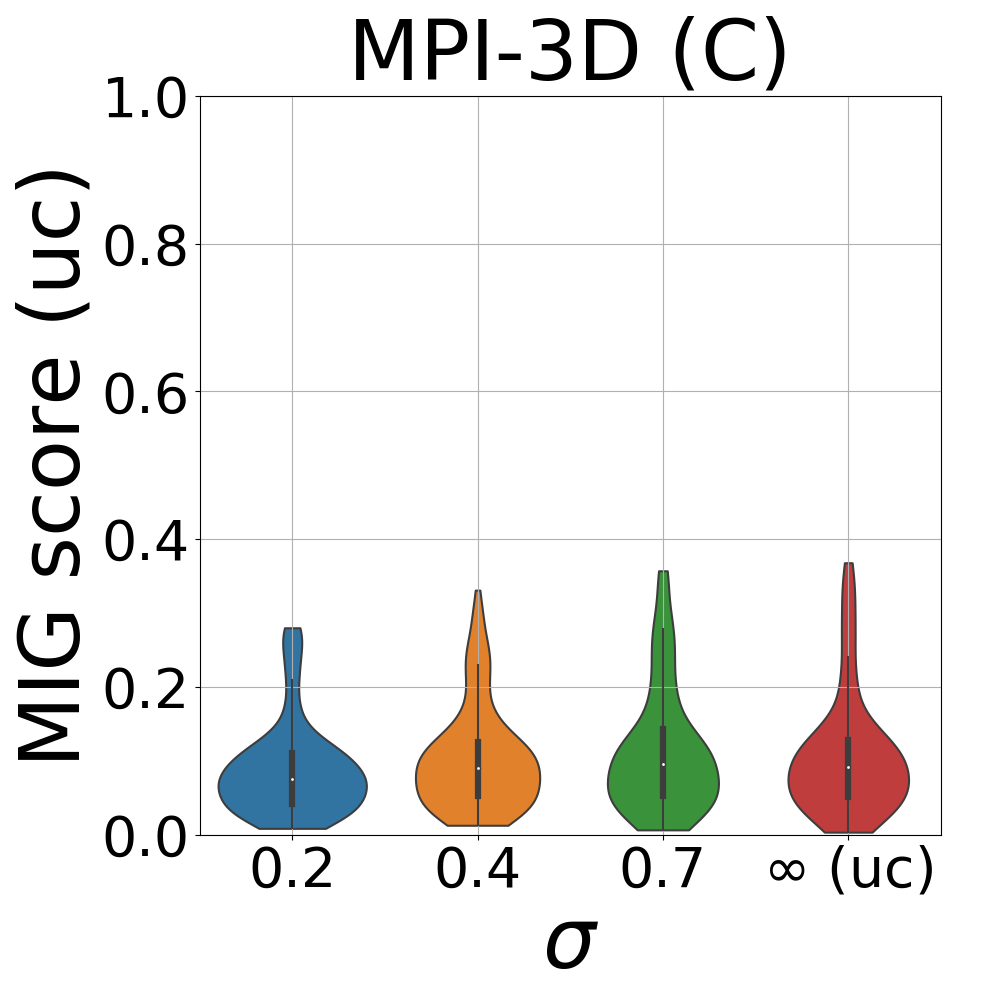}
\endminipage\hfill
\minipage{0.19\columnwidth}
  \includegraphics[width=\columnwidth]{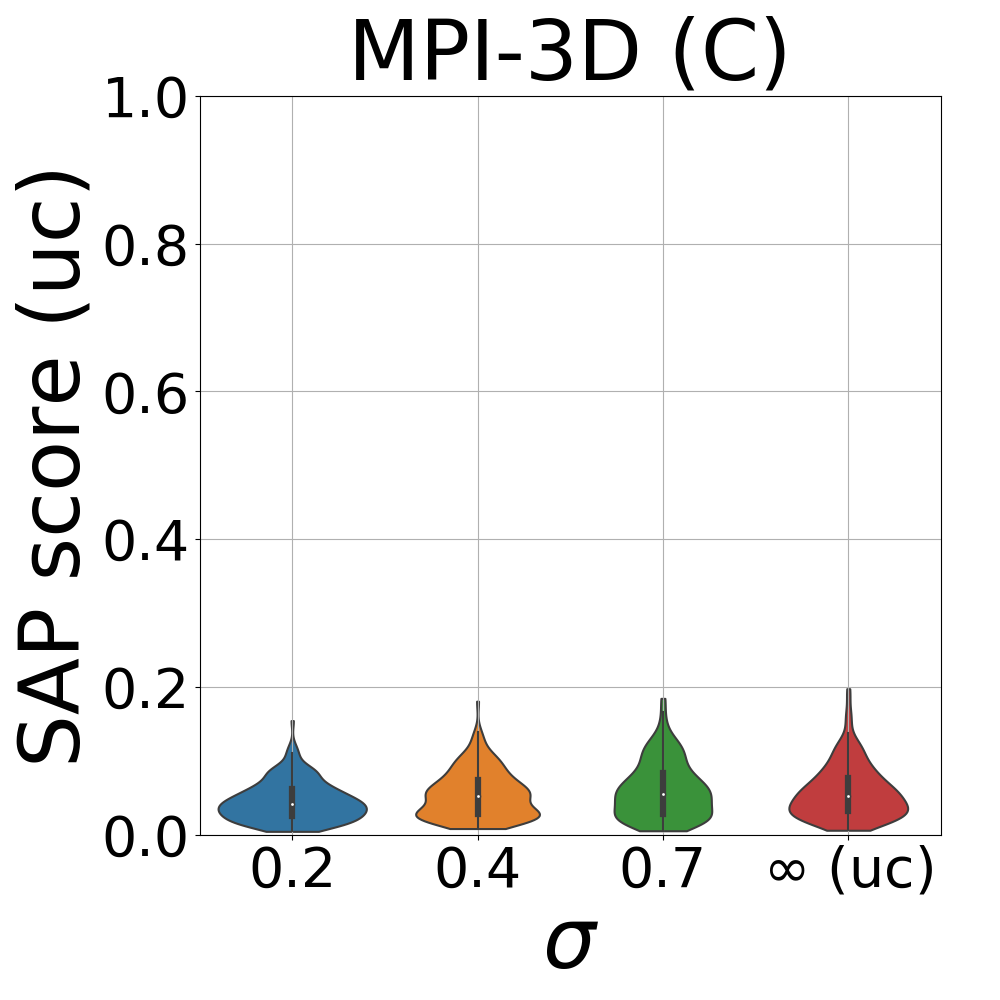}
\endminipage\hfill
\minipage{0.19\columnwidth}
  \includegraphics[width=\columnwidth]{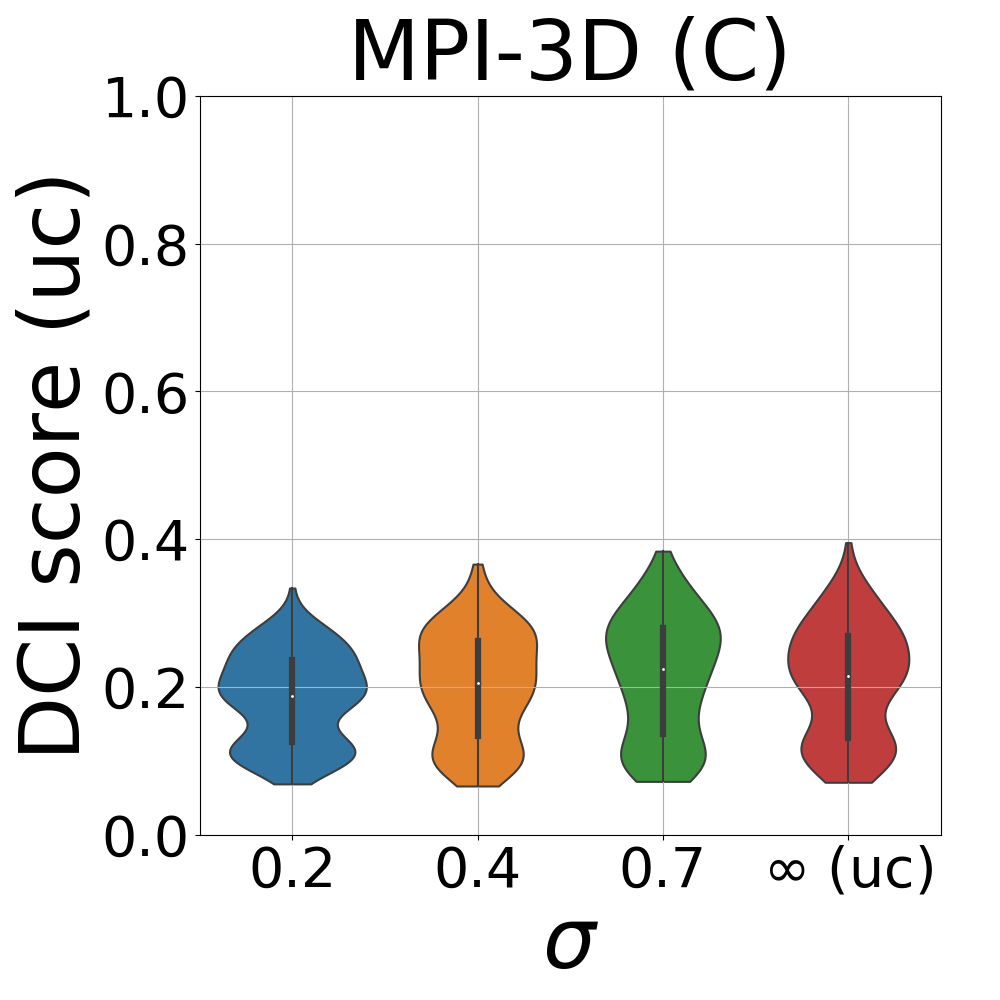}
\endminipage
\vskip 0.1in

\minipage{0.19\columnwidth}
  \includegraphics[width=\columnwidth]{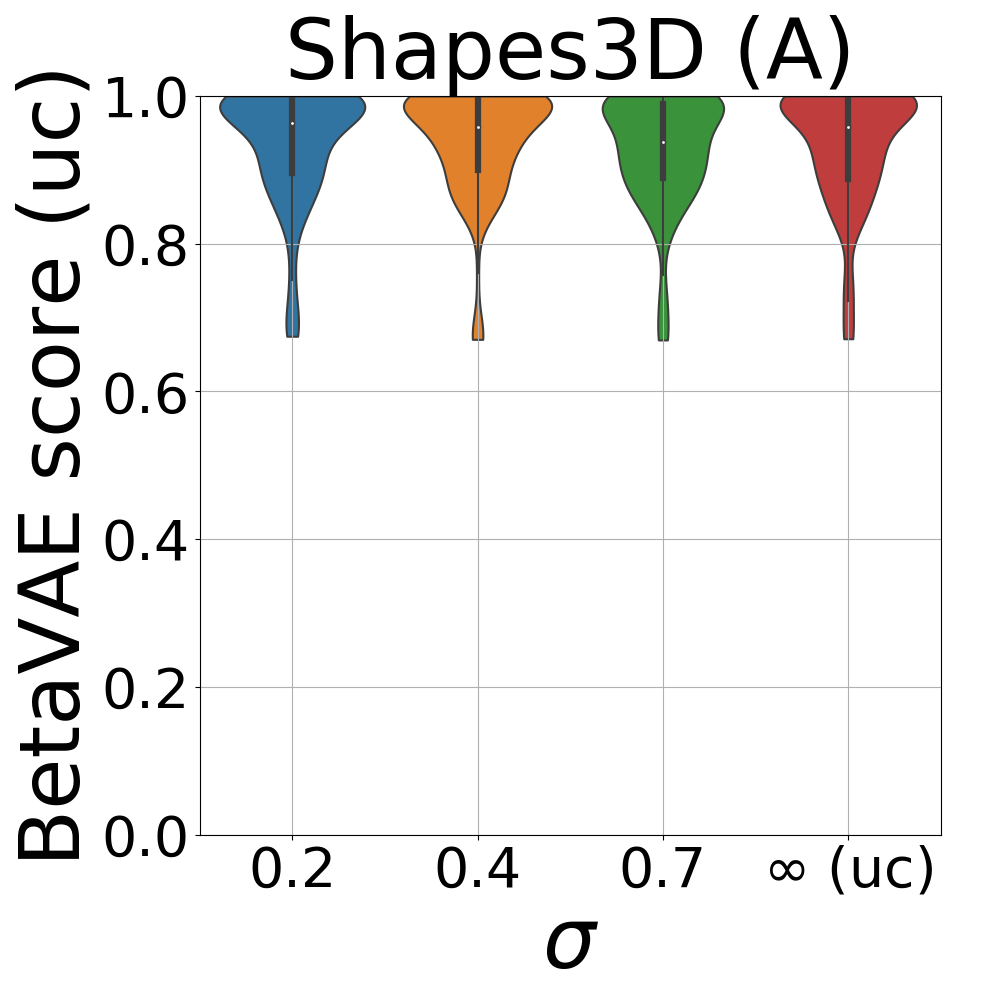}
\endminipage\hfill
\minipage{0.19\columnwidth}
  \includegraphics[width=\columnwidth]{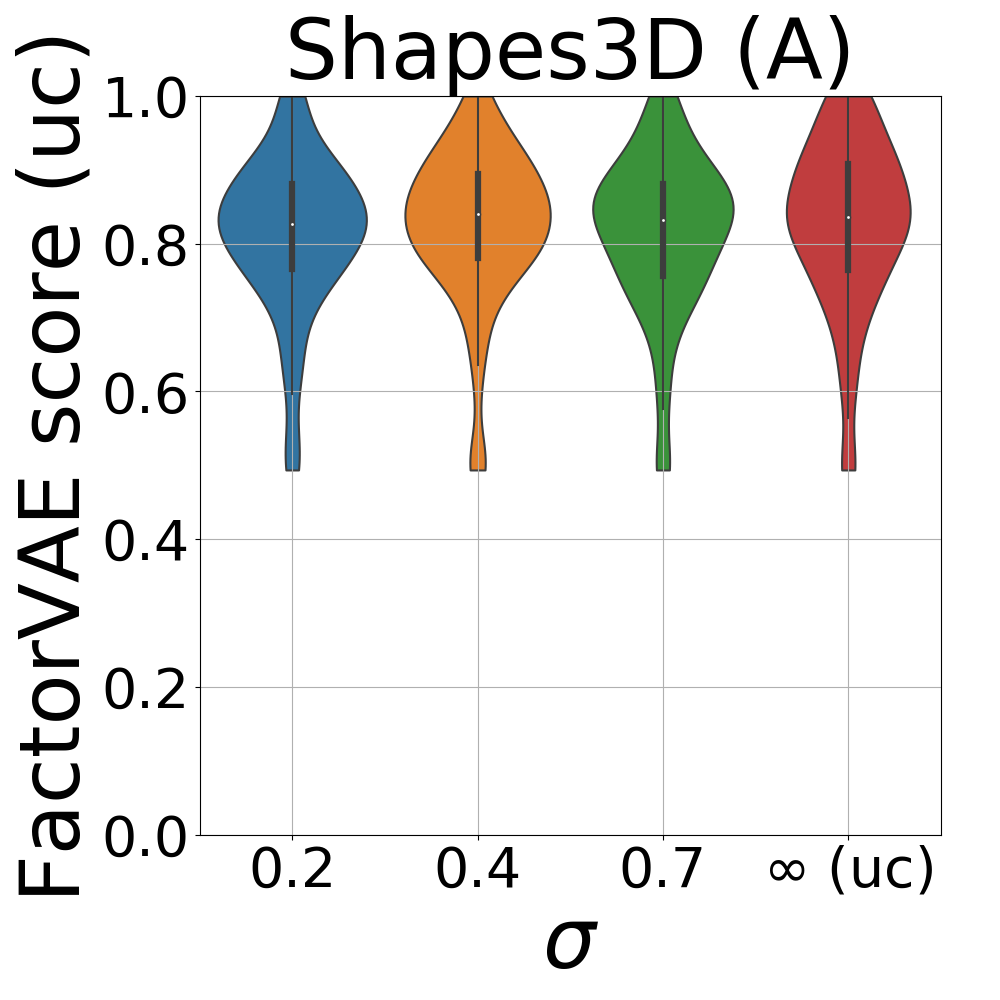}
\endminipage\hfill
\minipage{0.19\columnwidth}
  \includegraphics[width=\columnwidth]{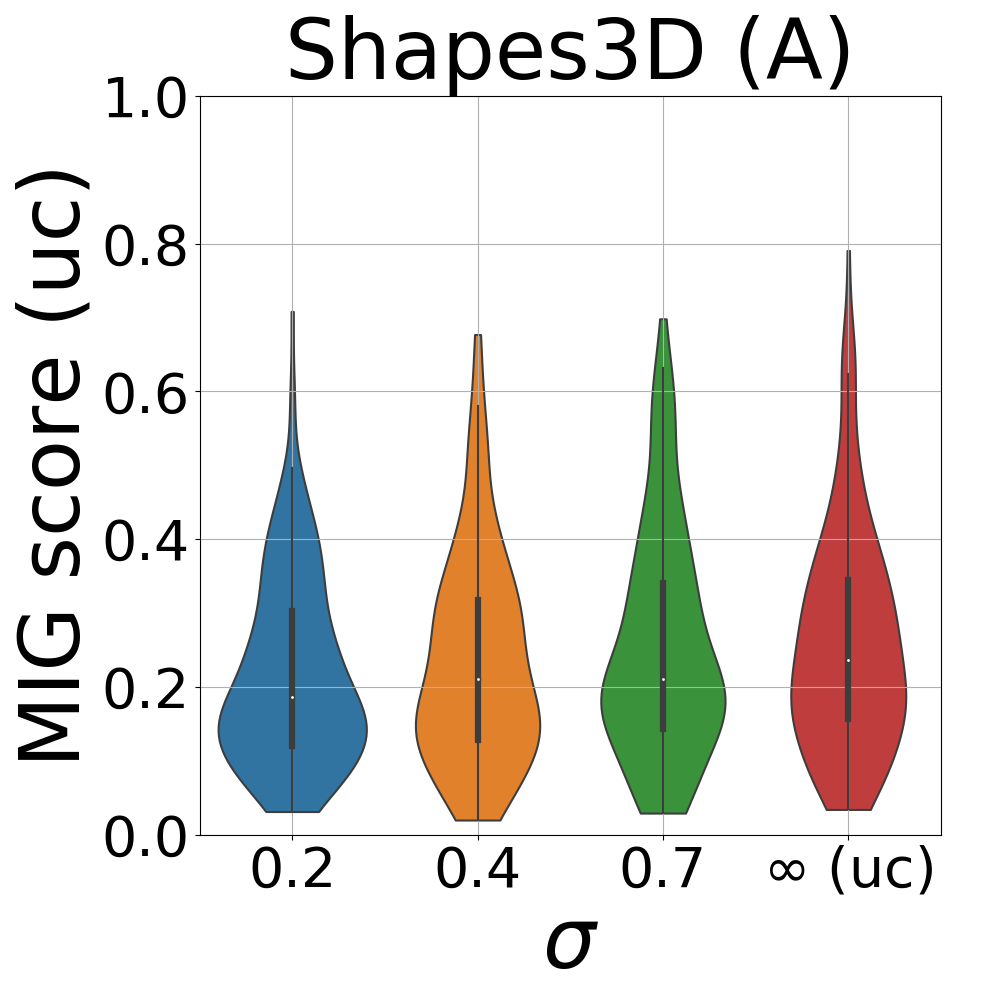}
\endminipage\hfill
\minipage{0.19\columnwidth}
  \includegraphics[width=\columnwidth]{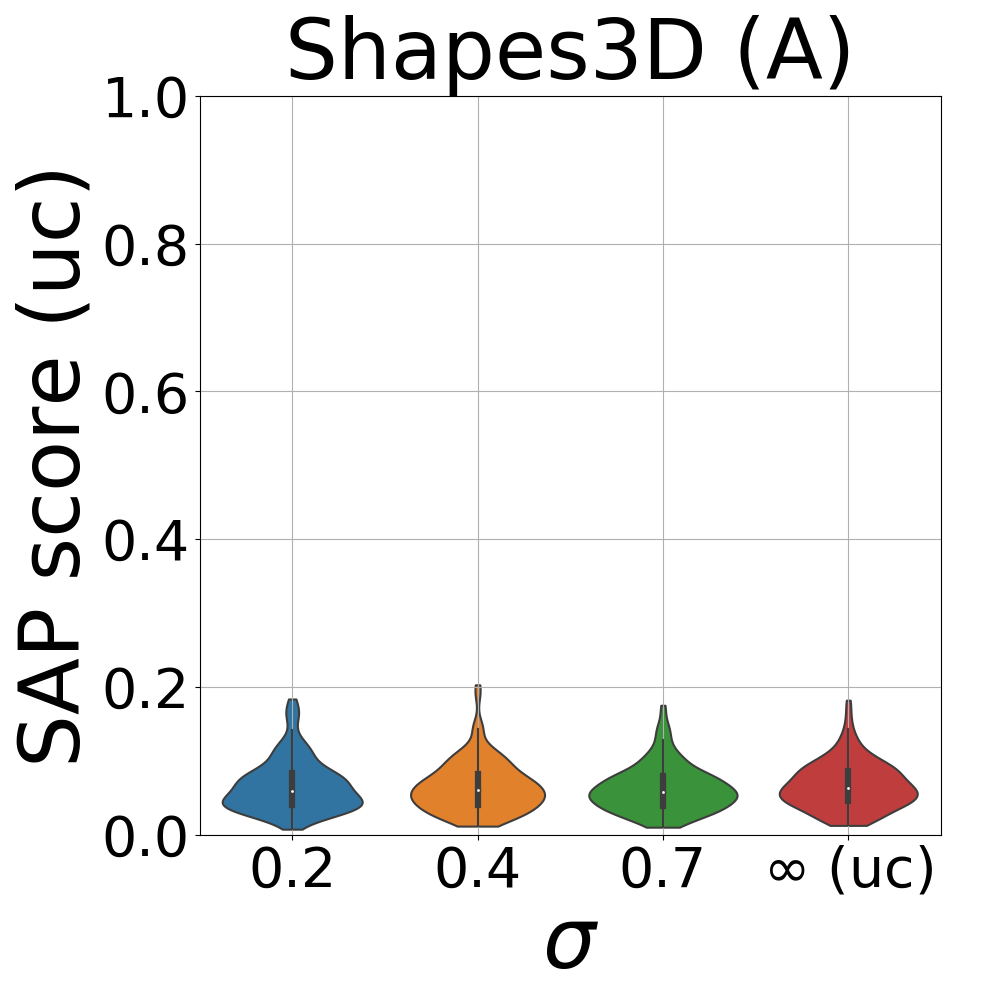}
\endminipage\hfill
\minipage{0.19\columnwidth}
  \includegraphics[width=\columnwidth]{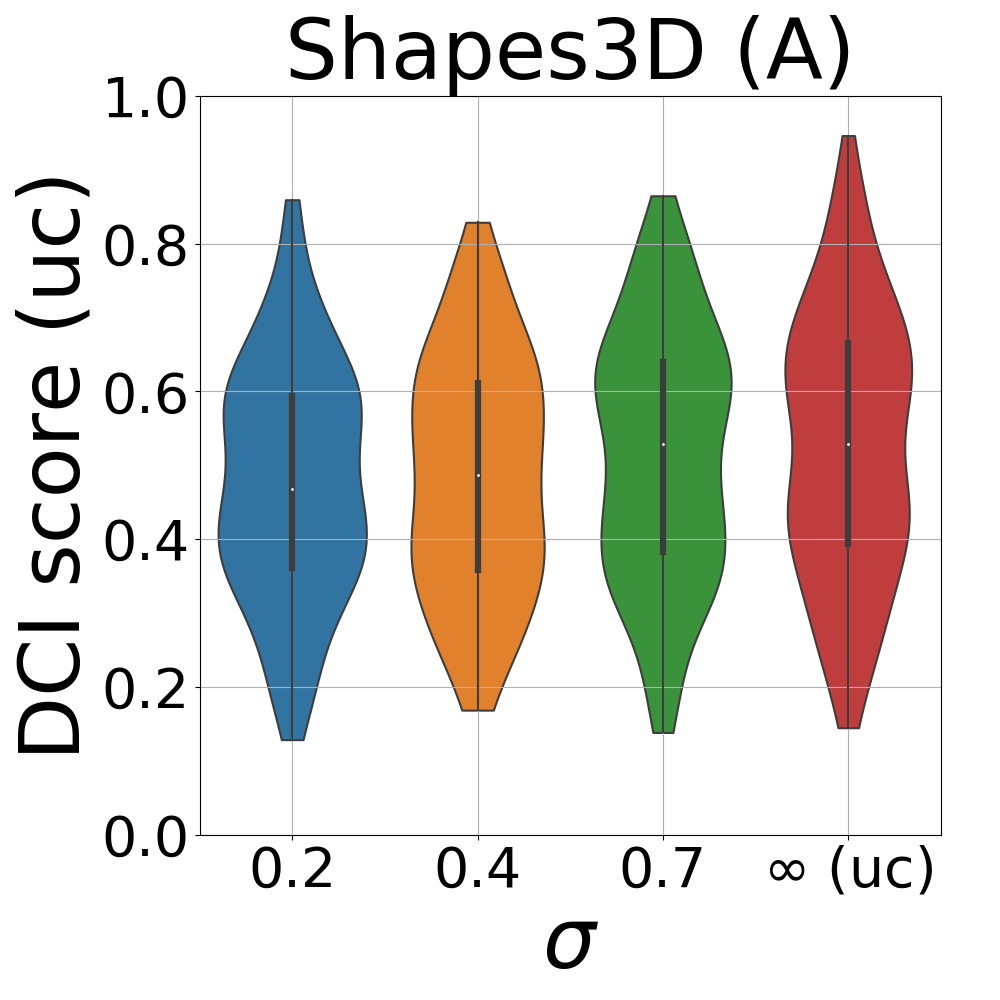}
\endminipage
\caption{Standard global disentanglement metrics evaluated on the uncorrelated (uc) dataset set showing no clear trend for different correlation strengths.}
\label{fig:disentanglement_metrics_unsupervised_appendix_uc}
\end{figure}

\paragraph{Shortcomings of existing metrics.} 
Following recent studies, we evaluate the trained models with the help of a broad range of disentanglement metrics that aim at quantifying overall success by a single scalar measure. 
Perhaps surprisingly, \cref{fig:disentanglement_metrics_unsupervised_appendix} and \cref{fig:disentanglement_metrics_unsupervised_appendix_uc} show no clear trend among all implemented disentanglement scores w.r.t.\ correlation strength.
The metrics have been evaluated by both, either sampling from the correlated data distribution or from the uncorrelated distribution. 
Given our extensive analysis of latent entanglements of the correlated FoV pair from above, we thus argue that common disentanglement metrics are limited in revealing those when correlations are introduced into the training data and we partly account this to the averaging procedures across many FoV with these pairs.
Note that regarding BetaVAE and FactorVAE this observed trend is to some degree expected as they would yield perfect disentanglement scores even if we would take the correlated ground truth factors or a linear transformation in the case of BetaVAE as the representation.

\begin{figure}
\begin{subfigure}{\columnwidth}\centering
\minipage{0.19\columnwidth}
  \includegraphics[width=\columnwidth]{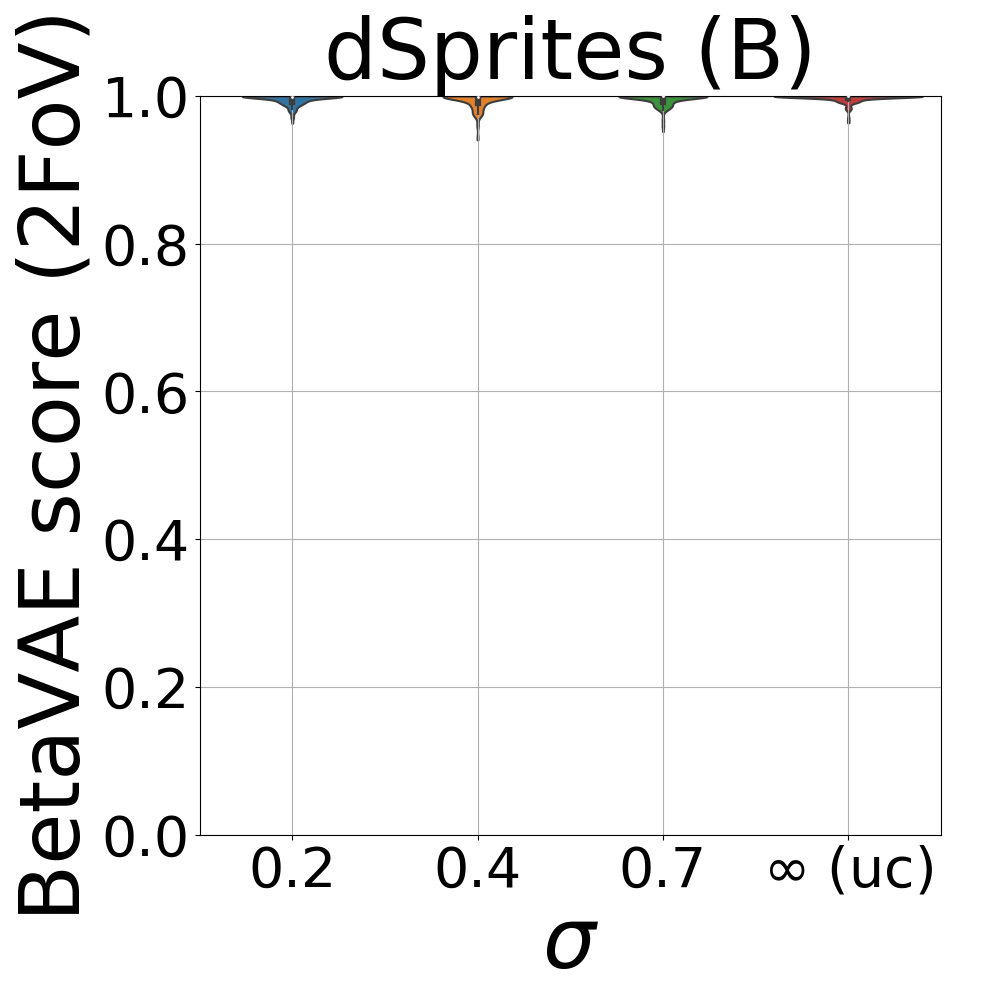}
\endminipage\hfill
\minipage{0.19\columnwidth}
  \includegraphics[width=\columnwidth]{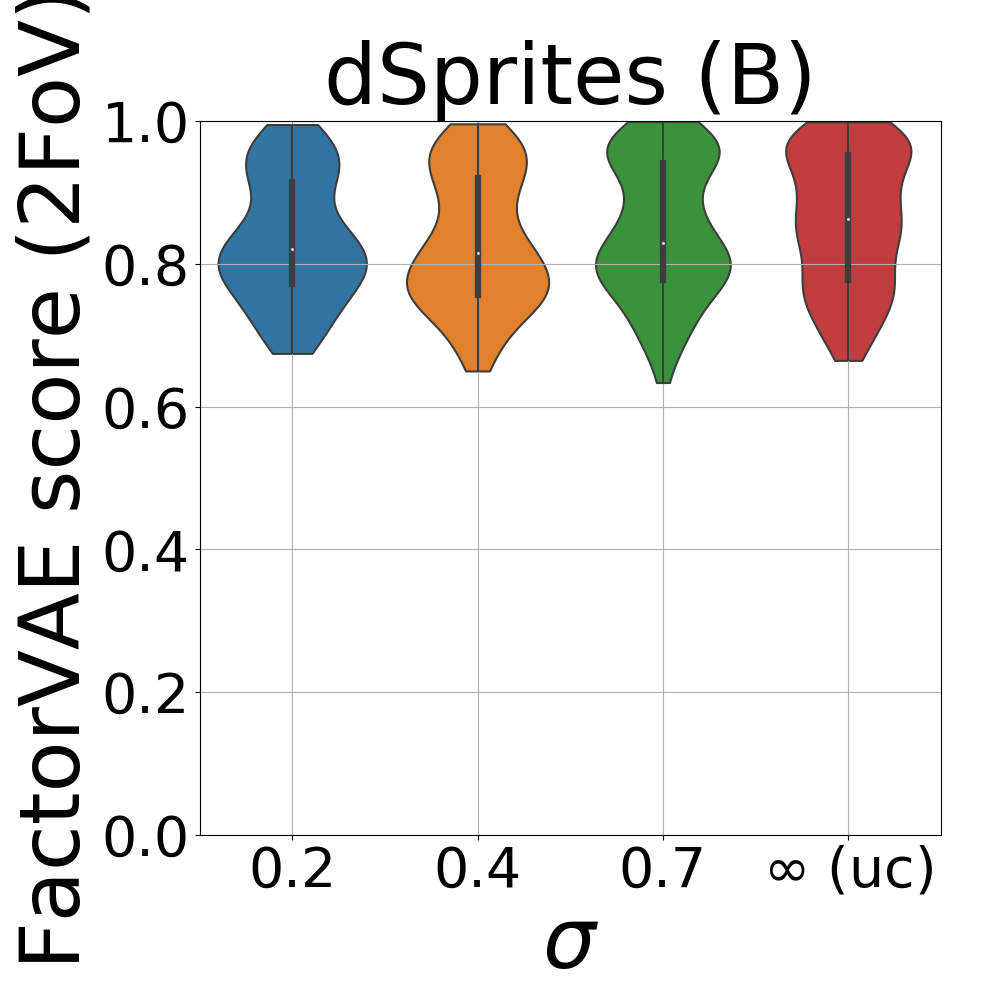}
\endminipage\hfill
\minipage{0.19\columnwidth}
  \includegraphics[width=\columnwidth]{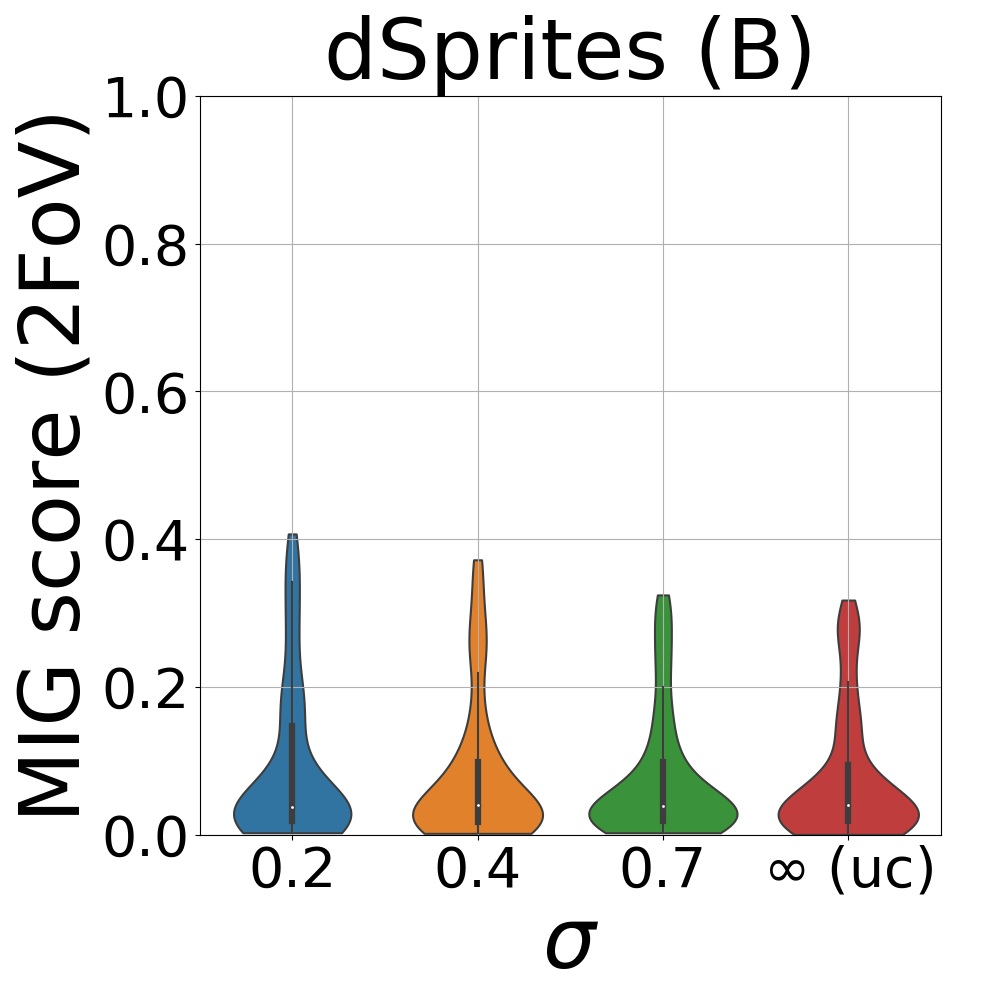}
\endminipage\hfill
\minipage{0.19\columnwidth}
  \includegraphics[width=\columnwidth]{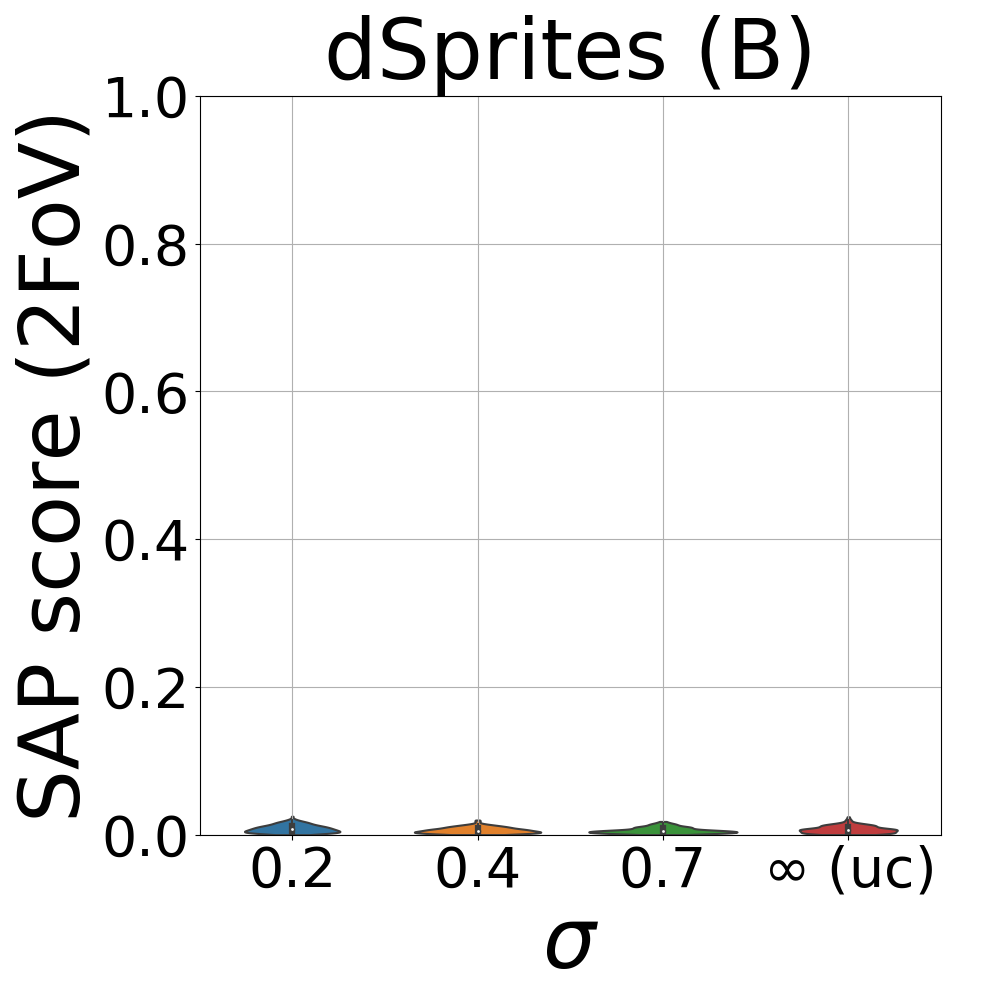}
\endminipage\hfill
\minipage{0.19\columnwidth}
  \includegraphics[width=\columnwidth]{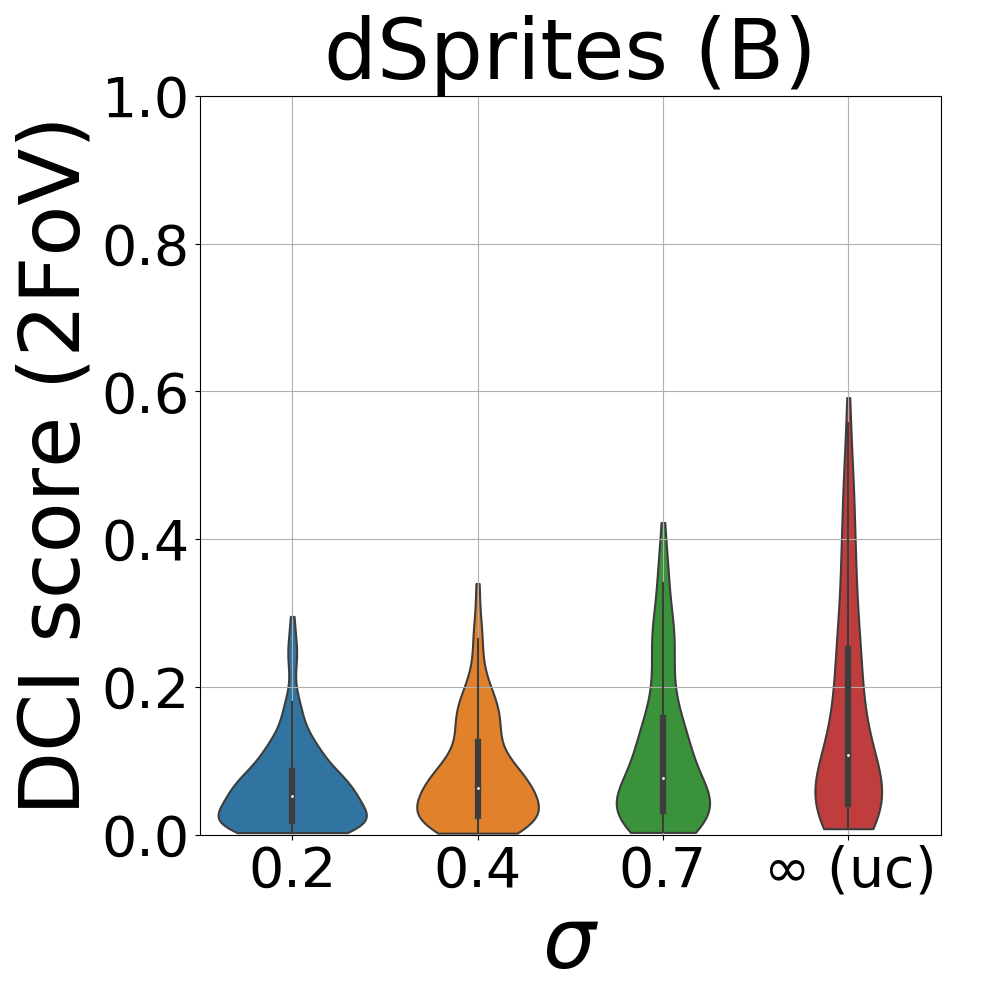}
\endminipage
\vskip 0.1in
\minipage{0.19\columnwidth}
  \includegraphics[width=\columnwidth]{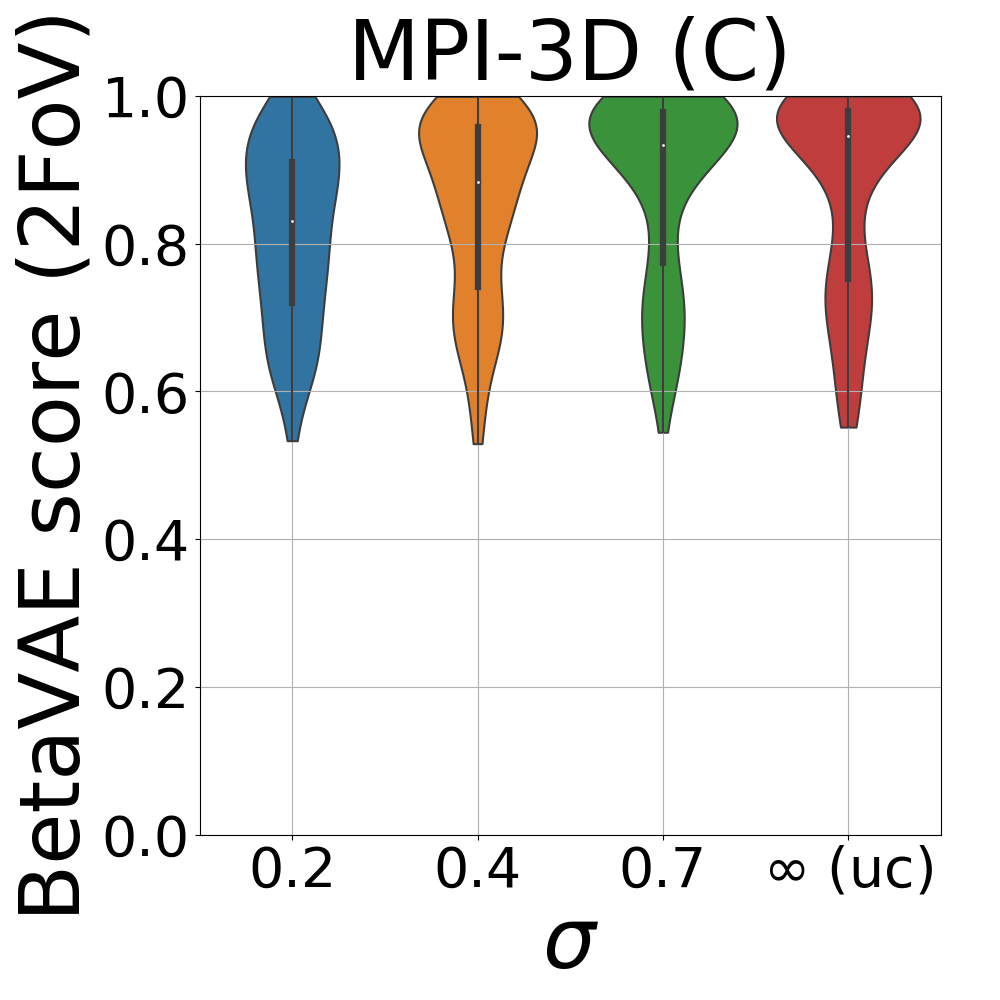}
\endminipage\hfill
\minipage{0.19\columnwidth}
  \includegraphics[width=\columnwidth]{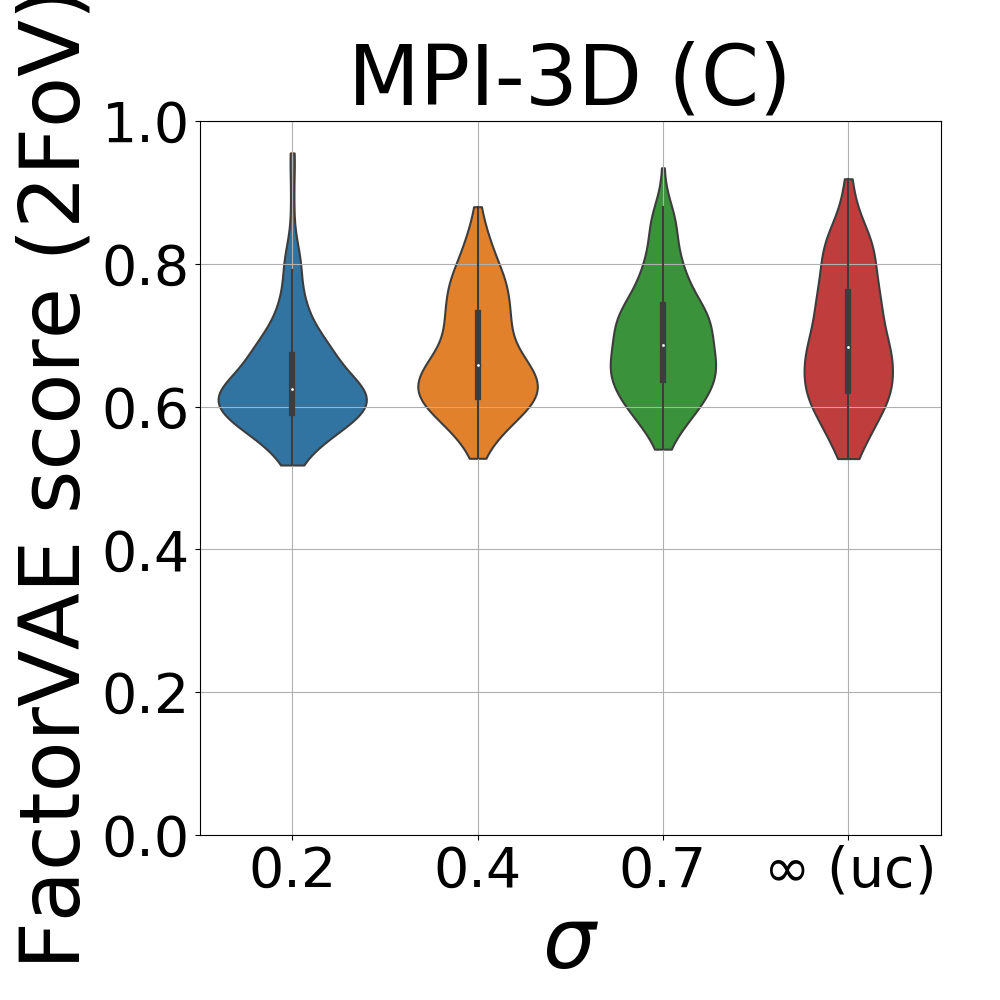}
\endminipage\hfill
\minipage{0.19\columnwidth}
  \includegraphics[width=\columnwidth]{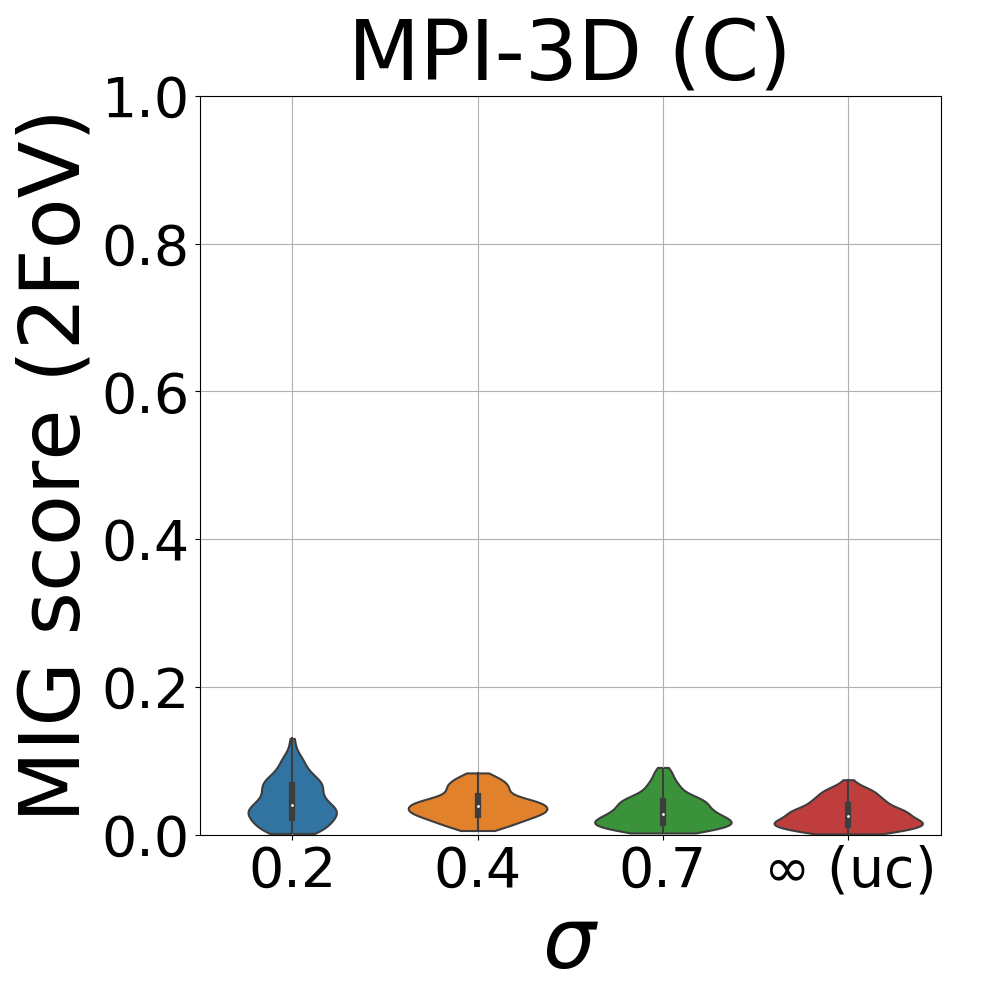}
\endminipage\hfill
\minipage{0.19\columnwidth}
  \includegraphics[width=\columnwidth]{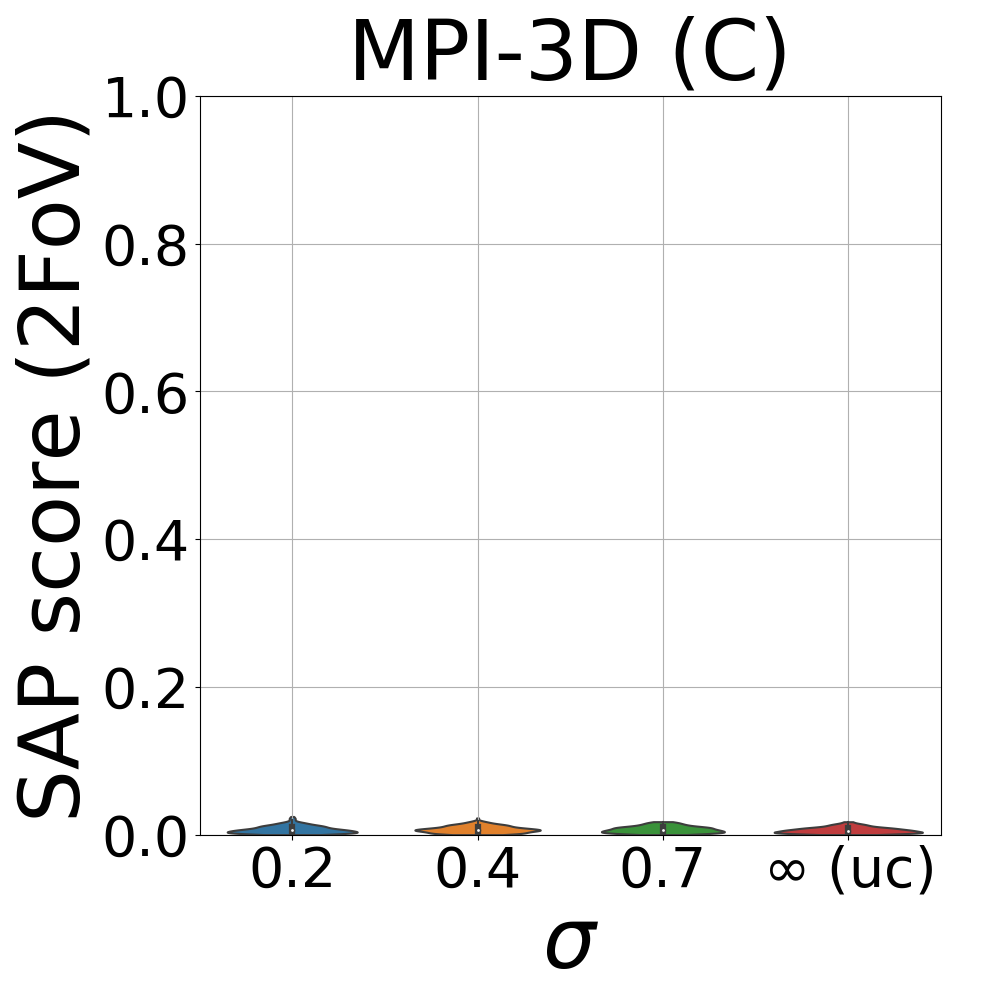}
\endminipage\hfill
\minipage{0.19\columnwidth}
  \includegraphics[width=\columnwidth]{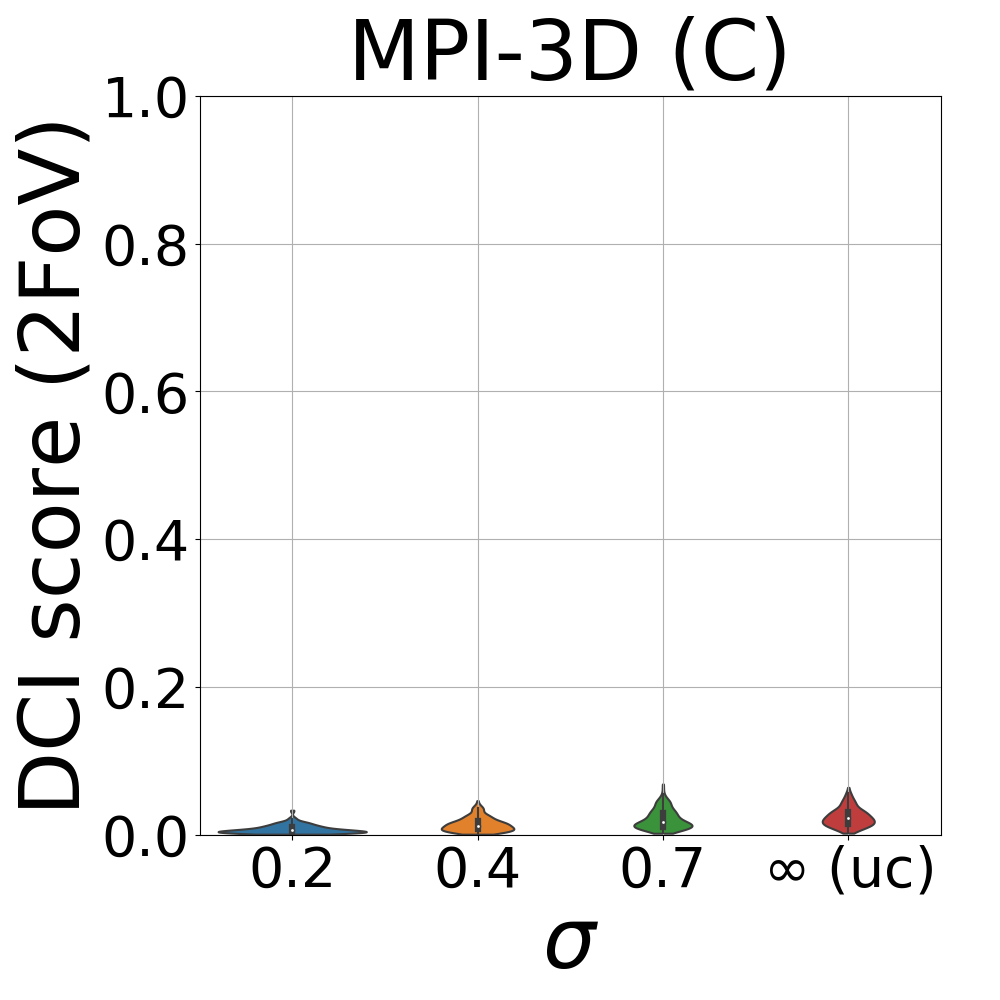}
\endminipage
\vskip 0.1in
\minipage{0.19\columnwidth}
  \includegraphics[width=\columnwidth]{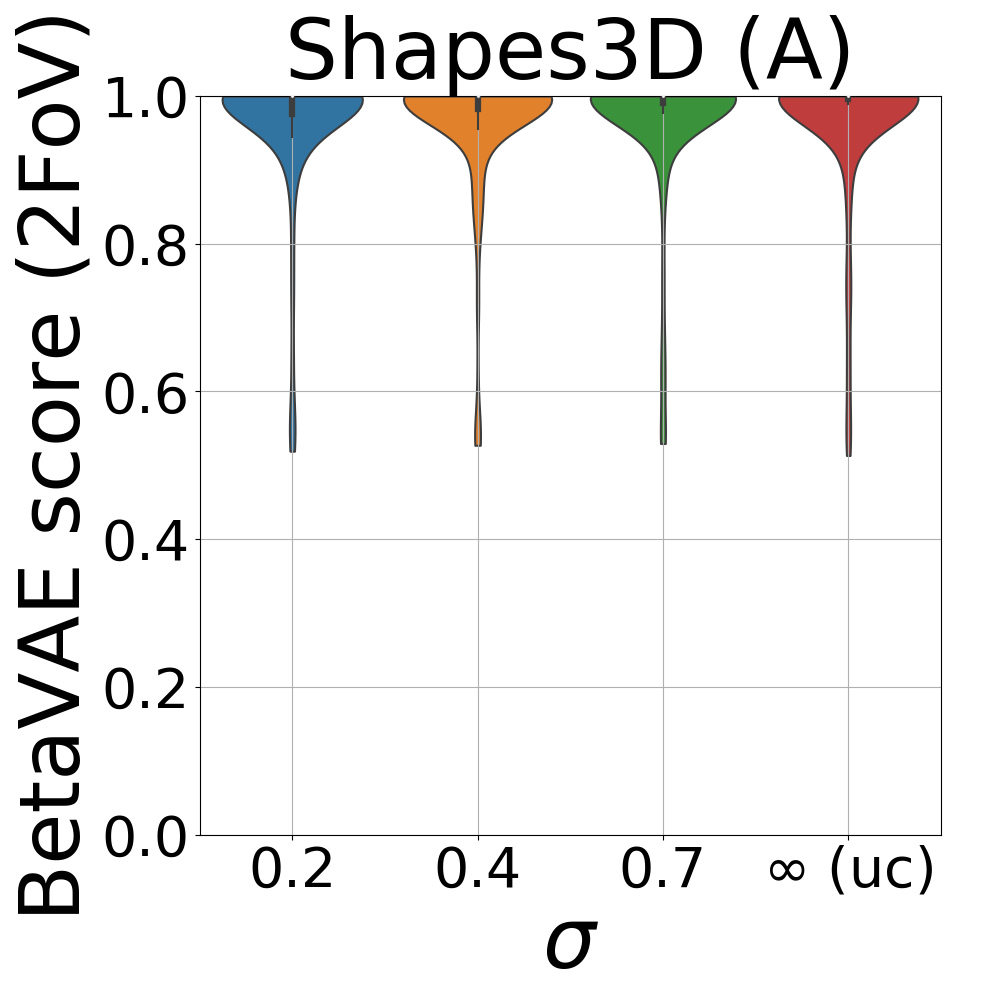}
\endminipage\hfill
\minipage{0.19\columnwidth}
  \includegraphics[width=\columnwidth]{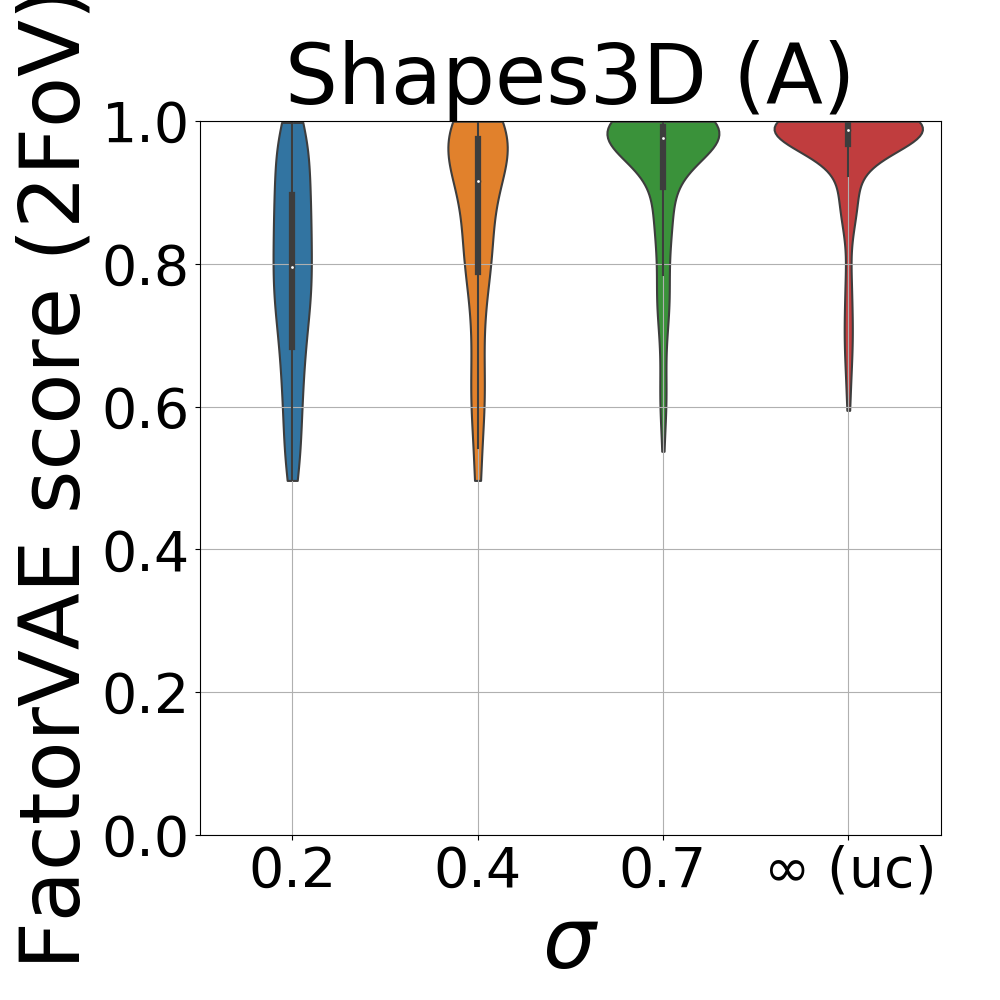}
\endminipage\hfill
\minipage{0.19\columnwidth}
  \includegraphics[width=\columnwidth]{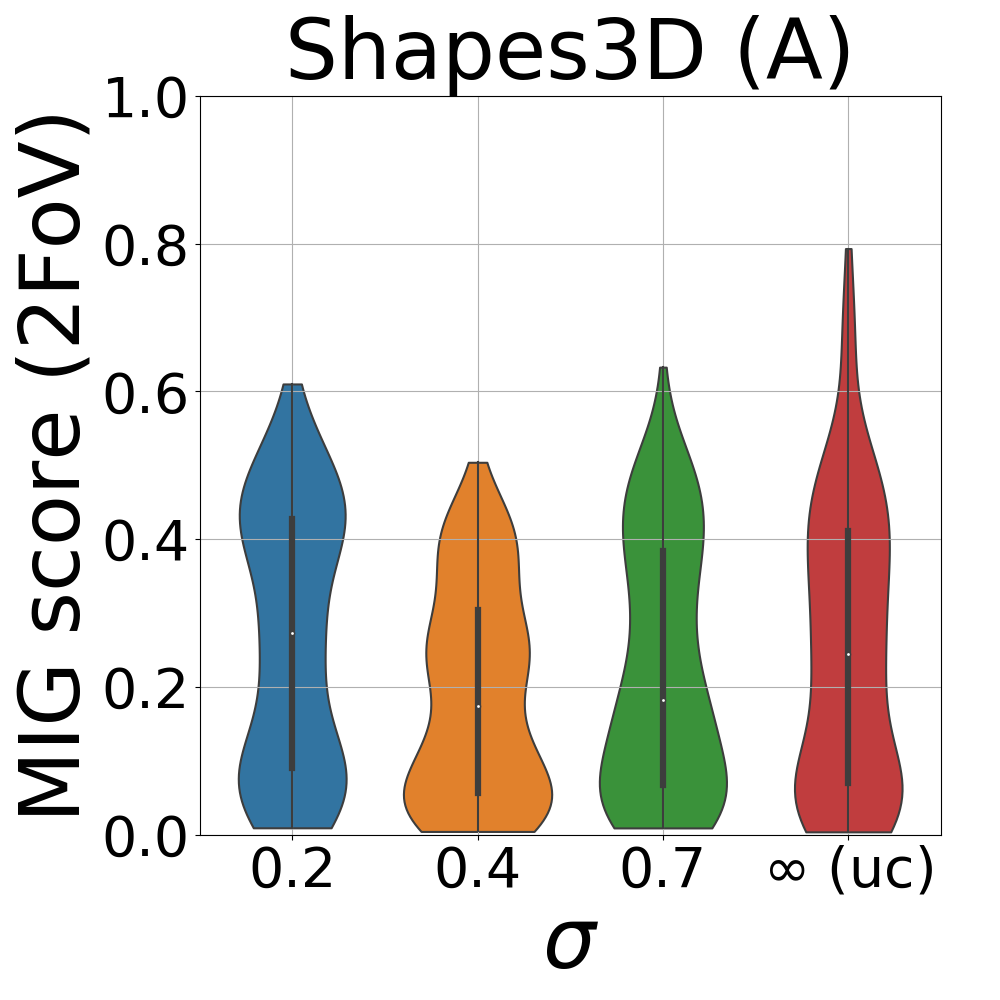}
\endminipage\hfill
\minipage{0.19\columnwidth}
  \includegraphics[width=\columnwidth]{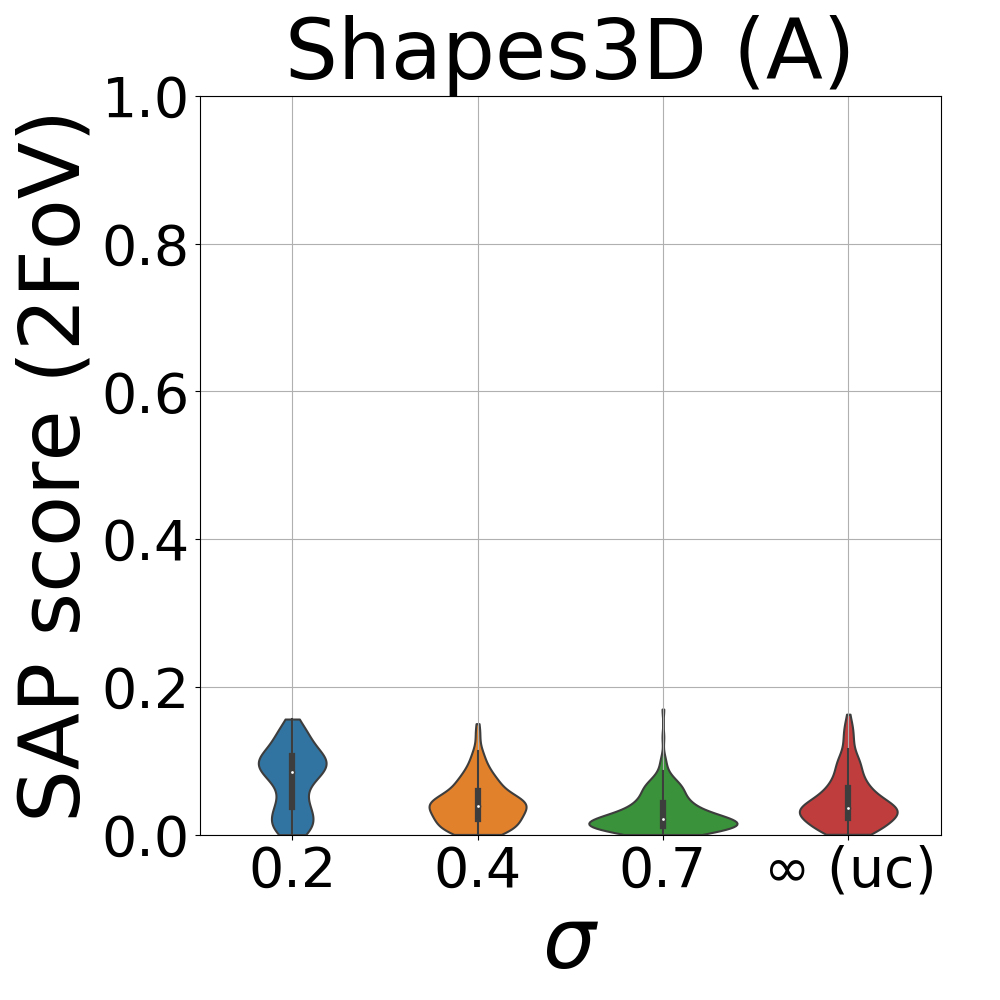}
\endminipage\hfill
\minipage{0.19\columnwidth}
  \includegraphics[width=\columnwidth]{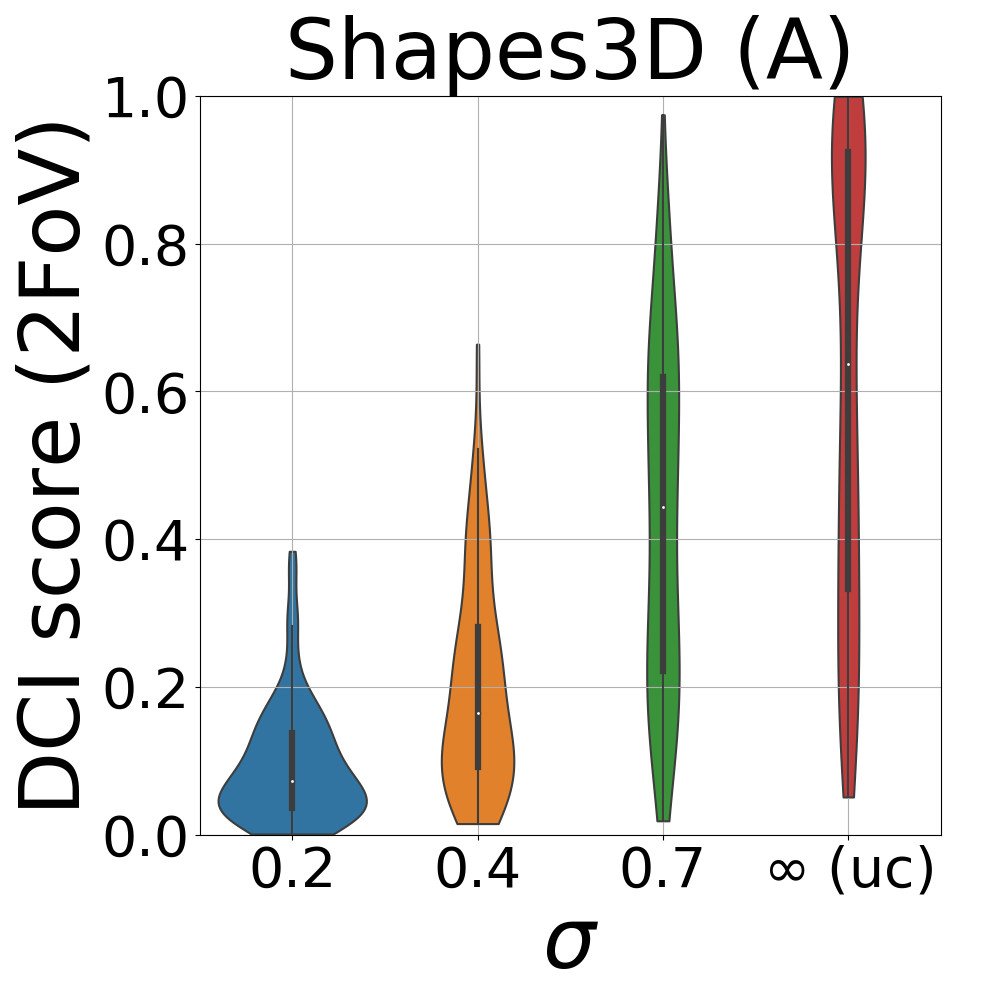}
\endminipage
\end{subfigure}
\caption{When disentanglement metrics are only evaluated regarding the 2 correlated FoV we can observe the still persisting entanglement in the latents using DCI score.}
\label{fig:disentanglement_metrics_unsupervised_appendix_2FoV}
\end{figure}
As these metrics are defined with respect to the whole set of underlying ground truth FoV and employ various averaging techniques to form a single scalar measure we want to investigate how much the observed latent entanglements are hidden though imperfect disentanglement of other factors. Thus, we evaluate the same metrics but only on the two correlated FoV excluding all other remaining factors. Indeed, as can be seen from \cref{fig:disentanglement_metrics_unsupervised_appendix_2FoV}, only DCI tracks the entangled latents under this reduced disentanglement score, while the others show no or only weak trends. We refer the interested reader to \citet{locatello2020sober}, where a detailed discussion is provided why MIG is not tracking the latent entanglement we observed.

\subsection{\cref{sec:generalization}}
\label{sec:generalization_properties_appendix}
\begin{figure}
\begin{center}
\includegraphics[width=\columnwidth]{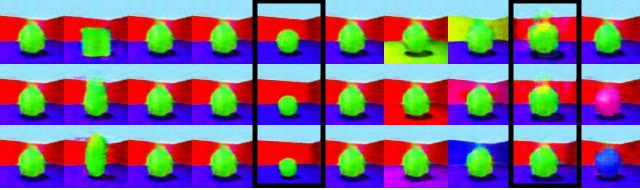}
\end{center}
\caption{Generalization capabilities towards out-of-distribution test data. Latent traversals from an observations the model has never seen during training.
The starting point corresponds to a factor configuration in point number 2 from \cref{fig:generalization_reconstruction}
Shown are the results of the model with highest DCI score among all 180 trained models  on Shapes3d (A) with a very restricted correlation strength $\sigma = 0.2$ in object size and azimuth}
\label{fig:latent_traversal_from_ood_sample}
\end{figure}

\paragraph{Generalization Properties}
In order to support our conclusion that disentanglement methods can generalize towards unseen FoV configurations we show in \cref{fig:latent_traversal_from_ood_sample} latent traversals originating from OOD point number 2 with smallest object size and largest azimuth. We observe that changes in the remaining factors reliably yield the expected reconstructions.

\begin{figure}[t]
\minipage{0.48\columnwidth}
  \includegraphics[width=\columnwidth]{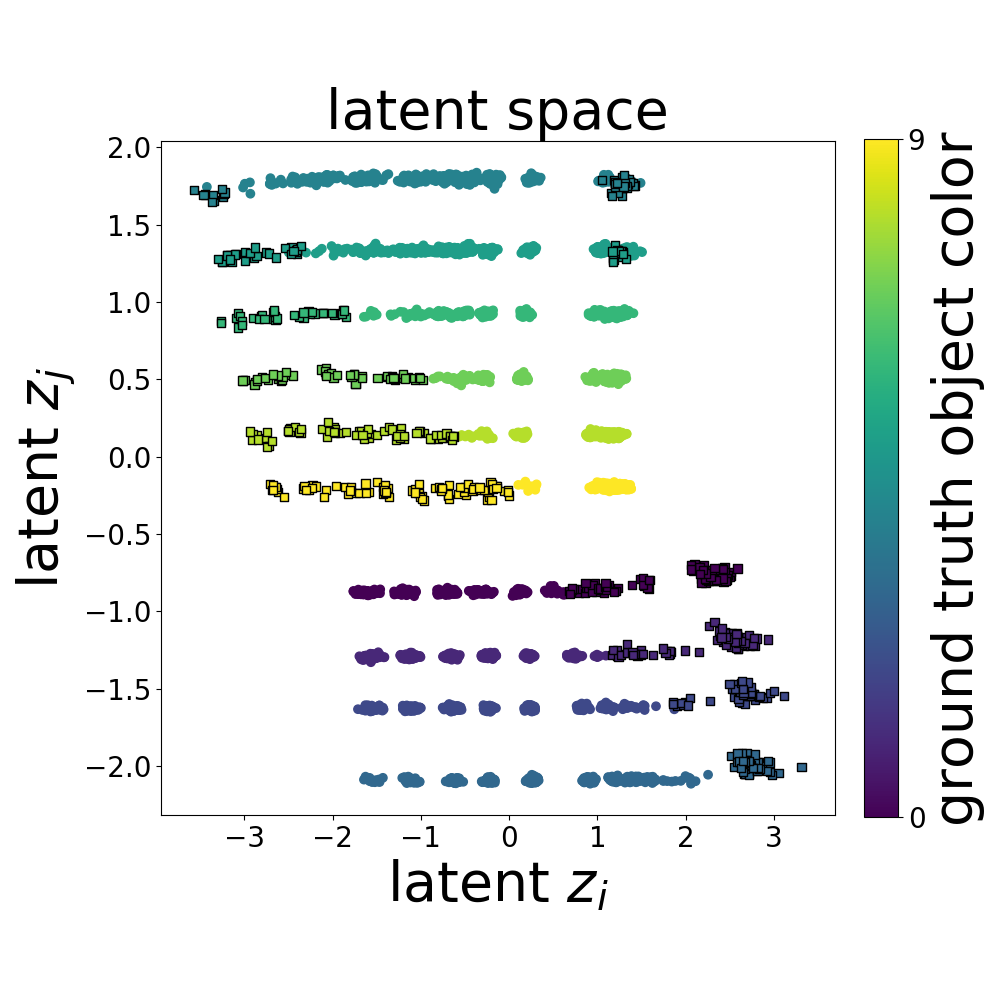}
\endminipage\hfill
\minipage{0.48\columnwidth}
  \includegraphics[width=\columnwidth]{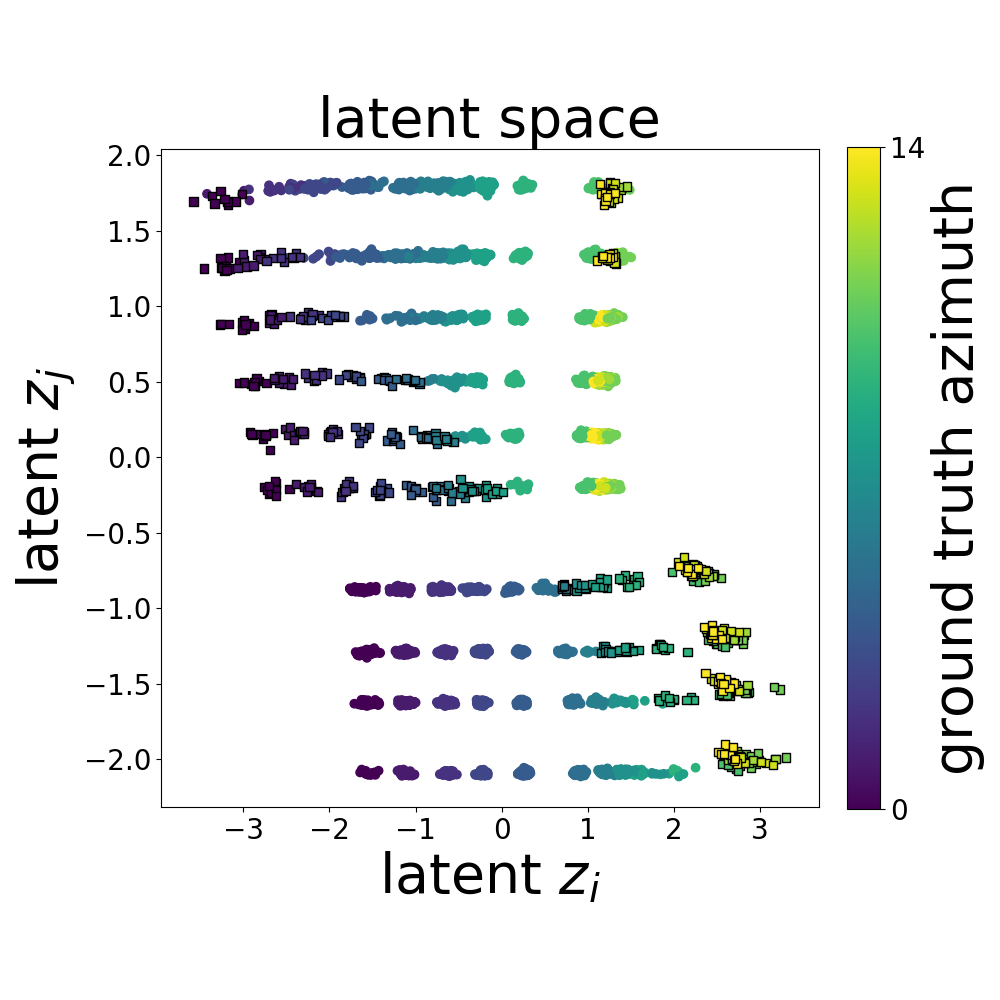}
\endminipage\hfill
\vskip 0.05in
\minipage{0.48\columnwidth}
  \includegraphics[width=\columnwidth]{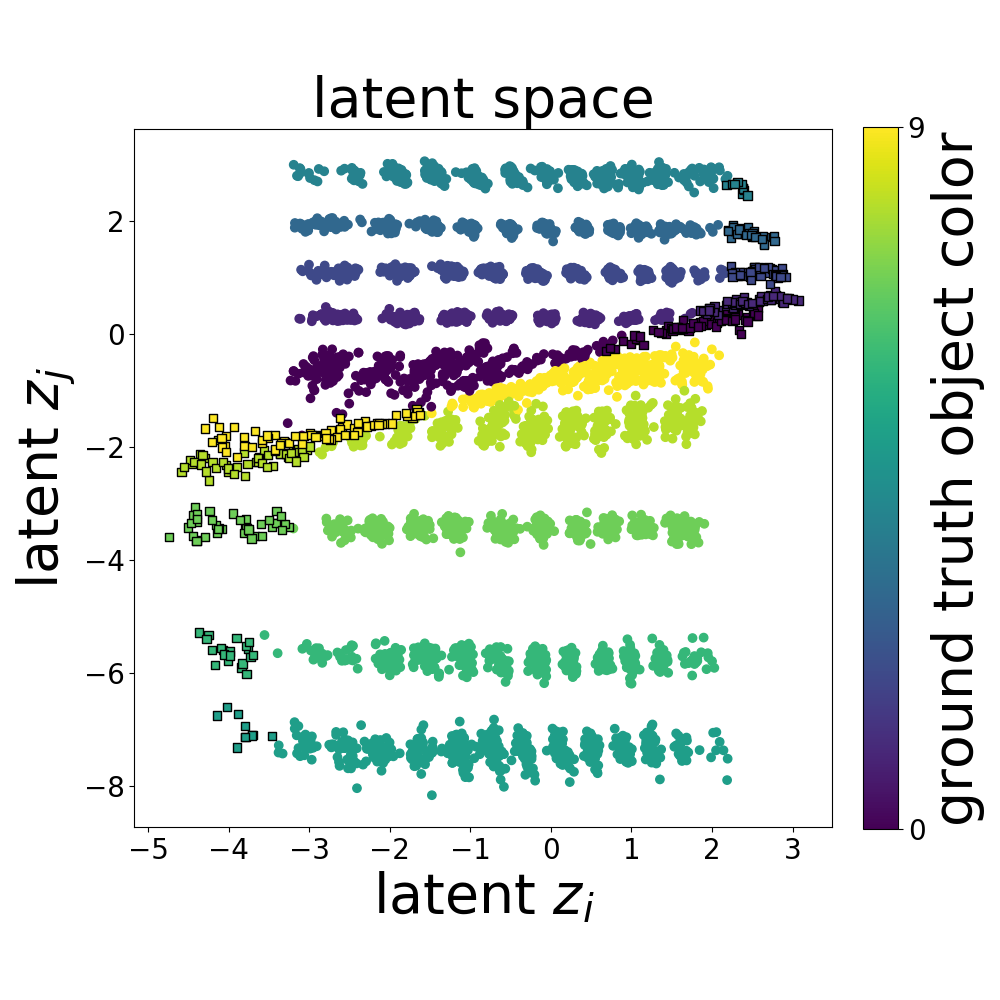}
\endminipage\hfill
\minipage{0.48\columnwidth}
  \includegraphics[width=\columnwidth]{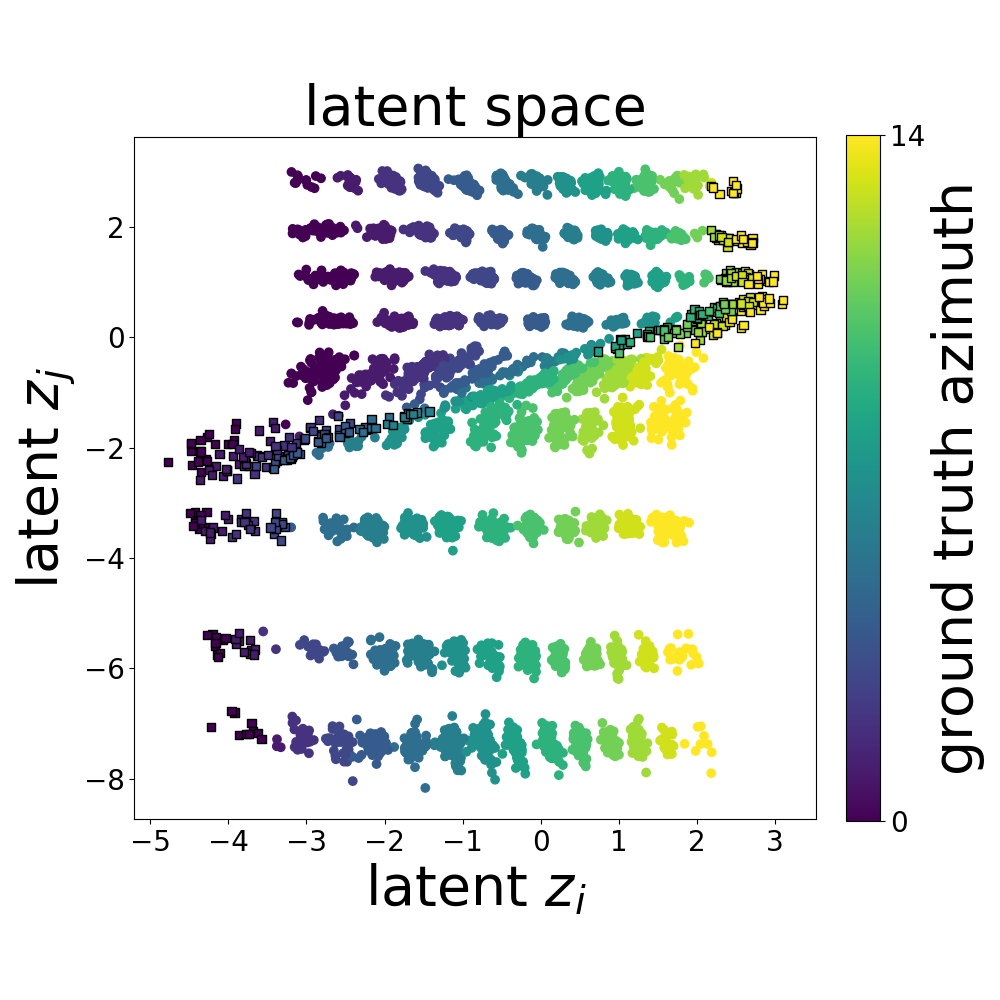}
\endminipage\hfill
\vskip 0.05in
\minipage{0.48\columnwidth}
  \includegraphics[width=\columnwidth]{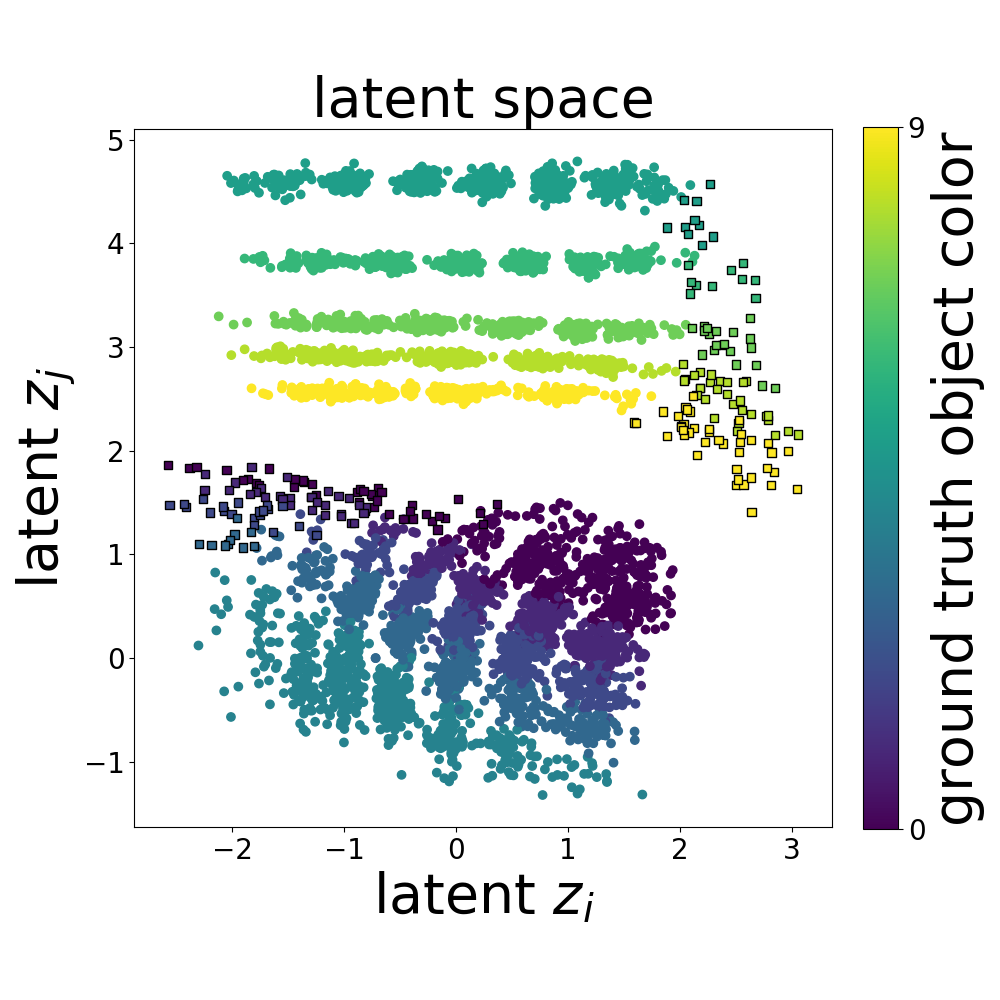}
\endminipage\hfill
\minipage{0.48\columnwidth}
  \includegraphics[width=\columnwidth]{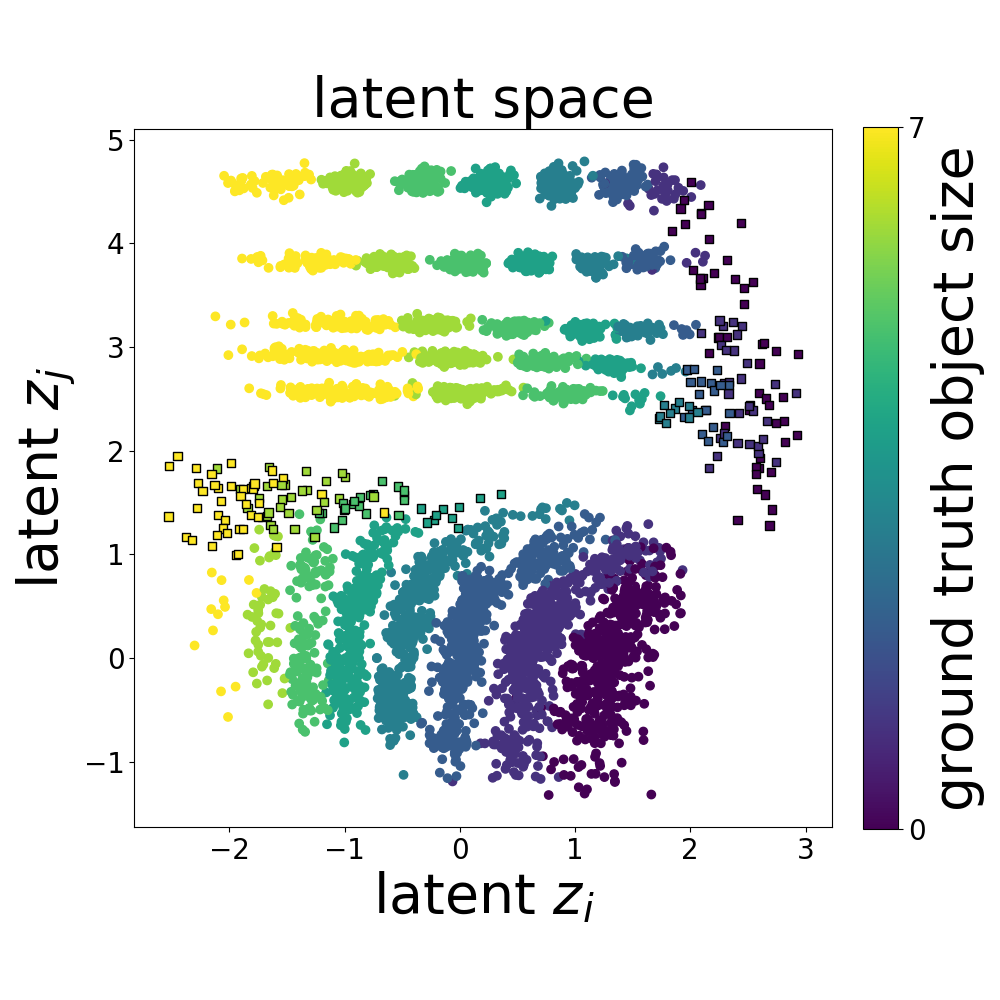}
\endminipage\hfill
\vskip 0.05in
\minipage{0.48\columnwidth}
  \includegraphics[width=\columnwidth]{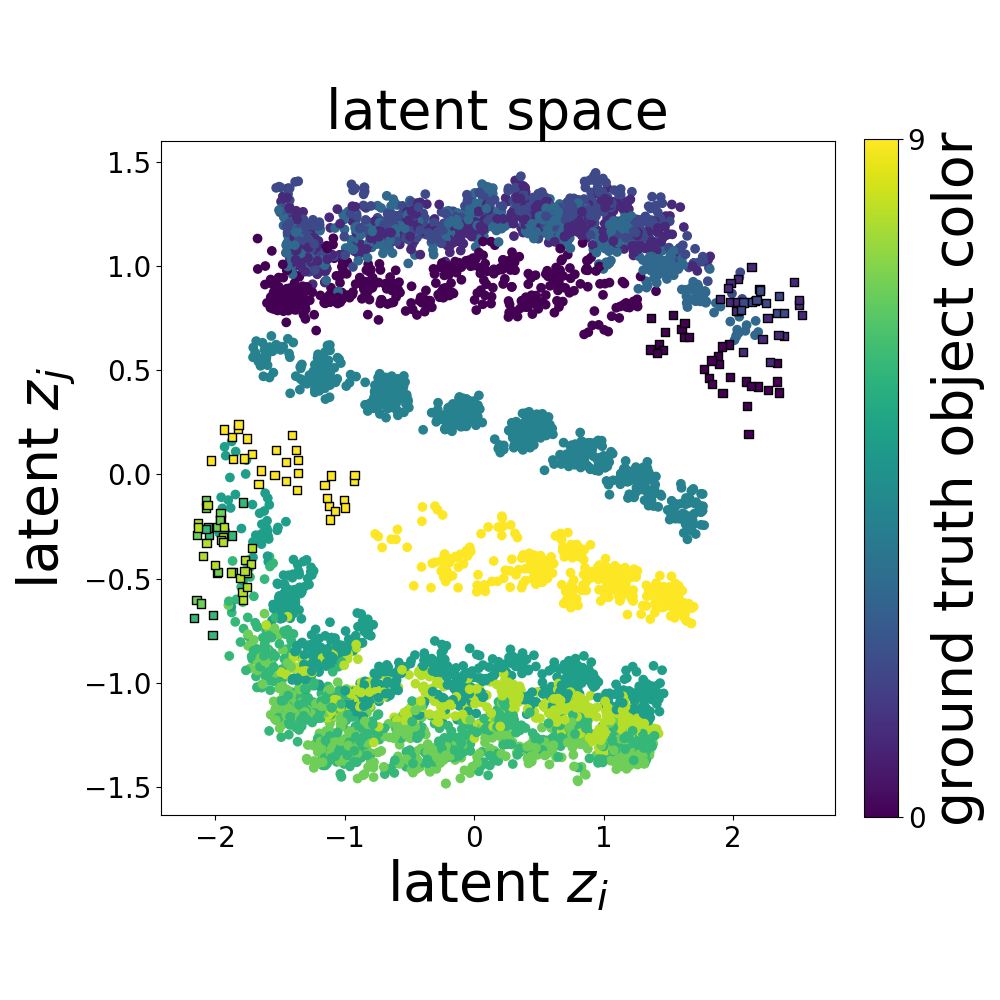}
\endminipage
\minipage{0.48\columnwidth}
  \includegraphics[width=\columnwidth]{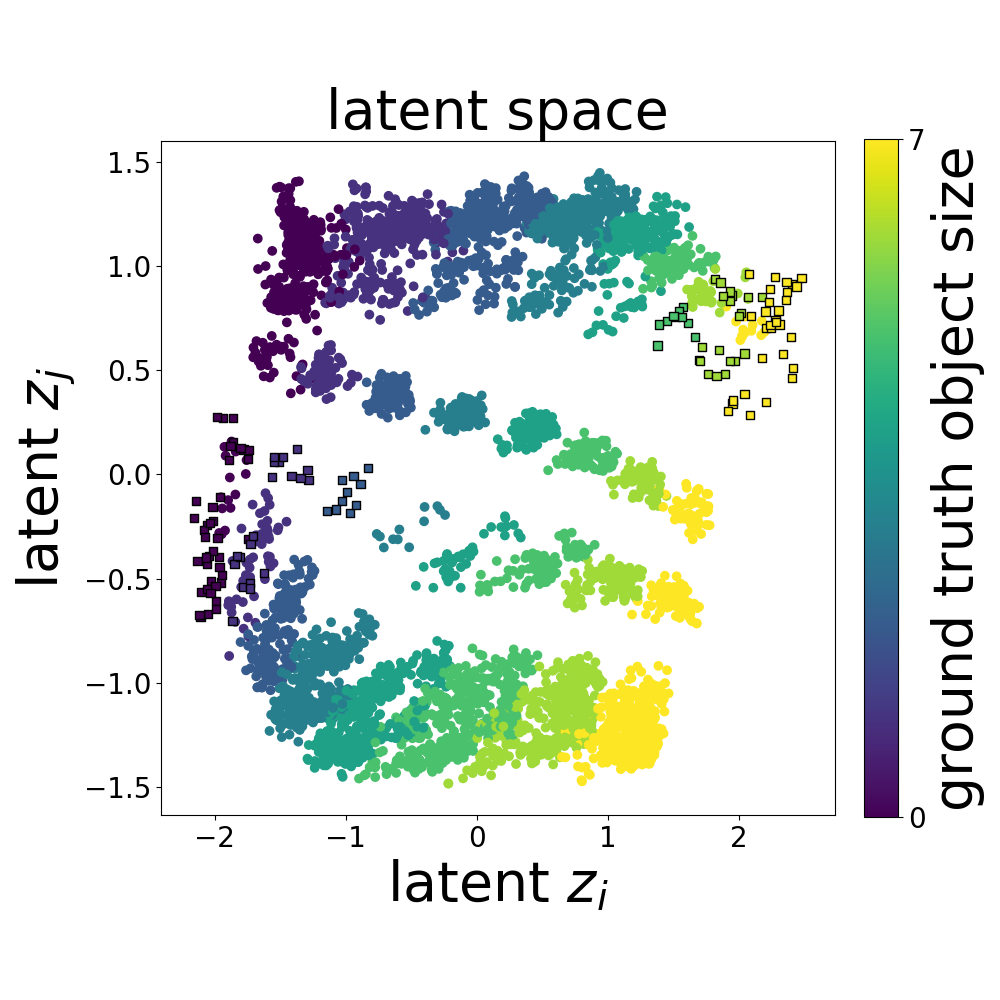}
\endminipage
\caption{Latent space distribution of the two entangled dimensions of the best DCI model in Shapes3d (E) with $\sigma = 0.2$ (top row), in Shapes3d (E) with $\sigma = 0.4$ (second row), in Shapes3d (D) with $\sigma = 0.2$ (third row) and in Shapes3d (D) with $\sigma = 0.4$  (bottom row). Latent codes sampled from correlated observations (circle without edge) and (2) latent codes sampled with an object size-azimuth configuration not encountered during training(squares with black edge). Each column shows the ground truth values of the two correlated factors by color.}
\label{fig:representation_space_unsupervised_study_w_fa_additional_models}
\end{figure}

Emphasizing the generalization results from the main paper,  we are visualizing the latent spaces with similar extrapolation and generalization capabilities of four additional models from the two strongest correlation dataset variants of Shapes3d (D) and Shapes3d (E) in \cref{fig:representation_space_unsupervised_study_w_fa_additional_models}. These latent spaces further support that OOD examples are meaningfully encoded into the existing structure of the latent space and that the decoder is equally capable of generating observations from such unseen representations.

\clearpage
\section{Additional Results Section \ref{sec:weakly_supervised_study_results}}
\label{sec:finding_right_factorization_result_appendix}

\subsection{Post-hoc alignment correction with few labels}
\label{subsec:fast_adaptation_appendix}
\begin{figure}
\minipage{0.33\columnwidth}
  \includegraphics[width=\columnwidth]{figures/representation_spaces/unsupervised_study/779/representation_space_fa_labels_0cvariable_1.png}
\endminipage\hfill
\minipage{0.33\columnwidth}
  \includegraphics[width=\columnwidth]{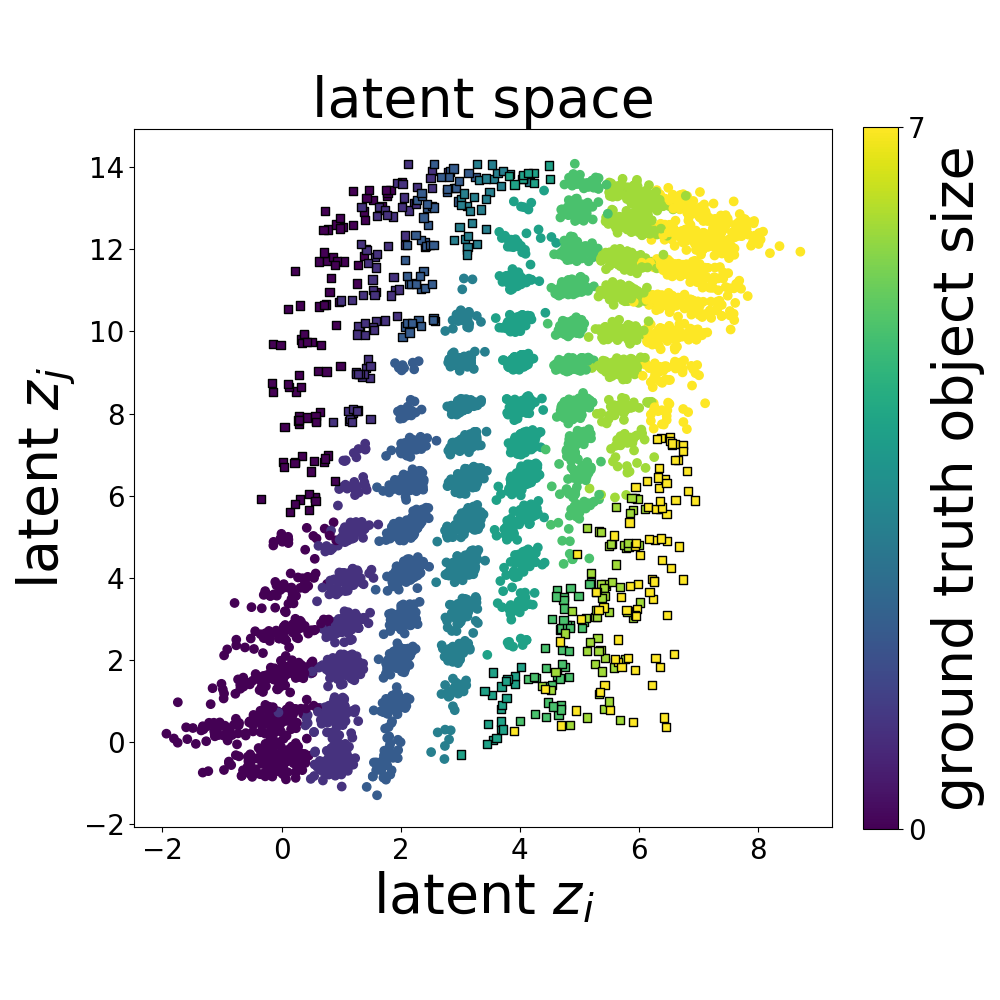}
\endminipage\hfill
\minipage{0.33\columnwidth}
  \includegraphics[width=\columnwidth]{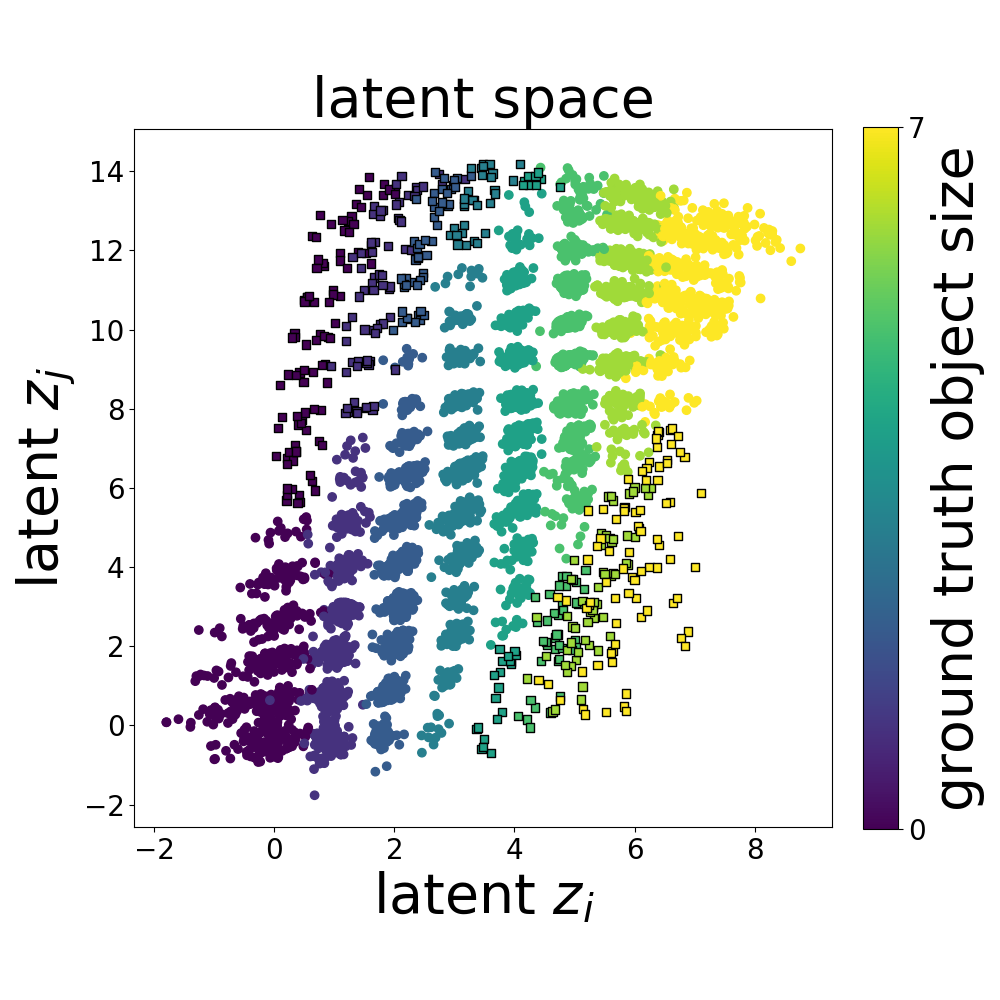}
\endminipage
\vskip 0.1in
\minipage{0.33\columnwidth}
  \includegraphics[width=\columnwidth]{figures/representation_spaces/unsupervised_study/779/representation_space_fa_labels_0cvariable_2.png}
\endminipage\hfill
\minipage{0.33\columnwidth}
  \includegraphics[width=\columnwidth]{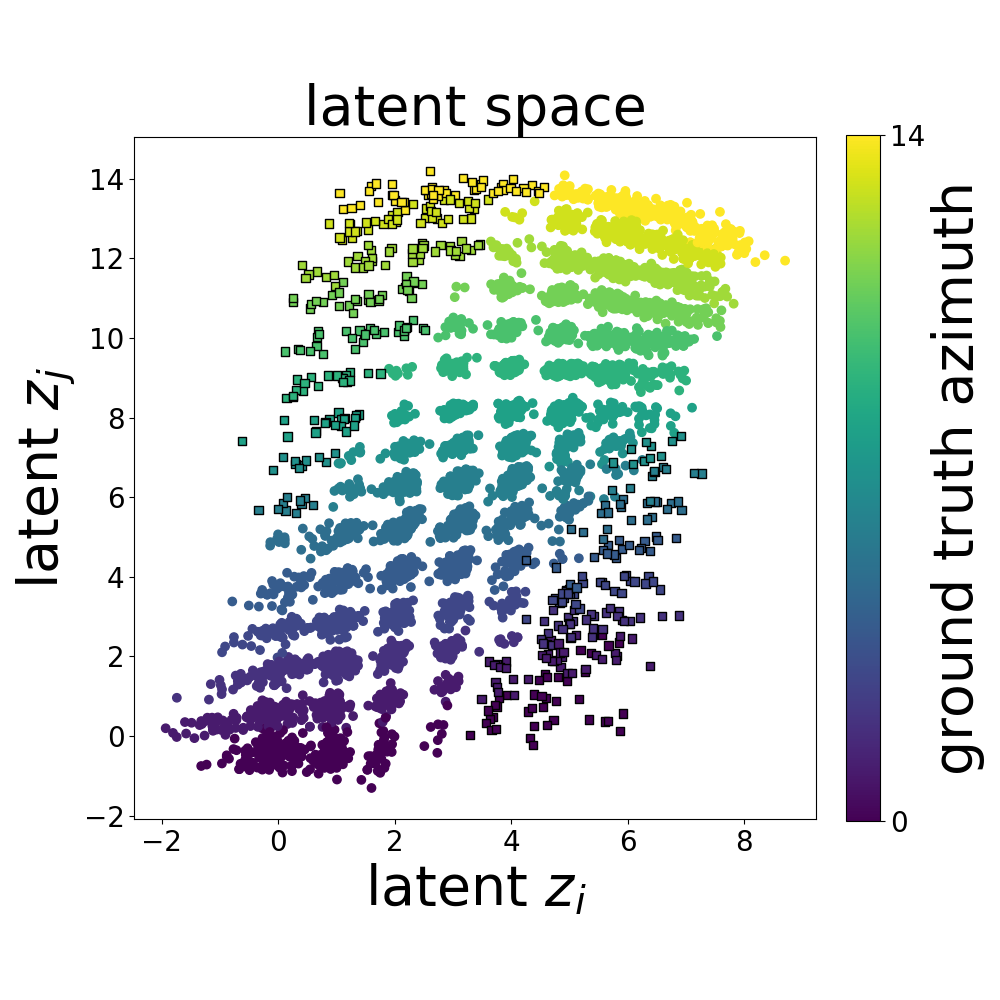}
\endminipage\hfill
\minipage{0.33\columnwidth}
  \includegraphics[width=\columnwidth]{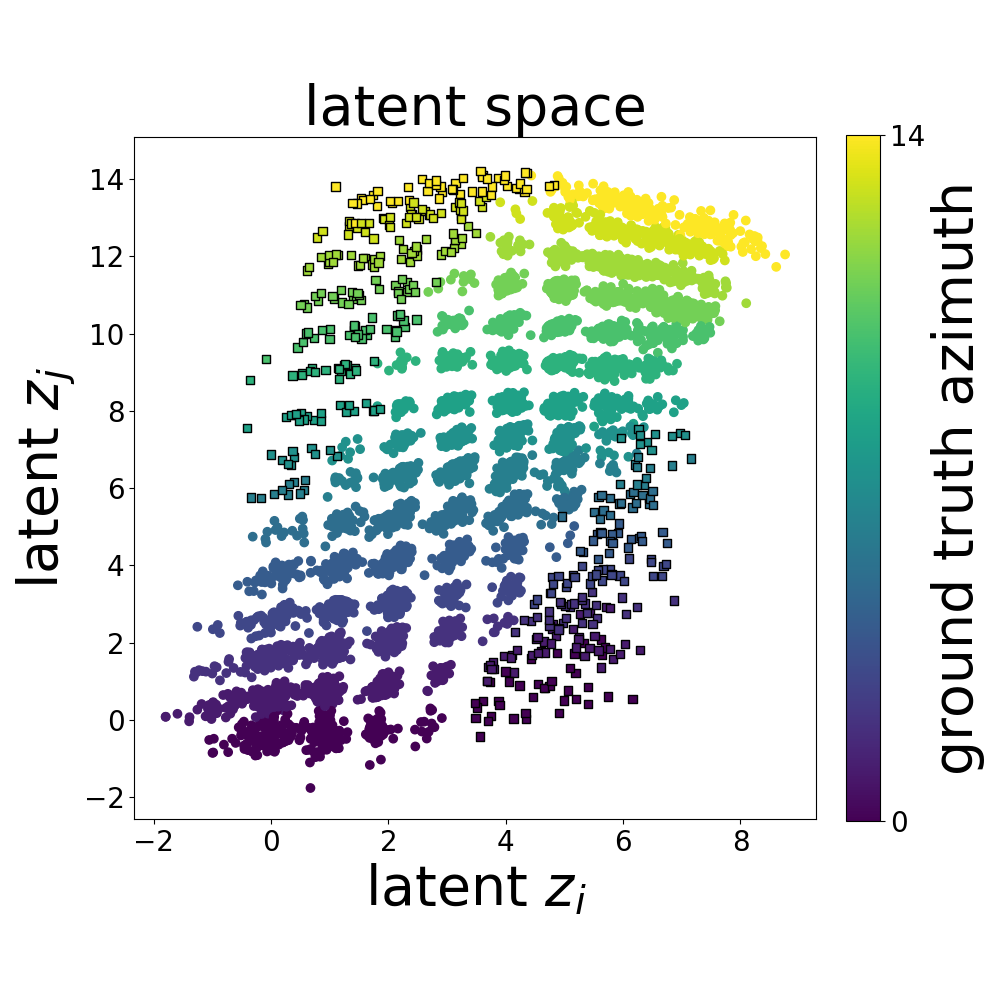}
\endminipage
\caption{Latent space distribution of the two entangled dimensions of the best DCI model in Shapes3D (A). Latent codes sampled from correlated observations (circle without edge) and (2) latent codes sampled with an object size-azimuth configuration not encountered during training (squares with black edge). Left column shows the latent space of the two correlated factors by color. Middle and right column show the fast adapted space using linear regression and 100 or 1000 labels respectively.}
\label{fig:representation_space_unsupervised_study_w_fa_figure}
\end{figure}

In \cref{fig:representation_space_unsupervised_study_w_fa_figure}, we see the axis alignment of the correlated latent space after fast adaptation using linear regression on a model trained on Shapes3D (A).
Fast adaptation with linear regression substitution fails in some settings: when no two latent dimensions encode the applied correlation isolated from the other latent codes, or when the correlated variables do not have a unique natural ordering (e.g. color or categorical variables).
Additionally, the functional form of the latent manifolds beyond the training distribution is unknown and in general expected to be nonlinear.
We test the possibility of fast adaptation in this case using as substitution function a one-hidden layer MLP classifier of size 100 on the correlated Shapes3D variants.
Under this method, we sample the few labels from a uniform independent distribution. A small number of such samples could practically be labeled manually. Using only 1000 labeled data points for our fast adaptation method shows a significant reduction in entanglement thresholds for the correlated pair (\cref{fig:entanglement_w_mlp_fa_GBT}).

\begin{table}
\begin{center}
\resizebox{0.7\columnwidth}{!}{
\begin{tabular}{llcc}
\toprule
 dataset & labels & 0 &   1000 \\
 \midrule
 \multirow{2}{*}{A ($\sigma=0.2$)} &\textcolor{red}{object size - azimuth} & \textcolor{red}{0.37} &   \textcolor{blue}{0.26} \\
 & median uncorrelated pairs              & 0.09 &   0.1  \\
\midrule
 \multirow{2}{*}{D ($\sigma=0.2$)} & \textcolor{red}{object color - object size} & \textcolor{red}{0.3}  &   \textcolor{blue}{0.16} \\
  & median uncorrelated pairs                   & 0.07 &   0.07 \\
\midrule
 \multirow{2}{*}{E ($\sigma=0.2$)} & \textcolor{red}{object color - azimuth} & \textcolor{red}{0.25} &  \textcolor{blue}{0.2}  \\
  & median uncorrelated pairs               & 0.1  &   0.11 \\
\bottomrule
\end{tabular}}
\caption{Mean of the pairwise entanglement scores for the correlated pair (red) and the median of the uncorrelated pairs (based on GBT feature importance) for all pairs of variables in Shapes3D (D) (top), Shapes3D (E) (middle) and Shapes3D (A) (bottom) all with correlation strength $\sigma = 0.2$. Each pairwise score is the mean across 180 models for each dataset and correlation strength.
First column is the unsupervised baseline without any fast adaptation and the second column shows that fast adaption using a one-hidden layer MLP reduces these correlations with as little as 1000 labels.}
\label{fig:entanglement_w_mlp_fa_GBT}

\end{center}
\end{table}

\subsection{Alignment during training using weak supervision}
\label{subsec:appendix_weak_supervision}
\begin{figure}
\minipage{0.3\columnwidth}
  \includegraphics[width=\columnwidth]{figures/experiment_weakly_supervised_observational/shapes3d_35/DCI_Disentanglement.png}
\endminipage\hfill
\minipage{0.34\columnwidth}
  \includegraphics[width=\columnwidth]{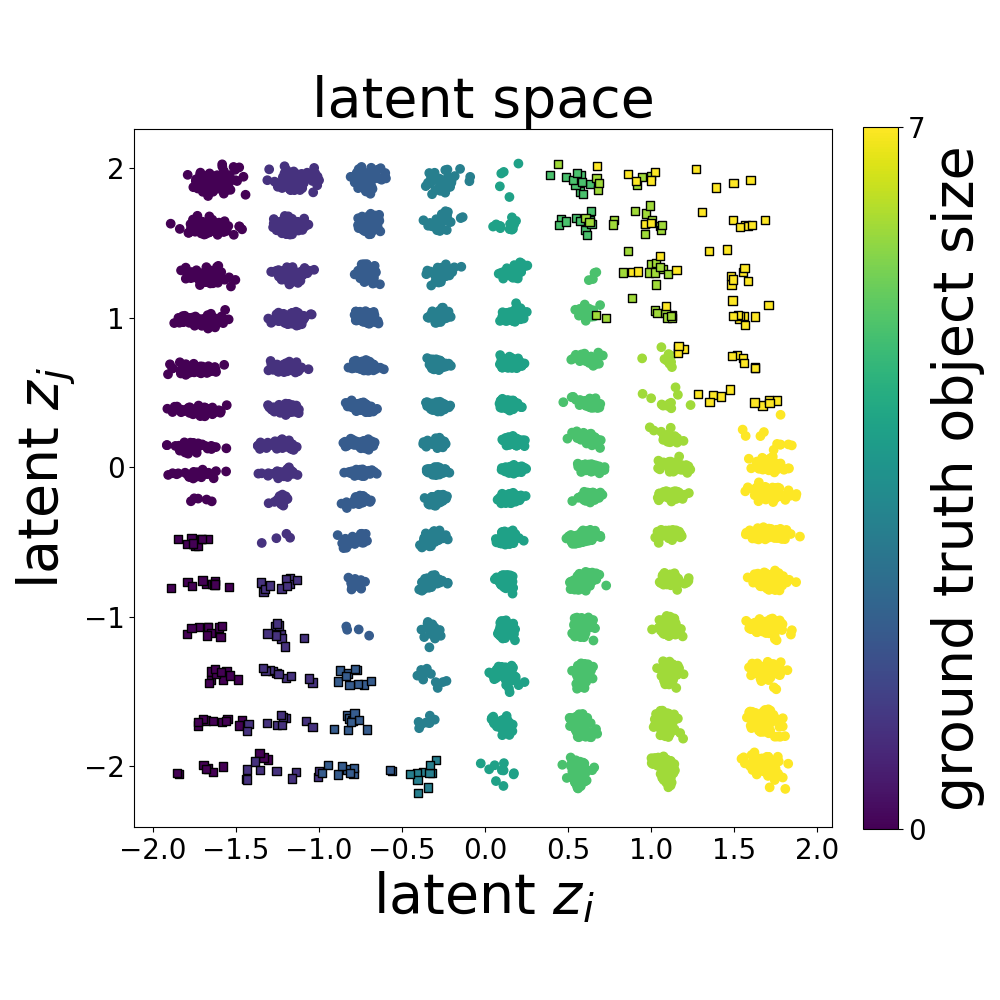}
\endminipage\hfill
\minipage{0.34\columnwidth}
  \includegraphics[width=\columnwidth]{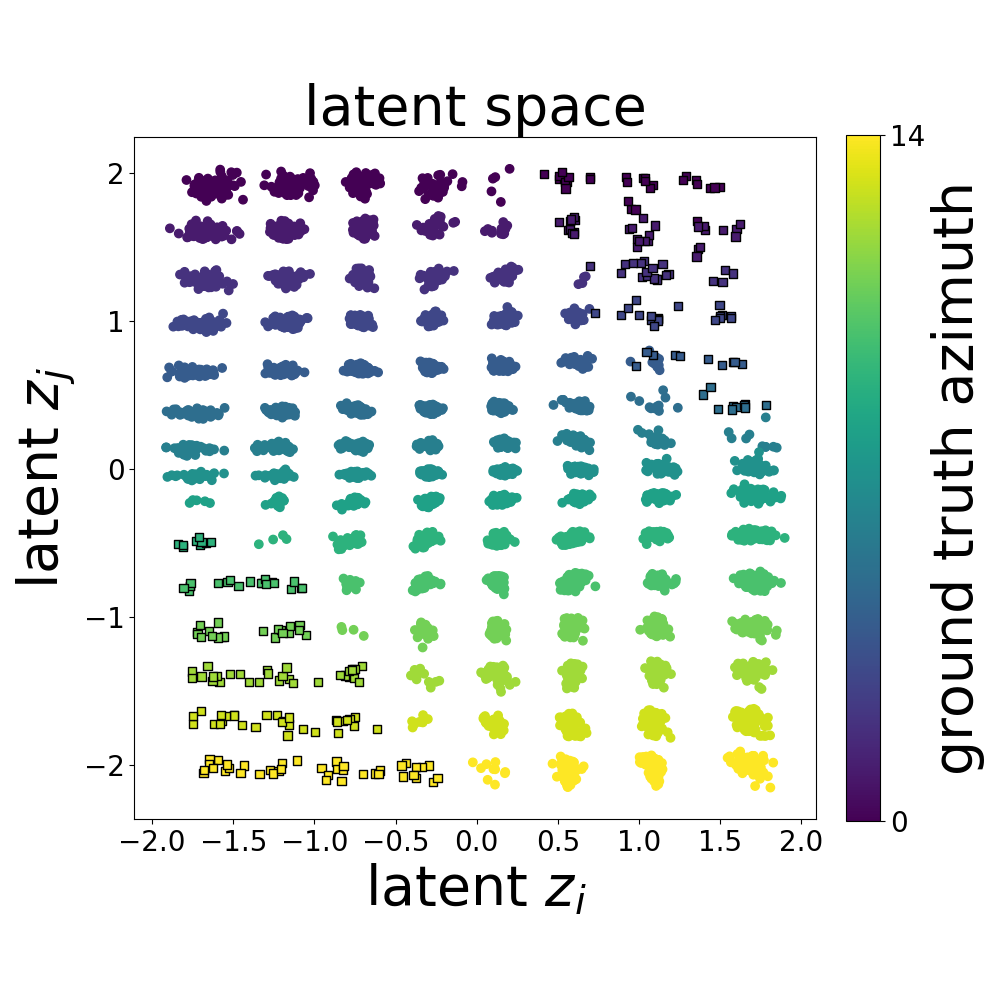}
\endminipage
\vskip 0.1in
\minipage{0.3\columnwidth}
  \includegraphics[width=\columnwidth]{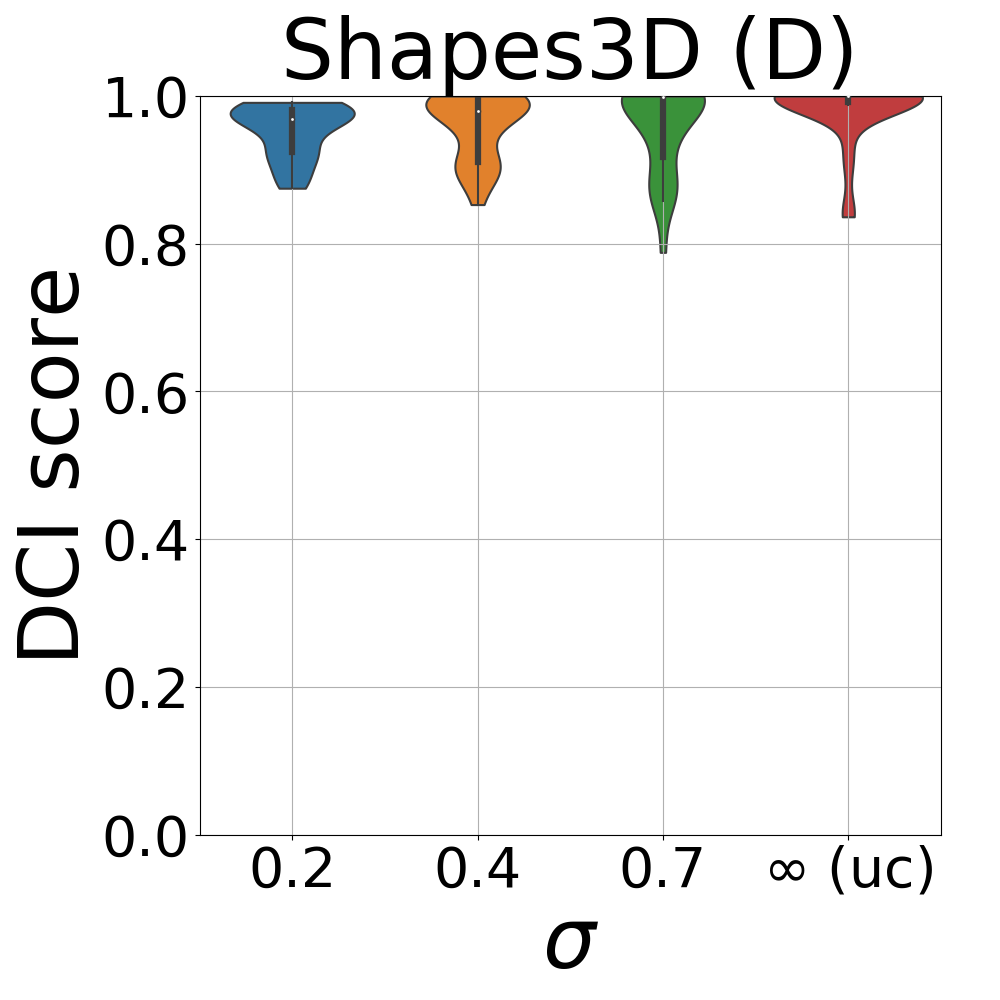}
\endminipage\hfill
\minipage{0.34\columnwidth}
  \includegraphics[width=\columnwidth]{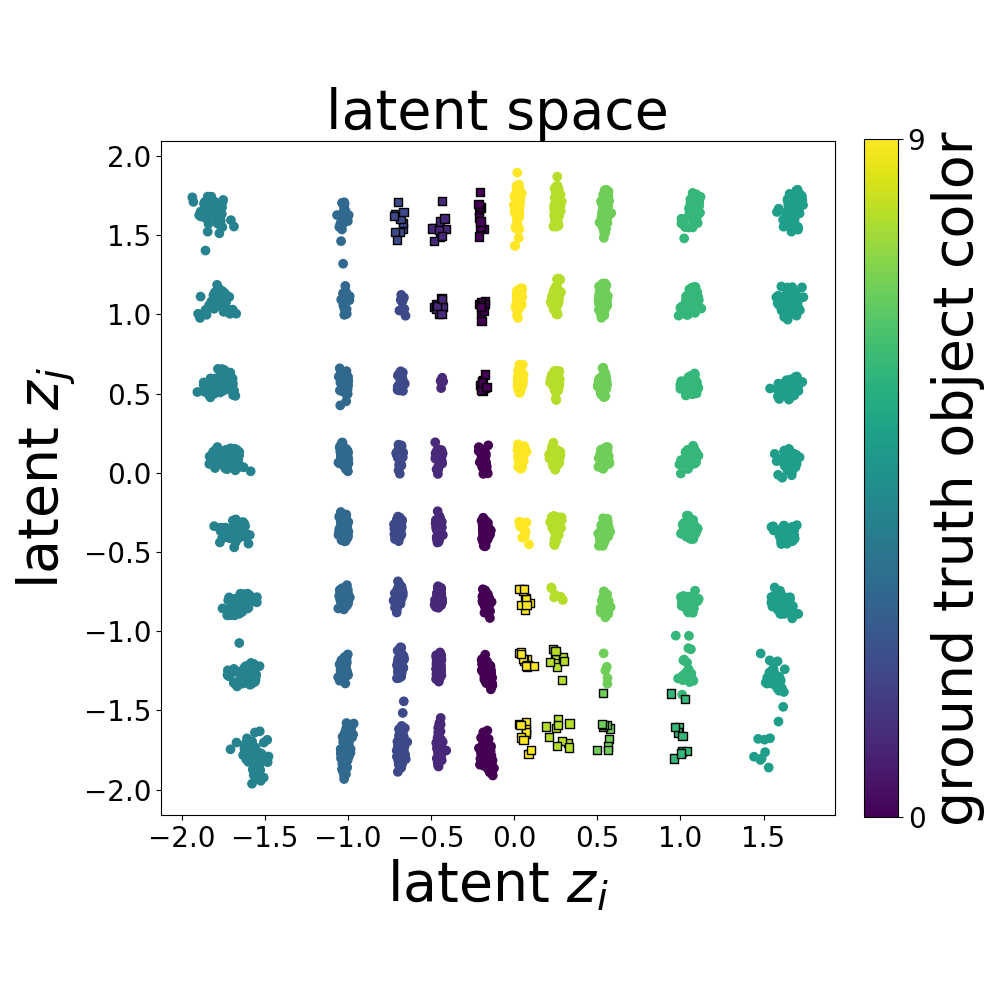}
\endminipage\hfill
\minipage{0.34\columnwidth}
  \includegraphics[width=\columnwidth]{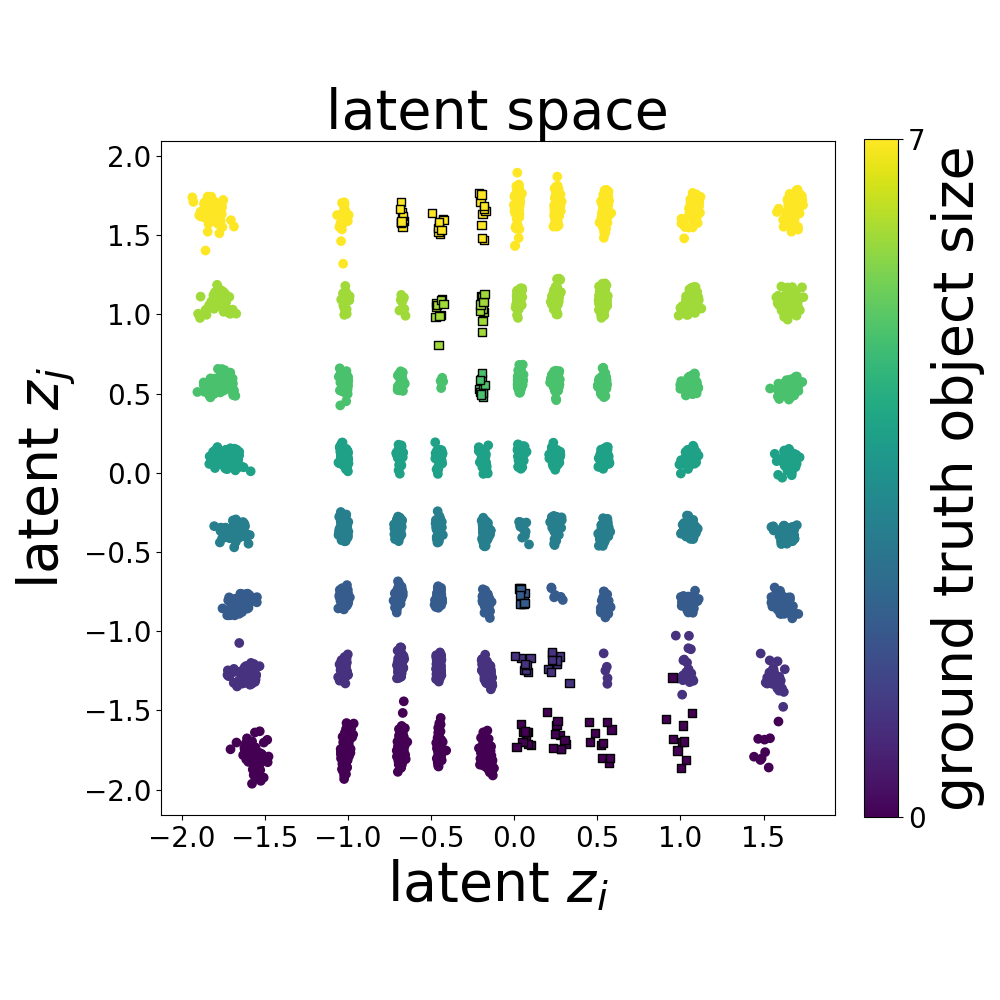}
\endminipage
\vskip 0.1in
\minipage{0.3\columnwidth}
  \includegraphics[width=\columnwidth]{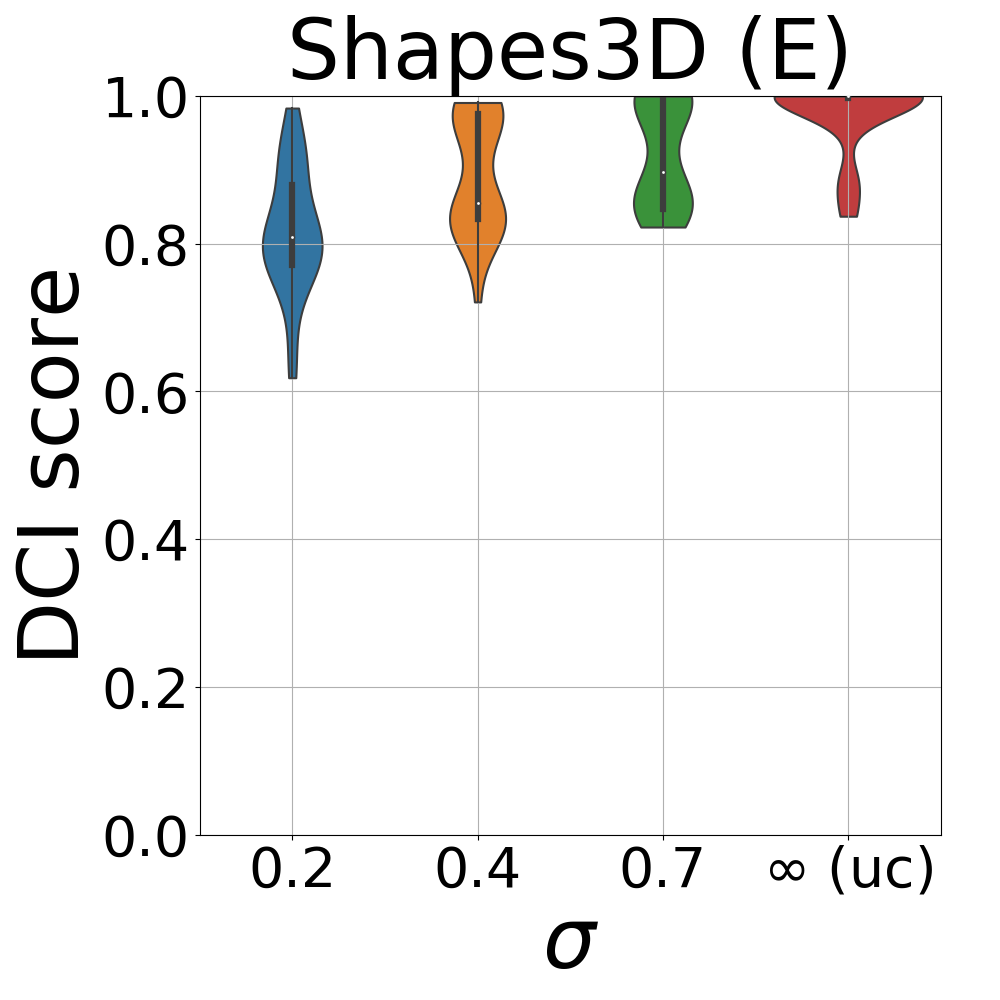}
\endminipage\hfill
\minipage{0.34\columnwidth}
  \includegraphics[width=\columnwidth]{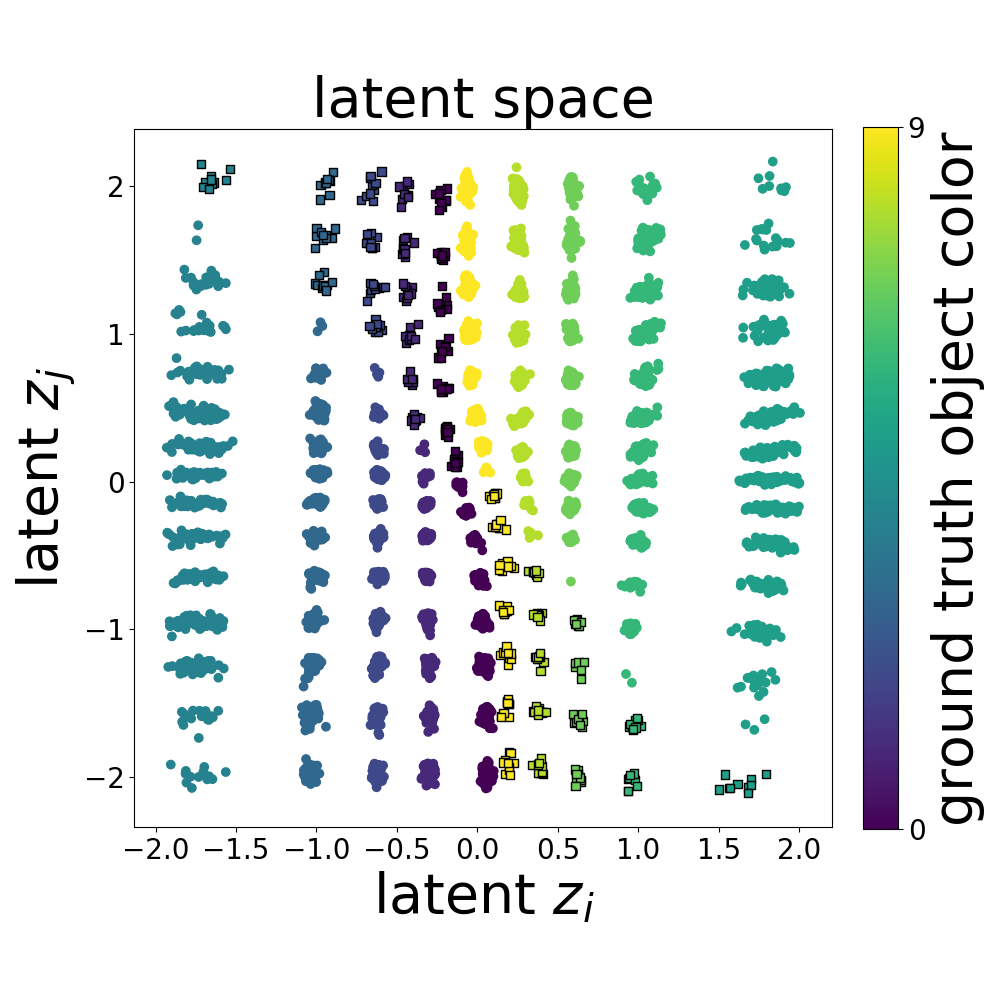}
\endminipage\hfill
\minipage{0.34\columnwidth}
  \includegraphics[width=\columnwidth]{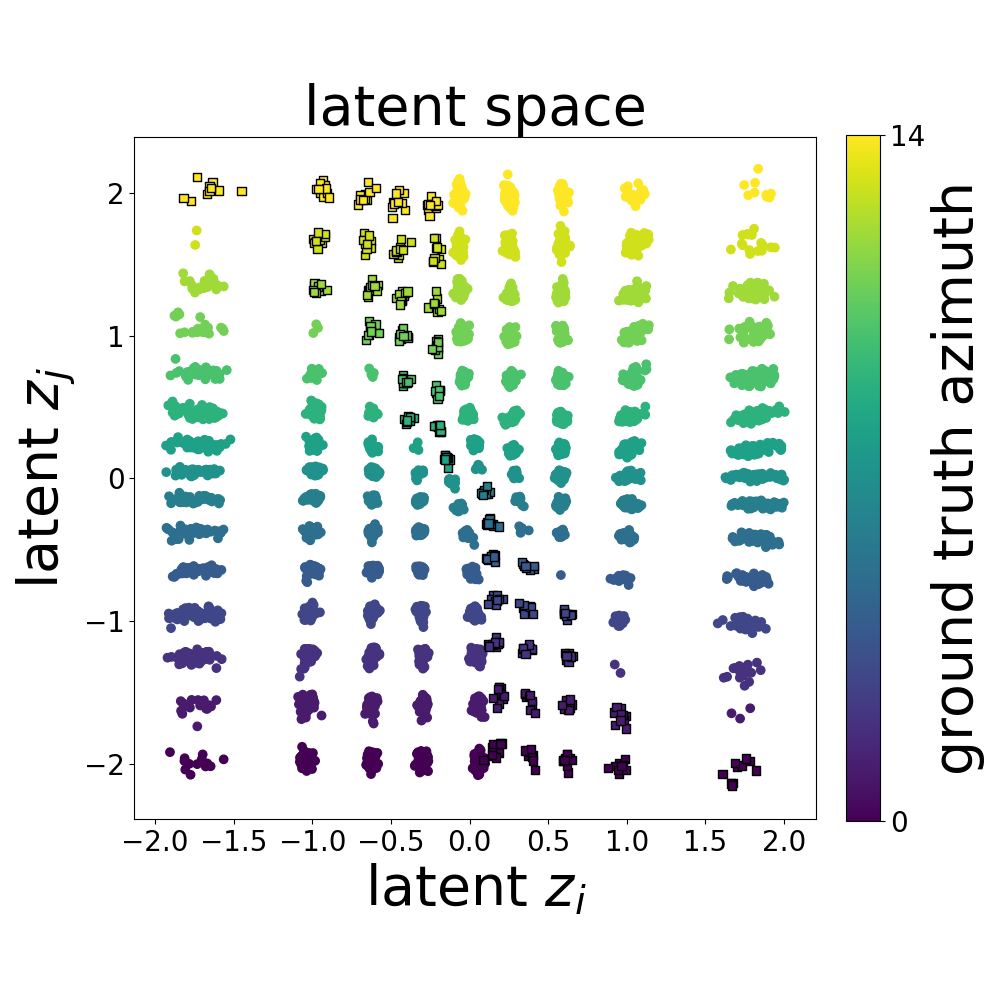}
\endminipage
\caption{Left: For the weakly supervised scenario using correlated observational data trained models on Shapes3D (A), (D) and (E) correlating object color and azimuth learn consistently improved, often perfect, disentangled representation across all correlation strengths. Right: Latent dimensions of a best DCI model trained on strongly correlated observational data. Representations are strongly axis-aligned with respect to both of the correlated variables ground truth values (right).}
\label{fig:o1_weakly_supervised_study_dci_and_latent}
\end{figure}

Using the studied weakly supervision Ada-GVAE method with $k=1$ from \citet{locatello2020weakly}, we showed that weak supervision can provide a strong inductive bias capable of finding the right factorization and resolving spurious correlations for datasets of unknown degree of correlation. Besides the results shown on Shapes3D (A) in the main paper, representative latent space visualizations that show strong axis alignment across all three correlation variants in Shapes3D (A, D, E) are shown in \cref{fig:o1_weakly_supervised_study_dci_and_latent}. This study contains a total of 360 trained models.

In addition to the experiment from the main paper where pairs are constructed solely from the correlated observational data, we want to study two scenarios where we have some intervention capabilities on the FoV to generate training pairs. The resulting distribution of FoVs (still exhibiting correlations) in these pairs depends on whether the correlation between two pairs is due to a causal link or due to a common confounder.

\textbf{Scenario I-1:} We assume there is a confounder (which is not among the observable factors in the data) causing a spurious correlation between the pair of correlated factors. Then, the correlation is broken whenever our interventional sampling procedure yields a pair where the changing FoV is one of the correlated ones.
In that case, the value of the changing variable is sampled uniformly in the second observation of the pair. Note that this still means that the vast majority of sampled pairs exhibit correlated FoV as in most cases the changing factor will be one of the other independent uncorrelated FoV.

\begin{figure}
\minipage{0.3\columnwidth}
  \includegraphics[width=\columnwidth]{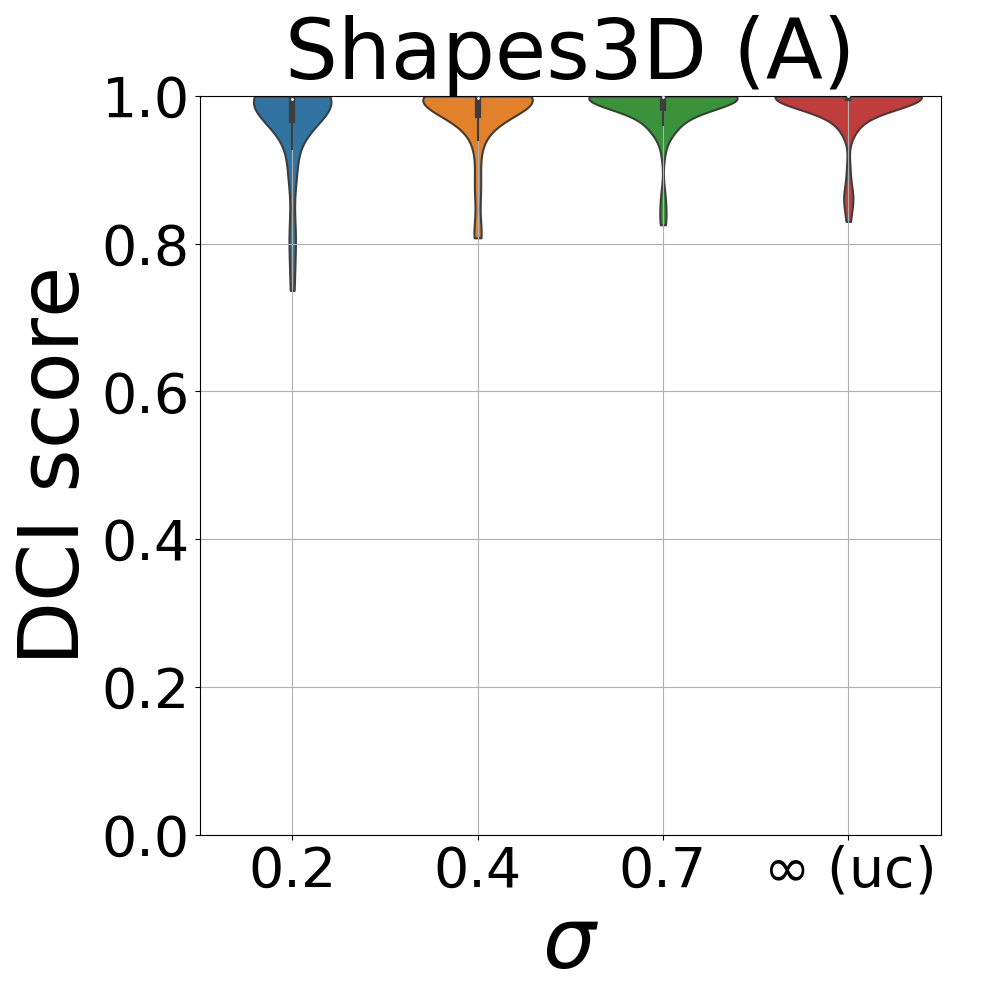}
\endminipage\hfill
\minipage{0.34\columnwidth}
  \includegraphics[width=\columnwidth]{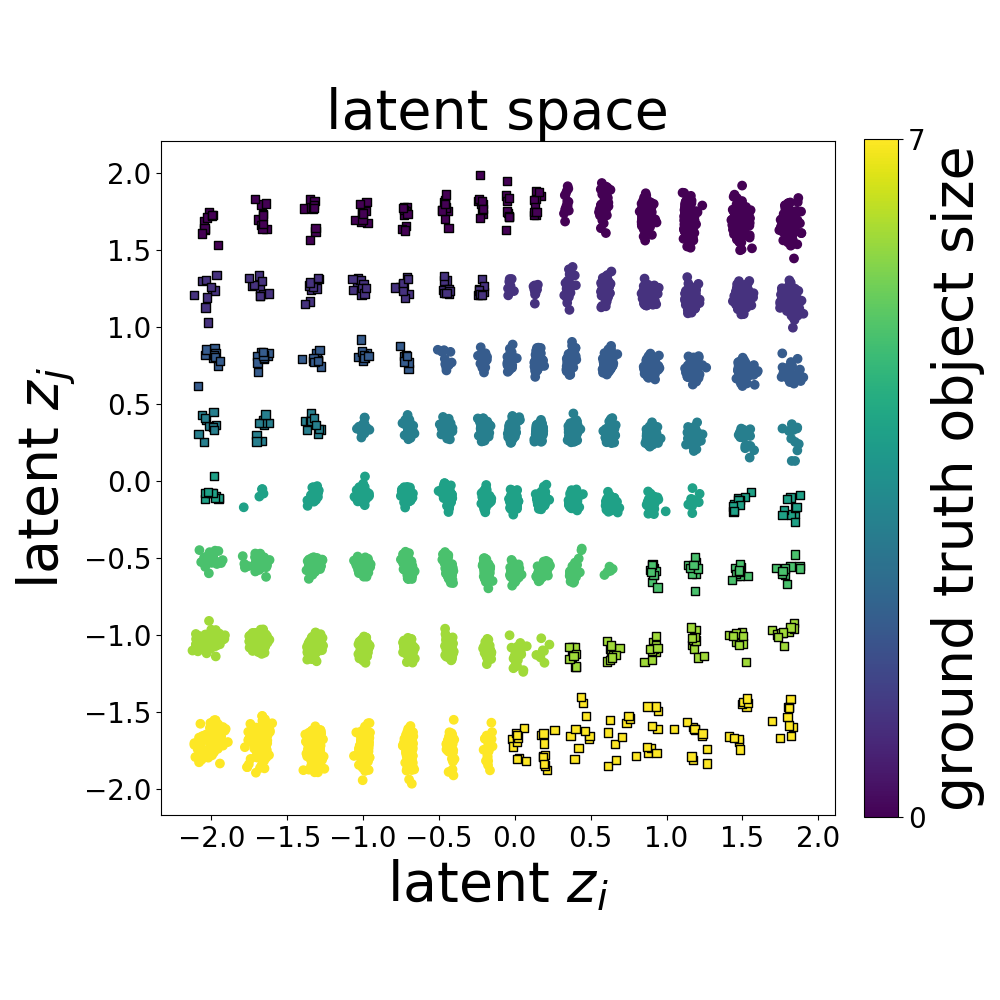}
\endminipage\hfill
\minipage{0.34\columnwidth}
  \includegraphics[width=\columnwidth]{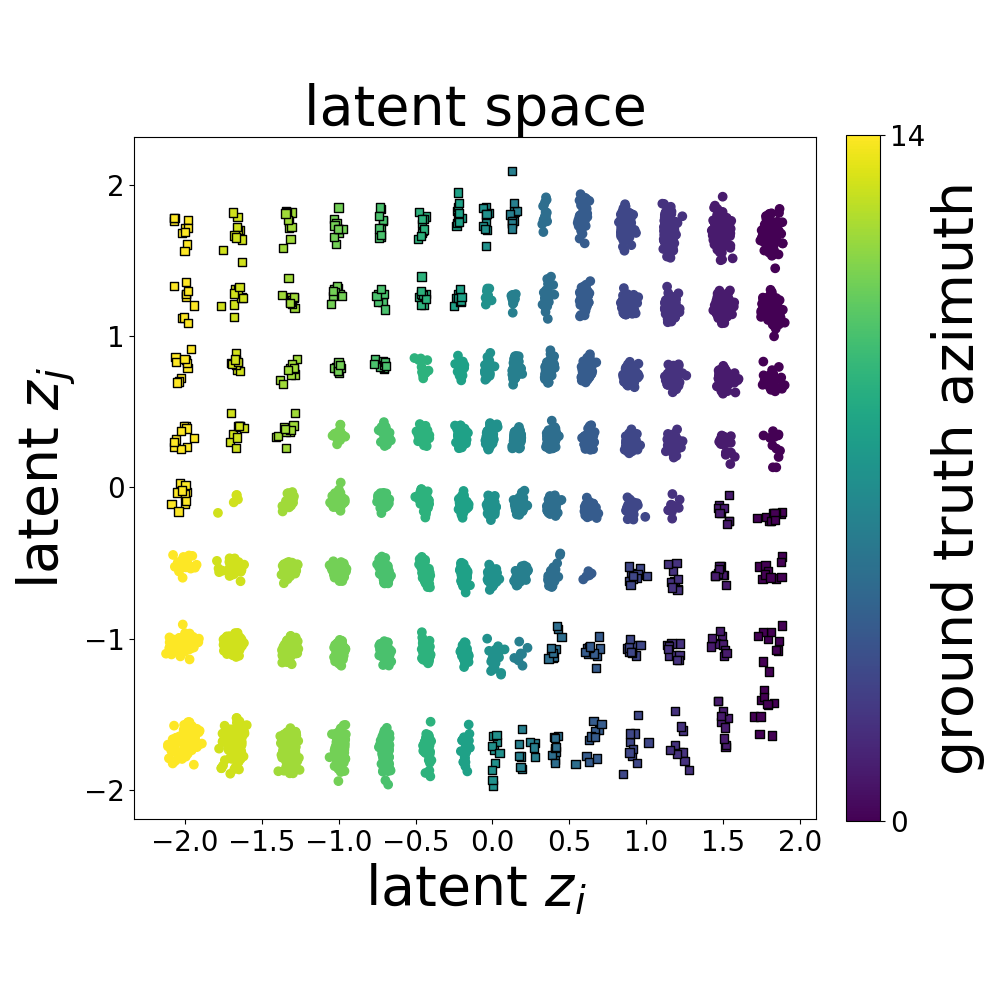}
\endminipage
\vskip 0.1in
\minipage{0.3\columnwidth}
  \includegraphics[width=\columnwidth]{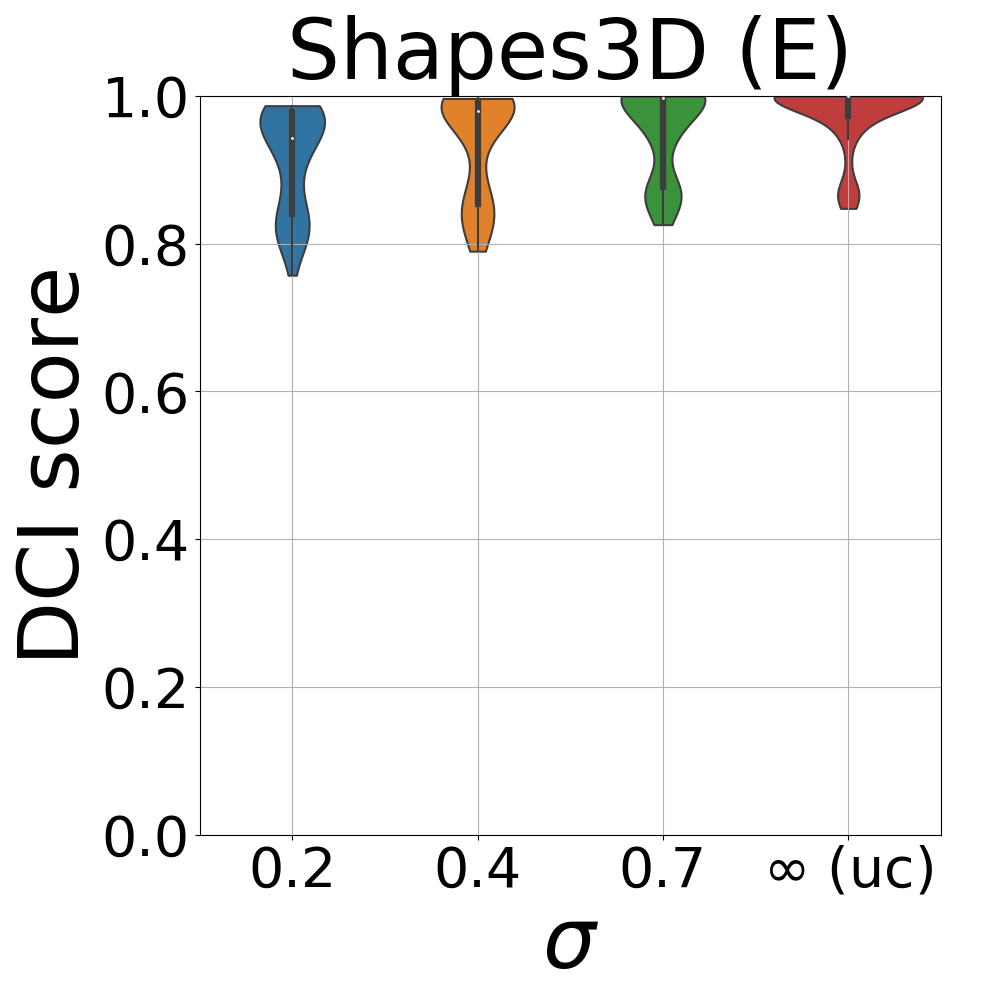}
\endminipage\hfill
\minipage{0.34\columnwidth}
  \includegraphics[width=\columnwidth]{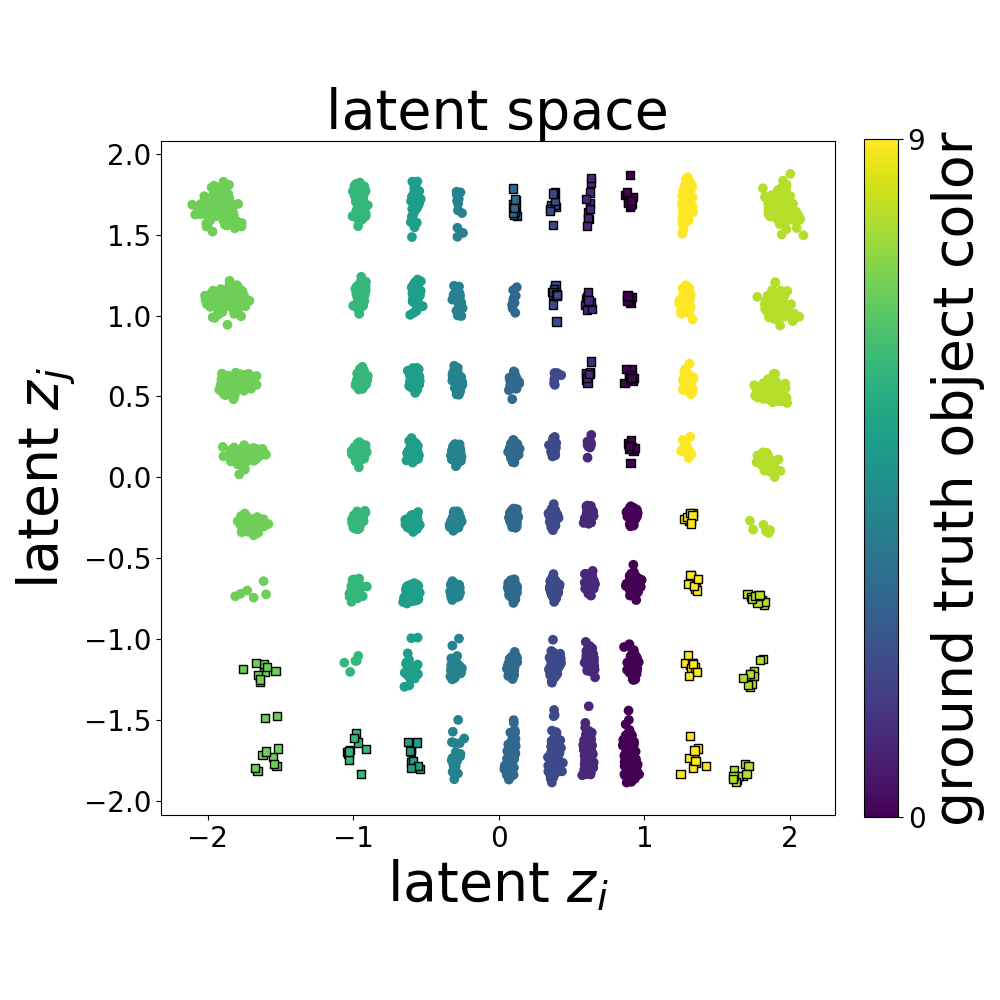}
\endminipage\hfill
\minipage{0.34\columnwidth}
  \includegraphics[width=\columnwidth]{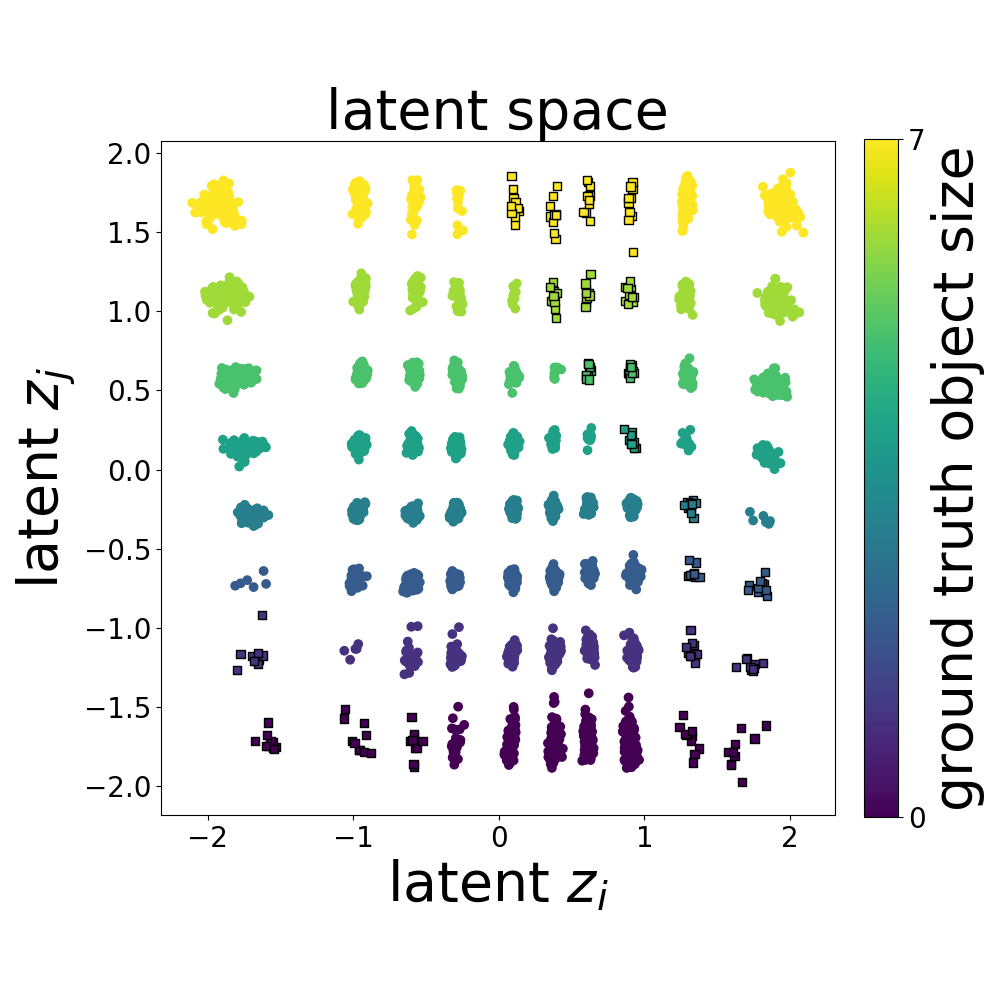}
\endminipage
\vskip 0.1in
\minipage{0.3\columnwidth}
  \includegraphics[width=\columnwidth]{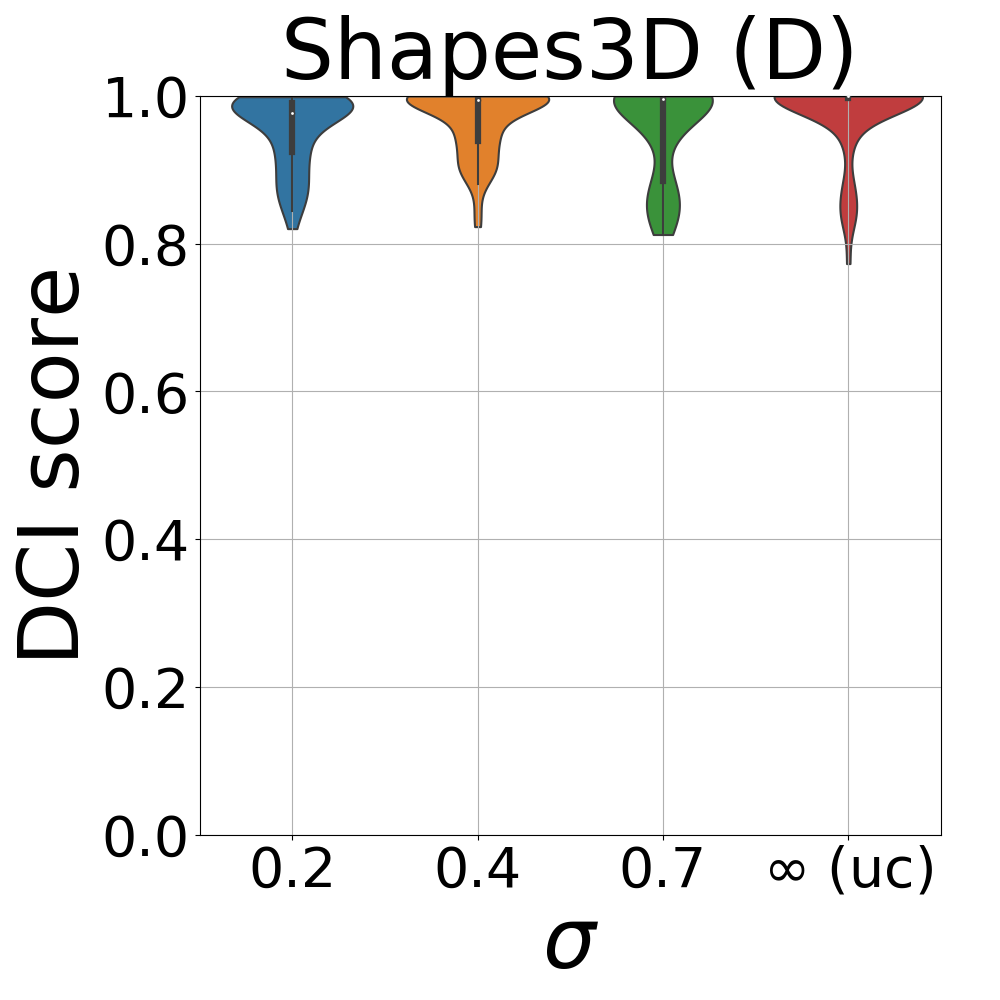}
\endminipage\hfill
\minipage{0.34\columnwidth}
  \includegraphics[width=\columnwidth]{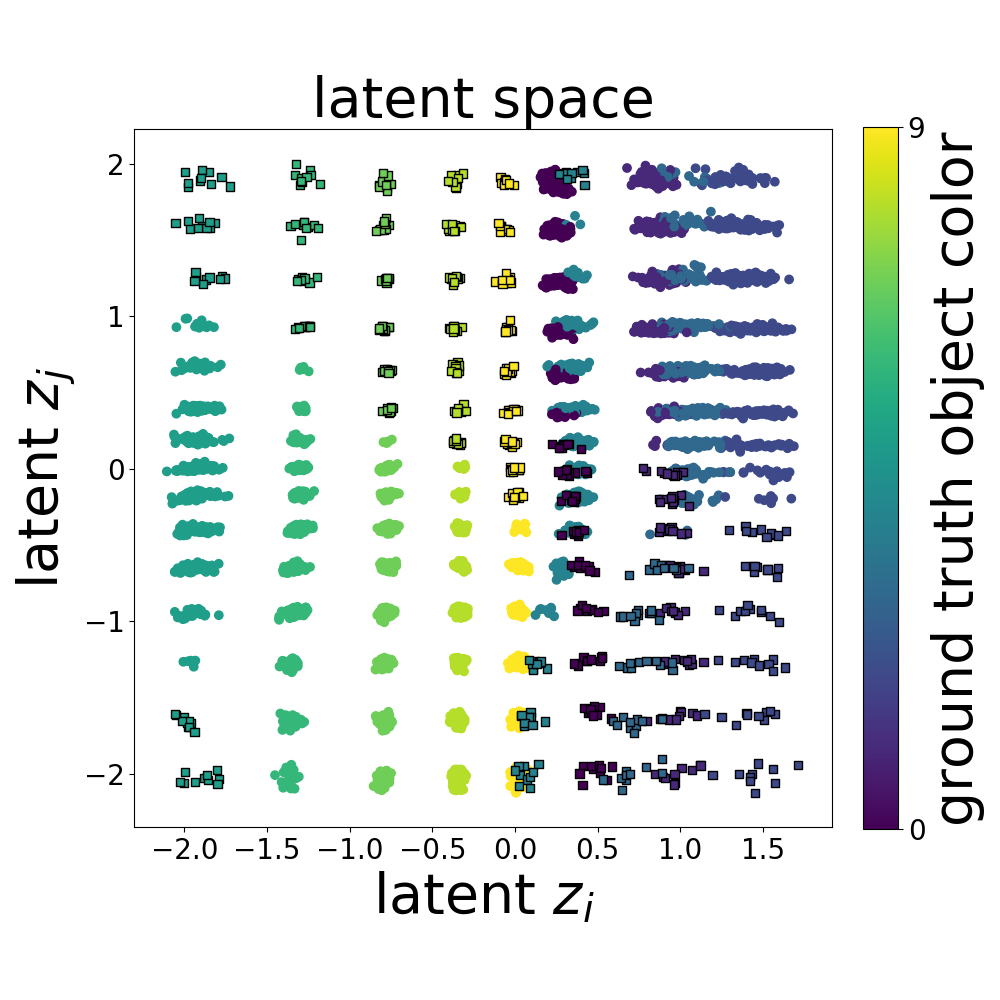}
\endminipage\hfill
\minipage{0.34\columnwidth}
  \includegraphics[width=\columnwidth]{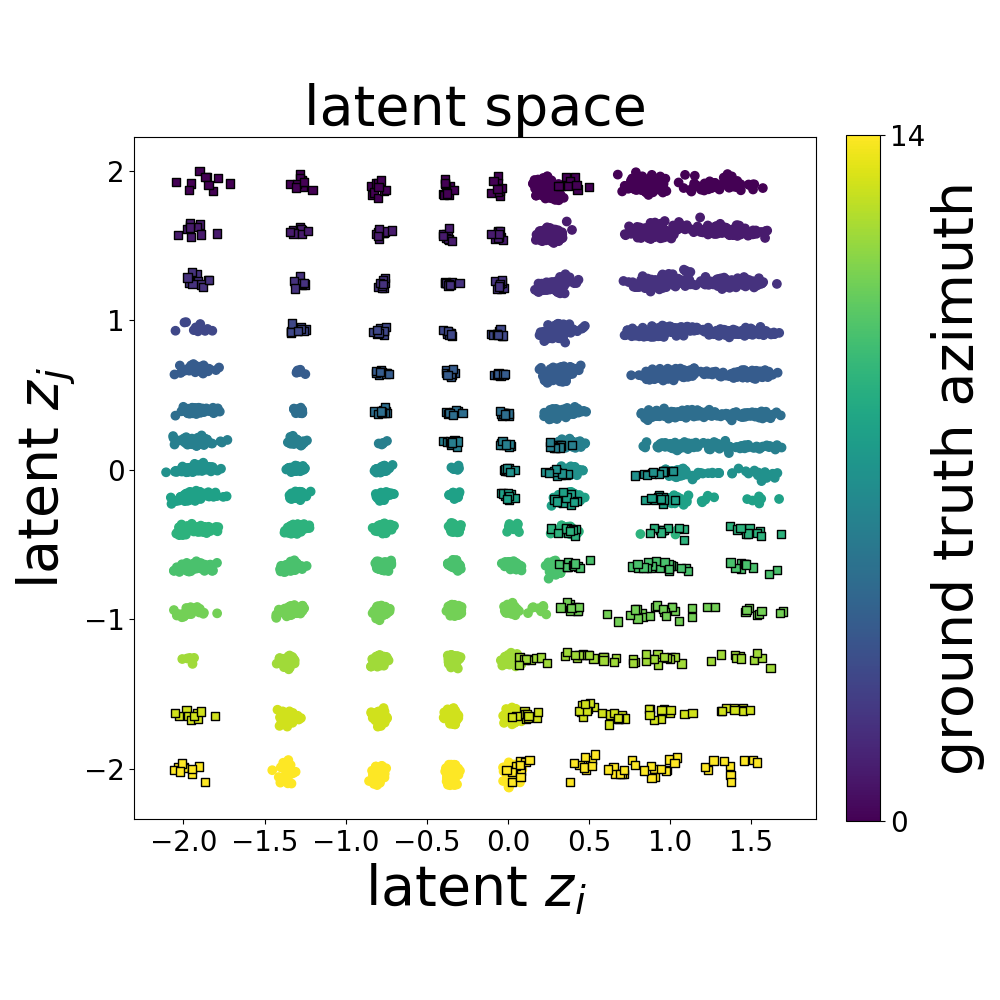}
\endminipage
\caption{Left: For the weakly supervised scenario with intervening capabilities (Scenario I-1) trained models on Shapes3D (A), (D) and (E) correlating object color and azimuth learn consistently improved, often perfect, disentangled representation across all correlation strengths. Right: Latent dimensions of a best DCI model with strong correlation ($\sigma=0.2$). Representations are strongly axis-aligned with respect to both of the correlated variables ground truth values.}
\label{fig:i1_weakly_supervised_study_dci_and_latent}
\end{figure}

As under the default scenario from the main paper, we consistently observe high disentanglement models, often achieving perfect DCI score irrespective of correlations in the data set. 
This is depicted together with some selected latent space visualizations that show strong axis alignment in \cref{fig:i1_weakly_supervised_study_dci_and_latent}.
The latent spaces of the correlated FoV in the train data tend to strongly align their coordinates with the ground truth label axis. We chose 10 random seeds per configuration in this study, yielding 720 models in total.

\begin{figure}
\minipage{0.3\columnwidth}
  \includegraphics[width=\columnwidth]{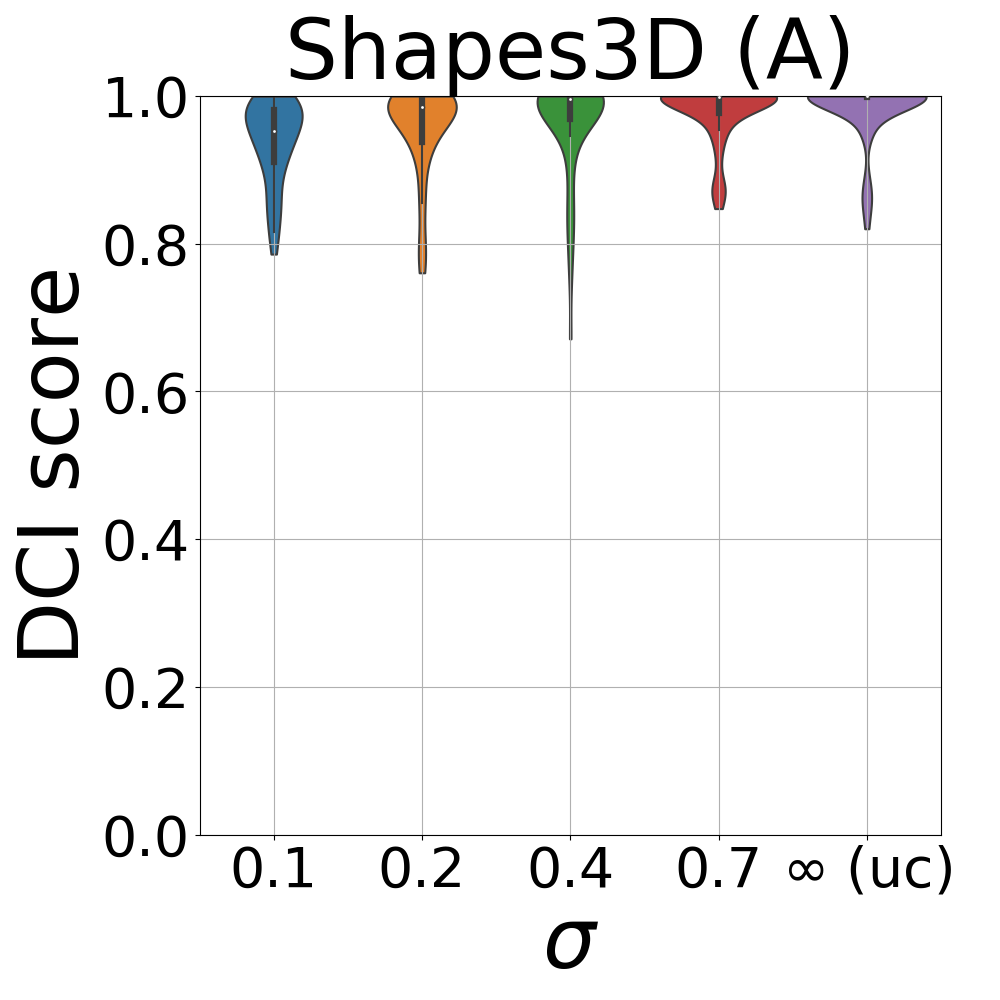}
\endminipage
\minipage{0.34\columnwidth}
  \includegraphics[width=\columnwidth]{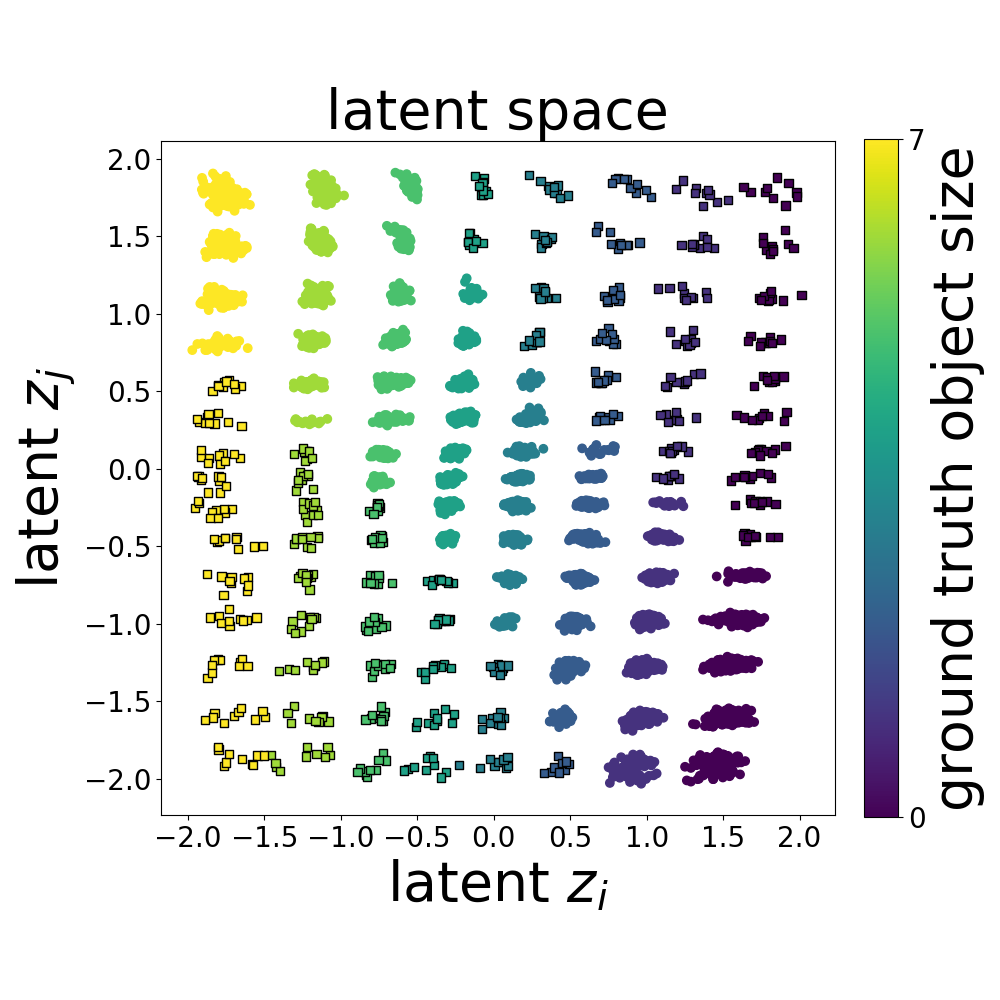}
\endminipage\hfill
\minipage{0.34\columnwidth}
  \includegraphics[width=\columnwidth]{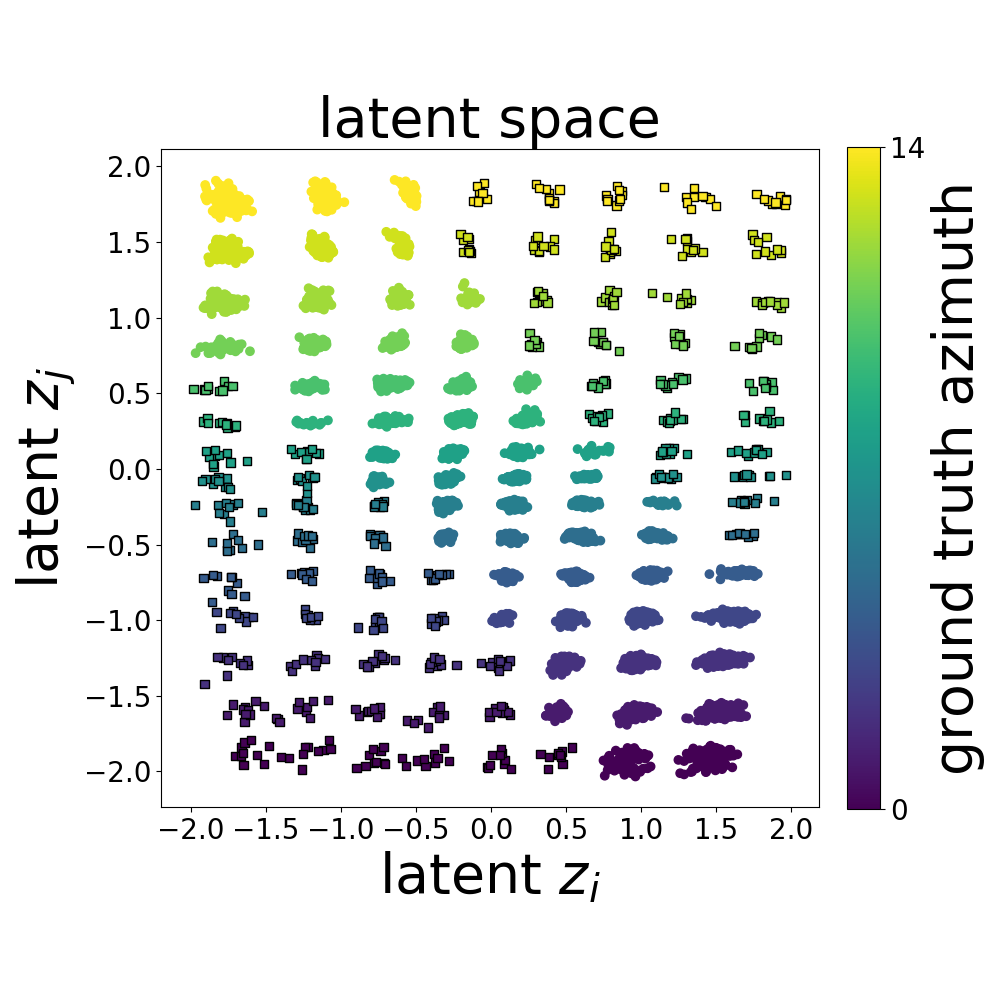}
\endminipage
\caption{DCI scores and latent spaces show strong disentanglement using weak supervision  with intervening capabilities (Scenario I-2) - even under the stronger assumption that sampling of observation pairs follow its causal generative model. We show the learned latent space encoding of the two correlated factors of variation for a model on Shapes3D with $\sigma = 0.1$.}
\label{fig:additional_results_causally_weak_supervision}
\end{figure}

\textbf{Scenario I-2:} Let us assume $c_1$ causes $c_2$ in our examples, which manifests as the studied linear correlation.
If we intervene on (or ``fix'') all factors except for the effect $c_2$, we cannot sample uniformly in $c_2$ as it is causally affected by $c_1$.
Intervening on all factors but $c_1$, however, allows us to sample any value in $c_1$ as it is not causally affected by $c_2$.
To test the hypothesis that this constraint also allows for disentangling the correlation, we trained on Shapes3D (A) and sampled pairs consistent with this causal model.
Besides observing visually disentangled factors in the latent traversals, we show a summary of our results in \cref{fig:additional_results_causally_weak_supervision} with the same significant improvements regarding disentangling the correlated FoVs. Besides the correlation strengths used throughout the paper, we additionally trained the same models using a very strong correlation of $\sigma = 0.1$. The study of scenario I-2 thus comprises 300 models.

